# A Theory of Language Learning

Robert Worden

Wellcome Centre for Human Neuroimaging, Institute of Neurology, University College London, London, United Kingdom

rpworden@me.com

## Abstract

A theory of language learning is described, which uses Bayesian induction of feature structures (scripts) and script functions. Each word sense in a language is mentally represented by an m-script, a script function which embodies all the syntax and semantics of the word. M-scripts form a fully-lexicalised unification grammar, which can support adult language. Each word m-script can be learnt robustly from about six learning examples. The theory has been implemented as a computer model, which can bootstrap-learn a language from zero vocabulary.

The Bayesian learning mechanism is (1) *Capable*: to learn arbitrarily complex meanings and syntactic structures; (2) *Fast*: learning these structures from a few examples each; (3) *Robust*: learning in the presence of much irrelevant noise, and (4) *Self-repairing*: able to acquire implicit negative evidence, using it to learn exceptions. Children learning language are clearly all of (1) - (4), whereas connectionist theories fail on (1) and (2), and symbolic theories fail on (3) and (4).

The theory is in good agreement with many key facts of language acquisition, including facts which are problematic for other theories. It is compared with over 100 key cross-linguistic findings about acquisition of the lexicon, phrase structure, morphology, complementation and control, auxiliaries, verb argument structures, gaps and movement - in nearly all cases giving unforced agreement without extra assumptions.

***Author's Note – May 2021****: This paper was written in 1996, and has been available in various places on the Internet since then – for instance, on ResearchGate since about 2012. I am now posting the paper on arXiv, with no changes except for re-formatting, for the following reason:*

*This paper shows how in any complex domain of cognition (such as language, navigation or 3-D spatial cognition), where knowledge can be represented in feature structures, it is possible to learn those feature structures from a few learning examples each – essentially as fast as is permitted by a Bayesian speed limit on learning. This result is now important because, after more than thirty years of research into neural nets, they are still not capable of fast learning in complex domains. It is time for renewed investigation of computational models which can learn rapidly, as animals and people do.*

*Some of the notation, notably the use of the term 'Trump links' now seems unfortunate. This term was motivated by the idea of trump cards in Bridge, and has not been changed.*

# 1. Introduction

Language acquisition has been high on the agenda of cognitive science for forty years, shaping theoretical linguistics and informing many child language studies. There are many theoretical ideas about parts of the acquisition process. Yet there are very few fully-articulated, workable models of first language learning — and even fewer have been compared with a wide range of data. A notable exception is Pinker's (1984, 1989) theory of language acquisition.

This paper describes a new broad-scope theory of language learning, with the following features:

- rapid, robust learning of language from unreliable and noisy data
- most features of adult language acquired by a single learning mechanism
- integrated learning of syntax, semantics and segmentation
- a working computational model, which can bootstrap to learn language from zero vocabulary
- a firm mathematical basis, linking linguistic theory and learning theory
- an evolutionary account of the origins of language and language learning
- good agreement with a lot of data on child language learning

The last point is the most important. I have compared the theory with over 100 key facts about language acquisition, finding good, unforced agreement in the vast majority of cases. In many cases the theory can give a clear, crisp account of facts which are puzzling in most current language learning theories. These comparisons are described in section 5. I believe no other theory of language learning can claim such broad agreement with the facts.

The theory of language learning is part of a larger theory of language evolution, learning and performance. I need to describe other aspects of this theory for three reasons:

1. To establish that what is learnt really is language, and not just a toy subset;
2. To motivate aspects of the learning mechanism.
3. Because understanding and production are the only windows to measure children's language learning

These descriptions of the non-learning aspects are kept as short as possible, to keep the focus on the learning theory (section 3), and the comparisons with data (section 5).

Theories of language learning have been polarised between two camps:

- Chomskian theories, in which the abstract structures of adult language are acquired by innate language-specific mechanisms.
- Broader frameworks such as cognitive linguistics, in which general social/cognitive mechanisms are used to learn language in stages of development.

This theory does not fit neatly in either camp. It does not posit *de novo* language-specific structures or learning mechanisms in the brain, nor does it rely on broad ill-defined learning mechanisms; it proposes a Bayesian learning mechanism, evolved specifically for primate social intelligence and extended for language, with a precise mathematical structure. This structure underpins the robustness, expressiveness and diversity of languages.

The mathematics of the theory are in three linked parts: (1) Script algebra, which is the discrete mathematics of feature structures, and will be familiar to many computational linguists; (2) M-script algebra, which extends this to functions on feature structures (as used, for instance, in categorial grammars) ; and (3) Bayesian learning theory, which is a simple application of probability theory. None of them are complex, or require anything beyond school maths; but the power and self-consistency of the learning theory hinges on them. This mathematical/computational basis is established in sections 2 and 3.

While the maths is elementary, it may be unfamiliar and inaccessible to some. Fortunately, many of its important consequences can be understood by a simple analogy to chemistry, which I shall develop alongside the maths, in highlighted paragraphs.

**Section 2** describes the theory of language understanding and generation. It is a unification-based theory, where sentence meanings are feature structures, built up by successive unifications of meaning elements. Many syntactic constraints are constraints on the unifiability of feature structures. The theory is comparable with other unification-based grammars such as LFG and GPSG, showing that it has similar power - and can handle complex features of many adult languages. Like them, it has a reversible model of language understanding and generation.

The theory is fully lexicalised; every word (or word sense) is represented in the brain by a structure called an m-script, which embodies all the syntax and semantics of the word. There are no separate phrase structure rules, transformations or parameters. Therefore if we can learn the m-scripts for words, we can learn a language.

Since our knowledge of child language acquisition comes only from studies of language production and comprehension, we need a theory of production and comprehension in order to compare the learning theory with data. Effects which have been attributed to learning limitations can often be understood as production effects - arising from children's strategies for speaking with limited vocabularies. The model of language production is a key part of the theory.

**Section 3** describes the process for learning the m-script for each word, and how this leads to `bootstrap' learning of a language. Many of the background assumptions are as in Pinker's theory - for instance, that the child learns by hearing sentences in contexts where he can infer their meaning non-linguistically. But the statistical and mathematical basis of learning is different.

It is a Bayesian learning theory, which can be shown to give optimum learning performance. The learning procedure projects out common structure from examples (rejecting random extra noise), and has a Bayesian criterion of sufficient evidence. This means it can learn the m-script for any word from a few noisy examples. It can gather implicit negative evidence and learn from it.

**Section 4** discusses the evolutionary origins of language, and the processes of historic language change; there are parallels between the two. I propose that the capacity to use scripts (which underlie language meanings) evolved to support primate social intelligence; so they have a 20 million year evolutionary history and require a fast, robust learning mechanism. M-scripts arose more recently, in part to support a primate theory of mind.

Language learning allows word m-scripts to reproduce and propagate through a population of speakers, and so to evolve (as a form of Dawkins' `memes'). They evolve to maximise the speed and efficiency of communication, and evolve much faster than the brains which use them. This accounts for many prominent features of language (such as approximate regularity, grammatical subjects, and the Greenberg-Hawkins universals) as the results of language change (m-script evolution) rather than innate features of the human brain.

**Section 5** is the longest section in the paper, and compares the predictions of the learning theory with observations. I first discuss some general properties of language acquisition (such as its speed, robustness, and approximate order of acquisition). I then discuss particular observations, in the order: acquisition of the lexicon, phrase structure, morphology, complement-taking verbs, auxiliaries, alternating verb arguments, pronouns and movement, and finally bilingual language acquisition.

For the majority of these 101 comparisons, the m-script theory is in good unforced agreement with the data, not requiring extra assumptions. Where extra assumptions are required, they do not strain credibility. I have found no major conflicts between the theory and the data. However, I have not been able, in the time and space, to examine the data as thoroughly as, for instance, Pinker (1984) does in his comparisons; much work remains to be done for a full evaluation of the theory. Nevertheless, the initial indications from these comparisons are positive.

**Section 6** compares this theory with other theories of language learning, I discuss Pinker's (1984,1989) theories (which have much in common with the m-script theory, and from which I have borrowed the treatments of some phenomena), then discuss Principles & Parameters theories, Connectionist theories, Slobin's Operating Principles, and Siskind's computational model of lexical acquisition.

**Section 7** concludes, summarising the main results of this work.

There are four appendices: (A) describing algorithms for the m-script operations which underpin the theory; (B) showing that Bayesian learning gives optimal performance; (C) deriving a fundamental theorem of language learning; and (D) describing the computer program which implements the theory.

# 2. Language Processing

Main points:

- *Every word is represented by an M-script, which is a function from scripts to scripts.*
- *In language understanding, meaning scripts are built up by applications of word m-scripts, done by m-unification.*
- *Language generation is done by m-unifications in the reverse direction.*
- *The script and m-script operations fit in a simple algebraic structure, much like set theory.*

This section steps back from the problem of language learning, to describe a theory of language understanding and generation. This defines what has to be learned, to learn a language. The theory is unification-based, with similarities to LFG (Kaplan & Bresnan 1981), categorial grammars (Oehrle et al 1988), and GPSG (Gazdar et al 1985), but with distinctive features of its own. It has a simple mathematical structure.

Using the mathematics, the essence of the theory of language, and of the learning theory, can be stated quite simply; so I ask the reader's patience in setting up this elementary mathematics (or, for the non-mathematical, the chemical analogy).

## 2.1 Representing Language Meanings as Scripts

In this theory of language, the meaning of every sentence is described by a symbolic structure called a *script*. A script is a kind of feature structure, as used in much computational linguists (e.g Shieber 1986; Pollard & Sag 1987) , and in lexical and conceptual semantics (e.g. Levin & Pinker 1991; Jackendoff 1991). For instance, scripts are directly comparable to the uncommitted f-structures of LFG.

I assume that when understanding a sentence, we form the script representation in our heads, before forming other representations such as mental images and mental models (Johnson-Laird 1983). Similarly to speak a sentence, we first form its script meaning structure in our heads [1]

It is proposed that the first script representation capacity evolved some 20 million years ago to support primate social intelligence (see section 4).

A script is a tree-like structure made of a few different types of nodes; each node contains information encoded in the values of various slots. They are pure trees, with no structure-sharing. However, slot values can be variables, and different slots on one script may have the same variable value - a form of value-sharing.

The basic form of the script representation derives from the work of Schank (1972,1977) and Nelson (1985). The choices of tree structure, slots and values used in this representation are broadly derived from the lexical semantics of Jackendoff (1991), Pinker (1989), Talmy (1985) and others (e.g. in Levin & Pinker 1991). There are many similarities to the structures used in cognitive semantics (Lakoff 1987; Langacker 1991).

**Warning**: The script structures used in this paper are intended to be illustrative, not definitive. I have not worked hard to find the very best all-round script representation, or to incorporate all the insights from Jackendoff, Pinker, and others. As this is an evolving theory, the choices of nodes and slots may even change a little between examples. I believe that the script structures used in the brain are actually rather more complex that those illustrated here. However, in most cases, the form of the learning theory, and the tests of it, do not turn on fine details of the script representation; where they do, it will be noted.

An example of a script encoding a simple meaning, `Charlie broke the doll' is shown in figure 1 below. This example is taken from a program, described in Appendix D, which implements the language theory described here.

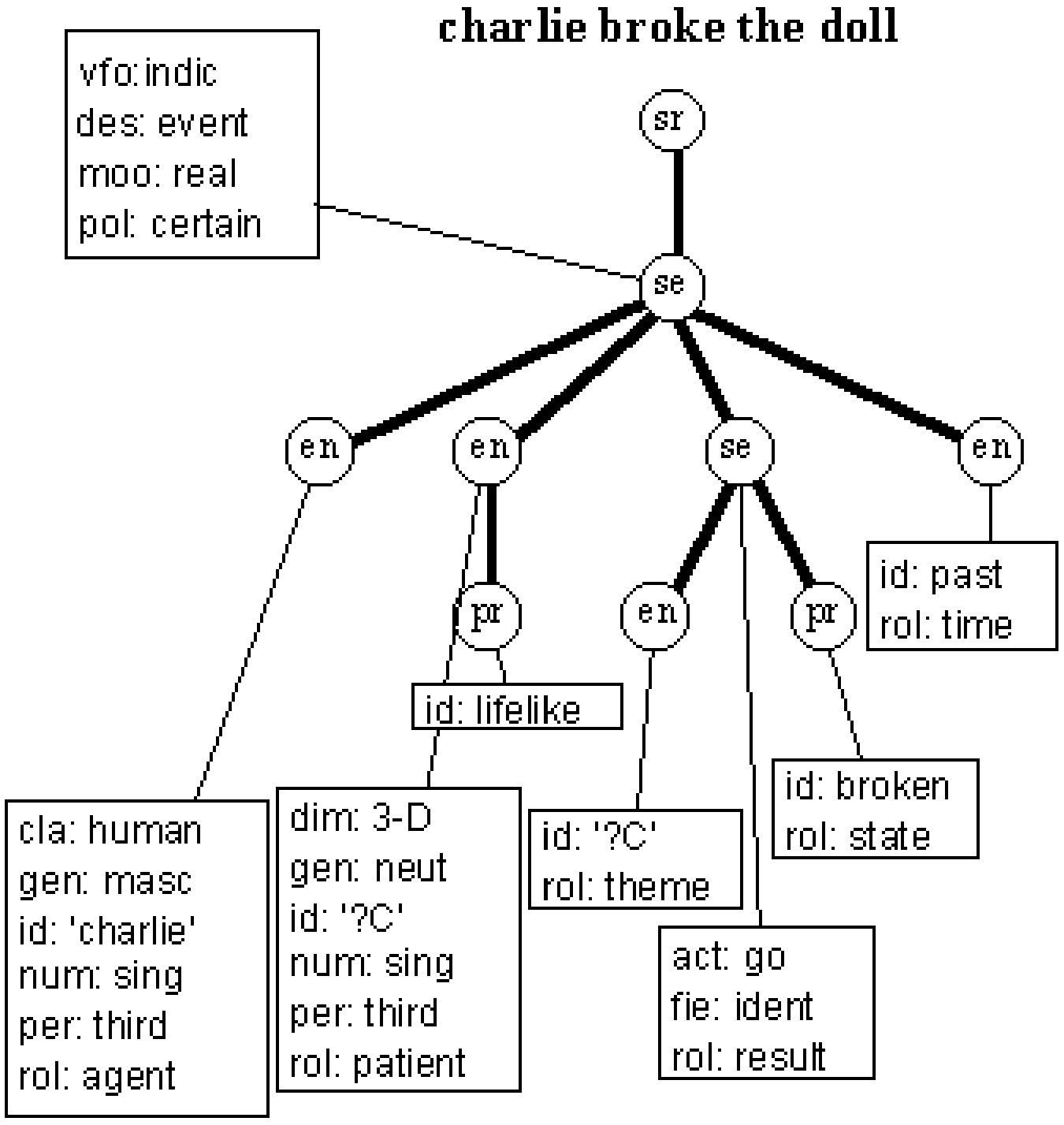

*Figure 1: A typical sentence meaning structure. Heavy lines denote the script tree structure. The light lines connect each script node to a box showing the information at that node, as slot:value pairs.*

Node types are marked inside the node circles, as follows - sr:script; se:scene; en:entity; pr:property. There are only these node types. Slots and their values are shown in the boxes attached to nodes. For most slots, the value is one of a small set of allowed values.

In this simple script, most of the action is defined in one scene, the top 'se' node, which designates an event [des:event]. This is a bounded transitive act [act: act2] which certainly happened [pol:certain] Reading off the nodes below that from left to right, they represent :

(1) Charlie - the agent in this scene, and also the topic; human, masculine, and third person.

(2) The doll - a three-dimensional object with the property of being lifelike. It is the patient of this scene. It is given a variable identity ?C in order to refer to it elsewhere.

(4) The scene which results from Charlie's action; in which the same doll (identity ?C) goes into a state of being broken. Breaking is a causative verb.

(5) The time at which this scene happens, in the past.

Because script trees can be indefinitely broad and deep, they can express an unbounded set of meanings - as is required for language.

Some aspects of meaning cannot be expressed directly in scripts, because there are no innate slots in the script representation capable of expressing them. For instance, the full description of an artefact such as a bicycle or a hammer cannot be expressed in a script. We assume that some script slots may contain links to other meaning representations about hammers - such as mental images, or the procedural skills of using them - which can be accessed after language understanding, or before production, but do not form a central aspect of the language faculty itself.

An important property of scripts (not illustrated in this example) is that a script may represent a sequence of events in a partial or total ordering. Then each event is represented by one `scene' node below a `script' node, with time-order constraints (represented by arrows) between the scene nodes defining the allowed time ordering. This time-ordering relates closely to constituent-order rules in grammars.

We adopt a simple model-theoretic semantics for these scripts, which does not address (for instance) some subtle distinctions of meaning needed for embedded attitude reports, nested quantifiers and so on, as are addressed by Montague semantics (Dowty et al 1981), situation semantics (Barwise & Perry 1983) and other formalisms. These distinctions are not important for many tests of the learning theory, but will need to be addressed at some future time.

In this simple model-theoretic semantics, each script represents a set of possible social situations. A simple script with one scene describes a social situation in which the events of that scene are going on, and in which other things, not mentioned in the script, may also be going on. Similarly, a script with two scenes linked by a time-order arrow describes an extended situation in which those two scenes occur in that time sequence, and other scenes may occur as well.

In this sense any script denotes an infinite set of possible social situations. This set is called the **scope** of the script; the scope of script S is written as σ(S).

**Note**: the three key script operations are denoted by symbols which should be like the set-theory operations of set inclusion, union and intersection, followed by subscripts s for script operations, m for m-script operations. For this hypertext version, I use >s for script inclusion, ∩s for script intersection, and Us for script unification. Similarly I shall use >m , ∩m and Um for the corresponding m-script operations. I shall also use > ∩ and U (without 'subscripts') for set operations.

There are three key operations on scripts (inclusion, unification, and intersection), which are defined in terms of the model-theoretic semantics, and can be done by symbolic manipulations of the tree structures. Their definitions are:

Script A **includes** script B (written as A >s B) if A contains all the information in B, and possibly more. Script inclusion is defined in terms of scope sets:

$$A >s B \text{ iff } \sigma(A) < \sigma(B) \qquad (2.1)$$

That is, A includes B if any situation described by A is also described by B. Intuitively, A must contain all the information in B, and may contain more; but it does not contradict any information in B. Therefore script inclusion is the inverse of *subsumption* for feature structures. With this inversion, the words `subsumption' and `inclusion' can be freely interchanged [2].

Although script inclusion is defined in terms of the infinite scope sets, it can be tested by looking at the tree structures of scripts A and B, to check that all the information in B is also in A. This algorithm is described in Appendix A.

The language learning theory is a Bayesian theory, depending on probabilities of events and prior probabilities of rules. These probabilities figure in the definitions of the two remaining script operations.

For any script S, there is a probability P(S) that the events described in the script take place within some time interval; this is a rough measure of the 'size' of the scope set $\sigma(S)$. We can also define an information content I(S), which can be calculated by adding up the information content of all slots on S. A simple slot `sex', with two equally likely values `male' and `female', adds 1 bit to I(S); similarly a time-order arrow adds 1 bit to I(S). The probability and the information content are approximately related by

$$P(S) = 2^{**}\text{-}I(S) \qquad (2.2)$$

The **unification** of scripts A and B, written as A Us B, is the simplest script which combines all the information in both of them. It is defined in terms of script inclusion:

C = A Us B is the script with smallest information content I(C) satisfying both C >s A and C >s B. (2.3)

So the scope set $\sigma(C)$ is a subset of $\sigma(A)$, and of $\sigma(B)$; it is the largest possible subset expressible as a script. If $\sigma(A)$ and $\sigma(B)$ do not overlap, C is not defined.

If it exists, what does C = (A Us B) look like ? C has all the structure - nodes and slots - of both A and B, but they are matched together to get maximum overlap of nodes and slots - making C as small as possible. Both A and B can be seen within C.

Script unification is very like unification in Prolog (Clocksin & Mellish 1979), or unification of feature structures (Pollard & Sag 1987), and involves some trial-and-error matching of nodes to get the best possible overlap between A and B. There is a fairly simple algorithm for script unification, described in Appendix A.

When two scripts A and B are unified, the information in the result is a set union of the information in the two inputs; any defined slot in either A or B must appear in the result.

***Chemical Analogy** : The unification of two scripts is like a chemical compound got by combining two atoms or molecules. The shared nodes and slots are like the shared valence electrons.*

*Chemical energy = - log(probability) = Information content. The two scripts (atoms or molecules) combine in the minimum-energy, minimum information configuration.*

*Minimum energy = maximum likelihood.*

The **intersection**[1] of scripts A and B, written A ∩s B, projects out all the information which A and B have in common and loses all other information. It, too, is defined from script inclusion:

D = A ∩s B is the script with largest information content I(D) satisfying both A >s D and B >s D. (2.4)

This means that the scope set σ(D) contains σ(A), and contains σ(B); it is the smallest scope expressible in a script which does this. It always exists, because a script with no information has the largest possible scope set, which contains all others.

Script intersection is like feature structure generalisation, or generalisation in Prolog [3]. It is done by a simple algorithm, described in appendix A.

Script intersection is the central operation of the language learning theory, so it is worth illustrating by an example, shown in figure 2.

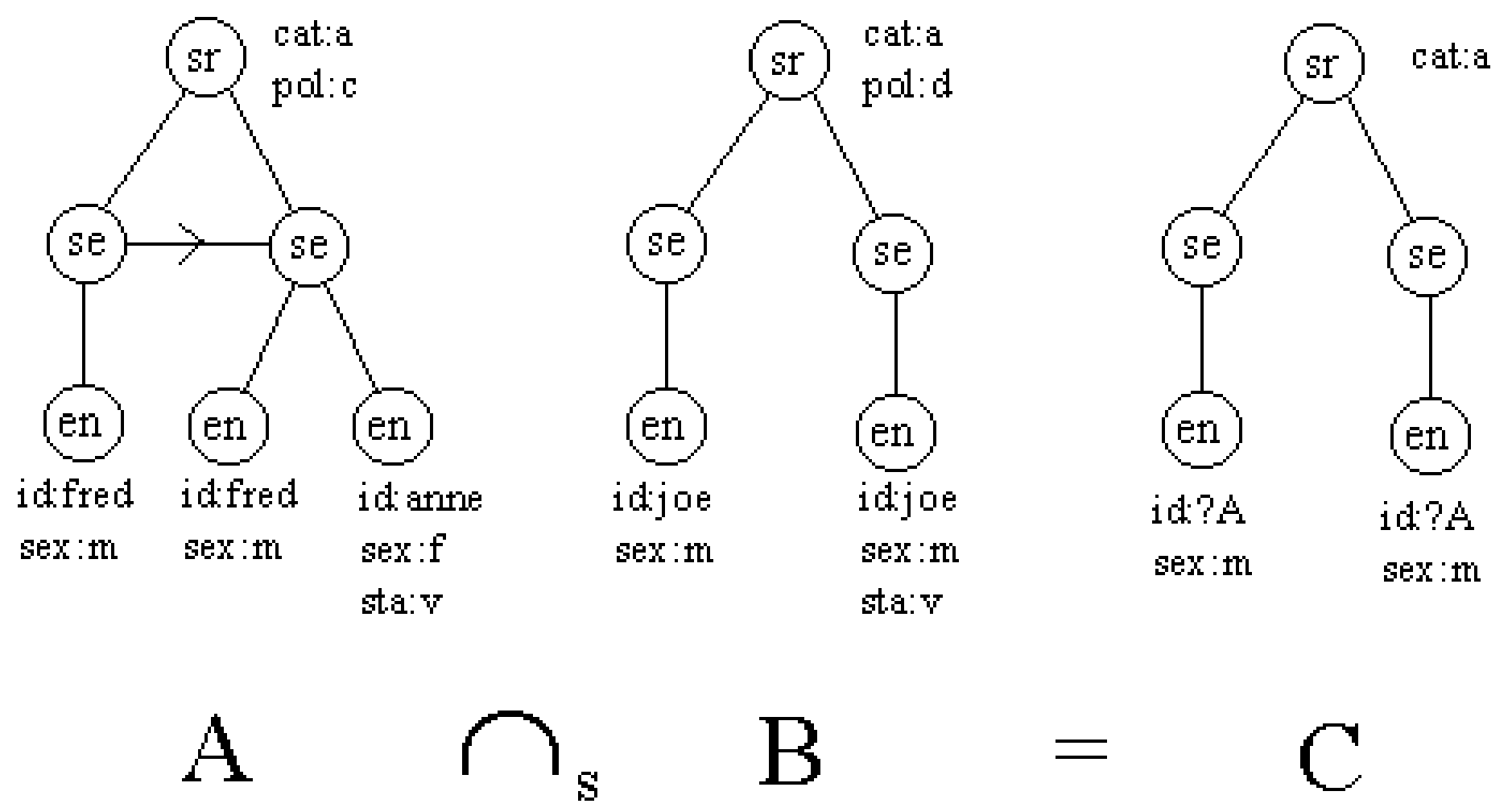

*Figure 2: Illustration of script intersection. Information on nodes is denoted as slot:value pairs.*

This shows the intersection of two simple scripts. It is done by matching them together node by node, retaining only information which is common to both, and trying out different node pairings to maximise the amount of retained information. Points to note are:

[1] Intersection is also called 'generalisation'.

- Only shared nodes appear in the result.
- Any piece of information in the result C appears in both the inputs.
- A has a `time-order' arrow between its two scenes, indicating that one must come before the other, but B does not; so the result C has no time-order arrow.
- Although the slot sta:v occurs in both scripts, it is not in the result, because it does not appear in the same position in the two input scripts.
- No values of the id (identity) slot match between A and B. However, in both A and B, the same id value occurs on two nodes. This information is retained in the intersection C, where the same (but unknown) id value is denoted by the variable `?A'. Intersection discovers these shared variables.
- The second scene node in A has two entity nodes below it. We have to choose which one of these matches with the entity node below the second scene of B, making this choice so as to maximise the amount of retained information.

Script intersection is used in language learning to project out the common meaning of a word from several instances of its use.

***Chemical Analogy** : As unification is like chemical synthesis, script intersection is like analysis.*

In intersection, two script/molecules are compared with each other, and the output molecule is their largest shared substructure.

The terms `include', `unify' and `intersect' have a simple link to set theory, as applied to the information in scripts. Because more information implies a smaller scope set, and vice versa, the relations to scope sets show an inversion. For inclusion, the inversion is evident in equation (A.1): script A includes B if the scope of B includes the scope of A. For intersection and unification, the corresponding relations are:

$\sigma(A \cup_s B) < \sigma(A) \cap \sigma(B)$ (2.5)

$\sigma(A \cap_s B) > \sigma(A) \cup \sigma(B)$ (2.6)

The inversion between $\cup_s$ and $\cap_s$ is evident in the equations.

These simple set-theoretic relations lead to important relations between results of script operations - similar to the relations of elementary set theory. Typical relations of this **script algebra** are :

$A \cap_s B = B \cap_s A$ (2.7)

$A \cup_s (A \cap_s B) = A$ (2.8)

$A \cup_s (B \cup_s C) \approx (A \cup_s B) \cup_s C$ (2.9)

$A \cap_s (B \cup_s C) \approx (A \cap_s B) \cup_s (A \cap_s C)$ (2.10)

Some of these relations are only approximate, as denoted by '≈' ; but in practical examples, the approximate relations turn out to be nearly precise, typically to within one or two slots or nodes difference. For most purposes we shall assume they are exact.

These three operations, of inclusion, unification and intersection (generalisation), are at the core of the theory of language. They are familiar from the study of feature structures and unification grammars (Shieber 1986; Pollard & Sag 1987), but their algebraic structure, in equations like (2.7) - (2.10), is perhaps not so well known. This algebraic structure is crucial for the self-consistency of the learning theory. It can be understood simply in the chemical analogy:

***Chemical Analogy*** *: The relations of the script algebra have analogues which are common-sense `theorems' of chemistry:*

1. *You can synthesise a complex molecule in several different orders (e.g. ABC can be synthesised in order A[BC] or [AB]C) and you will still get the same molecule.*
2. *You can analyse a complex molecule in several different orders, and you will still get down to the same constituents.*
3. *If you build up a complex molecule, then break it down into constituents, you will get back the constituents you started with.*

An important type of script is a **rule script**, used to describe some causal regularity. A rule script has at least a **cause scene** and an **effect scene**, with a time-order arrow between them; the rule states that if the cause scene takes place, then the effect scene is likely to follow with a certain probability P, defined as part of the rule. Variable slot values are important in rule scripts, to state regularities involving `the same individual' in both cause and effect scenes. If a particular script S is an example of a rule script R in action, then S >s R; all the variable identities in R are replaced by fixed identities in S.

Rule scripts are proposed to be an important element of primate social intelligence - the way primates represent important regularities of social life, and predict social outcomes. Examples of these rule scripts are discussed in (Worden 1996b). However, they are not powerful enough to express language rules, and it is worth describing why they cannot.

A rule script may be regarded as a function from scripts to scripts - a function which, given a script containing just the cause scene C, delivers a script containing just the effect scene E. If the rule script has some variable slot values (in both its cause and effect scenes), and the cause scene fixes these variables, they must then have the same values in the effect scene; thus the effect E depends on the cause C. This function from scripts to scripts may be written as E = f(C). However, it is not a very powerful function, since the only things that can change in the result E are the values of slots, each of which has only a small finite set of possible values. So there can only be a small finite set of values for the result E. As the function f can only deliver a finite set of values, whatever its argument, it is called a *bounded* function.

A key property of language is its unboundedness - its ability to express an unbounded set of meanings from a finite set of tokens. So if language meanings are delivered by some form of function application, it is clear that the required functions cannot be bounded functions. That is why rule scripts are not powerful enough, as functions, to perform the operations of language.

## 2.2 Scripts Functions to Build Language Meanings

In this theory, each word of a language is denoted by a function between scripts - a function which, given some scripts as arguments, delivers another script as result. The result script contains both the meaning of the word itself, and other meanings from the arguments. Each word script function combines these meanings to build up more complex meaning scripts.

Sentence meanings are built up by successive applications of word functions, which convert word sounds to meanings. For instance, to understand `Angry Fred shouts', first a word function f for the word `Fred' uses the sound `Fred' as argument, and delivers the noun meaning script F, denoting Fred: f(`Fred') = F. An adjective word function a for `angry' then acts on this noun meaning, a(`angry', F) = M, delivering a script M describing Fred in a state of anger; then finally, the function h for an intransitive verb `shouts' acts on M, h(M, 'shouts') = S, to deliver the meaning script S for the sentence `Angry Fred shouts'.

The succession of states of the sentence is then:

`angry' `Fred' `shouts' (state 1)

`Angry' F `shouts' (state 2)

M `shouts' (state 3)

S (state 4)

where a word script function converts from each state to the next - leading finally to the full meaning.

This functional view of language has been extensively developed in Categorial Grammars (Oehrle et al 1988).

There is always some `last' word function application, which delivers the full meaning of a whole sentence (typically the last word applied is the main verb). The set of possible sentence meanings is unbounded, so we need unbounded script functions to build up full sentence meanings.

The required script functions are called *m-scripts*. An m-script looks very much like a script, with one notational extension; but their semantics are rather different and their three main operations, while analogous to the three script operations of unification, inclusion and intersection, are more powerful.

Possible evolutionary origins for m-scripts are described in section 4 . Here we shall just describe the formal and mathematical properties of m-scripts, as required to describe their role in language.

There is a close parallel between scripts and m-scripts. The key to this parallel lies in defining a model-theoretic semantics for m-scripts, analogous to that for scripts.

Where a script represents a set of situations, **an m-script represents a set of scripts**. The set of scripts denoted by an m-script M is called its **scope**, σ(M).

An m-script looks much like a script, but may also have **trump nodes**, and **trump links**. Trump nodes are nodes labelled !A, !B etc., and a trump link is a curved arrow from one trump node to another.

To define what set of scripts is represented by an m-script M (i.e. to define its scope set), denote the nodes in its tree structure by i = 1,2,...m, and denote the subtree of M with its root at node i by M[i]. Similarly denote subtrees of scripts.

Consider an m-script M with one trump node , at node i. Script S is in σ(M) (i.e. S is in the set of scripts represented by M) if above the trump node i, S = M (i.e. they have exactly the same nodes, and the same information on each node), and the subtrees below the trump node obey

S[i] >s M[i] (2.11)

That is the definition of a trump node. So below the trump node, S must have at least as much information as M, but may have extra nodes and slots - any amount of them.

So the scope of an m-script with a trump node is an infinite set of scripts; the m-script denotes that infinite set. An example of an m-script with one trump node, and some of the scripts in its scope set, is shown in figure 3.

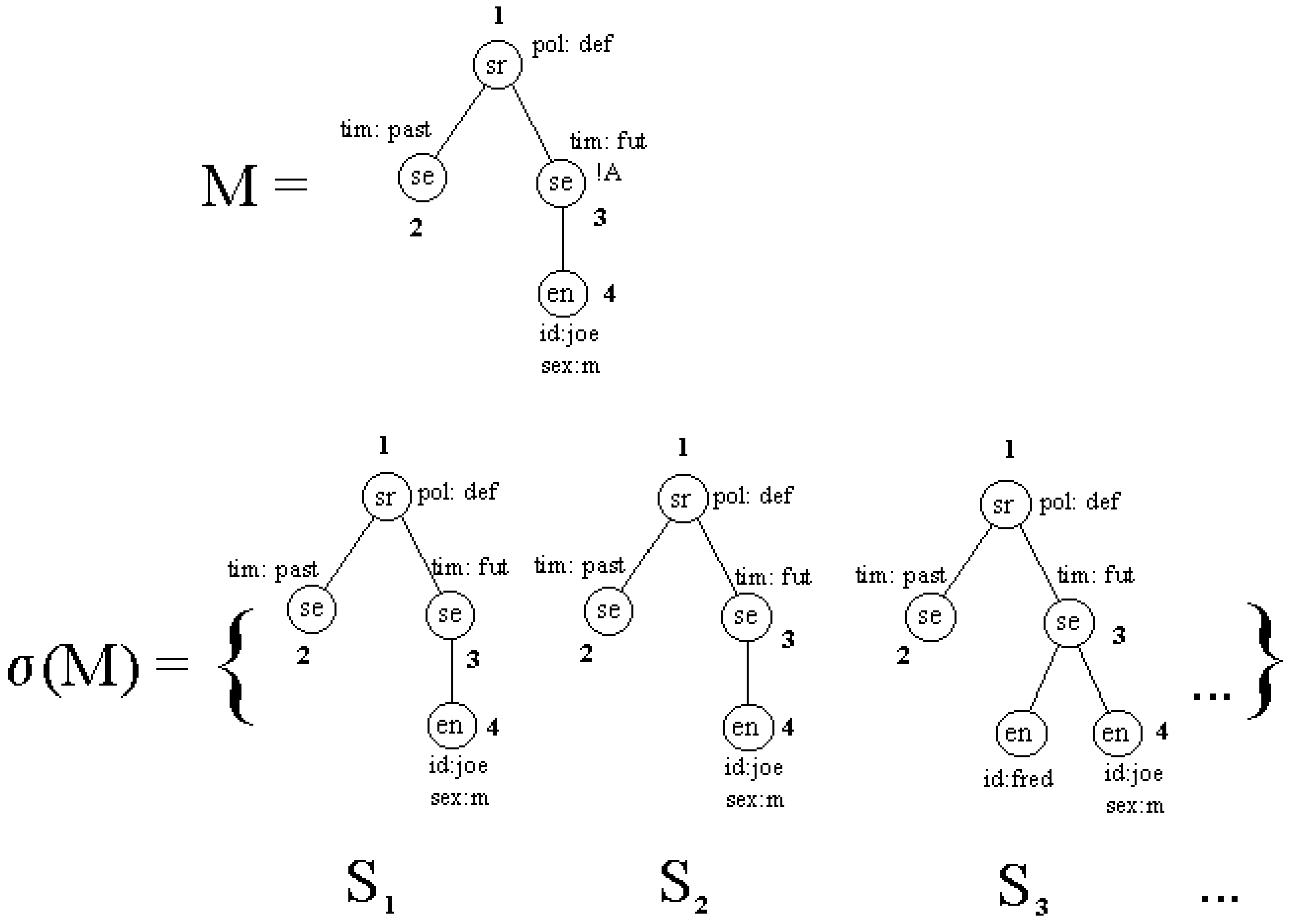

*Figure 3: a basic m-script with one trump node, and a few of the scripts which it denotes.*

This m-script has one trump on node 3. Therefore, for a script to be within its scope, the script must have no extra structure above that node, or below other nodes (such as 2) which are not below node 3. But it may have any amount of extra structure below node 3, as long as that structure includes the structure already below node 3 in M. All these scripts obey S[3] >s M[3], and therefore also obey S >s M.

In practice, m-scripts with a lone trump node are not of much interest for language, and this example was only used to introduce the idea of trump nodes. To be useful, they must be linked together.

For an m-script M with two trump nodes i and j, and a **trump link** from node j to node i, then for a script S to be in the scope of M, above the trump nodes S and M must, as before, be equal. Below the trump nodes, the two subtrees of S must obey S[i] >s M[i], S[j] >s M[j] as before. They must also obey

S[i] = S[j] Us M[i]                    (2.12)

that is the constraint implied by the trump link between nodes i and j. By this equation, each trump link requires a unification - which (in this theory) is what introduces the unification into unification-based grammars.

Trump links are central to the theory of language, so they will be illustrated by a simple example, shown in figure 3 above.

This example shows a simple m-script, M, and a few of the scripts $S_1$ , $S_2$ , $S_3$ ,... in its scope set σ(M) - the scripts that it denotes. The nodes of M have been numbered to make the scope definition clear. M has a trump link from node 2 to node 3, so any script S in its scope must obey S[3] = S[2] Us M[3], where S[3] is the subtree of S below node 3, etc. You can check that the three examples given obey this constraint.

This definition extends straightforwardly to an m-script with several trump links; see Appendix A for details.

So while S can contain extra information (beyond that in M) below both trump nodes, with a trump link the extensions below nodes i and j are not independent. The subtrees S[i]and S[j] must be related by a unification. This means that if the subtree below one trump node is fixed, then the subtree below the other end is also fixed by the trump link. This works in either direction; S[j] is a script function of S[i], or conversely S[i] is a function of S[j]. That is how m-scripts act as script functions.

There is one constraint on well-formed m-scripts. This is that an m-script must be within its own scope. Therefore the subtrees of any m-script must obey its own trump link equations.

The definition of trumps and trump links defines the scope σ(M) of any m-script M. Then, just as for scripts, we can define three key m-script operations in terms of the scopes:

An m-script G **m-includes** m-script H (written as G >m H) if any script in the scope of G is also in the scope of H; that is, if G contains more information about allowed scripts in its scope than H:

G >m H iff σ(G) < σ(H) (2.13)

The relation G >m H can only hold if G has all the structure (nodes and slots) of H, possibly more. The structure of H can be seen within G. But G cannot have any extra trump nodes, unless they are on or beneath trump nodes of H; and G may have extra trump links, not in H.

The **m-unification** of m-scripts M and N, written as M Um N, is the simplest m-script which combines all the information in both of them. It is defined from m-inclusion:

P = M Um N is the m-script with smallest information content I(P) which satisfies both P >m M and P >m N. (2.14)

The information content of an m-script is its usual script information content plus a term for trump structure. An m-script with bigger information content has smaller scope, and vice versa. If two m-scripts have identical script trees, the one with fewer and lower trump nodes, and more trump links, has a smaller scope set and so has greater information content. A 'top trump' on the root node has less information content than any other trump configuration, because it allows the greatest freedom the expand the structure.

Two m-scripts cannot necessarily m-unify, if their scope sets do not overlap. In particular, if plain unification (ignoring trump nodes) does not work for them, then m-unification will not work either.

If it exists, what does P = M Um N look like ? Just as for ordinary script unification, P has all the structure (nodes and slots) of both M and N; M and N can both be found within the structure of P. As for ordinary script unification, N and M are matched so as to maximise their overlap, making P as simple as possible. Any trump node of P must be on or below a trump node of M, and below a trump node of N. Finally, all the trump links of both M and N must appear in P. This means that P may have extra nodes and slots (not required in ordinary unification) just to make it satisfy all its own trump link relations.

The **m-intersection** of m-scripts M and N, written as M ∩m N, projects out all the information which M and N have in common and loses all other information. The m-intersection always exists. It is defined from m-inclusion:

Q = M ∩m N is the m-script with largest information content I(Q) which satisfies both M >m Q and N >m Q. (2.15)

Again, the result of an m-intersection looks much like the result of a script intersection. Q = M ∩m N only has those nodes and slots which appear in both M and N, but M and N are matched so as to maximise this overlap, making Q as big as possible. If either N or M has a trump node, then Q must have one in the same place; and Q can only inherit a trump link from M if N has one in the same place. However, crucially, m-intersection can `discover' trump nodes and trump links, creating them where they were not before:

- If, for some node of Q, the corresponding nodes of N and M differ in some slot or in number of descendant nodes, than that node must be a trump node in Q.
- If, for any two trump nodes in Q, the subtrees of N and M below those nodes both obey Q's trump link relation, then Q will have a trump link between those nodes.

This creation of trump links by m-intersection is a vital part of language learning.

Note the close analogy between these definitions and the definitions of the script operations. Because the m-script operations are defined set-theoretically, in the same way as the script operations, they obey algebraic relations precisely analogous to those of the script algebra; these form the **m-script algebra**.

***Chemical Analogy****: This goes over from script to m- script operations unchanged; m-unification is like chemical synthesis, and m-intersection is like analysis. Now we can bring the analogy closer to its use in language:*

- *Every word m-script is like an atom*
- *M-unification synthesises sentence molecules out of word atoms (used, as we shall see below, for both language production and understanding)*
- *M-intersection analyses sentence molecules down to learn word atoms. It analyses by comparing two or more molecules, to extract their biggest common submolecule.*
- *You can build up a molecule by m-unification in any order, and still get the same molecule*
- *You can analyse a molecule by m-intersection in any order, and still get the same constituents.*
- *If you build up a molecule, then analyse it, you get back the constituents you started from.*

While the m-script operations are defined set-theoretically in terms of their scopes, they can be calculated by algorithms operating on the m-script tree structures. These algorithms are more complex than those for script operations, but always start by doing the script operations. They are described in Appendix A.

Strictly, the m-script operations act only between m-scripts, not between scripts and m-scripts. However, by treating a script as a special m-script, we can m-unify scripts and m-scripts together.

## 2.3 Word M-scripts and Language Processing

In the theory of language, each word is defined by an m-script (call it W), which has two branches - a left branch and a right branch - hanging from a top script node. The left branch consists of one or more scene nodes (and all the structure below them) and the right branch is just the rightmost scene node with the structure below it. These two branches will be denoted by L(W) and R(W). There may be zero, one or more trump links, each one directed from a node in the left branch to a node in the right branch. An example with a trump link is shown in figure 6 below.

For a script in the scope of this m-script, if the left branch is known, then the right branch is fixed by the trump links; and vice versa. Therefore the m-script acts as a function from scripts to scripts. It is:

- a partial function, since not all scripts will unify with the left-hand branch to give a result
- a reversible function, which can be applied left-to-right or right-to-left.

- an unbounded function, which can deliver an unbounded set of results; there can be subtrees of arbitrary complexity below each end of a trump link.

Therefore m-scripts are functions from scripts to scripts, which can deliver, as results, the arbitrarily complex meanings of language. They are applied left-to-right in language understanding, and right-to-left for generation.

To apply an m-script M as a left-to-right function f on a script S, proceed as follows:

1. Check that S >s L(M); otherwise M cannot be applied to S.
2. Add a trump to the top node of script S, forming an m-script S(top) with scope σ(S(top)) = {T|T >S}. This means that structure can be added to S in m-unification.
3. Form N = M Um S(top). Then L(N) = S and R(N) =f(S). This m-unification adds a right branch R(N), which is the function applied to S.

To apply the same m-script M right-to-left to a script T, do the same, but in step (1) check that T >s R(M); then the m-unification P = M Um T(top) adds a left branch L(P) = f **-1 (T), calculating the inverse script function.

We can regard an m-script as a function with several arguments, X = f(A, B, ...). Here A, B, etc. are all scenes below a top script node of a script S, and the m-script is applied to S as before. This is how a verb m-script works with one argument each for the agent, patient, etc. When these m-scripts are applied right-to-left, they calculate all the `argument' scripts A, B, etc. from the one `result' script X.

I shall illustrate by an example, how (a) an m-script for each word embodies all the syntactic and semantic constraints on the use of that word, as well as its meaning, and (b) language understanding and generation are done by the repeated application of these word m-script functions, to build up or dissect meaning scripts.

First consider the simplest m-scripts in any language, the m-scripts for nouns and proper nouns. A typical one of these is shown in figure 5.

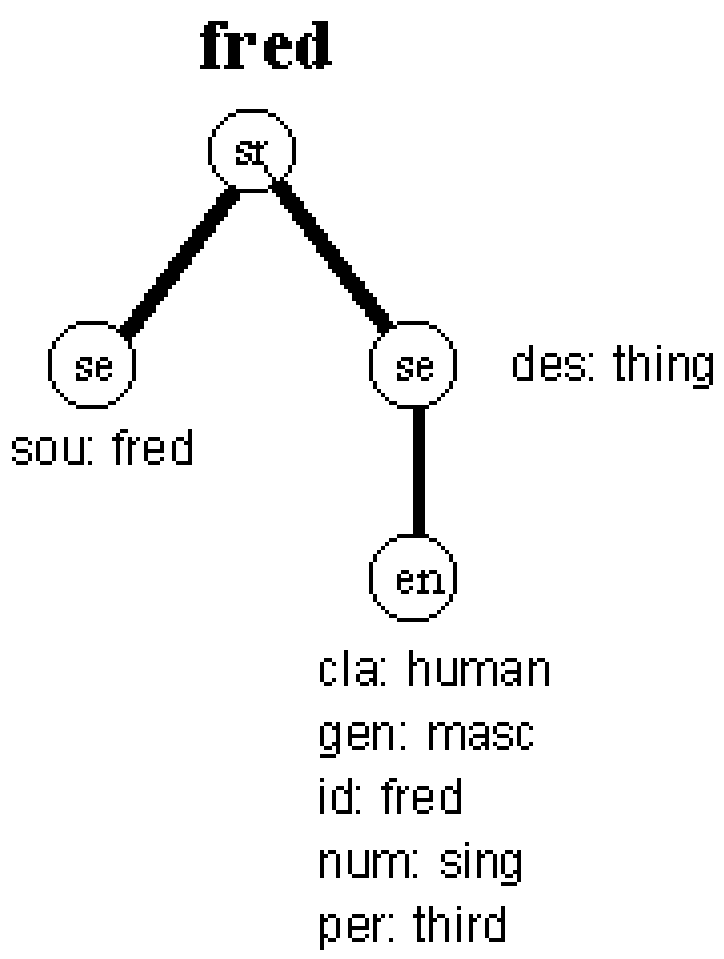

*Figure 5: M-script F defining the meaning of the name 'Fred'*

This has a left branch script and a right branch script, so it is a function from scripts to scripts - albeit a very trivial function. As it has no trump links, it is a bounded function, which can only deliver one result scene - the scene describing the individual Fred (its right-hand branch) As its argument (the left-hand branch) it has just a single scene containing the sound `Fred'.

This script can be used to convert from sounds to meanings in a simple, one-word, manner. When hearing the sound `Fred', applying the script function left-to-right delivers the meaning script F which denotes Fred: f(`Fred') = F. Alternatively, given the same meaning script F, the function can be applied right-to-left, to deliver the sound `Fred': f -1(F) = `Fred'.

Next consider the m-script function for the adjective `hungry'. This is shown in figure 6.

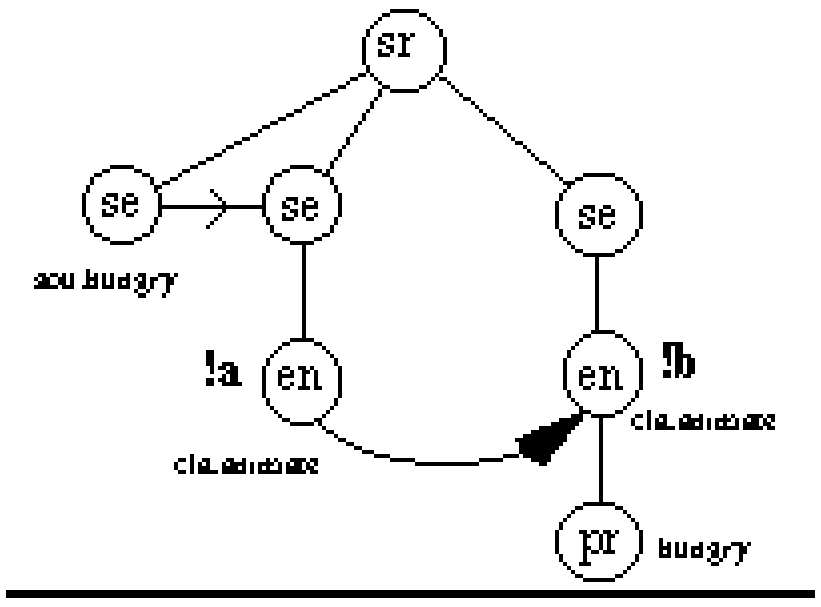

*Figure 6: M-script G for the sound "Hungry"*

This m-script, like all other word m-scripts, can be regarded as a function h from its left-hand branch to its right-hand branch. This time, however, as it has a trump link, it is a non-trivial function.

It is a partial function, and can only be applied to scripts which include its left branch. To do this, the argument (left-hand branch) must have a scene containing the sound *hungry*, followed by a scene containing an animate entity. There may be any amount of extra information below the animate entity node.

When the function is applied, the trump link requires a unification, between nodes a and b (see figure 6):

S[b] = S[a] Us M[b] (2.16)

The result S[b] combines what we knew before about the entity (in the argument script S[a]) with new information in M[b] (the property node for being hungry - see figure 6).

So the m-script for hungry can add the property node for hunger to a meaning script S (of arbitrary complexity) describing some animate entity.

We can now illustrate the whole process of understanding the phrase `hungry Fred', shown in figure 7.

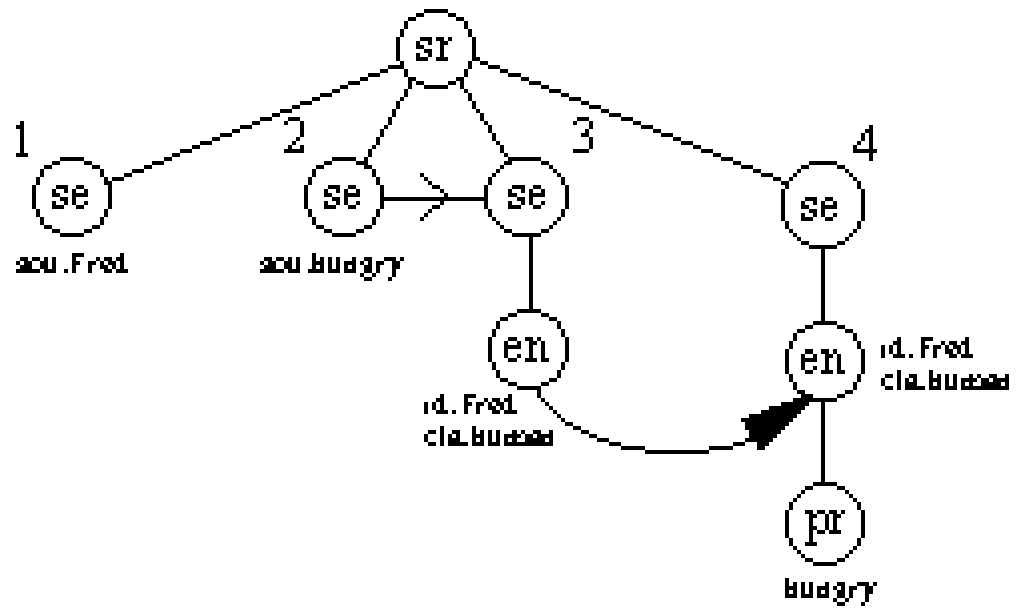

Z

*Figure 7: the script Z created in understanding "hungry Fred ". Z is a sentence-meaning structure (SMS).*

We start off having heard the sounds, with just two scenes, describing the sounds *hungry* and *Fred*. Understanding then proceeds by successive m-unifications (equivalently: successive function applications) to build up more complex meaning structures.

Denote this whole script by Z, and its four subtrees, below the four scene nodes, by Z[1] ... Z[4]. The full script structure Z is called a **Sentence-Meaning Structure** (SMS) since it contains both the word sounds of the sentence, and the meaning script.

Initially we have just the two heard sounds, Z[1] = *Fred* and Z[2] = *hungry*. Then we apply the m-script for the word `Fred', calculating Z' = Z Um F and so adding Z[3] = f(Z[1]) to the structure; and finally we apply the m-script for `hungry', calculating Z'' = Z' Um G and so adding Z[4] = h(Z[2], Z[3]). Both m-script applications simply add structure to Z, and both m-scripts can be seen in the final structure of Z (figure 7). Z[4] is the meaning script resulting from the understanding process.

For language generation, we start with the meaning script Z[4], and apply the same word m-scripts right-to-left. First we apply the m-script for hungry, calculating Z' = Z Um G and so adding Z[2] and Z[3]; then we apply the m-script for `Fred', calculating Z'' = Z' Um F and so adding Z[1]. This gives us scripts for two word sounds (Z[1] and Z[3]) which can then be said.

Note that **in both language understanding and generation, the same SMS (Z in figure 7) is built by m-unification**. In understanding, we start with just the sounds Z[1] and Z[3], and calculate the meaning Z[4]; in generation, we go the other way round, from the meaning Z[4] to the sounds Z[3] and Z[1]. The end result is the same in both cases.

***Chemical Analogy****, the SMS is a molecule that we synthesise by m-unification of word atoms. Each word atom can be seen within the structure of the molecule. Speaker and listener both build up the same molecule in their heads. You can synthesise the SMS molecule in various different orders (as the speaker and listener do) and still get the same molecule; communication is faithful.*

This example illustrates several features which remain true for understanding and generation of much more complex sentences in adult language:

- For understanding, we start with just a sequence of word sound scenes, and by m-unifying left-to-right with m-scripts for the words, build up ever more complex meaning scripts as new branches of the SMS, ending with the full meaning script.
- In language understanding, there is one m-unification left-to-right for every word in a sentence.
- For generation, we start with just a meaning script, and by m-unification right-to-left with word m-scripts, we break down the meaning into simpler parts and create word sounds which can be said.
- In generation, there is one m-unification right-to-left for each word in the sentence.
- The same SMS (molecule) - containing all the word sounds, the meaning script and the intermediate stages - is built up in generation as in understanding, but in approximately reverse order.
- Every word m-script (constituent atom) can be found embedded in the SMS. Denoting the SMS by Z, and a word m-script by W, for every word Z >m W.

It will not yet be obvious that this function application model can cope with the full complexities of adult language. To provide some evidence that it can, I shall do two things:

1. Show how the m-script formalism has close links with unification-based grammar formalisms such as LFG or HPSG, which can handle full adult languages
2. Describe how some of the complexities of adult language are handled, in a working program which uses the m-script method for both understanding and generation.

The m-script formalism can be compared to any unification-based grammar formalism, but for definiteness I shall use Lexical Functional Grammar (LFG) (Kaplan & Bresnan 1982). In terms of LFG, each word m-script embodies a phrase structure rule and a lexical entry, including all its functional equations. To see how this happens, consider the m-script for a more complex word, `gives', shown in figure 8.

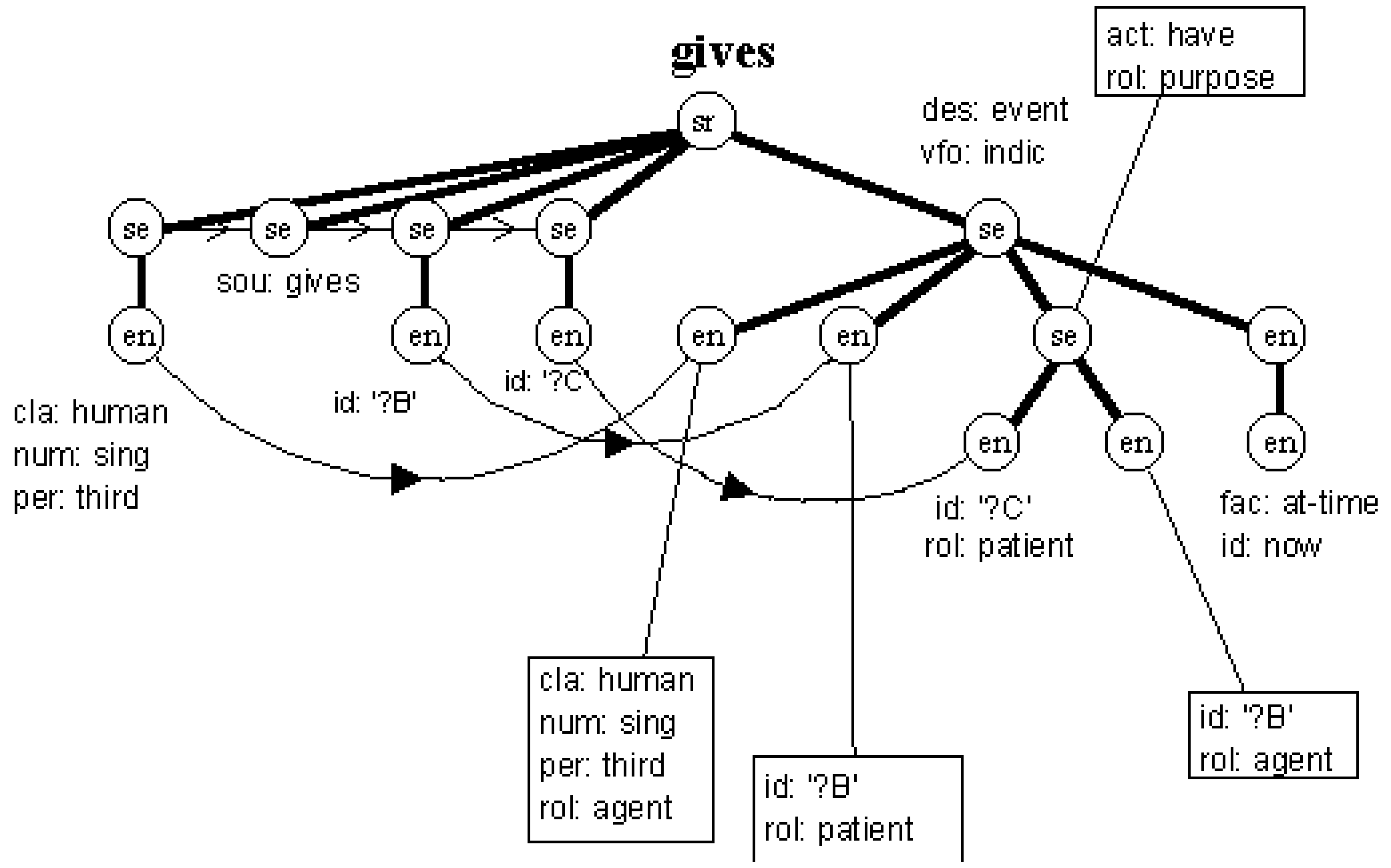

*Figure 8: M-script for the word "gives"*

Again, this m-script is a function from its left-hand side to its right-hand side - to be applied left-to-right for language understanding (to build meaning scripts) or the inverse direction for generation (to break them down). On its left-hand side, as well as the script for the sound `gives', it requires as arguments three meaning scripts - one each for the donor, the recipient and the gift.

The right hand branch expresses the meaning of `A gives B C' - roughly that A acts on B, with the result that B possesses C. The unifications required by the trump links mean that, whatever script structures appear in the arguments representing A, B and C, these same structures will appear in the full meaning script.

- The LFG *phrase structure rule* is embodied in the structure of the left-hand branch. The `immediate dominance' (ID) rule is in the script node with four scene nodes below it - specifying the verb and its three arguments. The `linear precedence' (LP) rule is embodied in the time-order arrows between these scene nodes.
- In the LFG lexical entry, the *uncommitted f-structure* (feature structure) is the meaning script for *gives* in the right-hand branch of the m-script.
- The *word sound* in the lexical entry is in the second scene of the left branch.
- The LFG *linking rules* for verb arguments are embodied in the trump links (curved arrows), which define that the first argument becomes the donor, the second the recipient and so on. Note that these linking rules work in terms of semantic roles (eg role slots on the entities) and argument positions, not the grammatical functions of subject and object.
- The *functional equations* of LFG appear in several guises. For verb-noun agreement, the fact that *gives* requires a singular agent shows in the `number' slot on the leftmost entity. Case-marking constraints on the nouns would be similarly reflected. Control equations for complement-taking verbs like *want* show as variable identity slots in the meaning, with the same variable used in two or more places (*gives* has a `hidden' control relation: that the patient is the one who later owns the gift. This is denoted by the shared variable ?B in figure 8).

This shows how the m-script formalism is roughly equivalent in power to LFG, and can be expected to handle the same range of language features. That supports the key claim of this paper, that if you can learn m-scripts, you can learn a language.

Each word m-script has one or more scenes in its left branch, and only one meaning scene in its right branch. These scenes can all be classified into three main types:

- `Word sound' scenes which just denote the sound of the word heard, or to be spoken.
- `Event' scenes which describe some event or state. These scenes have a slot [des:event].
- `Entity' scenes which describe some thing. These scenes have a slot [des:entity].

In terms of this classification, the description of the *gives* m-script above is:

[entity] *gives* [entity] [entity] <->[event]

We can discern some standard shapes of the different parts of speech in a language, as follows:

**Noun, pronoun**: *ball* <->[entity]

**Adjective**: *green* [entity]<-> [entity]

**Article**: *the* [entity]<-> [entity]

**Preposition**: [entity] *on* [entity]<-> [entity]

**Preposition**: [event] *to* [entity] <->[event]

**Adverb**: *boldly* [event]<-> [event]

**Simple transitive verb**: [entity] *hits* [entity] <->[event]

**Complement-taking verb**:

[entity] *tells* [entity] [event] <->[event]

**Auxiliary**: [entity] *can* [event]<-> [event]

These are summaries of the number and type of scene nodes in left and right branches of the word m-scripts. The orders of scenes in the left-hand branch is not significant, but has been given in English-like form.

***Chemical Analogy**: When combining word atoms to form SMS sentence molecules, an event scene on the left branch of one word must match with an event scene from the right branch of another; similarly for entity scenes. This means that the atoms and molecules have a kind of two-valued, directed valency which must be respected in synthesis.*

Like chemical valency, it is not always precisely satisfied. In particular, verb m-scripts are often m-unified (in language understanding) when some of their required entity scenes are not present, and these `gap' entities must be determined later.

You may wish to sketch for yourself the process whereby a sentence such as `Fred gives the boy a banana' is understood or generated (ignoring for a moment the articles). In understanding, all the nouns are processed first using simple m-scripts like that in figure 5; this assembles the three arguments required for the `gives' m-script above, which then m-unifies with them to build the full meaning structure. This looks like the right-hand branch of figure 8, with the three unknown entities filled in by unifications. Because the `gives' m-script has time order arrows on its left-hand side, the noun phrases for the three entities involved must appear in that time order in the sentence. For generation, the process goes in reverse, leading to the same SMS.

If, instead of `banana', we had `the lollipop he found in the park', the full meaning structure for this noun phrase would be assembled as the `gift' entity before applying the `gives' m-script. Being an unbounded script function, `gives' can pass over an arbitrarily complex entity script, as a description of the gift.

For a language with weaker word-order constraints, such as Latin, some or all of the time-order arrows on the left-hand side would be missing, allowing several word orders; but there might correspondingly be stronger agreement constraints (eg requiring that the donor be marked as an agent).

The m-script formalism differs from many unification-based grammar formalisms, such as LFG, in three respects:

- It has no separate phrase structure rules; each word m-script has its own phrase structure rule built into its m-script structure. This allows us to express highly irregular languages; but it also works well for nearly-regular languages, where many words have the same phrase structure constraints built into them.
- For agreement constraints, it relies on semantic roles such as `agent', `patient' and so on which appear in meaning scripts, rather than the grammatical roles of subject and object.
- It tends to use rather flat parse structures in which, for instance, a transitive verb sentence is analysed as S Agent Verb Patient rather than as S Agent VP, VP Verb Patient; and the extra levels of structure posited by X-bar syntax are not present.

Apart from these differences, which will be discussed later, it is equivalent in power to most unification-based grammars and can handle the same range of language constructs. How it does so is illustrated in the next subsection.

## 2.4 Processes of Understanding

The syntax and semantics of any part of speech can be expressed in m-scripts, as illustrated above. Sentences of any complexity can be understood by the same mechanism of repeated m-unifications, one for each word. Each m-unification adds to the SMS and builds a part of the meaning script by unifying elements of meaning together, and this process can build meaning scripts of arbitrary depth and complexity.

To handle typical sentences of adult language, m-unification is the core process, but is supplemented in various ways:

- Word m-scripts need not be m-unified in the order the words are heard; typically they are applied nouns-first to build up meaning structures from their simplest components.
- At any point in understanding a sentence, there are choices of which word m-scripts to m-unify, and of their arguments. A chart-parser-like mechanism keeps track of the options.
- It is often necessary to m-unify some word m-script (typically for a verb) when one or more of its argument scripts are missing - a `gap' to be filled in by later processing, as in questions or relative clauses. Filling these gaps may involve an element of pragmatic search through possible candidate entities.
- Finding the right referent for a pronoun or noun phrase may also involve some pragmatic search.
- In cases of structural or other ambiguity, two or more analyses of the same part of a sentence may exist. In this case, calculate the meaning scripts arising from all possible analyses, form their script intersection (this keeps the parts of meaning common to all of them) and use that for the rest of the understanding process. This avoids a combinatorial explosion of possible analyses, giving a partial meaning, which can be made up to the full meaning later by choosing one from each set of ambiguous senses.

- Coordinate constructs are handled like ambiguities; form the script intersection of the coordinated meanings, and use that for the remainder of the understanding process. Reintroduce the two (or more) coordinated meanings at some time later, for instance at the end of sentence processing.

These methods have all been tested in a program which can use a vocabulary of about 400 words to understand or generate a wide variety of English sentences, containing many of the complex constructs of adult language. Understanding time is roughly linear in sentence length, and there seem to be no hard limits on the vocabulary size or sentence complexity it could handle. More details of the program are given in Appendix D.

The synchronous processing of semantic and syntactic information, with no branching at ambiguities, has advantages of psycholinguistic plausibility. It enables us to form meaning structures in real time mid-way through hearing a sentence, in spite of structural ambiguities - as introspectively we know we do.

Since language is so often used in conditions of haste, high background noise and interference, the handling of ambiguity - without incurring a combinatorial explosion of analyses, and arriving reliably at the most likely interpretation - is a very important practical aspect of the theory. All types of ambiguity (structural, word sense, missing or unclear words) are handled by the same two-stage mechanism:

1. Take the script intersection of the two or more ambiguous senses, thus finding a subset of the meaning script which will be true in any full interpretation, and avoiding a combinatorial explosion of different senses.
2. At some later stage, choose between possible full meanings by Bayesian maximum likelihood inference. This can bring together all possible cues - syntactic, semantic and pragmatic - to make the choice.

A maximum likelihood choice of interpretations can be shown to give best average performance (Worden 1995, and Appendix B) and, as we will see below, has a very close link to the learning theory, which is also Bayesian.

***Chemical Analogy****: In cases of ambiguity, the listener may combine the same set of word atoms in two or more different ways to get different sentence molecules. Or for homonyms, he may have a choice of two word m-script atoms for the same word sound.*

In these cases, the listener may continually analyse the candidate molecules as they are forming, to find the shared common substructure of all candidates; and then later use other knowledge to find the preferred minimum energy configuration (the maximum likelihood interpretation)

# 2.5 The Process of Generation

Language generation is done by the same basic mechanism as language understanding (m-unification, to build up an SMS, but in approximately reverse order). In outline, the process is as follows:

Start from a meaning script, which is the meaning you intend to convey. Choose a word to m-unify with it, matching the meaning to the right branch of the word m-script. This adds new scenes corresponding to the left branch of the word m-script (including its word sound scene, and other meaning scenes). Elements of meaning are passed back into the new meaning scenes by the trump links. In turn, these new meaning scenes are consumed by m-unifying them with other word m-scripts.

In this way, a full SMS is built, consuming all the original meaning and creating word sound scenes for all the words to be said. The SMS contains the necessary time-order constraints between words, limiting the orders in which they can be said.

In one sense, generation is simpler than understanding, because one starts from a known meaning structure, and never has to cope with structural ambiguities or missing information. However, it poses two new problems:

1. In all likelihood, the initial meaning script cannot be precisely expressed using the words in one's vocabulary. So language generation is a process of successive approximation, at each stage finding words which say almost what one wants to say
2. One has to continually decide what can be left out, and what ambiguities the hearer can cope with.

The first problem is particularly acute for children with small vocabularies; how they handle it determines what they say, and so influences much of the data we have on child language learning. It is therefore important to understand the generation mechanism, and its impact on child language learning data.

The challenge facing the child is to say `the truth, the whole truth and nothing but the truth' (although `truth' may be intended meaning, rather than literal truth on some occasions!). We can characterise these three criteria in script terms:

**(A) The truth** : At each stage, the meaning script of the selected word should not contradict the intended meaning script - e.g. they should not have different values of the same slot.

**(B) The whole truth**: The meaning script should not omit any nodes or slots in the intended meaning, unless those nodes and slots are below a trump node, and so will be `passed across' on trump links to be expressed by other words later in the generation process.

**(C) Nothing but the truth**: The meaning script of the intended word should not add any nodes and slots which are not in the intended meaning.

With a finite vocabulary, these three criteria cannot be precisely satisfied at once. At any stage in generation, for any candidate word, we can make a measure of the mismatches between the word and the intended meaning of types (A), (B) and (C), in information-theoretic terms (i.e the number of bits of mismatch of each type), and choose the word which minimises a weighted sum of these mismatch scores.

Thus at each stage, the speaker chooses the best-matching word according to these criteria, and applies it by m-unification. There is a further criterion in each word choice; each step of m-unification strips out as much meaning as possible from the meaning passed across on trump links, and which remains to be said; but it is also possible to use words which re-introduce some of this stripped-out meaning to say some things redundantly.

Thus language generation is a process of successive approximation, with word choices to be made at each stage. I have described a simple `discrepancy minimisation' model of how these choices are made, which seems to work well in the computer implementation of the theory (it generates acceptable versions of all the sentences which the program can understand, and behaves gracefully when its vocabulary is inadequate; see Appendix D).

However, in reality the choice of words will be guided by many pragmatic factors, above all by those elements of meaning which the speaker most needs to convey accurately. For children with small vocabularies, these factors may have important effects on production-based measures of language learning.

Because words do not match the intended meaning exactly, language generation may modify the original meaning script - adding to it in places, and leaving out some of its meaning. Nevertheless, the SMS which is formed by generation contains all the m-scripts for the words spoken, and is the same as the SMS formed by the hearer.

## 2.6 Procedural Skills of Language

The main thrust of this paper is that every language consists of a set of word m-scripts; therefore, if we can somehow learn word m-scripts, we can learn a language.

For this to be strictly true, we would require two conditions to hold:

1. Word m-scripts embody all the syntax and semantics of a language
2. The procedures for using word m-scripts in language production and understanding are not complex, and are universal across languages; therefore they may be assumed to be innate.

I believe that (1) holds to a good approximation, and I have illustrated it by simple examples above. (2) is also a moderately good approximation, because:

- The basic procedure for both production and understanding is m-unification, which we may assume is innate.
- In practice there is only a small choice of possible words to m-unify at any stage of production or understanding, because very few words will match the current script structures.
- Most of the work of handling ambiguities in understanding is done by a simple method of script intersection, which we may assume is innate.

However, (2) is certainly not 100% accurate. There is also a `periphery' of procedural skills in language use, which we may not assume are innate or universal, and which may therefore need to be learnt. Some of these are:

- Finding referents for pronouns, anaphors, gaps and noun phrases
- Deciding what can be left out, referred to by a pronoun, etc.
- Finally resolving ambiguities

- Choice of what to say
- Choice of words to say it - whether to under-express one's meaning, or risk over-expressing it.
- Pragmatic discourse skills

So there is quite a diverse list of these procedural skills. While some of them may be subject to innate constraints, we may also be fairly sure that some or all of them are learnt, and improve with practice. The mechanisms for learning them are expected to be rather different from the mechanism (described in section 3 below) for learning the declarative structure of word m-scripts.

The theory does not yet address the learning of these procedural skills; in a few of the comparisons with language learning data (section 5), where additional assumptions need to be made, these assumptions are often about the procedural skills and their acquisition.

# 2.7 Segmenting the Sound Stream

It might seem, from the description so far, that this model of language requires some separate solution to the problem of word segmentation, so that word m-scripts with a single `sound scene' describing the sound of the word, like those in figures 5,6, and 8, may be used for language understanding and generation. However, this is not the case. Suppose that we have some way of segmenting the sound stream into smaller units (such as syllables, phonemes, diphones or whatever) which in turn can be put together to make words. Then a word m-script, in stead of having just one `sound scene' describing the sound of the word, may have a sequence of contiguous sound scenes describing its sequence of phonemes.

The models of generation and understanding will work just as well with this modified sequence-of-phonemes form of m-script; and crucially, the sequence-of-phonemes form solves the problem of sound segmentation for language understanding. The child need only observe a sequence of phonemes, and word m-unification automatically segments then into words. As we shall see below, the sequence-of-phonemes representation also solves the problem of sound segmentation in learning.

Therefore the sequence-of-phonemes form of m-scripts is probably the form used in the brain. However, for notational simplicity, I shall continue to use the `single word sound' notation for most word m-scripts.

# 2.8 Retrieval of Words from Memory

For both generation and understanding, it is necessary to retrieve rapidly and efficiently just a few relevant word m-scripts from the many thousands stored in the brain.

For language understanding, this is not a problem; some kind of associative retrieval, based on the sound of words, will do the job.

For production, this will not do, as the word sound is not initially known. However, if word m-scripts are stored in a subsumption graph (a tree of scripts, where the script stored at any node subsumes any scripts stored on its descendants - i.e. is included in them) then at any stage of production, the search for good candidate words can be done by starting at the root of the graph, and descending it to find successively better approximations to the intended meaning.

This graph search takes time of order log(N), where N is the size of the vocabulary, and so is fast even for large vocabularies. This enables us to choose words to express what we mean (minimising the mismatch from the intended meaning, as discussed above) in real time.

# 3. Language Learning

- *Bayesian learning gives optimal fitness, so natural selection leads to near-Bayesian learners*
- *A Bayesian mechanism can learn any script structure reliably from around six learning examples*
- *A simple extension can learn any word m-script from the same number of examples*
- *This enables children to learn the m-scripts for nouns first, then verbs, then any part of speech.*

I have sketched how the m-script model of language, being equivalent to unification-based grammars, can describe the production and understanding of typical adult language - including complex sentence meanings and diverse syntactic constructs in a wide range of languages. It does this in a way which depends principally on m-scripts and m-unification.

If we assume that the ability to m-unify is an innate capability of the human brain, then to learn a language is to learn the m-scripts of its words. If, therefore, we can find out how a child can learn word m-scripts by attending to adult language, we have a working theory of language learning to compare with the evidence. This section describes the proposed method for learning any word m-script.

## 3.1 Bayesian Learning gives Optimal Fitness

I propose a Bayesian probabilistic theory of language learning. The reasons for using Bayesian learning, rather than any other learning framework, are:

- The capacity for language learning arose by natural selection (whether specifically for language learning, or for some other form of learning which was then co-opted for language).
- Selection tends to lead to cognition with greatest fitness
- Lifetime fitness depends on average performance over all situations (rather than, say, worst-case performance)
- Bayesian learning maximises average performance
- Therefore we expect a language learning capacity to be near-Bayesian.

Appendix B gives a general proof that Bayesian learning, with prior probabilities which match actual probabilities in the environment, gives best possible fitness. Here I shall give an intuitive argument that it does so.

Consider an animal which learns rule scripts in order to predict outcomes of social situations and to choose social actions. For instance, a monkey may learn the rule script `If I bite X and X has higher rank than me, then X will bite me back'; he may then use this rule script to know when he should, or should not, bite other monkeys. Knowing the rule script helps him to maximise the benefits of biting, and minimise the penalties.

He can learn such a rule script by observing some number of occasions when the prediction of the rule is confirmed. Suppose L occasions must be observed in order to learn the rule.

How large should L be? (i.e., how many examples are needed to learn a rule?) Two things may go wrong:

1. The method may learn too slowly, requiring too large L. In this case, the learner fails to exploit the rule on occasions when he could have done so.
2. The learner uses too few examples (L too small) and so comes to believe many spurious rules on the basis of `evidence' which was just coincidental; so the learner makes bad decisions on the basis of false rules.

Each possible rule script R has a prior probability $P_A(R)$ of being true. (A is a subscript; see formula below) Since there are very many possible rule scripts, this prior probability is rather small for a typical rule. To learn a rule R, we apply Bayes' theorem to the two alternatives: R is true, or R is not true:

$$P(R|D) = \frac{P_A(R)P(D|R)}{P_A(R)P(D|R) + P(D|not\ R)} \qquad (3.1)$$

Here D stands for the available sense data, which consists of a sequence of scripts (describing actual events) which the learner observes [1]. Initially, when no events have been observed, P(D|R) = 1 and so P(R|D) = $P_A$(R) = small; the learner should not believe the rule R.

Suppose that rule R makes a striking (and otherwise unlikely) prediction. For every event in D which confirms this prediction, P(D| not R) is multiplied by a small number (the small probability of the prediction, if the rule were **not** true), whereas P(D|R) is not diminished (as the rule predicts the event). So after even a few such events confirming the rule, the `handicap' of small $P_A$(R) in equation (3.1) is overcome; P(R) P(D|R) becomes much larger than P(D| not R), so P(R|D) grows very close to 1. The learner should then believe the rule and act as if it were true.

The Bayesian learning criterion is to choose L (the number of positive examples required to learn the rule) just so that P(R|D) in (3.1) is close to 1 - above a threshold [2] (1-d). This minimises the aggregate penalty of (1) and (2) above:

1. Because each event in D which confirms the prediction diminishes P(D| not R) by a large factor, even if d is quite small it does not take many examples to exceed the threshold for P(R|D) in (3.1); so L is typically quite a small number (e.g. 3 -8 examples), and the `slow-learning' penalty (1) above is quite small.
2. By choosing the threshold quite close to 1, we make the chance of believing `spurious' rules (arising from coincidences in the sense data D) quite small, and so keep the `spurious learning' penalty (2) small.

The penalty (1) for slow learning rises linearly with L, and the penalty (2) for spurious learning falls off exponentially with L. As is proved in Appendix B, the Bayesian criterion chooses L to precisely balance these two penalties, minimising their sum and achieving best overall fitness. No other learning criterion does this.

For instance, if a rule script R has a very small prior probability $P_A$ (R), then by (3.1) it will take longer to learn this rule; more examples L are required to overcome the handicap of small $P_A$ (R). But because $P_A$ (R) is so small, this rule very rarely holds; so the fitness penalty of learning it slowly is on average a small penalty. Bayesian learning is tuned to learn important rules fast, other rules more slowly.

The Bayesian criterion means that most rules are learnable from 6 - 12 examples (which is very fast learning, compared for instance to many neural net models, which require thousands of examples). The performance of vervet monkeys in learning social regularities (Cheney & Seyfarth 1990) shows they can learn complex symbolic rules, like scripts, from a few training examples, as the Bayesian model would predict.

The optimality of Bayesian learning reflects itself not only in very fast learning - as fast as is possible without rampant errors - but also in very robust learning, which is highly immune to random errors and noise in the data. From the more detailed Bayesian treatment of script learning below, we shall see how this robustness emerges. It is a very important property for biological fitness, and for a child learning a language from noisy, error-prone data.

# 3.2 Testing for Rule Scripts

We define a model for Bayesian learning of rule scripts as follows:

We treat the time dimension as discrete; the learner observes a sequence of factual scripts F(t) describing events at times t = 1...N. The average amount of information in each event script F is $I_{avg}$ (e.g. for definiteness, take $I_{avg}$ = 30 bits).

A rule script R describes a causal relation between the events at times t and (t+1). R is a script with two scenes - a `cause' or `condition' scene C(R) and an `effect' scene E(R). If at time t, the conditions of the rule are satisfied, then with probability s(R) the effects of the rule will follow at time (t+1). This is computed as follows:

1. The conditions of the rule are satisfied if F(t) >s C(R).
2. If so, calculate G = F(t) Us R. G is the rule script `enriched' with the information in F(t), and like R has two branches C(G) and E(G).
3. With probability s(R) the rule `fires', in which case F(t+1) >s E(G).

The rule conditions are tested by script inclusion (step 1). For instance, the condition C(R) = `I bite ?X' is satisfied by a factual script F(t) = `I bite Caesar'. The consequences of the rule are calculated by unification (step 2). A variable identity in the rule such as `?X' may be fixed by this unification, and the same fixed value will appear in the effect E(G). This is how a rule such as `if I bite ?X, ?X will bite me' can predict `Caesar will bite me'.

If the rule fires, the factual script for the next time step, F(t+1), will reflect the consequences of the rule E(G), but may also have other information unrelated to the rule (step 3). The actual outcome may be `Caesar bites me and takes my banana' - which includes the rule effect E.

The rule probability s(R) (the probability that the rule fires, if its conditions are satisfied) is typically large. It is not necessarily 1; the cause does not always lead to the effect. s(R) should not be confused with the prior probability $P_A(R)$ that the rule holds at all. For most rules, this prior probability is very small.

Suppose we know what rule script R we are testing for. Given a sequence of factual scripts F(1) ...F(N), how do we test whether R holds ? We apply Bayes' theorem, equation (3.1), to the two possibilities `Rule R holds' and `Rule R does not hold'. Here D stands for the sequence of factual scripts F(t). To calculate the terms in (3.1):

**To calculate P(D|not R)** : If R does not hold, then the probability P(D|not R) is simply 2**(-N $I_{avg}$) - because the probability of some event script F(t), with information content $I_{avg}$, occurring at random is just 2**(- $I_{avg}$) , and it happens N times over.

**To calculate P(D|R)** : Group the factual scenes F(t) into pairs for successive time intervals, forming two-scene scripts H(t) =[F(t),F(t+1)]. If a rule R fires between time steps t and (t+1), then H(t) s R.

Suppose there are Nc occasions where F(t) s C(R) - i.e. where the conditions of the rule were satisfied; and that on Nce of these occasions, H(t) s R - i.e. the conditions were satisfied, and the rule fired. The best estimate of the rule probability [3] s(R) is then s = (Nce / Nc) .

For each of the Nce `successful' occasions, the contribution to P(D|R) is a factor s*2**(Ie - $I_{avg}$). For each of the (Nc - Nce) `failed' occasions, the contribution to P(D|R) is a factor (1-s)*2**(-$I_{avg}$). For each of the (N - Nc) occasions when the rule conditions were not satisfied, there is a factor 2**(- $I_{avg}$). These are just the probabilities of the observed outcomes on each occasion [4].

**To calculate $P_A$(R)** : We need a simple model of the prior probabilities $P_A$(R) which somehow penalises more complex rule scripts (because there are so many more of them) and ensures that the expected number of true rules is finite. To do this, we can use the information content I(R) of the rule scripts, choosing

$$P_A(R) = C2^{-\lambda I(R)} \qquad (3.2)$$

Using l 2-3 penalises complex rule scripts sufficiently to ensure convergence; with finite C, the sum of P(R) for all rules (the expected average number of rules which hold) is finite.

The total information content of a rule script can be written as a sum of the information contents of its cause and effect scenes; I(R) = I(C) + I(E). Typically these two parts are roughly equal.

We can then multiply these factors together, calculating both terms $P_A$(R) P(D|R) and P(D|not R) of the Bayes formula, and compare them. If $P_A$(R)P(D|R) is the larger of the two, then there is sufficient evidence to believe the rule holds; otherwise, any apparent evidence is more likely to be a fluke.

This comparison can be envisaged in a simple graphical form, which I call a *crossover diagram*, shown in figure 9. This plots the log (to base 2) of the two terms in the denominator of (3.1), calculated as described above, against the number of occasions Nc where the conditions of the rule are satisfied. By inspection of this diagram, comparing the slopes of the lines, you can see which term dominates the denominator, and therefore which hypothesis (Rule or no Rule) is to be believed, as a function of Nc. As soon as the `Rule' line is well above the `No Rule' line, the rule should be learnt.

Note that the two lines in this diagram *both* apply to the case where the rule actually holds; but each line applies to a different hypothesis, of `Rule' and `No Rule', entertained by the learner.

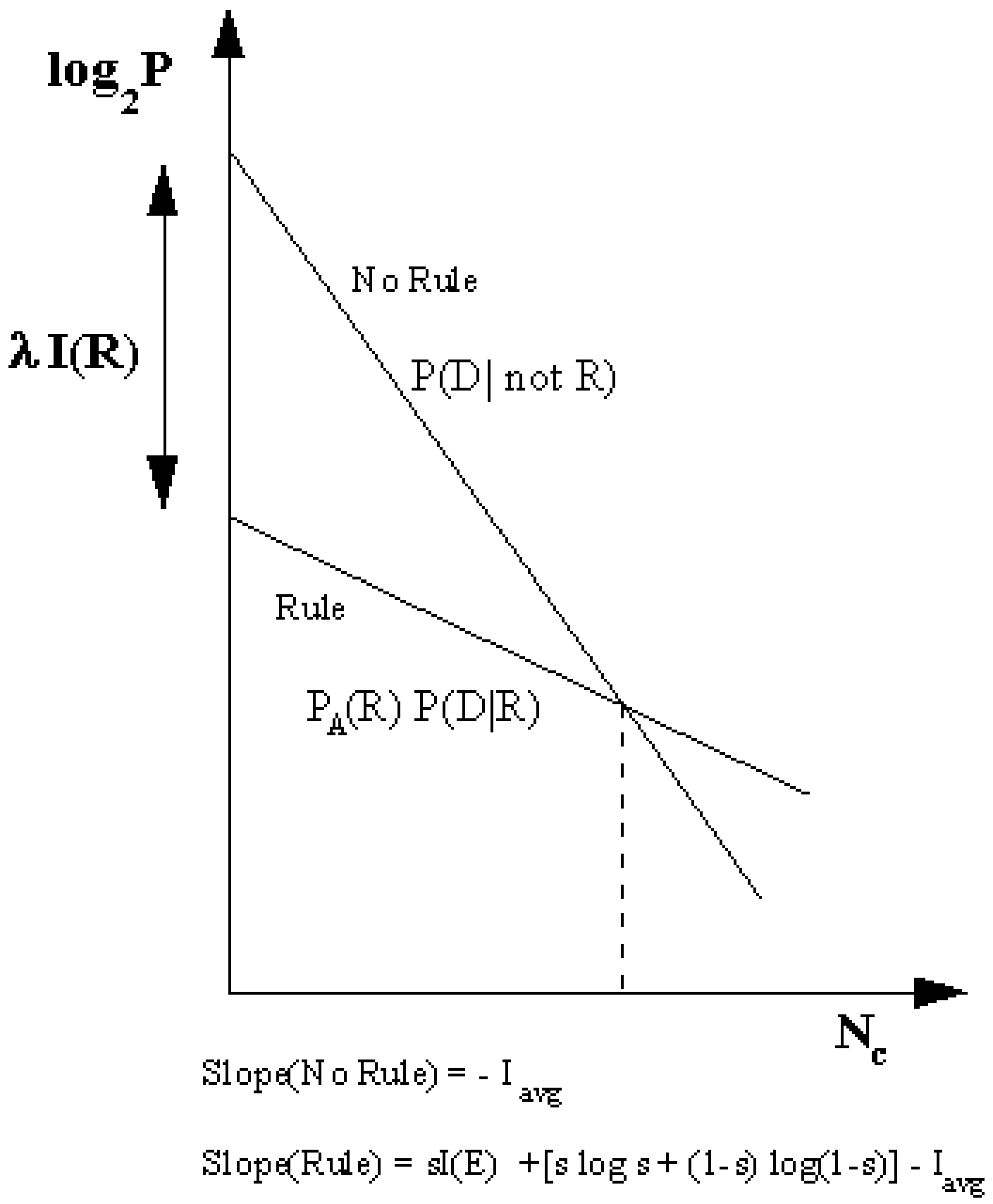

*Figure 9: Crossover diagram for two hypotheses, `Rule' and `No rule' to calculate how much evidence must accumulate before the rule is believed.*

At first, the `rule' line has the handicap of the small prior probability of the rule, which produces the initial gap between the two lines. As more cases Nc accumulate (where the rule conditions are satisfied), so evidence for the rule accumulates, until the two lines cross over; after that, the rule should be believed, and therefore learnt.

The `no rule' line has a slope - $I_{avg}$ because every observed script has a probability 2**[- $I_{avg}$]. The `rule' line is a weighted average of two cases - where the rule fires (i.e. its effect ensues) or does not fire. These two cases contribute respectively s[log s + I(C) - $I_{avg}$] and (1-s)[log(1-s) - $I_{avg}$] , so the difference of the slopes of the `rule' and `non-rule' lines is sI(C) + [s log s + (1-s) log(1-s)]. Of these terms, if I(C) is moderately large (10 bits or more) s I(C) will dominate, except for very small s.

The crossover between the two lines (the point beyond which the rule can be believed) will occur when

$$N_c = \frac{\lambda\,[I(C) + I(E)]}{sI(E)} \qquad (3.3)$$

or equivalently where

$$L = N_{ce} = \frac{\lambda\,[I(C) + I(E)]}{I(E)} \qquad (3.4)$$

Nce is the number of positive examples, where the conditions of the rule are satisfied and the rule fires. (3.3) shows the important result that *the number of positive examples required to learn a rule is independent of the rule probability s(R)* - therefore it is independent of the number of negative examples, where the rule does not fire. Highly `unreliable' rules, with small s(R), can be learnt.

So in a typical case where I(C) + I(E) 2I(E) , lambda =3, around six positive examples are required to learn any rule - regardless of how simple or complex it is, and regardless of the number of negative examples in between. For any rule, six positive examples are enough to provide definite evidence for the rule - overcoming its small prior probability [5].

`Six positive examples' is a useful rule of thumb for remembering the speed of Bayesian learning. However, it depends crucially on l, the coefficient in the simple model (3.2) of prior probabilities of rules. In practice, we expect the form of the prior probabilities to be more complex than that, so `six' is just a ball-park; the number may be anywhere in the range 2 - 10.

The `6 positive examples' rule of thumb is approximately independent of s; but it does not work down to very small s. In particular, for the case s = 2**[-Ie], the precise form of the crossover formula shows that the rule can never be learned. This is because, at this value of s, the probability of the effect occurring is identical, whether or not the rule holds; so the rule is vacuous.

***Chemical Analogy**: The vertical axis of the crossover diagram is -(Chemical energy). You learn a new rule/atom when the energy cost of introducing it is more than offset by its energy benefits across the molecules you have observed - giving a lower energy (= more likely) interpretation of the data as a whole.*

This completes the Bayesian criterion of sufficient evidence for a rule; but it does not yet tell us how to discover likely candidate rules. There is a simple algorithm to do this.

# 3.3 Discovering New Rule Scripts

If , for a rule R, on several occasions t its conditions C(R) are satisfied and the effect of the rule ensues, then, as described above, H(t) s R. To discover a rule script from a sequence F(1)...F(N) of factual event scripts, construct all the event pair scripts H(t) = [F(t),F(t+1)] and form all their pairwise intersections, $M_{tu}$ = H(t) ∩s H(u) = [F(t) s F(u), F(t+1) s F(u+1)].

Usually, when you intersect together two random event scripts F(t) and F(u), most of the information in the two does not coincide, so the result F(t) ∩s F(u) has a small information content, in the region of $I_{avg}$ /4. Whatever it is, this `background' level $I_{back}$ can be determined empirically, and random fluctuations above it will have exponentially decaying probability P(I) 2**( $I_{back}$ - I). Random fluctuations more than around 6 bits above $I_{back}$ will be rare.

If, for a pair of times t and u, the rule R was in action at both times, then H(t) ∩s R and H(u) ∩s R. From this it follows (via script algebra identities) that their intersection $M_{tu}$ also obeys $M_{tu}$ ∩s R. Therefore M must have at least the information content I(R), as well as an information content (in both scenes) of $[I_{avg} - I(C)]/4$ from random coincidences.

So if we look at the information content of all the pairwise intersections, there will be a clear gap of 3/4 I(R) between the haves and the have-nots - those which have the same underlying rule script R, and those which do not. The probability of a `have-not' looking like a `have' by chance - of it having an information content I(R) larger than it should - is of order 2**[-3/4 I(R)], which for information-rich rule scripts (say I(R) > 20 bits) can be safely ignored.

If $M_{tu}$ has large information content, and $M_{uv}$ also has large information content, this implies that scripts H(t), H(u) and H(v) all reflect the action of the same underlying rule R. In this way we can easily discover groups of scripts in which the same rule R was in action, and form the intersection X = H(t) ∩s H(u) ∩s H(v) ∩s H(w). By the script algebra, this must still satisfy X ∩s R, and if we are following the `six positive examples' heuristic above, it is likely that X will be a very good approximation to R; the chances of some superfluous piece of information (eg the value of some slot) appearing identically on the 6 scripts H(t), H(u),... are very small indeed.

Therefore the procedure of taking pairwise intersections $M_{tu}$ and then grouping them together is a reliable way of finding the rule scripts which are operating.

While pure script intersection is very robust against extra information in any of the scripts, it is not so robust against missing information in any of them; if one script happens to have a node or slot missing, it will be missing in the resulting rule script. So in practice we should move a little towards `majority voting' for features on the rule script - requiring, for instance, that at least two scripts in the set lack a feature before we remove it from the result. This makes a trade-off between the two kinds of robustness - robustness against extra information, and robustness against missing information.

It is then simple find how many F(t) obey F(t) >s C(R), that is, to find how many times the condition of the rule was satisfied (but its effects did not necessarily follow). In this way we can estimate the rule probability s(R).

There may be cases where two rules R1 and R2 have the same `condition' part C(R), but distinct `effect' parts E(R1) and E(R2). Such a pair of rules in effect says `If C holds, then either E(R1) or E(R2) will follow, with probabilities ...'. In language learning, this is relevant to homonyms, where the same word (cause part) has different meanings (effect part).

If there is such a pair of rules, then some of the intersection pairs $M_{tu}$ will have large information content $I(C) + [I_{avg}-I(C)]/4$ in their condition part, but random information content $I_{avg}/4$ in their effect part - because, for instance, R1 was acting at time t but R2 was acting at time u. This is easily recognisable in the pairs $M_{tu}$ ; in this case, use only $M_{tu}$ with large information content in both cause and effect parts, to define the groups of scripts to intersect together. Then both rules will be discovered.

At first sight, this algorithm seems to require time of order $N^2$ to form all the pairwise intersections $M_{tu}$ ; therefore, the amount of processing to be done at each new time step (generalising H(t) with all previous H(t')) grows linearly with time, which would be unacceptable. However, this problem is easily solved. If all the previous H(t) are stored in some kind of associative memory, then it is only necessary to retrieve the few most likely matches, rather than the whole set. In this respect, a subsumption graph of H(t) would be suitable. For learning word scripts, associative retrieval based on the sound of the word is also possible.

So the algorithm can learn rule scripts from experience with constant computational cost per unit time, requiring about six positive examples per rule.

***Chemical Analogy***: *When the same structure is seen in six or more supposedly random molecules, that structure is likely to be an elementary constituent.*

# 3.4 Speed and Robustness of Learning

This learning algorithm is robust in several ways:

- Script intersection combines several scripts to find the common information they all share. Any of the input scripts can have large amounts of extra information on them (i.e. extra nodes and slots), and this extra `noise' will be reliably removed by the intersection operation.
- The learning criterion can be tuned to give joint robustness against both extra information and missing information in some scripts.
- The algorithm can learn rules with small rule probability s(R), down to, say, s(R) =0.05. Therefore even if true instances of a rule are outnumbered by spurious counter-examples - where the conditions of the rule appear to hold, but the effects do not occur because of noise or poor observation - these just reduce the apparent probability s(R); it can still learn the rule, given around six positive examples.
- Since the rule discovery method starts from examples which each contain more information than the rule, and then narrows them down by script intersection, it has no innate tendency to over-generalise (i.e. to derive a rule with less information content than the true rule).

For language learning, this means that the child only has to hear a word correctly and infer its correct referent from non-linguistic evidence on a minority of occasions; the word's meaning script can still be inferred reliably.

The major issue not yet tackled is how the algorithm behaves when there are two similar rules - in particular, a general rule and an exception rule. This important case is addressed in the next subsection.

The Bayesian derivation makes it clear that six positive examples is about the minimum needed for reliable learning. Anything using less would tend to be unreliable, for at least two reasons:

1. An apparent rule might just be random noise in the data.
2. With only one example, there is no way to distinguish between the true, repeatable core of a rule, and random things which also happen to be occurring on one occasion. Script intersection projects out the common core from two or more examples; but it needs more than two, to reject all random noise reliably.

So one-shot learning would require very special constraints to make it robust and reliable.

## 3.5 Top-Down and Bottom-Up Learning

There is an important question concerning the order of learning: suppose there is a general rule R, and a number of special-case rules R1 , R2 etc. They are special cases, in that their conditions are stronger than the conditions of R; that is, C(Ri) >s C(R) for i = 1,2. The subcase rules are *exceptions* to the general rule; either their effect parts may be different E(Ri) not = E(R), or they may give rise to the same effects with different probabilities s(Ri) not = s(R).

The question is: which is learnt first, the general rule or the special cases ? If the general rule is learnt first, we call this *top-down learning*; but if the special cases are learnt first, we shall call it *bottom-up learning*.

The Bayesian learning theory gives an answer to this question, but it is not the same answer in all cases; it depends on the form of the rules.

The question of top-down versus bottom-up learning is a very important one for language learning, related to the issues of learning general rules of syntax, over-generalisation, and so on. It turns out that for language learning, the theory predicts that bottom-up learning predominates; but there is also some top-down learning in a few cases.

We first sketch the mechanisms for top-down and bottom-up learning, in the Bayesian framework:

- **Top-Down Learning**: The general rule R is learnt first, by the method described above. Then, as evidence accumulates for each special rule, it too is learnt - this time, against a `null hypothesis' of the general rule (rather then no rule, as in the previous analysis). This change in null hypothesis may alter the number of learning examples needed to learn an exception rule, but the number is in principle calculable from the theory.
- **Bottom-up Learning** : Each special-case rule Ri is learnt as evidence for it accumulates, against a null hypothesis of `no rule'. This takes approximately six positive examples per rule. Then, when several similar special-case rules have been learnt, the general rule is heuristically induced by script intersection of those rule scripts: R = R1 ∩s R2...Rn. If a non-trivial rule can be made in this way, it is then tested to see whether it gives a better overall account of the data - i.e. , larger P(D|Rules) - than just the special case rules on their own. If so, it can be learnt. (This is the secondary learning algorithm of section 3.9)

Why should one of these happen, rather than the other ?

Since the conditions of the general rule are less restrictive than those of the special-case rules, its conditions are satisfied more frequently than they are (every time some C(Ri) is satisfied, then automatically C(R) is satisfied). So the learning examples accumulate faster for the general rule than for any special case. On these grounds, we expect the general rule to be learnt first, and the top-down mechanism to dominate.

However, it may not be possible to recognise instances of the general rule in the data. Recall that the learning algorithm takes pairwise intersections $M_{tu}$ of learning examples H(t) and H(u) - and that when the rule R is operating on both occasions t and u, the intersection $M_{tu}$ is readily distinguishable from a `background' $M_{tu}$ , by its information content. But this only works if the rule R has large information content (say, greater than 10 bits) in both its cause and effect parts.

Rules with small information content in their cause or effect parts cannot be recognised directly in the data. The only way to find such rules is by the bottom-up mechanism described above. In this mechanism, each special rule Ri has already been `cleaned up' by script intersection when it was learned; do it does not have a lot of random noise in it. The small information content of the general rule R can be found in the intersection of the Ri .

We conclude that **the top-down mechanism dominates for rules with large information content; but rules with small information content can only be found by the bottom-up mechanism**.

For language learning, the m-scripts for open-class morphemes have large information content on both sides (the `sound' side and the `meaning' side), so they can be learnt direct from the data. M-scripts for broad syntactic rules, or for closed-class morphemes, all have small information content. So they can only be learnt by the bottom-up, secondary mechanism, after several narrower rules have been learnt (as described in section 3.9). So **the bottom-up mechanism dominates for language**.

Nevertheless, there are occasions where top-down learning of exception rules is used in language learning, to correct over-regularisation of irregular forms. So I shall briefly describe how that works in the Bayesian theory.

## 3.6 Learning Exceptions from Negative Evidence

Consider a general rule R and an exception $R_e$, where $C(R_e)$ >s C(R).

If the `effect' parts of the two rules E(R1) and E(R2) are easily distinguishable, then the examples of each rule are easily distinguishable; each rule can be learnt individually as described above, using of the order of six positive examples.

However, consider the case where the `effect' parts of R and Re are identical, $E(R_e) = E(R)$, but this effect occurs with different probabilities in the two cases. The probability of firing is s for R ,and t for $R_e$. For instance, s might be near 1, and t near zero; this means that in the more general case, E is very likely to happen, but in the special case, it will probably not happen. Suppose that R is learnt first (top-down learning).

If s = t, the exception rule predicts exactly the same as the general rule; in effect, there is no exception to learn. When s t, the exception rule makes a different prediction from the general rule, but this could not be detected from any one example.

To learn the exception rule, when t < s, one needs to observe several examples where the conditions of the exception rule $R_e$ are satisfied (and so the conditions of R are also satisfied), and to realise that the effect sometimes does not occur when predicted by the general rule R. In other words, one has to learn from implicit negative evidence.

The Bayesian analysis tells us how many examples must be observed to do this. As before, we can calculate the probabilities P(D|S) for the two different hypotheses `R only' (no exception) and `R and $R_e$ ' (exception). These are input to Bayes' theorem (3.1) and the crossover analysis done as before. Again, the two hypotheses give straight lines on the crossover graph.

Whereas for the simple learning case (section 3.2) the difference between the slope of the `rule' and `no rule' lines was typically large (or order 10 or more), for the `rule and exception' case the difference turns out to be small - typically 1 or less. In other words, evidence in favour of the exception rule can only accumulate slowly.

Where the two lines cross (the number of examples needed to learn the exception) depends on their zero intercepts - which depend on the Bayesian prior probabilities built into the learning mechanism. With the simple model (3.2) of independent priors for each rule, this difference is lambda*I($R_e$). For a typical rule with I($R_e$)= 20 bits and lambda=3, lambda*I(Re) =60. With a slope difference less than 1 in the crossover diagram, this would require more than 60 examples to learn the exception.

However, rules often have exceptions. That means that the simple model of prior probabilities in (3.2) is not realistic for exception rules. A more realistic model might be P(R and $R_e$ | R) = 2**lambda [I(Re)-I(R)]; this allows frequent exception rules, and by penalising the exception less in Bayes' theorem, allows faster learning - in, say, 5 - 10 examples.

Monkeys can learn exception rules from implicit negative evidence. Cheney and Seyfarth (1990) have shown how monkeys habituate to a particular peer's alarm call (but not to all peers' alarm calls) if that peer's call proves, on a number of occasions, to be false alarm. The general rule might be `If any individual ?X gives an alarm call, there is an eagle about with probability 0.8' and vervets can learn an exception `If Brutus gives an alarm call, there is an eagle about with probability 0.1'. The monkey has to learn this by observing occasions where Brutus gave a call and the eagle (expected by the general rule) failed to materialise; nobody explicitly gives him the information `that was a false alarm'.

Cheney and Seyfarth's results show that vervet monkeys can learn just these exception rules from implicit negative evidence, and can learn them from a fairly small number of examples - as the model above predicts [6]. The same process for learning an exception from implicit negative evidence is available for language learners.

# 3.7 Learning Noun Meanings

Since the m-scripts for nouns have no trump links, they are identical in form to simple scripts; so they can be learnt by the script learning mechanism which we have described. This gives a computational theory of how children learn the meanings of nouns.

Figure 6 illustrates a typical noun script which is to be learned. To apply the script learning theory to this problem, consider the left branch of a noun m-script (the branch containing just the sound of the word) as the `cause' part of a rule script; and treat the right branch (which contains the relevant meaning script) as the `effect' part.

In this model, the child hears people speaking, in circumstances where she can see what is happening and construct one or more representations of it, in script form. On each such occasion (labelled by t) she forms `sound/meaning scripts' H(t) whose left branch represents just some sequence of contiguous phonemes (sequences rather longer than words give efficient learning) and whose right branch is some piece of script meaning inferred from the scene - specifically, the script description of some entity involved. There may be several H(t) formed for one sentence/occasion. Then the H(t) are input to the learning algorithm as described above. A typical H(t) is shown in Figure 11.

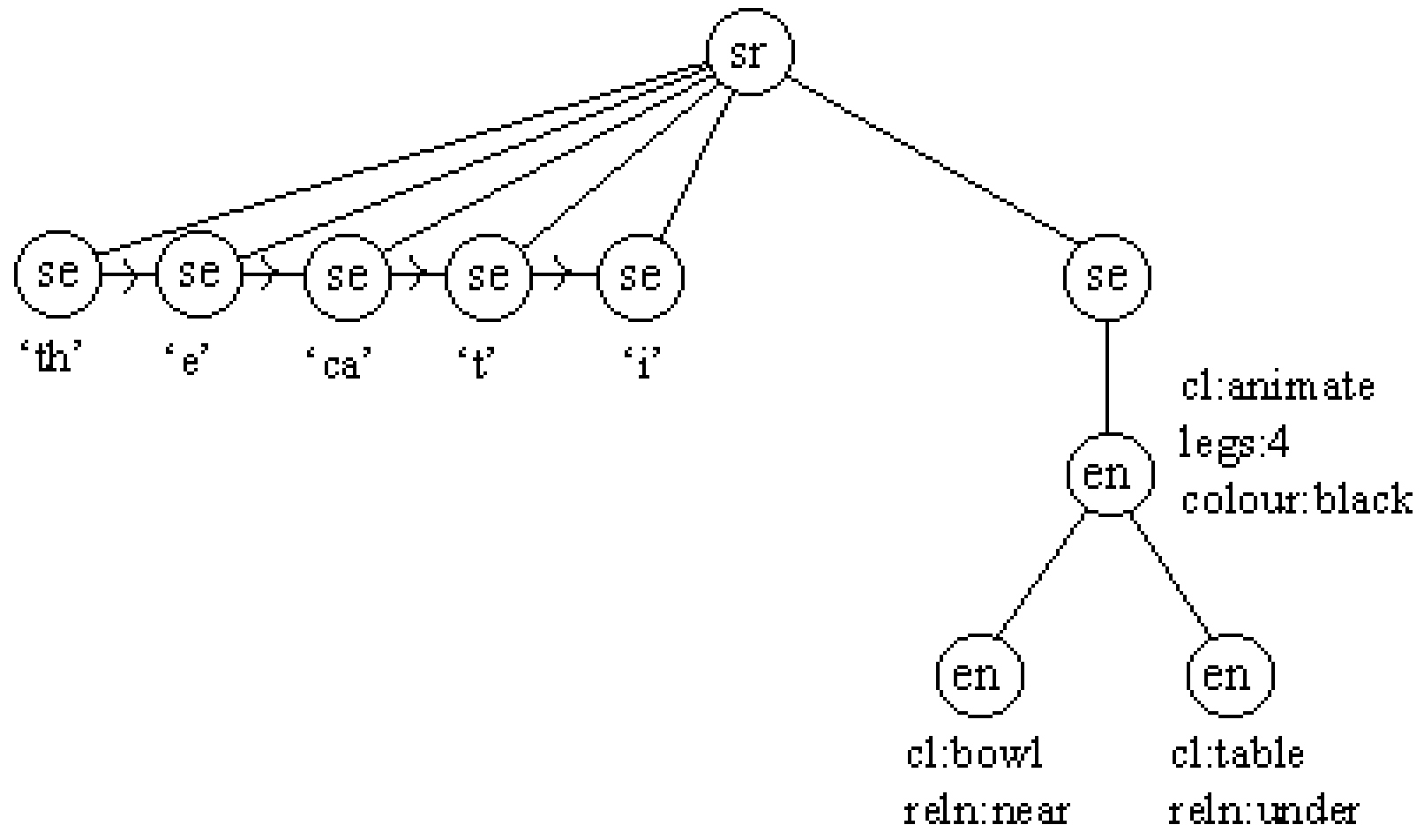

*Figure 11: Script H(t) formed by a child observing a parent say `The cat is in disgrace', and input to the noun learning algorithm.*

Pairwise script intersections $M_{tu}$ are formed, keeping only those which have significant information over the random noise level $I_{back}$ in both the left branch (i.e. which have the same sequence of phonemes occurred on both occasions) and the right branch (some significant overlap in meaning).

On the occasion shown in figure 11, the child hears her parent say the phonemes `th', `e', `ca', `t' and so on, in that sequence, as represented by the left-hand branch of the script (the precise way in which the child divides the sound stream into sections such as phonemes is not important, as long as it is consistent). She also observes a black quadruped, which happens to be under the table and near a bowl - as represented by the right-hand branch of figure 11.

On taking pairwise intersections $M_{tu}$, the only other H(t) which give above-noise information content on both sides of $M_{tu}$ will be those in which the phonemes constituting `cat' are present in the left branch, and the black quadruped is present in the right branch. For instance, the sentence `Look at my shoes' will give high information on the left branch (mimicking `cat') but no high information content on the right branch.

Figure 12 shows the H(u) resulting from a second occasion; and Figure 13 shows the script $M_{tu}$ which is the script intersection of figures 11 and 12.

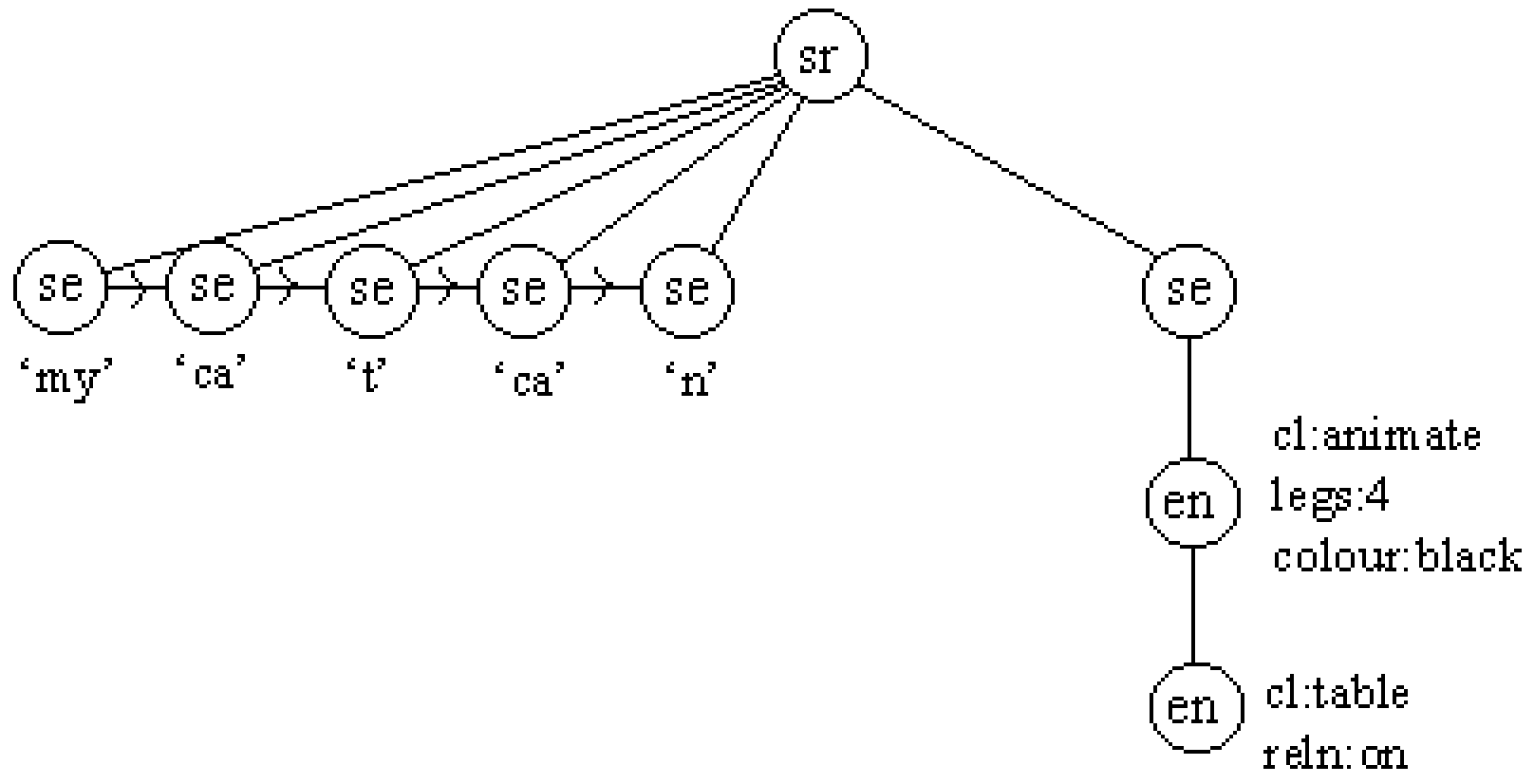

*Figure 12: Script H(u) formed by the child hearing someone say `My cat can catch mice' and observing the nearby scene.*

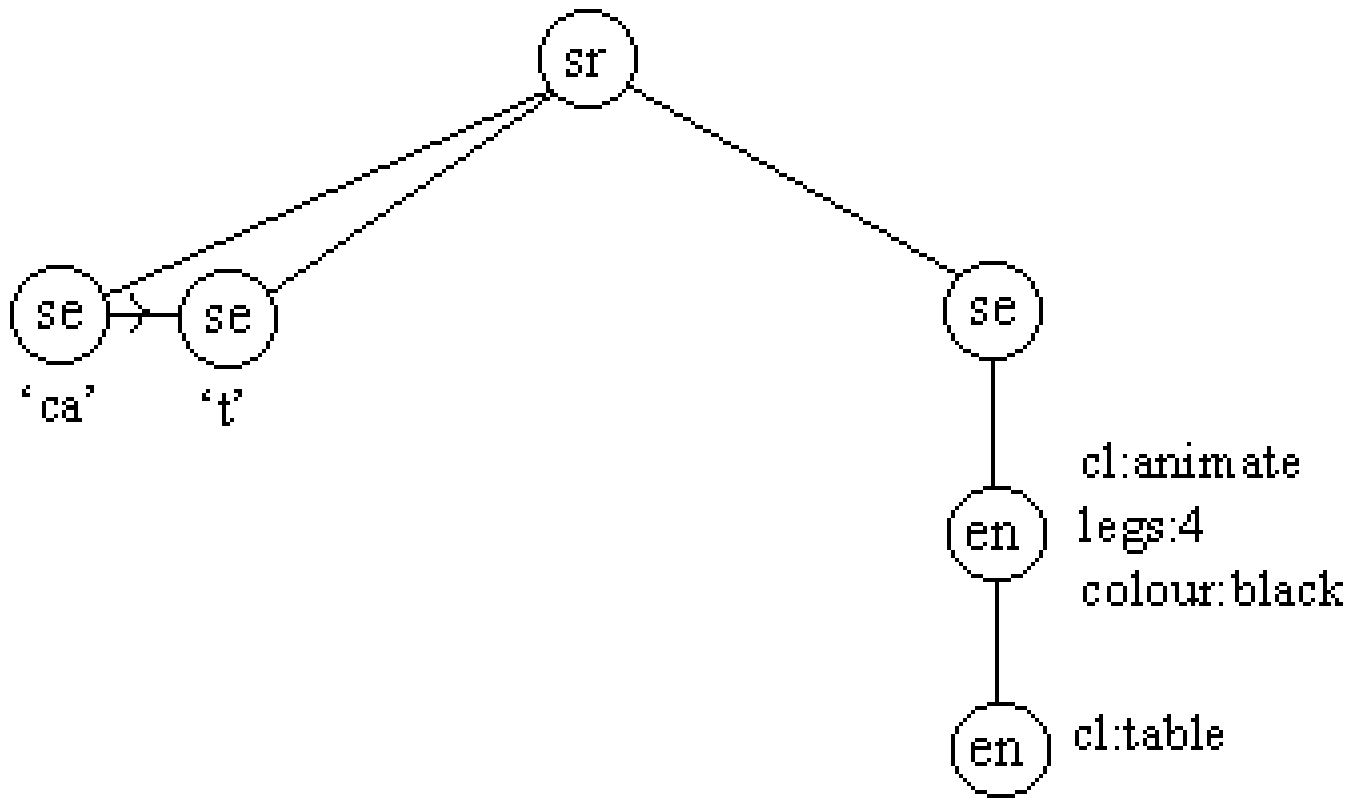

*Figure 13: Script $M_{tu}$ formed by intersection of the two scripts from figures 11 and 12.*

Note that this $M_{tu}$ has high information content in both branches, but still contains some information not germane to the concept `cat' - information about a colour `black' and some relationship to a table.

Those $M_{tu}$ with high information content on both branches define groups of occasions t, u, v, ... to be intersected together. Combining about six examples in this way will give a clean form of the meaning of the noun `cat'. In the left branch, the result will only contain the sound `cat' because all other phonemes (before or after) are lost in the intersection; and in the right branch, any meaning not central to the concept `cat' (such as the information about colours and tables in figure 13) will also be rejected by the intersection process.

Script intersection prunes a number of noisy examples, to associate the core sound with the core meaning. As long as both the left branch and the right branch of the underlying word m-script have adequate information content (say, 10 bits or more), the useful pairings $M_{tu}$ (where both H(T) and H(u) contain both the core sound and the core meaning) will be clearly separated from other pairings (where either the sound or the meaning is missing) by the information-theoretic criterion, that the information content of $M_{tu}$ be well above background noise in both branches.

This procedure can reliably learn noun meaning scripts of any complexity; it is not restricted to simple meaning structures like that in the right branch of figure 13.

Homonyms will give some pairs $M_{tu}$ whose left branch has significant structure (the sound of the homonym) but whose right branch is noise (because two different senses were intersected together). However, as described above, only those $M_{tu}$ where both branches have high information content are used to define the groupings for intersection; in this way the different senses of a homonym can all be learnt independently.

For a homonym such a `bank' or `palm', each distinct sense is learnt with a rule probability s(R) which depends linearly on its frequency of occurrence. A rare word sense will have s(R) near zero, and a common word sense will have s(R) near 1. When choosing between senses of a homonym, we use Bayesian inference to find the maximum likelihood interpretation; the rule probability of each word sense enters into this likelihood calculation, leading us to favour common word senses. (this model is a little over-simplified; we should also allow for context-dependence of the probability s(R), distinguishing s(bank|finance) from s(bank|fishing) )

This form of learning is memory-intensive, requiring the child to store hundreds or even thousands of scripts H(t). However, I do not regard this as a drawback. As adults, we have the capacity to remember tens of thousands of words, and a similar number of places we have visited. Even rats can neurally encode memories of hundreds of distinct places (O'Keefe 1987). Why should not a similar memory capacity be used in childhood for such an important purpose ?

As discussed previously, the H(t) should be stored in some kind of associative memory (or a subsumption graph memory) so that not all the pairings H(t) s H(u) need be literally formed; only promising candidates (eg those where several phonemes in sequence match) need be made, to keep the computational costs acceptable.

In this way the script intersection mechanism solves two hard problems together: segmenting the sound stream into words, and factoring out core word meanings from random extra meaning present on particular occasions.

The robustness of the rule learner carries over to this application. Even if the child frequently misinterprets the scene, or mishears the word, still a few clean examples, (interspersed amongst any number of spoiled ones) are sufficient for learning. Even if on every occasion, the child adds much irrelevant occasion-specific detail to the meaning, still the intersection mechanism will reliably project out the common core meaning.

Robustness follows basically from statistics and the independence of different learning occasions. The chances of some spurious phoneme appearing five times over in the left branch, or of some spurious element of meaning appearing five times over in the right-hand branch, are very small indeed. Both would have to happen in the same five examples to affect the result. Therefore taking a script intersection of five or more occasions eliminates spurious noise with high reliability (while retaining the core sound and meaning). And on the very few occasions when it does not, a couple more examples will do so.

This mechanism can learn not just the meanings of nouns, but of several other classes of word as well. For instance, it can learn the bare meaning of a verb (without, however, learning how this meaning is linked to the meanings of the verb's arguments); similarly for an adjective or adverb. Such early learning of bare verb meanings is not, in this theory, a necessary prerequisite to learning the full verb m-script.

It cannot, however, learn anything about closed-class morphemes - because their meanings have an information content (typically 1 - 5 bits) which is too small to operate the learning mechanism. Recall that it requires rule scripts to have large information content (say, 10 bits or more) in both their `cause' and `effect' branches in order to clearly distinguish operation of the rule script from random noise; closed-class morphemes fail on one or both counts.

# 3.8 M-script Learning

Learning bare meanings of isolated words - nouns or otherwise - is only a start to learning a language. It tells the child nothing about the syntax of the language, how sentences hang together, or how complex meanings are constructed out of simple ones. To know these things, the child needs to learn the unbounded script functions by which language meanings are constructed. She must learn the full m-scripts for words.

While functionally, an m-script is much more powerful than a script (it is an unbounded function from scripts to scripts), structurally, an m-script is very much like a script; it is just a script with a few trump links. If you can learn a script, and then learn its trump links, you have learnt an m-script. The only thing you need to learn about a trump link is where to put its two ends. It turns out that this is a rather simple learning problem, compared to the problems we have solved already.

Just as rule scripts are discovered by script intersection, so m-scripts are discovered m-intersection, described in section 2 and Appendix A. To m-intersect two scripts (or m-scripts) together, first you do a script intersection, and then do some further simple tests to discover any trump links.

There is a general procedure to learn a word m-script from a few sentence examples where (a) you know the meanings of all other words in the sentence, and (b) you can infer the meaning of the sentence from non-linguistic clues. It is a fairly direct extension of the noun-learning procedure we have already described. The procedure is:

1. Form an initial SMS from the sounds you hear, including known words and the unknown word.
2. Go as far as you can in understanding portions of the sentence, using the words you know. This builds the left-hand end of the SMS H(t).
3. Form the right-hand branch of the same H(t) from the full meaning of the sentence, inferred from non-linguistic information.
4. Collect several such H(T) with the same unknown word, and use them in the learning procedure as before - forming pairs $M_{tu}$ = H(T) ∩m H(u) and retaining only those $M_{tu}$ with high information content in both branches. (Use m-intersection in stead of script intersection.)

M-intersection projects out the word meaning structures and phoneme sequences as before, but it also discovers the trump links which give m-scripts their power as script functions. It learns the full word m-script for the unknown word.

We shall show how this works out by an example, learning an intransitive verb. Suppose the child has learnt the noun m-scripts for a few common and proper nouns, by the script intersection mechanism above; she then knows m-scripts for *Mummy*, *Fred*, *girl* and a few other nouns. Suppose she then hears the sentence *Fred shouts* in a context where she can hear who is shouting.

She forms an initial SMS from the sounds she hears. She then applies the m-script for *Fred* to add an entity node representing Fred, and adds the full meaning which she has inferred non-linguistically as a right branch. The resulting SMS is shown in figure 14.

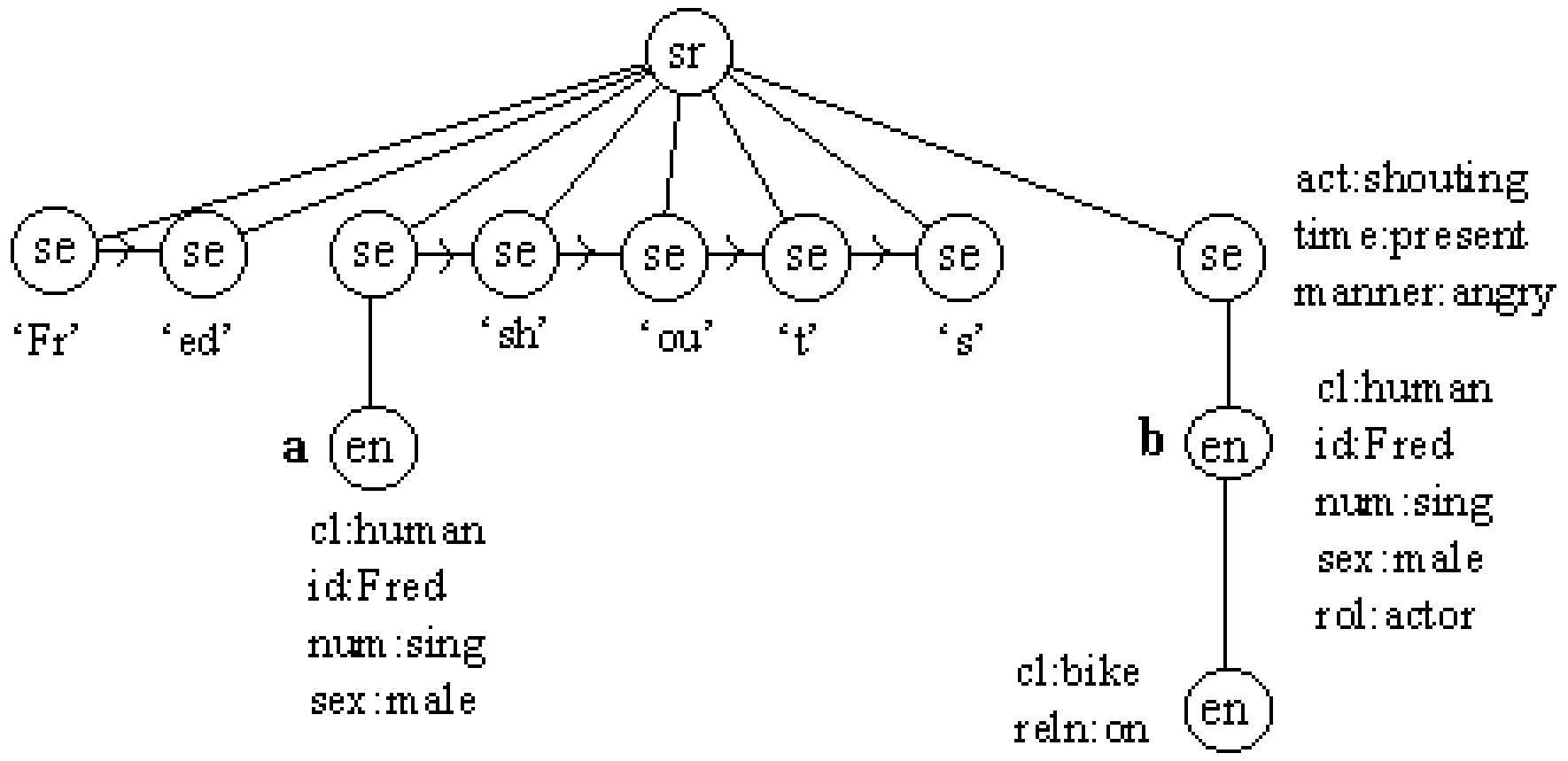

*Figure 14: SMS H(t) constructed by the child on hearing the sentence `Fred shouts' and using her knowledge of the word `Fred' to partially understand it.*

A second example *Mummy shouts at Fred* gives rise to the SMS H(u) in figure 15.

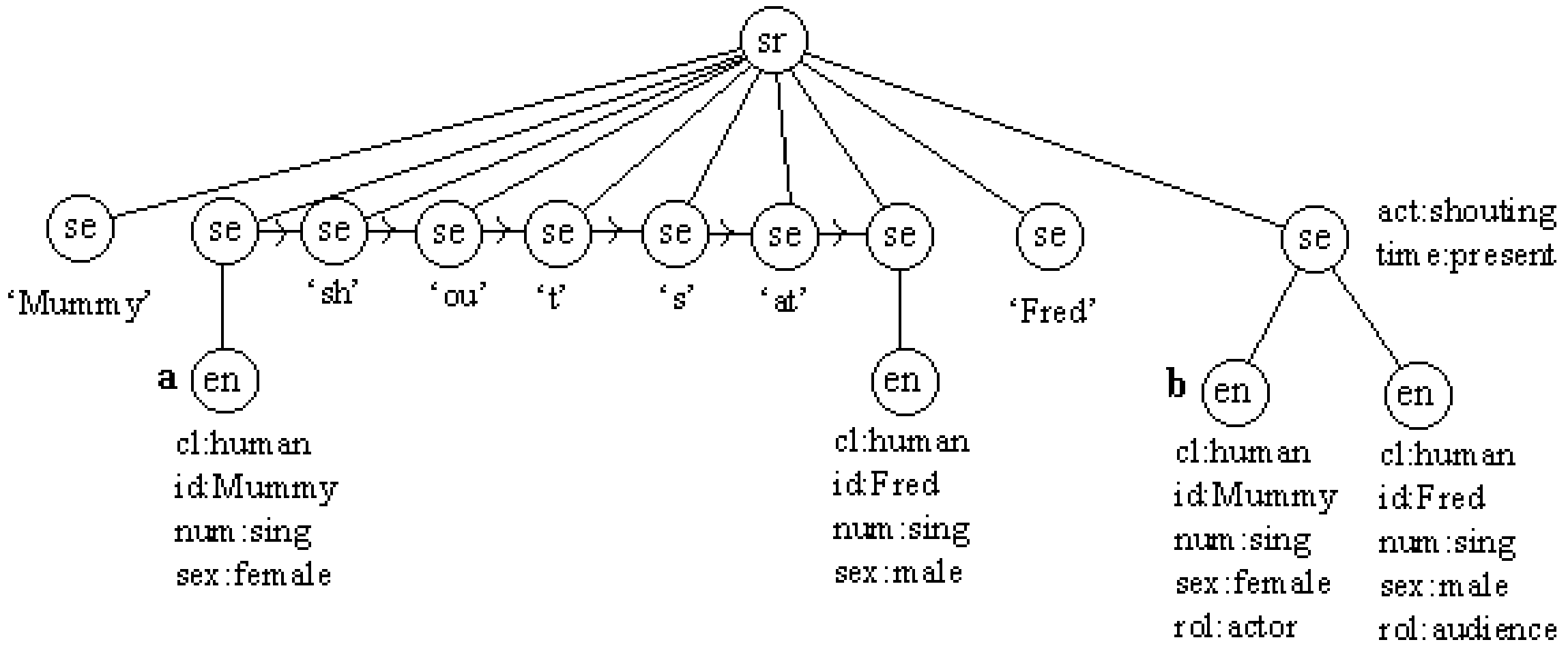

*Figure 15: Another example, when the child hears the words `Mummy shouts at Fred', using her knowledge of the words `Mummy' and `Fred' to partially understand it.*

The result of m-intersecting these two SMS together is shown in Figure 16.

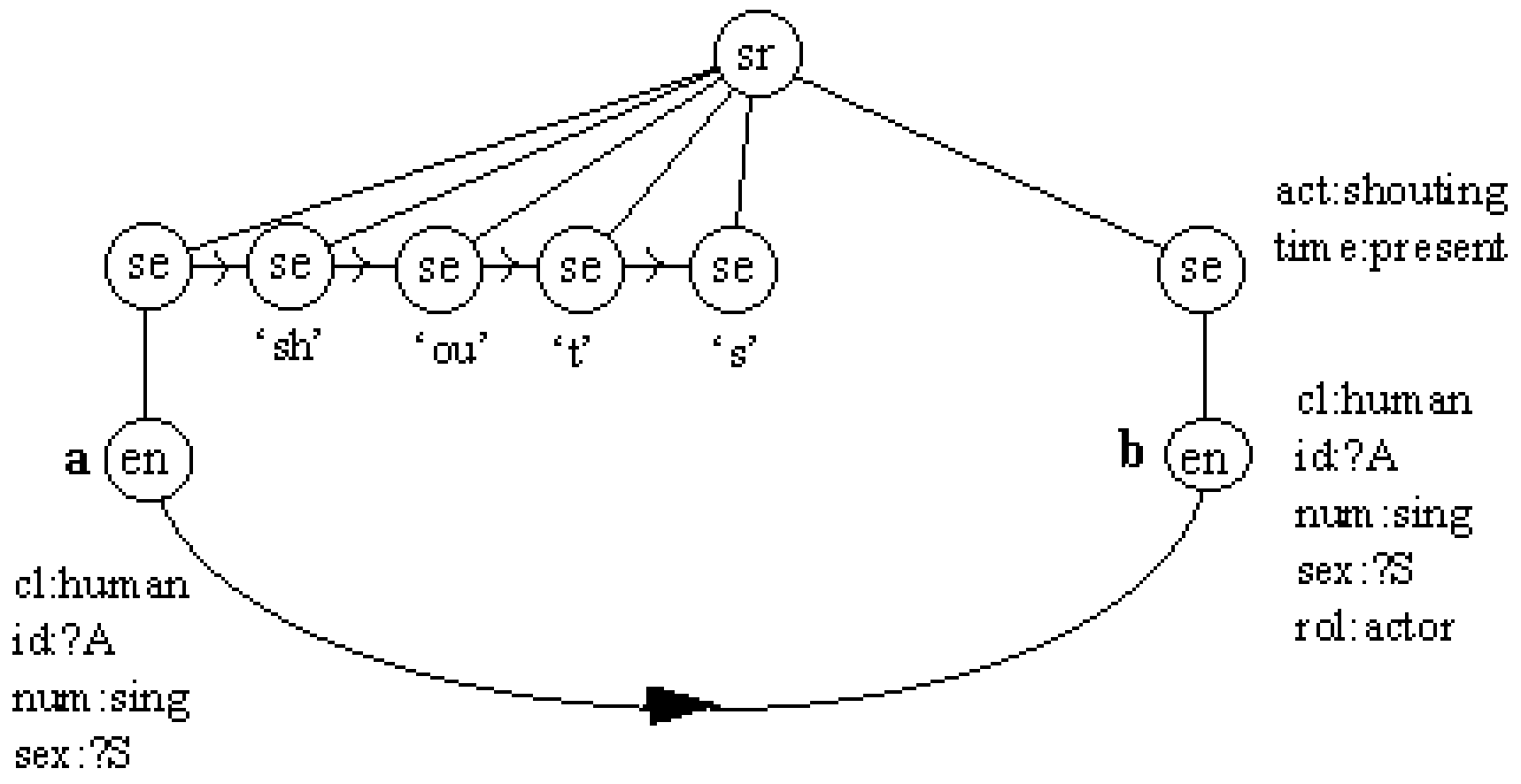

*Figure 16: The result of m-intersecting the two scripts of figures 14 and 15, which is the m-script for the word `shouts'.*

M-intersection, like script intersection, projects out the common structure from two or more scripts, and rejects all the rest. Thus it discovers the shared meaning in the right branch (the core meaning of *shouts*), and the shared sound sequence `shouts' in the left branch - rejecting non-shared sounds such as `Mummy'. It also discovers shared variable identities [7] such as `?A' in figure 16. However, most importantly, it discovers the trump link (the curved arrow) which enables the argument of *shouts* (e.g. Fred or Mummy) to be incorporated in its meaning script.

If an m-script W has a trump link from node **a** to node **b**, then all the scripts H within its scope (the learning examples) should obey the trump link relation H[b] = H[a] Us W[b]. However, the child observes not just the speaker's intended meaning, but may observe other irrelevant aspects of the scene - so she may observe a meaning script N which includes the intended meaning M, so that N s M. This means the learning examples do not necessarily obey H[b] = H[a] Us W[b], but must obey a weaker relation H[b] >s H[a].

Therefore to learn a word m-script, we use a variant of m-intersection which discovers a trump link whenever the relation H[b] >s H[a] is obeyed by all the learning examples.

This looser form of m-intersection can discover trump links even in the presence of noise in the meaning scripts. Can it do so reliably ? Suppose it erroneously put a trump link between two nodes **c** and **d**. It would only do this if the relation H[d] >s H[c] were obeyed in all the (6 or so) SMS used for the m-intersection. For two random script subtrees H[c] and H[d], the probability of having H[d] >s H[c] by chance is of order 2**[-I(H[c])]; for even one input script this probability is usually quite small, and the chances of it happening for all six SMS H(t) are very small indeed. So the m-intersection algorithm has only a small chance of discovering a spurious trump link.

M-intersection reliably discovers trump links, and thus reliably discovers the ways in which word m-scripts link their meanings to the meanings of their arguments. It discovers full word m-scripts; figure 16 is the full word m-script for `shouts', containing all its syntax and semantics. Once the child has learnt this m-script, she can reliably use the word `shouts' both for understanding and language generation.

As before, each possible word m-script W has a prior probability 2**[-lambda*I(W)], and a crossover analysis is used to find how much evidence is needed to overcome this small prior probability, and hence to `believe' that the m-script exists. As before, the result of this analysis is that about six clean learning examples are needed - which is also the number required to determine reliably what nodes, slots and values are part of the m-script, and to determine its trump links.

The same procedure works to find more complex m-scripts, such as that for the word `gives' in figure 8. Even though `gives' has three separate trump links (one each for the donor, the recipient and the gift) it does not take more than about six good examples to define all the trump links and all the information in the m-script.

# 3.9 The Primary Learning Algorithm

One further elaboration is required for the algorithm of the previous section, to make it capable of learning any open-class word (eg noun, verb, adjective or adverb) in a language. The algorithm of 3.7 is good for finding any verb which may be the main verb of a sentence, because that is the last m-script to be applied in constructing the meaning, when understanding the sentence; the SMS H(t) are made by partially understanding all the verb arguments (to make their left branches) and by inferring the full meaning from non-linguistic clues (to make their right branches).

Consider the problem of learning the m-script for *red* from example sentences such as *Peter burst the red balloon* and *Madge bought a red lollipop*. Suppose all other words are known, but *red* is unknown. Applying the m-script for *red* is not the last step in understanding the sentence, so we cannot just do all the other understanding steps and then pair the result with the full sentence meaning. We need a way to isolate the unknown meaning portions *red balloon* and *red lollipop* as input to the learning algorithm.

The way to do this is to use both partial sentence understanding and partial sentence generation in constructing each SMS. Given a sentence *Peter burst the red balloon,* given a non-linguistic understanding of the full sentence meaning F, and given word m-scripts for all words except `red', the algorithm is as follows:

1. As before, form an initial SMS from the word sounds, and apply m-scripts for known words to understand it as far as possible.
2. Add the full meaning script F as a right branch of the SMS, and generate a sentence as far as possible, using known words which have not been used for partial understanding in (1) above. In this case, m-unify F with the m-script for *burst*, which breaks out the meanings of *Peter* and *red balloon* (the two arguments of *burst*). This forms the full SMS H(t).
3. As before, collect several of these SMS from sentences where *red* is the only unknown word. M-intersect these together in pairs to form $M_{tu}$, and group together those H(t) whose pairings $M_{tu}$ have large information content in both branches (thus separating homonyms such as *read*).
4. These SMS are all scripts within the scope of the m-script for *red*. M-intersect together all the H(t) in such a group (about six of them) to form the m-script for *red*, shown in figure 19.

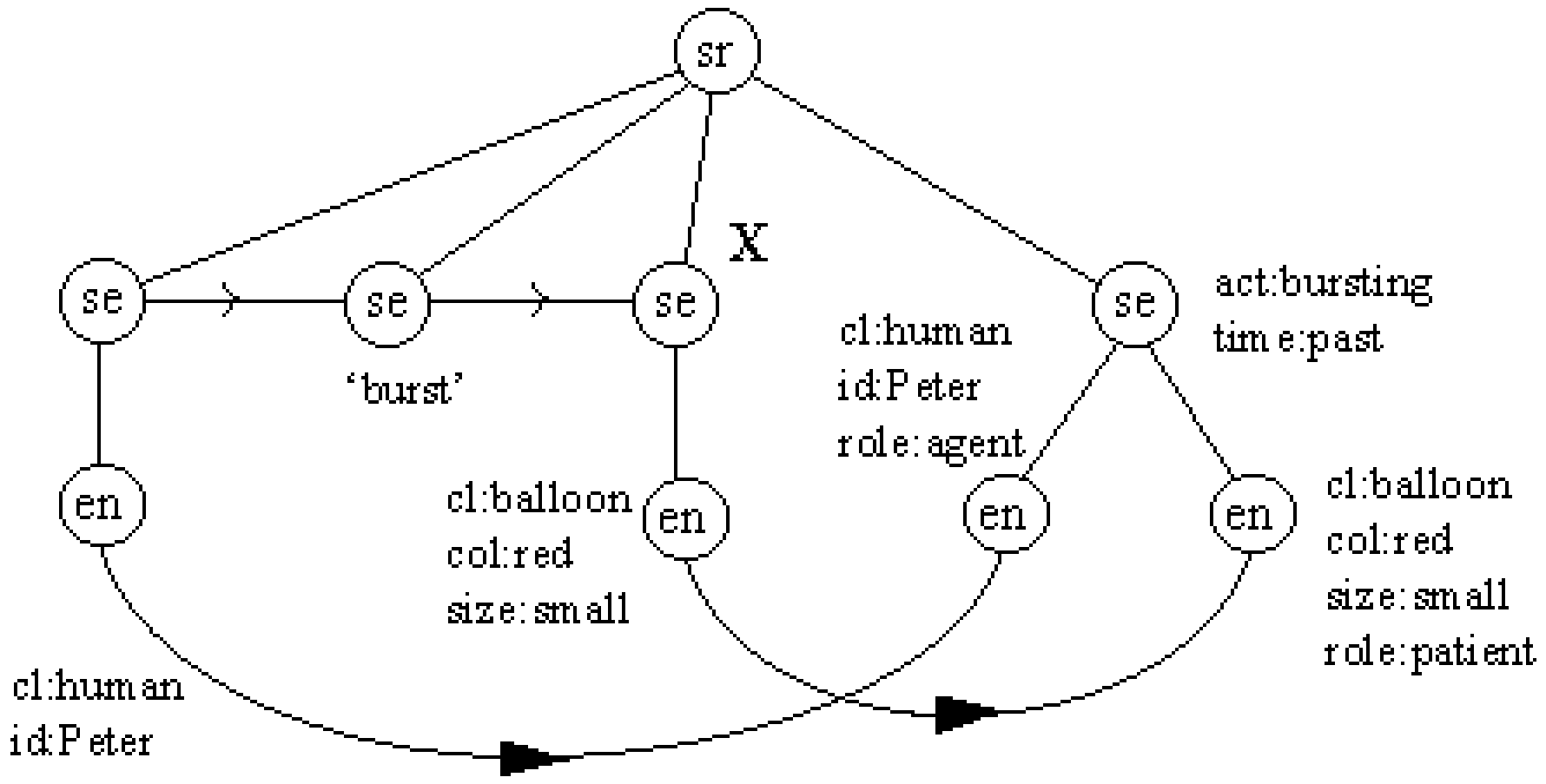

*Figure 17: First stage of generation from the meaning script describing `Peter burst the red balloon'. The m-script for `burst' is applied right-left, to create the left branch, passing meanings back along the trump link arrows. The scene marked `X' is then used to help learn the word `red'.*

In these examples, the script meaning representations are simplified, and the word sounds are represented as single scenes, rather than as one scene per phoneme, for simplicity. Also, some part of the SMS which obviously differ between examples, so will be eliminated by the m-intersectoin, have been stripped off from the figures. The learning algorithm is not affected by these simplifications.

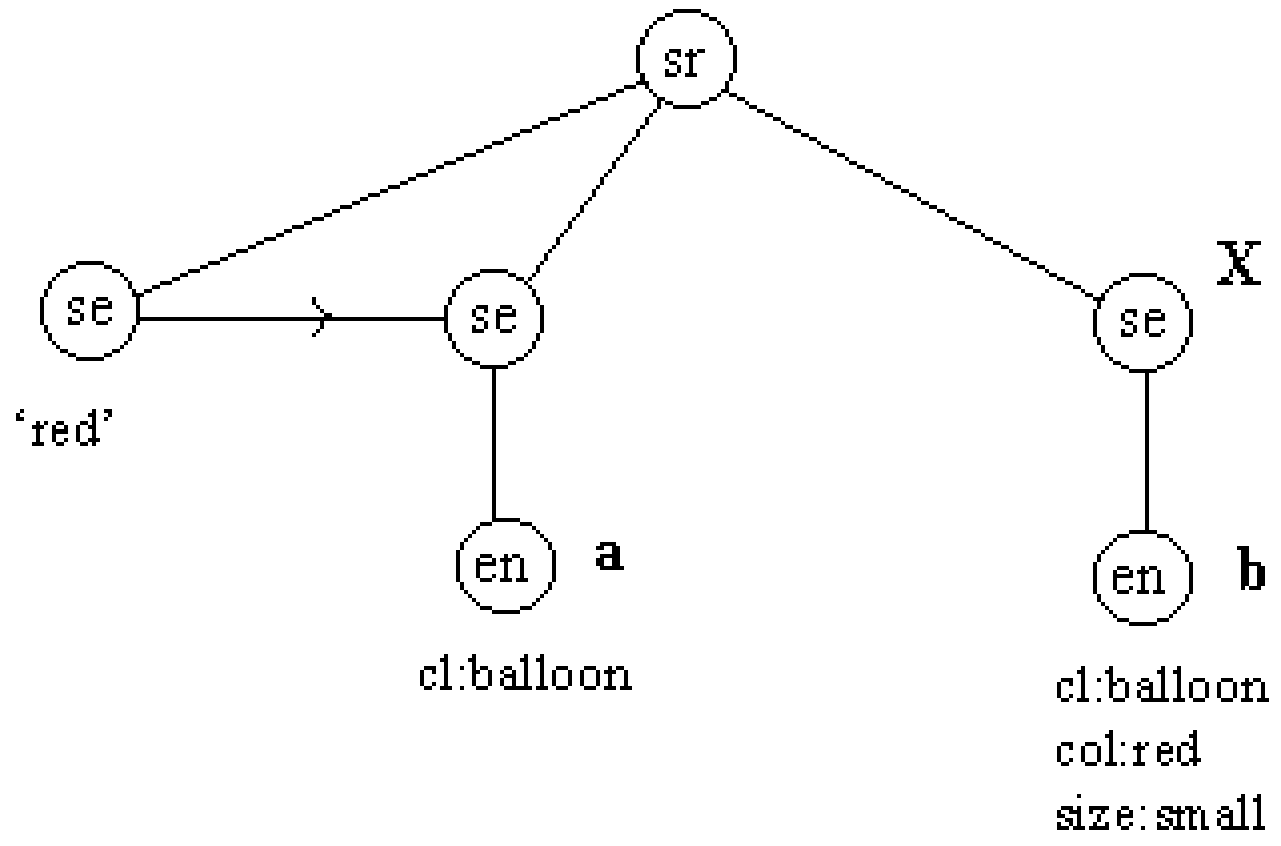

*Figure 18: A learning script H(t) obtained by pairing the meaning scene X from figure 17 (right branch) with the partially understood phrase `red balloon' (left branch)*

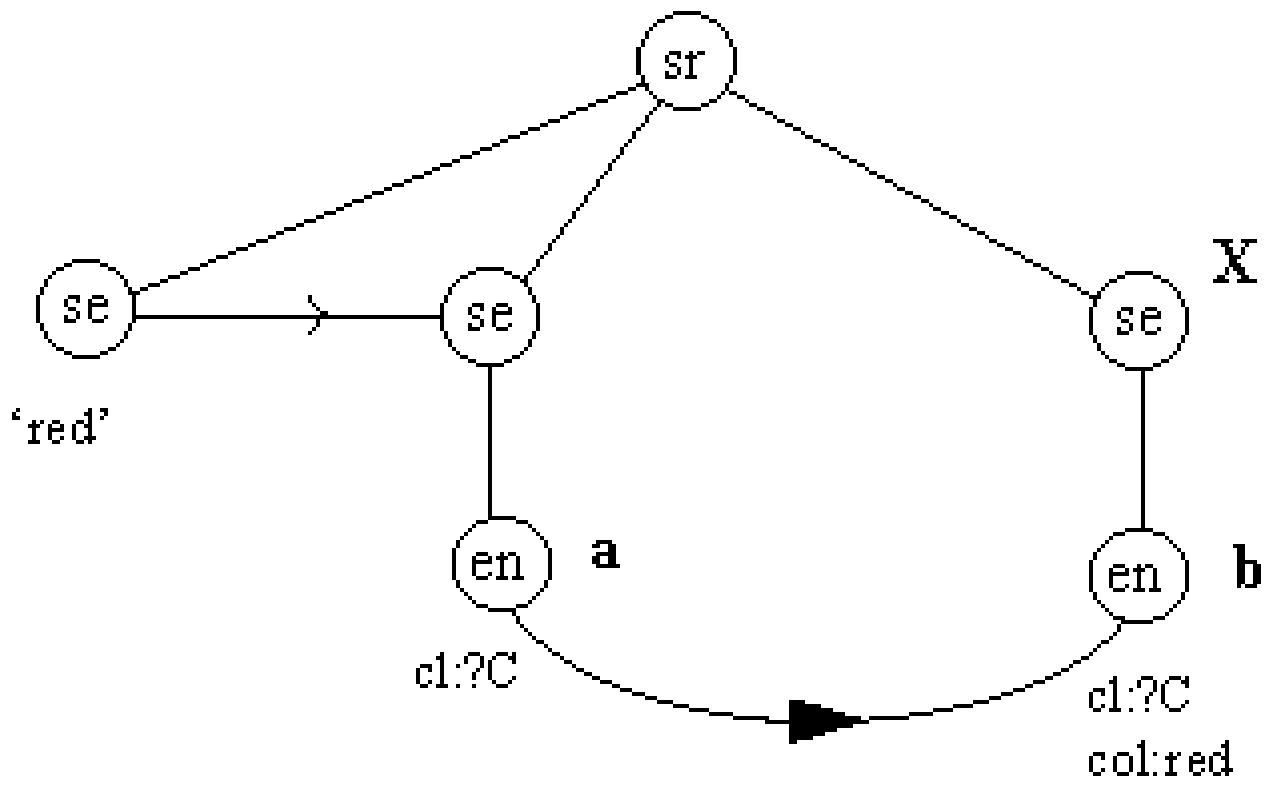

*Figure 19: M-script R for the word `red', found by m-intersecting several learning scripts H, including that shown in figure 18.*

Since the learning script H(t) in figure 18, and all the others H(u) ... like it, obey the relation H[b] s H[a], the m-intersection R = H(t) ∩m H(u) ∩m H(v) ... discovers the trump link from node **a** to node **b**. This trump link enables the script function for *red* to convert the meaning script for *balloon* into that for *red balloon*; the meaning script for *wooden sailing boat* to that for *red wooden sailing boat* and so on.

We can now see that the previous learning procedures - the first one for nouns, and the second one for verbs - are special cases of this primary learning procedure. For nouns, there is no partial understanding stage necessary; for main verbs, there is no partial generation stage necessary; but in general, one can use both partial understanding and partial generation to narrow down to the unknown word, followed by m-intersection of several examples to find its m-script.

In normal language use, speaker and listener form the same SMS in their heads. The SMS contains the word m-scripts for all words in the sentence. If you know the m-scripts for all the words, and know the meaning script, this SMS is defined redundantly; every scene within it is constrained in two independent ways. So even if you do not know the m-script for one of the words, you can still reconstruct the same SMS. Children do this, and m-intersect the SMS to find the m-script for the unknown word.

To see why each SMS is defined redundantly, regard an SMS as a chain of the form:

[word sounds]<->[word m-scripts]<->[sentence meaning]

Every scene along this chain is redundantly specified; each scene is

**either**: (a) A heard sound, and in the left branch of a word m-script

**or**: (b) The right branch of a word m-script, and part of the left branch of another word m-script

**or**: (c) The right branch of a word m-script, and the full meaning script.

So in each of (a), (b) and (c), a scene of the SMS is defined in two independent ways. Because the SMS is redundantly specified, **even if any one word m-script is missing, the**

**SMS can still be constructed**. The scenes of the missing word m-script will each be defined just once (in one of (a), (b) or (c)) rather than twice; but it will be the same SMS. The examples above illustrate this.

***Chemical Analogy**: Normally, speaker and listener synthesise the same SMS molecule from opposite ends; the speaker from the `meaning script' end and the listener from the `word sound' end.*

If a child does not know the m-script for one word, but can infer the meaning script by other means, she can still synthesise the same SMS molecule, working inwards from both ends.

Collecting a set of these SMS molecules with the same unknown word, and analysing them by m-intersection, she can project out the missing word atom.

This learning procedure uses linguistic knowledge powerfully to narrow down the range of things the speaker might be referring to. From the example *Peter burst the red balloon*, knowledge of the meaning and syntax of *burst* is used to deduce that *red* can refer only to the thing which Peter burst; then knowledge of the word *balloon* is used to further constrain the possibilities - *red* must be some attribute of a balloon. Also, the only phonemes left unaccounted for are those in the word *red*, which are bounded on either side by known words.

To use an analogy: at first, the child must pick pieces of meaning out of her environment with clumsy fingers, and try to link them with word sounds; but later, her growing knowledge of language is like a pair of tweezers which enables her to pick out much more finely the meanings that adults are referring to.

So as the child's vocabulary grows, she can use this stronger learning procedure - which uses both partial understanding and partial generation to narrow the range of possibilities - to learn each new word from a few clean learning examples, rather than from many noisy, contaminated ones. This linguistic `cleaning' of the signal allows the procedure to learn some quite `small' words, which are phonetically not very salient and have small information content in their meanings. For instance, prepositions such as *in*, *above* etc. may now be learnable by this procedure. However, it is not the only way available to learn closed-class morphemes; for these, another procedure is also available.

Intuitively, the communicative fidelity of language depends on generation and understanding being (in some sense) the reverse of one another, so that hearers recover the same meanings as speakers start from. In this theory, the reversibility arises because generation and understanding use the same process of m-unification, run either `forward' or `backward'. In the secondary learning process, learners use both together to isolate learning examples for new words.

# 3.10 Gathering Negative Evidence

This full primary learning process has another essential use for the learner, in that it **creates occasions for negative evidence**. It is known that explicit negative evidence is rare and unreliable (Marcus 1993), and implicit negative evidence is a hard concept to define. For instance, if adults never say *goed* for six months, how much negative evidence does that constitute ? How many times might an adult have said *goed* in that period? On what occasions does the child notice the lack of *goed*? On this basis, it seems that any rare word or construct risks being unlearned by implicit negative evidence.

In this theory, the child can receive definite, usable negative evidence on a countable number of occasions, as part of the primary learning process. Suppose she believes that *goed* is a word, with the obvious meaning. On some occasions, she will hear an adult speaking, infer non-linguistically what is being referred to, and start to generate a sentence describing it (step (2) above). This `silently generated' sentence contains *goed*; the child may then record that the adult did not say *goed* but in stead said *went*. So this one occasion can serve both as positive evidence for *went* and negative evidence for *goed*.

By accumulating enough negative learning examples against the *goed* m-script (probably more than six examples, as discussed in section 3.6) the child may learn that *goed* is not a word.

I shall call this the *primary unlearning process*. It is an explicit computational realisation of one of the operational principles in Slobin's (1985) model of language acquisition: *OP (REVIEW): MONITORING. Compare utterances you hear with forms that you would produce in the same situation. Store mismatches and attempt to accommodate your grammar to unassimilated input forms....*

Theories of language learning have gone through many contortions, because it has been believed that children receive no negative evidence. The primary learning mechanism enables the child to capture pieces of negative evidence almost as easily as she captures positive evidence.

# 3.11 The Secondary Learning Algorithm

There is a second learning procedure which may be used for learning `bound' closed-class morphemes, such as the regular morphology of nouns and verbs - m-scripts which may have rather small information content, so cannot be discovered by the primary learning process.

Suppose the child has learnt several forms of the same verb - the m-script *gives* of figure 8, together with other forms such as *give* and *giving*. For each one, we shall suppose the sound scenes are phoneme-encoded, with a sequence of contiguous sound scenes for *g*, *iv s* and so on. Each verb form describes the same basic giving action, but particular forms have extra information - for instance the form *gives* specifies that the time is the present, and that the donor is third person singular (this information can be found in the m-script of figure 8).

Now if we m-intersect together the different m-scripts for of *give*, *gives*, *giving* , two things happen:

- In the left-hand branch of the result, only the stem of the verb survives; other phonemes such as the *-s* of *gives*, or the *-ing* of *giving*, do not appear in all forms, so do not appear in the m-intersection.
- In the right branch of the result, only the core meaning of *give* survives; for instance, the information (time:present) only applies to certain forms, so does not appear in the m-intersection.

Suppose the child has also learnt m-scripts for the third person singular form of several verbs - *gives*, *cuts*, *eats* etc. All of these specify (time:present), and specify that the actor is third-person singular; but the actions in the verbs are diverse. If we m-intersect all these m-scripts together, then two things happen:

- In the left branch, all verb sounds must be preceded by some entity representing the agent, who must be third person singular, and all verbs end in *-s* ; but otherwise there is no commonality, and nothing else survives in the m-intersection.
- In the right branch, all actions occur at time:present and all have a singular agent; but otherwise, all the particular verb meanings are lost in the m-intersection.

By taking these m-intersections, therefore, we have factorised verb m-scripts into `stem' and `inflection' parts. We can also show that if we m-unify the stem m-script for some verb with an inflection m-script, we recover the full m-script for the inflected verb. Therefore full inflected m-scripts need not be stored; they can be reconstituted by m-unification.

Denote full inflected verb m-scripts by V, verb root m-scripts by R, and inflection m-scripts by I. The root and inflection m-scripts are derived by m-intersections such as:

R(*give-*) = V(*gives*) ∩m V(*give*) ∩m V(*giving*)....

= V(*gives*) ∩m X

I(*-s*) = V(*gives*) ∩m V(*hopes*) ∩m V(*lets*) ...

= V(*gives*) ∩m Y

Then an identity of the m-script algebra shows that the reconstituted inflected verb script [R(*give-*) m I(*-s*)] cannot contradict the original inflected verb m-script:

R(*give-*) Um I(*-s*) = (V(*gives*) ∩m X) Um (V(*gives*) ∩m Y)

<m V(*gives*)

Therefore the reconstituted script cannot contain more information than the original, and cannot contradict it; but it still might contain less information. However, we know that [R(*give-*) Um I(*-s*)] must contain all the root verb information in R(*give-*), and it must contain all the third-person present information in I(*-s*); so we can show by enumeration that it contains all the information we need in the original inflected form. To all intents and purposes it is equivalent. It is usually identical.

This secondary learning process enables the child to learn reliably the m-scripts for closed-class morphemes such as *-s*, *-ing*, *-ed* and so on - even though the sound scene in their left-hand branch is not very salient, and the amount of meaning in their right-hand branch may be very small (too small for the primary learning process to work). It can do this because the input for secondary learning is not `noisy' observations of everyday scenes, but consists of word m-scripts which have already been cleaned up by an m-intersection process. Secondary learning starts with clean, low-noise data, to extract a small signal.

The secondary learning process has several distinct and important uses in the theory:

1. **Economy of storage**: separate m-scripts need not be stored for all inflections of regular words; only the stems and a few inflections need to be stored, and can be reconstituted by m-unification
2. **Economy of learning**: all inflections of regular words need not be learnt separately. One can learn one or two inflections, then generate the others productively.
3. **Storage structure for generation**: As discussed in section 2.8 above, language generation can be done efficiently if word meanings are stored in a subsumption graph - a tree of scripts, where the script on each node of the tree subsumes (is included in) the scripts on its descendant nodes. This means we can navigate rapidly down the tree from the root, finding words whose meanings approximate better and better to the meaning we want to express. The secondary learning process creates the intermediate nodes of this tree structure.
4. **Handling ambiguities** : We often hear sentences in which words are unclear, or are made so by other noises. On hearing *I'm [CRASH]ing your supper* , we need some way to interpret it in spite of the lack of a verb. Secondary learning from m-intersection of many transitive verbs gives a `vanilla transitive verb' m-script which can be used to make a preliminary interpretation of this sentence - fixing the agent, patient, tense and modality before other processes are used to disambiguate it.

The secondary learning of broad pseudo-words gives us two ways of handling word ambiguities, which are mutually consistent. If we thought the missing verb might be *cooking* or *bringing* , we could try both verbs and take the script intersection of the resulting sentence meanings; or if we have no clue, we just use a `vanilla transitive verb', got by secondary learning, whose meaning is a script intersection of many transitive verb meanings. In either case we get a temporary meaning which is a script intersection of the true meaning with something else - so is guaranteed to be a subset of the true meaning.

The primary and secondary learning procedures are part of a continuous spectrum. There are intermediate points along this spectrum, which can lead to learning the same word. This is illustrated in Figure 20.

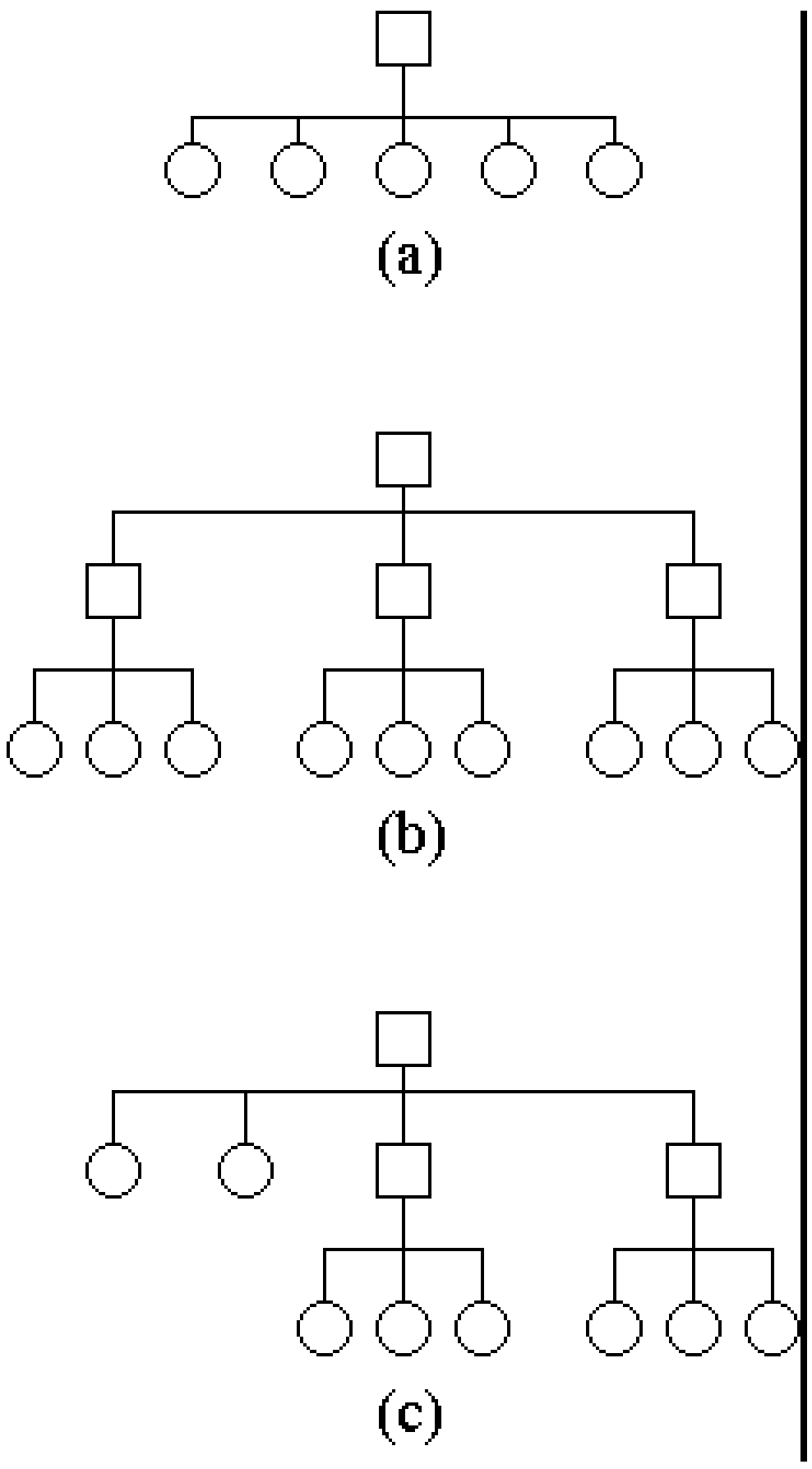

*Figure 20: (a) Primary learning, (b) secondary learning and (c) hybrid primary/secondary learning*

Figure 20(a) shows the primary learning process, in which a number of learning example SMS (circles) are m-intersected together to form a word m-script (square).

Figure 20(b) shows a pure secondary learning process; several primary learning processes deliver m-scripts, and they in turn are input to a further m-intersection.

Figure 20(c) shows a hybrid learning process. To m-scripts from primary learning, and a number of new SMS, are m-intersected together.

Because m-intersection is commutative and associative, it is hard to distinguish between these processes; they all give the same result.

***Chemical Analogy**: Very light word atoms cannot be reliably detected in noisy large sentence molecules. In stead, the sentence molecules are first analysed down to heavier `pseudo-atoms' which are actually small molecules. These (purified) small molecules are later analysed further, to reveal the light atoms of closed-class morphemes. The analysis can be done in different orders and still gives the same result.*

## 3.12 A Fundamental Theorem of Language Learning

We can prove an important self-consistency property of this set of learning procedures. It is the equivalent (for language) of the stability of DNA replication as the basis of life.

This fundamental theorem of language learning is:

**If speakers compose sentences by m-unification of word m-scripts, and people learn words by m-intersection of sentences they hear, then a set of word m-scripts (a language) will propagate stably through a population**.

The sketch of the proof is as follows: consider a word whose m-script is W, which is used by speakers in a population. They compose sentence-meaning structures S by m-unifying W with m-scripts for other words U, V. etc.: S = U Um V Um W Um... = W Um X. (Here S is an SMS like that in figure 7, which speakers use for generation, and learners reconstruct as learning examples). Somebody learns his own version W' of W by m-intersecting these SMS together, W' = S ∩m S' ∩m S"... = (W Um X) ∩m (W Um Y) ∩m...

A simple result of the m-script algebra then implies that W' >mW. Therefore W' is consistent with W; it contains all the information (nodes, slots, trump links) that W contains, and may contain more if there are random coincidences between the learning examples S, S' etc. However, if enough learning examples are used, these random coincidences will be vanishingly rare; and if they accumulated they would tend to make a word too narrow in scope to be useful, and speakers would tend to broaden it again. Therefore the learning process faithfully transmits the m-script W through a population of speakers.

Since this applies to any individual word m-script W, it also works for an arbitrary collection of word m-scripts W, X, Y...... It works, for instance, in the presence of homonymy and polysemy - when the same word sound has several distinct meaning scripts (i.e. distinct m-scripts). So it guarantees that any language will be faithfully transmitted by the learning process.

***Chemical Analogy**: Each language has its own distinctive stock of word atoms. Speakers synthesise these together by m-unification to form sentence molecules. Learners re-synthesise the same molecules by attending to inferred meanings, and analyse them down to word atoms. They can only recover the same word atoms which the molecules were originally made from.*

This shows that the m-intersection learning mechanism propagates word m-scripts stably from one generation to the next, just as DNA replication propagates genotypes stably. As for DNA replication, there must be mutation-like processes which lead to language change; but these happen against a background of stable, faithful replication.

# 3.13 Summary of the Learning Theory

Every word (more precisely, every word sense) in a language is described by an m-script with two branches. This m-script is a script function from its left branch to its right branch. The left branch describes the word sound and other meaning scenes which are arguments of the function; it defines the syntax of the word (e.g. it defines order and agreement constraints between the arguments). The right branch describes the full meaning of the word including the meanings of its arguments. Trump links from left to right branches convey meaning scripts of unbounded complexity from the arguments into the full meaning.

Each word script function is applied left-to-right, to build up complex meaning scripts, in language understanding. It is applied right-to-left, to break down complex meaning scripts into their component parts, in language generation. This applies to all parts of speech.

In both generation and understanding, word application is non-destructuve - adding to the sentence-meaning structure rather than changing. Speaker and listener build up the same SMS in their minds, which underpins the faithful communication of language.

While the `function' view of word m-scripts is most appropriate for language generation and understanding, the m-script algebra view is more appropriate for learning. If a word with m-script W is used to construct several SMS Z, then they all obey Z >m W.

The strategy for learning a word m-script is to collect a few example SMS (X, Y, Z...), then to m-intersect them together forming W' = X ∩m Y ∩m Z.. M-intersection of the SMS constructs the m-script W' with the smallest possible scope which contains all of them, which must (by an m-script algebra identity) obey W' >m W ; given six or so examples, coincidental similarities between X, Y, Z.. are rapidly weeded out, and W' is a very good approximation to the word m-script W.

Different strategies for collecting SMS are used at different stages of learning:

1. For learning early nouns, the left branch of an example SMS is a sound sequence containing the sound of the noun; the right branch is a meaning script which may describe the entity being referred to.
2. For learning early verbs, some nouns must be known, to partially understand a sentence and build the SMS from the left, supplying the verb argument entities; the right branch is a sentence meaning inferred by non-linguistic means.
3. When more words are known, both partial generation from the full meaning script, and partial understanding from known words, are used to build the SMS from both ends.

The strategies (1) and (2) can be regarded as special cases of (3), which constitutes the most general method for primary word learning.

The secondary learning procedure also uses m-intersection; but as input it uses word m-scripts rather than raw sentences and meaning scripts. This allows it to extract the m-scripts for closed-class morphemes by a `factorisation' process.

These learning mechanisms together give a child the means to learn robustly the m-script for any word in the language, and thus to learn a full adult language. They make many predictions about the nature and course of the learning process; these are compared with observations in section 5.

# 4. The Evolution of Mind and Language

- *Scripts and script learning evolved to support primate social intelligence.*
- *The extension from scripts to m-scripts may have evolved for several purposes, including a theory of mind*
- *Word m-scripts evolve as they reproduce through generations of speakers*
- *Many important features of languages (such as partial regularity and the Greenberg universals) arose from language change, and are not innate in the human brain.*

The m-script theory gives a working computational model of language learning and use - a theory of `where we are now' in language, differing from others in important ways. For instance, in this theory a language is not defined by a few principles and parameters, and a child has no innate parameter setting mechanisms in his head.

No theory of where we are now is complete without a picture of how we got here - of the processes which led to m-script based language. These processes operate over two timescales:

1. Processes of biological evolution, operating over millions of years, which gave our innate capacity to form scripts and m-scripts in our heads
2. Processes of language evolution, operating over hundreds and thousands of years, which lead present-day languages to have the form they do.

By understanding these processes, we can see how the approximately regular, parametrised forms of many languages arose from historic change rather than evolution, so are not an innate endowment of our species. This has important implications for theories of language learning.

## 4.1 Primate Social Intelligence

I have proposed (Worden 1996) that the cognitive faculties underlying language evolved to support primate social intelligence. As this hypothesis can help us understand the nature and robustness of language learning, I summarise here the key arguments for it:

1. Language is used for social purposes; therefore it is likely to have arisen from social intelligence.
2. One facet of our social intelligence is a theory of mind, to represent what others may be thinking. A theory of mind is necessary for language.
3. Language meanings have many properties in common with the internal representations of social situations, which are required for social intelligence.
4. Social intelligence requires a fast, robust learning mechanism, which may then have been co-opted for language.
5. Because of an evolutionary speed limit (Worden 1995) there can be at most 5 Kilobytes of new innate design information in the human brain created by evolution since our divergence from chimps. This is too little for the complete design of a language faculty; therefore language must be largely based on pre-existing faculties. Social intelligence is the best candidate.

We know from many observations that most primates have an acute social intelligence, not found in other land mammals (see e.g. Cheney and Seyfarth 1990; de Waal 1982; Tomasello & Call 1994). They recognise one another as individuals, know all about each others' kin and alliance relations, and can rapidly learn regularities about who will do what in what circumstances. To have this social intelligence, all primates need:

1. To represent in their minds information about social situations past and present - facts about their peers and their social actions.
2. To learn and represent the causal regularities whereby one social situation leads to another - regularities such as “if X screams and Y is X's mother, then Y will react”
3. To combine a knowledge of the present with their knowledge of causal regularities to predict what may happen next, and so to choose actions which further their own ends of stronger alliances and increased rank.

So primates need a mental representation of social situations and causal regularities. In order to be effective, representations in the brain should match the properties of the things they represent (Marr 1982; Johnson-Laird 1983). The social representation in the primate brain should match the properties of social situations, which are:

1. **Structured**: A social situation consists of a number of individuals with attributes (identity, sex, rank, mood...) and relationships or interactions (mother-of, grooming, threatening...). The structural way in which these are combined is important; it matters who is grooming whom.
2. **Complex and Open-ended**: There may be several individuals in one incident, in a variety of relationships; and several incidents together may constitute a particular situation; the set of possible social situations is a very large set.
3. **Discrete-Valued**: Many of the important variables which characterise social situations are discrete-valued (e.g. identity, sex, rank, kin and alliance relations)
4. **Extended in Space and Time**: The incidents which make up a social situation may take place over several days or more, at different places
5. **Dependent on Sense Data of all Modalities** : Important information about the social situation may come from vision, hearing, smell, movement or bodily feelings; the social representation must be connected in the brain to all these sense data.

This list bears a remarkable resemblance to the properties of scripts, and of language meanings. The meanings we can express in a script, or in a sentence, are structured, complex and open-ended, discrete-valued, extended in space and time, and may involve sense data of all modalities. This leads to the hypothesis that *scripts evolved to support primate social intelligence, and were then co-opted for language*; it applies not just to the script representation, but also to the learning and inference mechanisms.

I have built a computational model of primate social intelligence using scripts and compared it with diverse data (particularly from Cheney and Seyfarth, 1990) on primate social intelligence (Worden 1996). This model uses script intersection for learning, and unification for inference. It gives a simple account of many observations of primate social intelligence - such as the learning of alarm calls, habituation to false alarms, use of facts about rank and kinship, and attachment behaviour. These comparisons show that:

- Scripts are a viable internal representation of social situations, to support primate social intelligence.

- Script unification is sufficient to make the inferences needed for most primate social intelligence (excluding `theory of mind' intelligence)
- Bayesian script learning gives rapid, robust learning of social regularities - using just a few examples per rule script - just like the rapid, robust social learning seen in vervet monkeys.
- Monkeys can learn social regularities and their exceptions using implicit negative evidence, as the Bayesian learning model can.

It seems likely that strong, sustained selection pressure for social intelligence has led to the near-optimal Bayesian form of social learning.

Therefore general primate social intelligence, as seen in vervet monkeys, is a likely evolutionary origin for the script representation and operations. However, monkeys' social intelligence can be modelled using scripts, not m-scripts; this does not yet account for the unbounded script function capability of m-scripts, which is also essential for language.

# 4.2 From Scripts to M-scripts

Why did the more powerful m-script facility (needed for language) evolve? Answers to this question are of necessity conjectural, and are not an integral part of the theory of language learning (it need only assume that an m-script capability evolved for some reason); but if there are plausible answers, they lend some support to the theory.

I assume that the m-script capability evolved well before language, and did not evolve just to support language. There are two possible selection pressures (needs for a more capable social intelligence) which may have driven the evolution from scripts to m-scripts, well before language. These are (a) the need for re-construal of social situations, and (b) the need to support a theory of mind.

**A. Re-construal of social situations**: Any social situation can be regarded from several different points of view - for instance, with different entities or agents `foregrounded' as initiators of change. One may construe the same situation (verbally) as *Lucy cried* or as *Charlie made Lucy cry*; in another example, as *The green bottle broke* or *Charlie broke the green bottle*. These different verbal construals correspond to different script meaning structures, which are different viewpoints on the same situation.

The more different construals one can make of the same situation, the more one is able to calculate interesting consequences, and so to do something about it. There is a selection pressure to make multiple construals. If there is any automatic way to create second and third construals, it will be selected for.

Given a script representation A of the situation *Charlie broke the green bottle*, there is a fairly automatic transformation to a script B for the second construal *The green bottle broke*. Script B is a script function of script A, and this function can be represented as an m-script (not as a simple script). To get from script A to script B, the whole subtree representing *the green bottle* (or any other breakable entity) must be moved from one place to another in a script tree; trump links are the feature of m-scripts which enables them to do just this.

Linguistically, such re-construal m-scripts are important in the analysis of alternating verb argument structure. One of these m-scripts, for the locative alternation, is shown in Figure 5.2 in section 5.7. However, the important point is that re-construal is an entirely non-linguistic operation; it is needed as a part of general primate social intelligence, independent of language. This may be a selection pressure that led to the evolution of m-scripts, before language.

**B. The theory of mind**: It appears that most primate species (such as monkeys) have no theory of mind. In predicting each others' social actions, they seem to act like behaviourists. A monkey seems to know social rules like `If I do A, then monkey X will do Y' (e.g. if I groom X, then he may help me in a later fight with Z); but they do not seem to know social rules of the form `If I do A, then X will know B, so he will do Y'. Monkeys can reason about how other monkeys will act, but not about what other monkeys perceive, know, want or plan. One monkey's mental states are opaque to another monkey.

It is unclear whether chimpanzees and other great apes have a theory of mind, but it is very clear that human beings do, and use it practically every moment of the day. We use it as a part of our social intelligence, to analyse what others will think about our actions; and it is an essential prerequisite for language. If we did not realise that other people know things and do not know things, there would be little point in talking to them.

This implies that (a) a working theory of mind has evolved at some stage in our ancestry (possibly since our divergence from chimps) as a facet of social intelligence, and (b) the theory of mind has a close link to language, if only as a pragmatic guide to what is worth saying.

I shall argue that the link to language is even closer; that the formal, computational extension of script-based social intelligence, required for a theory of mind, is just the introduction of m-scripts - and that this m-script capability was then co-opted as the computational basis of language.

What extensions to script-based social intelligence are needed for a theory of mind ?

It seems that monkeys require script trees of depth about 3-4 nodes, to represent scenes with other individuals, their actions and attributes. A theory of mind requires deeper script trees, to represent `X knows Y' - where the top of the script tree represents `X knows...' and the rest of the tree is Y - standing for any script which X may know. Script trees of depth about 8 can represent a basic `X knows Y', but greater depths would be needed to represent `X knows that Z knows that Y' and so on.

A more fundamental extension is also needed. To account for vervet monkey intelligence, rule scripts with simple variables, such as `?X', to represent `any individual', are sufficient. As explained in section 2, these rules act as simple, bounded script functions. To reason with a theory of mind, you need to represent not only facts such as `X knows Y', but also general theory-of-mind rules, such as `If X sees A, then he will know B' or `If I know rule R, then X also knows rule R'. Without these general theory-of-mind rules (and the ability to learn them), your theory-of-mind inferences will be very limited.

In rules such as `If X sees A, then he will know B' and `If I know rule R, then X knows rule R', the variables A, B and R stand not for individuals (which have only a finite number of possible identities), but for whole script subtrees, describing scenes or causal regularities. There is an unbounded set of these subtrees. So theory-of-mind rules need to be unbounded script functions, with unbounded variables in their arguments and results. Such variables can be represented by trump nodes, joined by trump links.

I suggest that the extension of social intelligence to contain a theory of mind required (and to some extent drove) the evolution of scripts to m-scripts. The script operations of unification and intersection were extended to m-unification (for applying unbounded rules) and m-intersection (for learning them). This extension (made possibly within the last 5 million years) was an evolutionary line of least resistance to give a working theory of mind. It was then a small step to co-opt these mechanisms to support language, as described in sections 2 and 3. The theory of mind supports not only the pragmatic `what to say now' aspects of language, but also the computational mechanisms by which we say, hear and learn it.

This evolutionary account is not unique or certain; but if it is true, it explains:

- **How language meaning structures evolved**: to support social intelligence
- **How the m-script language operations evolved**: for re-construals of social situations, and for a theory of mind
- **Why m-script learning is rapid and robust**: it evolved through the selection pressure of intense social competition, which requires rapid robust learning.

In this picture, although the script and m-script faculty for social intelligence may be autonomous (capable of learning and inference by its own mechanisms) **it is not isolated**, in the sense which seems to be implied by the phrase `the autonomy of syntax'. The social representation of scripts and m-scripts is richly interlinked to other representations in the brain, such as mental images and body schemata. It has to be, or it would serve no biological purpose.

In this theory, syntax is embodied in m-scripts, and m-scripts are richly connected to other non-symbolic meaning representations. Aspects of language (such as metaphor) which depend on other meaning representations are not excluded from this theory.

The rich mental faculties which underpin language evolved in response to strong, sustained selection pressure over 20 million years - not in a rapid burst of recent evolution (which could only produce 5 Kbytes of new design) or a freak mutation (which could produce even less). If this economical account can fit the facts of language, it is to be preferred.

## 4.3 M-script Evolution

The m-script theory gives a robust and flexible model of language. Individual word m-scripts can compose together the meanings of their arguments in powerful ways, to build complex meanings in several stages; and each word m-script can propagate robustly and stably through a population of speakers, by a learning mechanism which evolved to learn arbitrary symbolic regularities (of primate social life) from noisy, fragmentary data.

The theory is fully lexicalised. There are no separate phrase structure rules; each word effectively carries around its own phrase structure rule in its m-script, so the set of phrase structure rules could be completely irregular and word-specific.

The partial regularity of languages is a well-established fact, and has been at the base of many theories of language. The m-script theory should give some account of that partial regularity - for instance, the fact that languages seem to have a few core phrase structure rules, or a few core parameters. If, as far as the innate language faculty is concerned, languages do not need to be regular, why do they turn out to be nearly so ?

The answer lies in the process of language change, which can be envisaged (by analogy with biological evolution) as a process of evolution of word m-scripts.

The m-script for each word is a small information structure, with typically 10-100 bits of information. By use and learning, these structures propagate through a speaking population, like parasites in the human brain; they are an example of Dawkins' (1976) memes. By the fundamental theorem of language learning (section 3.12) they propagate stably, without basically changing their form from one generation to the next. They can be regarded as a simple parasitic life form, subject to variation, selection and evolution. It is this process of m-script evolution which leads to the quasi-regular structure of languages we see today.

# 4.4 Domains of Regularity in Language

If language structure results from selective pressures on individual word m-scripts, why should those changes lead to partial regularity ?

Language is used by people to communicate. To do this effectively, it should be expressive, robust, economical and learnable. If any word m-script tends to make the language more expressive, robust, economical or learnable, that word will tend to be favoured by speakers, used more frequently, and therefore learnt more by listeners. This is the basic selective force which leads the populations of different word m-scripts to wax and wane - which leads to language change.

The fitness of any word m-script depends not just on that m-script alone, but on how well it works together with other word m-scripts. M-scripts tend to hunt in packs, and to hunt best in regular packs; that is the selective force which builds the partial regularity of language.

As a first example, consider case-marked versus unmarked languages. Every language needs devices to distinguish between the two arguments of a simple transitive verb - to distinguish *Fred hits Joe* from *Joe hits Fred*. Languages have two main ways to do this:

- **Word-order constraints**: these show as time-order arrows between the scenes in the left branch of the verb m-script, as in figure X. Then, for instance, the noun or noun phrase which precedes the verb must become the agent in the meaning structure.
- **Case markings**: There are no time-order arrows in the verb m-script's left branch, but the entity scenes in the left branch have `semantic role' slots which, to unify, must match with slots on the entities denoted by nouns. Case markings on the nouns give different values of these role slots, forcing a unique assignment of entities to verb arguments. (Section 4.6 describes in detail how this works for nominative/accusative and ergative/absolutive languages)

In both cases, verb m-scripts and noun m-scripts should match up: *either* the verbs have time-order arrows and the noun meanings have no semantic role slots, *or* the verbs have no time-order arrows and the nouns have semantic role slots. A mismatch leads either to redundant meaning in the nouns, or to rampant ambiguity.

This means that if a few prominent verbs in a language go one way (eg require case markings), then there is a strong selection pressure on noun m-scripts to conform to that need; so all the noun m-scripts will tend over time to `line up' with those verbs, having the required case markings. This in turn puts the same selection pressure back on all the verbs of the language - they will tend to use case markings rather than word order. So over time, small incremental changes tend to line up all verbs and nouns with the argument matching system of the dominant verbs.

This interaction back and forth between the nouns and the verbs of a language provides a 'weak force' which tends to 'align' all the nouns and verbs in the same 'direction'. This force is strongest for nouns which are semantically similar, since they tend to be the most interchangeable (those nouns tend to be used with the same verbs); the force is a local force in the space of words, but also has a long-range component across the whole language.

This is much like the local inter-atomic forces in a ferromagnetic solid (interactions between magnetic fields of different atoms) which tend to line up the magnetic moments of all atoms in the same direction. Such a solid typically consists of small regular crystalline domains with irregular boundaries between domains; within each domain, the magnetic moments of the atoms are all aligned, but different domains have different alignments at random.

Similarly we would expect the weak forces of language change to produce local **domains of regularity** in any language Ñ sets of words of similar meaning and with similar syntax. Each domain tends to be self-sustaining and stable against change; but different domains may have different orientation, allowing overall irregularities in the language. When languages collide, domains are broken up and rearranged.

For instance, amongst case-marked languages, the choice between nominative/accusative and absolutive/ ergative markings also makes a force for regularity, tending to line up nouns and verbs on the same choice. Here, the penalties for mixing are weaker, and mixed languages exist; but still the mix is not random, and there are large domains of regularity.

# 4.5 Greenberg Universals and the Head Parameter

Perhaps the most important selective force for regularity concerns word order, and is the force leading to the Greenberg-Hawkins universals and the so-called `Head parameter' of languages.

Recall from section 2 that structural ambiguities are handled (in language understanding) by taking a script intersection of the two ambiguous meanings, thus avoiding a combinatorial explosion of possible readings of a sentence. In practice (as shown by the computational implementation of m-script-based understanding) this works well; but it only does so for languages which obey the Greenberg-Hawkins universals

This can be seen from an example, involving Greenberg's universal number 2, that *"in languages with prepositions, the genitive almost always follows the governing noun, while in* *languages with postpositions it almost always precedes."* This is statistically one of the most reliable of the universals discovered by Greenberg.

Consider the sentence “John saw the lid of the box on the table”. The noun phrase can be read in two different ways, as (the lid of (the box on the table)) or ((the lid of the box) on the table). Because English obeys Universal no. 2 - having prepositions and the genitive following the governing noun - both of these readings refer to some kind of lid. If, however, \`on' were a postposition and so \`A on B' referred to some kind of B, then the first reading would be a lid, while the second reading would be a table.

Taking the script intersection of the two senses preserves their common information that it is a lid, helping in further analysis of the sentence; but if English had postpositions, taking the script intersection (of a \`lid' script and a \`table' script) would leave little useful information for the rest of the analysis.

So to make this particular structural ambiguity easier to handle, the word m-scripts of English pre/postpositions and possessives must evolve together to obey Greenberg's universal no. 2. Any postposition or reversed genitive (in a dominantly prepositional language) would give rise to hard ambiguities, where script intersection does not work and multiple senses must be handled in parallel. Such an m-script would be strongly selected against by speakers and usage. After a generation or so of restricted use, any such \`reversed' m-script would die out, leaving Universal no. 2 true again.

Similar considerations apply to other kinds of headedness of phrases, giving selection pressures on word m-scripts to make them obey the other Greenberg universals. Since, without the universals, structural ambiguities would be very hard to handle, this is a strong selection pressure on word m-scripts, and it tends to line up the whole language in one domain of regularity. There is a strong tendency for languages to be either head-first (like English) or head-last (like Japanese).

Thus the Head parameter of languages does not reflect a fundamental constraint in our capacity to use or learn languages. It did not evolve as part of the human brain to make languages learnable, but evolved in the m-scripts of each language, to make ambiguities easy to handle. A similar analysis may apply to other parameters in the \`principles and parameters' model of language.

# 4.6 Subject and Object

In this theory, matching of verbs with their arguments is done purely in terms of semantic roles, and the grammatical functions of subject and object emerge for other reasons. I shall first describe how, in case-marked languages, verbs match up their arguments with semantic roles; then describe how subject and object emerge, and how they relate to semantic roles.

The account of nominative/accusative and ergative/ absolutive case-marked languages is as follows: entity nodes in scenes typically have two slots: an \`actor' slot (with values act: = yes/no) which denotes whether that entity initiated the action; and a \`change state' slot (with values chs: = yes/no) denoting whether that entity undergoes a change of state. In their use with transitive and intransitive verbs, these slots have values:

Georgy [act:yes, chs:no] *kisses* the girl [act:no, chs:yes].

Georgy [act:yes, chs:yes] *runs away*.

These values are an intrinsic part of the meaning scripts we form, pre-linguistically. Every language needs at least a binary marking (or position) to distinguish the two arguments of a transitive verb. Nominative/ accusative languages define the case marking by the action slot on the noun entity (nom = act:yes, acc = acc:no), and morphologically ergative languages define it by the `change state' slot (erg = chs:no, abs - chs:yes). In each type of language, for economy the more marked case tends to be the rarer (the `1 out of 3' cases in the two examples above - act:no and chs:no).

When sentences are constructed by m-unification, verbs are matched to their arguments directly by these semantic slots, rather than through the grammatical functions of subject and object. Therefore there is no need for `linking rules' between grammatical functions and semantic roles. But the m-script theory still owes us an explanation of how grammatical subjects and objects have arisen in languages.

The answer is driven by the requirement of economy of expression in language Ñ in particular, by the need to avoid unnecessary repetitions of noun phrases. (The need to express pragmatic information also plays a part.)

People tend to say several things in succession about the same thing - particularly when building up a complex meaning script, where entity nodes for the same thing may occur repeatedly. In these cases, it is uneconomical to use the same noun phrase repeatedly to refer to one thing. Languages have evolved to have a whole battery of economy devices to avoid doing so:

E1. Reflexive pronouns, referring to the nearest entity in certain roles

E2. Pronouns, typically referring to some recently-mentioned entity

E3. Omission of a noun phrase in coordinate constructions

E4. Omission of a noun phrase in relative constructions

E5. Omission of a noun phrase in complements

E6. Cross-referencing on verbs, so that semantic roles which are not explicitly given may be identified more easily.

E7. Switch-reference in strings of statements about the same thing

All of these devices achieve economy of communication by avoiding repetition of 'obvious' noun phrases Ñ making it somehow easy for the hearer to work out (by a convention) what is being referred to, without hearing it described explicitly.

Economy devices need rules or conventions to help listeners pick out the thing not described. These rules need to be simple and consistent across any language, in order to minimise the difficulty of learning and using them. This provides another force to make domains of regularity.

To make consistent economy devices across a language, it is useful to have a particular 'privileged' entity node in any verb meaning script, which any economy device may then use without risk of misunderstanding; listeners rapidly learn to use that entity node in consistent ways to construct the intended meaning.

In this respect, the entity node attached directly to the top scene node of a verb meaning script -the entity node with a slot [act:yes] - is particularly useful, because:

- Every verb (transitive or intransitive) has one
- It is easy to locate it in the verb meaning script
- Being the entity which initiates and controls the action, it is always important (and so is likely to be multiply referred to)

These entity nodes are the subject of the verb, and the ways in which the economy devices E1 - E7 use them are the syntactic criteria for identifying subjects in a language.

For any language, some subset of the devices E1 - E7 exist, and use the subject entity in the way described above. While this usually provides a good economy device, it is not the only way, and any language might use some different device (not involving the subject entity) for any of them. Therefore there is no consistent subset of E1 - E7 which precisely defines the notion of subject in all languages (Andrews 1985).

The economy devices are particularly involved in complex constructs such as complementation and coordination, and subjects are central to most economy devices; this is what leads to the autonomous 'grammatical' functions of subjects, independent of semantic roles, and to the historic importance of subjects in the study of syntax.

In the development of child language, on this account we would expect the semantic roles, and their identification in verb argument frames, to come first, and the more complex economy devices, with their use of the notions of subject and object, to come later.

# 4.7 Competing Explanations of Regularity

The m-script theory explains the partial regularity of languages as a consequence of historic language change - the evolution of word m-scripts to achieve greater economy and disambiguity of communication.

This explanation has the advantage that, not depending on innate biological structures in the brain, it does not demand complete regularity. Any account of language which starts from regularity, linking it to structures in the brain, will sooner or later find irregularity an embarrassment. How do brain structures designed for regularity cope with irregularity [1] , and why do they tolerate it ?

On the m-script account, however, m-scripts evolve to build `domains of regularity' in a language; but where two different regular domains collide (as, for instance, where two languages collide) there must be an irregular border between them. The brain copes easily with this irregularity, because the brain requires no regularity.

One might feel that this is just another, competing, theory of language regularity. Regularity might have evolved innately as part of the human brain, or it might have evolved by m-script evolution - how can we decide between these two accounts ?

We can do so by considering the speed at which the two competing processes take place. Both are processes of evolution and selection - on the one hand, selection of human brains to handle regular languages, and on the other hand, selection of word m-scripts to be approximately regular.

We can see, both theoretically (from the evolutionary speed limit in Worden 1995) and empirically, that the evolution of m-scripts is much, much faster than the evolution of the human brain. The form of a language can change radically in a few hundred years - whereas the design of the brain has not changed significantly in tens or hundreds of thousands of years.

If you have two competing mechanisms for the same effect - in this case, for the partial regularity of language - then there are two strong reasons to believe only the faster mechanism:

- The faster process gets there first; m-script evolution can make a language partially regular, while the required evolution of the human brain is still on the starting blocks.
- The faster process actually removes the selection pressure which would have driven the slower process. Since languages change to be partially regular, there is no selective advantage in a brain which prefers regular languages; a general-purpose brain can use regular language, but can also handle the irregularities which are inevitable, so always has a selective advantage.

M-script evolution, being the faster mechanism, is clearly preferred. We would need to show that m-script evolution is either incoherent or fails to fit the facts, to justify a belief in the alternative.

# 5. Comparisons With Observation

In this section I compare the theory with the evidence on language learning. These are summary comparisons which cannot do justice, in a small space, to the great wealth of evidence on child language learning which has been gathered in recent years. However, they show that the theory agrees well with the main known facts - in most cases, in a natural and unforced manner.

I have compared the theory with about 101 empirical facts about language learning, under the following headings:

A. Key Facts of Language Learning

B. Early Word Learning

C. Phrase Structure Rules

D. Morphology

E. Complementation and Control

F. Auxiliaries

G. Verb Argument Structures

H. Pronouns, Gaps and Movement

I. Bilingualism and Language Change

Comparisons under these headings are numbered (A1), (A2), etc. I have classified 4 different possible outcomes when comparing the theory with the evidence:

1. **Distinctive Agreement**: (DA) where this theory agrees with the data, in a way not shared by most other theories, or a way which depends on a distinctive feature of this theory.
2. **Unforced Agreement**: (UA) where this theory accounts for the data in a direct manner, without additional assumptions; but other theories also agree, so the agreement is not distinctive.
3. **Agreement with Extra Assumptions**: (AA) where the theory does not naturally predict the finding, and it is necessary to make extra assumptions to account for these data
4. **Contrary Evidence**: (CE) Where the data go against a direct prediction of the theory, or the extra assumptions needed to account for the data seem highly *ad hoc* or implausible.

The comparisons which will be made are summarised in the lists below:

**GENERAL PROPERTIES OF LANGUAGE**

Languages are highly expressive

Languages are very diverse in structure
Languages are partially regular, but all have irregularities
Diverse languages are stable over time
Languages are learnt rapidly
Language learning starts slowly, then accelerates
Word segmentation is necessary for language learning
Language learning is very robust
We learn only structure-dependent rules
Language learning is lexically-based and conservative
Comprehension precedes production
Children make many types of transient errors, and correct them all
Languages change continually through intermediate forms
There are no major differences between the acquisition of sign language and of spoken language

**LEARNING WORD MEANINGS**

Word meanings are very diverse and rich in structure
Words are matched only intermittently with their meanings
The learner may observe many things which are not part of a word's meaning
Each word meaning is acquired from limited evidence
Children tend to link words to whole objects, and to types rather than thematically-related objects
The tendency to label whole objects and types is a bias, not a constraint
Some early words are very context-specific
Word meanings change gradually through intermediate forms
Children's over-extensions of word meanings may have a prototype structure
Some word meanings have a prototype structure
The set of meanings grammatically encoded in any language is quite limited
Children separate mutually less relevant elements of meaning into distinct words
True synonyms are very rare
Children apply a uniqueness bias in learning new words
Nouns predominate in the first 100 words learnt
Verbs and adjectives are learnt more rapidly after the first nouns are learnt
Learning closed-class morphemes accelerates at 400-600 words vocabulary
Early meanings tend to be over-specialised rather than over-generalised
Children over-extend some words in production (less so in understanding)
Names for basic-level categories are learnt first
Word meanings change by metaphor and metonymy
Children confuse names for parts of arms and legs
NP-type nouns describe social routines
In disambiguating homonyms, we favour common word senses
Gender has little to do with sex

**PHRASE STRUCTURE**

The syntax of any part of speech can be represented and learnt
Early syntax centres on verbs
Children link arguments to verbs correctly from the start of verb learning
Syntactic constraints also guide verb learning from an early age
There are no sharp, language-wide transitions observed in language learning
There is no dissociation between syntax and vocabulary size

Early analysed noun vocabulary predicts later syntactic ability; rote production does not
There is scant evidence for unmarked parameter values in language learning
There are no major blind alleys in language learning
Languages have regularities captured in X-bar syntax
In assigning agent roles, cue strength depends on overall cue validity
In many languages, cue strengths change dramatically between ages 6 and 16
Meaning elements encoded locally in the sentence are learnt most easily
Children tend to mark individual meaning elements explicitly and separately
Negation, interrogatives and conditionals are moved outside clause boundaries
Double-marking of negation is a later transient error

**MORPHOLOGY**

Individual word morphology is learnt before productive morphology rules
Productive regular inflections can be learnt even if regular forms are not in a majority
We learn which dimensions of meaning are encoded by inflectional morphemes
The speed of learning of inflectional morphology varies between languages
Productive inflections are learnt faster in agglutinating languages than synthetic languages
In agglutinating languages, inflections are learnt from the outside inwards
In agglutinating languages, children make no errors in ordering affixes
Irregular forms are initially learnt correctly
There is transient over-regularisation of irregular forms
English noun plurals and past tense verbs are over-regularised with low frequency
Specific Language Impairment affects regular morphology
High use of closed-class morphemes at 20 months leads to low use at 28 months
Ergative and accusative case markers are initially under-extended

**COMPLEMENTATION AND CONTROL**

Children acquire some complement-taking verbs early
Errors of control in complement-taking verbs are very rare
`Tough-movement' complements are acquired more slowly
There are mistakes in inflection of embedded verbs
Verbs acquired with missing complementisers are slow to acquire them
Verbs with optional complementisers are correctly learned
The `wanna' contraction is not made over a gap
The rare `promise' control structure is learnt more slowly

**AUXILIARIES**

Highly irregular English auxiliaries are learnt reliably
Over-generalisation of auxiliaries does not occur
Errors of Auxiliary control almost never occur
Children Often Fail to Invert Subjects and auxiliaries in Wh-questions
Complement verbs are sometimes overtensed

**ALTERNATING VERB ARGUMENT STRUCTURES**

Alternate argument structures for the same verb are learnt early and without confusion
Alternations of argument structure are in broad classes, yet respect narrow-range rules
Children use the alternations productively
Over-generalisations which violate the narrow rules occur, but are corrected

There are `indiosyncratic' non-alternators, which children learn
Children learn passives of `action' verbs before others

**LONG-RANGE MOVEMENT PHENOMENA**

Anaphors and pronouns have complementary binding domains
Reflexives are used correctly before pronouns
Pronoun reference principles have irregular edges
Some constraints on long-range movement are known from an early age
Ross' Island Constraints are obeyed from an early age

**BILINGUALISM AND LANGUAGE CHANGE**

Two or three languages can be learnt simultaneously
Children learn overlapping vocabularies for two languages
There is no evidence for a single grammatical system early in learning two languages
The course of bilingual language learning is very similar to the course of monolingual learning
Code-switching is done most frequently with nouns
Neighboring languages do not completely intermix
Creoles form very rapidly from Pidgins
Creoles use simple analytic forms to express meanings
Tense, Mood and Aspect appear in order TMA for creoles, MTA for most languages

In the comparisons which follow, I shall refer to sections of the paper which describe the relevant part of the theory, by giving a section number in square brackets, as e.g. [3.4]. I shall make brief comparisons in passing with some other theories of language acquisition; but the main comparison with other theories is deferred to section 6.

# 5.1 Key Facts of Language Learning

I first survey some of the broad, well known facts about language and language learning, and describe their interpretation in the theory. While these key facts are all well known, they are nonetheless remarkable and stand in need of explanation; many theories of language and language learning have difficulty accounting for them.

**(A1) Languages are highly expressive**: This is perhaps the central mystery of language learning - that languages can express an infinite range of complex meanings, using only finite resources; and that as children, somehow we all learn how to do this. That is what makes language learning seem so far beyond any form of animal learning, and has made it a central challenge for cognitive science.

This theory provides a working computational answer to that challenge. An unbounded set of language meanings are represented by tree-like feature structures (scripts) [2.1]. Each word meaning is a script, and language combines these word meanings by function application. Every word is a script function (m-script) which combines its own intrinsic meaning (its right branch script) with the meanings of its arguments [2.3]. This process can build up an unbounded set of meanings, or generate sentences from them - as has been shown by a working computer program [2.6]. We use a robust, general method of learning the word m-scripts [3.8] to learn this unbounded language capability. All this has been demonstrated in a working computer program.

**(A2) Languages are very diverse in structure**: (Shopen 1985) documents the great syntactic and expressive diversity of the world's languages. I have not found any forms in that survey which are not expressible in word m-scripts - which suggests that word m-scripts can capture the syntax of any language. This conclusion is supported by the correspondence between the m-script formalism and Lexical Functional Grammars [2.4] - which have been successfully applied to a very wide range of languages (e.g Austin & Dalrymple 1995).

Since there is a robust, general method to learn any word m-script [3.2 - 3.13], this gives us a way to acquire the syntax of any language - in spite of their huge syntactic diversity.

**(A3) Languages are partially regular, but all have irregularities**: Data in (Shopen 1985) and other surveys confirm that no language is completely regular, and equally, no language is completely irregular. The regularities have been the source of elegant syntactic theories, to which the irregularities have been annoying exceptions.

In this theory, the syntax of any language is embodied in the m-scripts of its words, which can embody complete regularity, complete irregularity, or anything in between. By m-script learning we can acquire an arbitrary set of word m-scripts, anywhere along this spectrum [3.8]. So irregularity is not a problem for the theory. Nor is regularity; the partial regularity in languages can be understood, as arising from historic changes in their population of word m-scripts, rather than from any structure in the human brain [4.3-4.7].

**(A4) Languages are stable over time**: The remarkable fact is that such a diverse range of languages are all individually stable over hundreds or thousands of years (e.g. Renfrew 1994). It might not be so; many conceivable learning mechanisms might, over the generations, `funnel' any language towards one of a few standard forms, apart from vocabulary variations.

In the m-script theory, this stability of language diversity follows from two facts:

1. The fidelity of the basic learning mechanism; speakers may use an arbitrary set of word m-scripts to compose sentences, and children will faithfully learn just those word m-scripts and no others [3.11].
2. The domains of regularity in a language provide a selective force which tends to prevent alien word m-scripts, which do not fit in with the regularities, from invading a language [4.4].

**(A5) Languages are learnt rapidly**: From children's' peak learning rates of several words per day (Ingram 1989), it seems that each word must be fully learnable (in its syntax and semantics) from just a few exposures. In this theory, each word m-script can be completely learnt from (of the order of) six clear examples of hearing the word in use [3.2]. From everyday experience and observations of young children, it seems clear that children are exposed to at least this number of learning examples.

Therefore the m-script theory can account, to within an order of magnitude, for the observed speed of language learning. Some other theories (such as simple neural nets, whose training times are measured in `epochs', or thousands of examples) clearly do not.

**(A6) Language learning starts slowly, then accelerates**: Many studies have shown early learning rates of only one word every few days, some 50 times less than the later peak rates of order 10 words/day. There is widespread evidence that many (but not all) children undergo a `vocabulary burst', or rapidly increased learning rate, when their productive vocabulary is in the range 50-100 words. However, the effect seems to be a steady acceleration rather than a takeoff point (Bates et al 1993).

Many different mechanisms have been proposed to account for this burst. In this theory there is one obvious factor which may account for some, if not all, of this rapid acceleration.

Initially, children have few clues to segment the sound stream and to guess what is being referred to; so their initial learning must involve many wrong guesses, picking a signal out from a high noise level [3.7]. The learning mechanism can do this, but slowly. Later on, children can partially understand many sentences, giving strong clues as to what sounds are new words, and what they refer to [3.9]. The partial understanding process gives powerful constraints on what an unknown word may refer to; it is like having a pair of tweezers to pick up unknown meanings, in stead of clumsy fingers. This gives the child a much cleaner learning signal; so learning is expected to accelerate rapidly.

On this account, we would expect the number of clean learning examples per day to increase linearly (or perhaps even more rapidly) with vocabulary size V. As learning rate is proportional to the number of learning examples, this gives dV/dt = lV, which implies an exponential vocabulary growth, approximately as is observed (Van Geert 1991; Bates and Carnevale 1993).

**(A7) Word Segmentation is Required for Language Learning** : Most theories of language acquisition need to assume a separate mechanism for word segmentation, and these segmentation theories are not yet very satisfactory. Few of them address the problems of segmentation for sentences containing novel words. Some theories depend on particular prosodic cues or styles of caregiver speech, and these cues and patterns are not cross-culturally universal.

In this theory, provided the child can segment the sound stream into some kind of sub-word units such as phonemes, their grouping into words is learnt directly by the m-intersection learning mechanism, as described in [3.7, 3.8]. Just as m-intersection projects out the script meaning of a word (in the right branch of its m-script) from a large amount of extra meaning in learning examples, so it projects out the sound of the word (in the left branch of its m-script) from a large amount of surrounding sound in learning examples. Whatever is not common to all learning examples (meaning or sounds) is efficiently pruned out by the m-intersection mechanism. A separate word segmentation mechanism is not required.

An attractive feature of this account is that it does not depend on any special properties of the sound stream, and so can account for word segmentation in the acquisition of sign language just as well as in spoken language, by the same m-intersection mechanism.

**(A8) Language learning is very robust**: Children learn the correct meaning and syntax of words from a few examples which (a) may be interspersed between many noisy or misinterpreted cases and (b) each have a large amount of irrelevant information present; and (c) they can learn without any explicit prompting or instruction.

The social learning mechanism, which is the basis of language learning, evolved to do a very similar task - to learn social regularities robustly from examples sparsely spread amongst noise [4.1], with no explicit instruction.

The Bayesian learning mechanism evolved to meet this need. It can learn a rule or m-script which applies to only 5% of all qualifying cases [3.2-3.4] from around 6 good learning examples, where each example has a large amount of extraneous information. We can show mathematically that the Bayesian learning mechanism is capable of this performance [3.4]. Thus children's' learning examples may be heavily loaded with other information, interspersed with a large number of false examples, and they will still successfully learn word m-scripts.

This robustness applies not just to the learning of individual word meanings; since the syntax of a language is embodied in word m-scripts, the learning of syntax is equally robust. Since syntax is distributed across many word m-scripts, its learning is more robust than if it were embodied in a few parameters.

**(A9) We learn structure-dependent rules**: Children do not seem to make errors which would follow from learning a surface-order dependent rule in the place of a structure-dependent rule (Chomsky 1991); for instance, it has been shown that when children ask questions, they use forms which are a structure-dependent (rather than surface-order-dependent) modification of the indicative forms (Crain 1991).

In the m-script theory, both statements and questions are generated directly from script meaning structures [2.5]. The meaning script for a question is a structure-dependent alteration of the indicative meaning structure (it can only be structure-dependent, because meaning scripts are structures; they have no surface order [2.1]). Therefore we can only form questions in structure-dependent ways.

More generally, the way we generate language is fundamentally structure-dependent [2.5], and learning any word is centrally dependent on meaning structures [3.7]. The whole theory is built on script meaning structures rather than surface order; it would be extremely difficult, in this theory, to force some surface-order dependent error.

**(A10) Language Learning is Lexically-Based and Conservative**: For many years, linguistic theory has been dominated by the study of syntax and its language-wide approximate regularities, and some underlying cross-language universals. Child language research has been led by an expectation that the prime learning task is to acquire these language-wide syntactic regularities (implicitly, as fast as possible) perhaps in the process revealing cross-language universals.

These expectations have not, in broad terms, been met. In stead, the empirical picture of child language learning has repeatedly been one of lexically-based learning - grammatical structures being learnt in close association with individual words, with only cautious, conservative generalisations beyond this . For instance Maratsos (1983), in summarising the evidence, concluded:

*A recurring finding of the last years is that children often make highly specific analyses of combinations, and apply possible generalisations cautiously, rather than rapidly making highly general ones which are productively extended immediately.*

Similar points are made by Tomasello (1992) and Pinker (1996).

This broad finding fits well with the m-script learning theory in which the entire language is lexically based, the learning process centres on discovery of m-scripts for individual words [3.8], learning is mainly bottom-up, from narrow rules to broader generalisations [3.5], and where any further generalisations (secondary learning) must all pass a test of statistical significance [3.11]. Therefore the m-script theory predicts exactly the kind of lexically-based, conservative learning which children have repeatedly shown.

**(A11) Comprehension Precedes Production**: It is a robust cross-linguistic finding that vocabulary for comprehension grows well in advance of productive vocabulary. Between 12 ans 18 months, comprehension vocabulary exceeds productive vocabulary by typically a factor of 4 or more, with wide variations between children (Bates et al 1993).

This might seem to be an obvious fact, which must be true in any theory of language learning, but it is not so. If, for instance, language learning proceeded by doing - by trying out words to see if they work - it might be the case that production was closely matched with comprehension. Compared to many other skills - where we can only learn by doing - it is remarkable that language learning proceeds very effectively almost entirely by observation, with hardly any doing.

In the m-script theory, the learning mechanism is based on a mixture of comprehension and `silent' internal generation [3.9]; however, in the earliest stages (e.g. to learn nouns) comprehension dominates [3.7]. This means that, if children are under some selective pressure to learn language as fast as possible, it is comprehension which is under pressure to advance most rapidly. Although a word m-script, once fully learned, can be equally used for comprehension or production [2.3-2.6], we would expect the child at any stage to have a large number of word m-scripts partially learned - well enough to be used in comprehension (with contextual help) and to give valuable clues when learning other words - but not known with enough confidence for use in production.

The other obvious factor in explanation is that the child is exposed to (and so automatically learns) many words which she just has no interest in using for speech. This remains true into adulthood.

In summary, the lead of comprehension over production has no single neat explanation, but equally poses no problems for the m-script theory.

**(A12) Children make many types of transient errors, and correct them all** : Although the word `errors' may reflect an over-simplistic, normative view (Givon 1985), nevertheless there are many distinct ways in which children's' speech transiently differs from adult speech - many distinct `error types' which are universally eventually corrected.

We have reliable cross-linguistic evidence on an increasing number of these error types - including over-regularisation of morphology, over-extensions of word meanings, alternations of verb argument structure, and pronoun usage (see e.g. Slobin et al 1985). The ways in which these errors arise (while others do not) are diverse and provide many insights into the learning mechanism, which will be discussed individually in the sections which follow.

These errors are all, of course, corrected before adulthood. Learning theories have struggled to explain how each individual error type is corrected; in each case it has been a struggle, because explicit negative evidence is known to be absent or insufficient to support learning (Marcus 1993). However, the fact that they are *all* corrected (and we may find many more transient error types, also corrected) is itself a big experimental fact which calls for a simple explanation. We cannot carry on constructing piecemeal accounts of all error corrections.

In this theory, there is a straightforward mechanism for gathering and using negative evidence. When a child hears an adult sentence, and can infer the meaning, the normal learning process [3.9] is to partially understand the sentence, and also to partially generate a sentence from the inferred meaning. This gives the child the opportunity to observe "where I might have said X, an adult said Y" - a piece of negative evidence about construct X. The adult has supplied an implicit correction without intending to [3.10].

If the child gathers enough of this negative evidence about X (enough evidence to be statistically significant) she may conclude that in certain circumstances, construct X cannot be used [3.6]. This offers a general account of the correction of many distinct types of transient errors. Some of the specific instances are discussed below; but having an effective general mechanism for a very general phenomenon is an asset for the theory.

**(A13) Languages change slowly through intermediate forms**: While a few languages seem to have settled on highly regular and stable forms (e.g Turkish: Slobin 1982), the majority of the world's languages seem to be mixtures, or in a state of transition, or both (McMahon 1994).

This is problematic for some theories, where much of the structure of a language is carried in a few phrase structure rules or parameters. For these theories, it is hard to understand the intermediate forms of language which must persist for at least a few generations during the course of language change (eg between one parameter value and another). How do these languages work (in production or in understanding) and how do children manage to learn a language with mixed or intermediate-valued parameters? Theories which rely on a few regular discrete structures or parameter values cannot easily accommodate incremental changes.

In this theory, where the structure of the language is embodied in a large set of word m-scripts [2.3], a language in transition is defined simply by a mixed set of m-scripts - some for the initial state, and some for the final state. A mixed language contains m-scripts from both parent languages. There are then no difficulties either about how the language works either in production [2.5]or in comprehension [2.4] (since these processes can use an arbitrary mixture of word m-scripts), or in acquisition (since this process can reliably acquire an arbitrary set of word m-scripts [3.12] ). The m-script theory is entirely compatible with mixed and transitional languages.

**(A14) There are no major differences between the acquisition of sign language and of spoken language**: Sign language has all the syntactic complexity and expressive power of spoken language, and is acquired in a very similar manner. For instance, as summarised by Pettito (1993*) "deaf children acquiring signed languages from birth and hearing children acquiring spoken languages from birth achieve all linguistic milestones on an identical time course."*

In this theory, language learning is based on a general mechanism of social intelligence, which evolved to learn causal regularities about sounds, gestures and data of any sensory modality [4.1]. So it is no surprise that language learning can be coupled to acoustic and visual channels with equal ease. Theories which postulate a more recent, language-specific origin for the learning mechanism might have difficulty with this fact.

The m-script learning mechanism gives a practical answer to the `poverty of the stimulus' argument which has motivated much linguistic theory. If you have a robust Bayesian learning mechanism, evolved over millions of years to learn complex social regularities in a noisy social milieu, then there is no lack of stimulus to learn language:

- At the first stage of learning, the six or so examples needed to learn each noun may easily appear over a period of several days or weeks; many misleading non-examples in between will not prevent learning.
- At the stage when children are learning ten words per day, it is quite possible for them to hear sixty sentences per day in circumstances where they can work out, from linguistic and non-linguistic clues, what an unknown word refers to. The learning examples for one word may be spread over an extended period.

The Bayesian learning mechanism is provably capable of this performance, and it is now clear that children of all cultures get adequate inputs. They hear many hundreds of words per day, in situations (such as well-rehearsed family routines) where they know what is going on, and what is being referred to. What seemed like poverty of the stimulus was in fact poverty of learning mechanisms.

Points (A1) - (A14) are some of the most salient and remarkable facts about language learning - facts which have made it one of the major challenges for cognitive science. The m-script learning theory seems to give a natural and unforced account of all of them.

# 5.2 Learning Word Meanings

The acquisition of word meaning has been called *`perhaps the deepest mystery in the study of language acquisition'* (Bloom 1993) because, while syntax and phonology might be argued to be closed or parameter-driven systems, word meanings span an open set of concepts - possibly the whole of human cognition.

Points (B1) - (B4) list some of the key reasons why this has seemed such an deep mystery, and describe briefly how the m-script theory can gives an un-mysterious, practical answer to each of them.

**(B1) Word meanings are very diverse and rich in structure**: The size of any dictionary reminds us of the great diversity of word meanings, while introspection confirms that many word meanings are highly structured things - which could not, for instance, be captured as points in any small-dimensioned space. The set of possible word meanings is systematic and productive (singular things can be made plural; things that can be done may be undone or re-done; and so on). The meaning of a word may refer to almost any situation which we can conceptualise, and to sense data of any sensory modality. Somehow we can internally represent these diverse, richly structured word meanings, and learn each one of them from only a few examples. This is a major challenge for theories of language.

In the m-script theory, word meanings are scripts [2.1], symbolic structures which derive from primate social intelligence [4.1]. Primate social situations are diverse, highly structured, multi-sensory and systematic; so the script representation has evolved have these properties. Being tree structures of unlimited depth and breadth, scripts can express a systematic set of meanings of arbitrary complexity - just as is required for word meanings.

Although the basic script representation can capture much of the complexity of word meanings, scripts are not an isolated representation in the brain. For social cognition, they must couple to other internal representations, such as mental images, or representations of procedural and physical skills (Worden 1996b). These links to other representations extend the representational power of scripts. For instance, the word for a physical skill may be represented by a script which links to a mental image, or to a dynamic encoding of the skilled movements.

Each time a word is heard in a sentence, its meaning is embedded in the sentence meaning [2.3]. If the child can infer this meaning non-linguistically, for a few learning examples of each word, the operation of script intersection will rapidly project out the word meaning script from example-specific extra meaning [3.7] -learning the meaning of the word.

Therefore the script representation has sufficient expressive power to represent the meanings of words, and script intersection can rapidly learn word meanings from examples.

**(B2) Words are matched only intermittently with their meanings**: In the learning input of the young child, there is only (in Gleitman's (1990) phrase) a *`fitful fit of word to world'*. Studies of corpora of utterances heard by children show that for typical early verbs, the verb's action is visible, or the things which are the verb's arguments are in view, on less than 50% of the occasions when the verb is spoken (Beckwith, Tinker and Bloom 1989). It has been found (Golinkoff 1986) that mothers immediately understand their children's' wishes on only about 50% of occasions - so how could the children do any better ? With so much dilution of the learning signal, how can a learner find out the true (and possibly complex) meaning of a word ?

The analysis of the Bayesian learning algorithm in [3.2] shows that a signal:noise ratio of 50% can be handled easily; the mechanism can still pick out a sound/meaning regularity even when it only applies 5% of the time. The key criterion is the number of positive learning examples, irrespective of how many false examples there are to dilute them. Therefore the m-script learning method is highly robust against dilution of the learning signal.

**(B3) The learner may observe many different things which might be part of a word's meaning**: Again to quote Gleitman (1990) , it seems that *`an observer who notices everything can learn nothing, for there is no end of categories known and constructable to describe a situation'*. This problem has also been famously elaborated by Quine (1960).

In the m-script theory, the answer to this problem comes in two parts: first, our primate social intelligence pre-disposes us to construe the situations around us in a limited number of ways, expressible as scripts (see (B5) below) [4.1] ; and second, the script intersection mechanism [2.1, 3.7] is a highly efficient way to project out the common meaning from two or more scripts, rejecting noise.

To illustrate its efficiency, suppose the child observes 200 bits of information in a typical situation, and that some word has a meaning with information content of 40 bits. Each learning example for this word has a superfluous `noise' information content of 160 bits, almost drowning the signal; but intersecting two such example scripts together prunes the tree, and removes at least 3/4 the superfluous `noise' information (sparing only the slots where the two examples coincide accidentally), leaving the noise at less than 40 bits. As the learning examples are uncorrelated, intersecting three examples leaves under 10 bits of coincidental noise; four leaves only about 2 bits of noise, and after that there is probably only the pure word meaning script left. Script intersection rapidly projects out the true word meaning from large amounts of noise.

**(B4) Each word meaning is acquired from limited evidence** : Word meanings of unbounded complexity are acquired from a small number of examples per word, at peak learning rates often over ten words per day (Ingram 1989). At this speed there is not time to get very many learning examples per word.

The analysis of the script and m-script learning algorithms showed how any meaning script can be acquired reliably from around six (i.e 3-10) learning examples. This is enough to establish that the apparent word meaning is not just coincidence between the examples, and to remove any other meaning coincidences between examples. It seems reasonable that the child can find 3-10 learning examples per word, at a peak rate up to 10 words/day. This learning mechanism has a likely biological origin [4.1], and can be shown mathematically to give adequate performance and noise rejection [3.3-3.8].

**(B5) Children tend to link words to whole objects, and to types rather than thematically-related objects**: Markman (1990) has shown that children have a learning bias to associate word meanings with whole objects rather than their parts, and to extend word meanings taxonomically (rather than thematically). They do not show the same biases for non-linguistic tasks. For learning language, however, children have some effective solution to Quine's (1960) `gavagai' problem.

If words are learnt by a mechanism which evolved for social learning [4.1], we expect biases in word learning to reflect the needs of the social domain - which is mainly concerned with whole objects rather than parts, and involves taxonomies (eg types of food). Thus we expect a learning bias towards whole objects, towards labels for individuals (precocious learning of proper names - Bloom 1990), and to and taxonomies or types.

In practice, the needs of the social domain have shaped the script representation, which then shapes language learning. The biases are reflected in the architecture of scripts. The most important properties (for social reasoning) are expected to be near the roots of script meaning trees, so that social learning (by script intersection) preserves them best.

For instance, we would expect whole objects to be represented by nodes near the roots of script trees, and their component parts by subordinate nodes below each whole-object node. Then the learning mechanism of script intersection [2.1, 3.3] will preferentially preserve whole-object nodes, giving a learning bias towards whole objects - a bias which originates in social cognition, but is reflected in language learning.

**(B6) The tendency to label whole objects and types is a bias, not a constraint**: As argued by Nelson (1988, 1990), children do not always make the taxonomic choice, and they should not (and do not) always choose whole-object meanings. However, the social learning process is intrinsically probabilistic, not categorical; so its learning biases are just biases (in Bayesian priors, reflecting the social environment) rather than constraints [3.1] ; they can be overcome by sufficient evidence.

The learning mechanism normally requires of order 6 examples to learn any word meaning script, simple or complex. This is because the prior probability of a simple meaning script is higher than that for a complex one; but a complex meaning script accounts for more data, so can gather confirmation from the learning examples faster [3.2], catching up after 6 examples. However, if a child is called upon to guess a meaning on the basis of an insufficient number of examples (e.g. one example, as in some of Markman's experiments), the prior probability bias to simple meaning scripts will win - causing the child to prefer the simpler whole-object hypothesis.

**(B7) Some early words are very context-specific**: Nelson (1985) and others have noted how some of a child's very first words are formulae which seem to denote some whole routine (or event description) rather than any adult word meaning.

The script intersection learning mechanism approaches the true meaning script for a word from above, rather than from below; early approximations have more nodes and slots that the true meaning, and successive script intersections remove these extra nodes and slots [3.3,3.4].

If, then, adults tend to use some words in stereotyped and highly context-specific ways, script intersection will learn the words with those narrow, context-specific meanings; this is what we expect from a bottom-up learning process [3.6]. The child will tend to learn and use context-specific rote formulae rather than true words. Only later can he broaden the meaning by further script intersections, learning words by a hybrid of the primary and secondary learning mechanisms [3.11].

Scripts are very closely related to Nelson's event descriptions [2.2, 4.1], and may well be the same thing; so the script learning mechanism, which learns early words, is expected to learn some words as complex event descriptions [3.3, 3.7].

In Nelson's evidence, context-specific, formulaic words tend to be used more by `expressive' rather than `referential' children. In this analysis, the expressive/analytic distinction may depend more on parental input than on the innate dispositions of the child.

**(B8) Word meanings change gradually through intermediate forms**: It is a well known fact of language change that the meaning of a word may drift over hundreds of years, to become completely different from its original meaning. The incremental changes of whole languages were explained, in the m-script theory, as changes in the population of word m-scripts which constitute a language (A14), but the slow change in meaning of an individual word seems to pose a bigger challenge for the theory. If each word meaning is a script - a discrete tree structure with discrete-valued slots on its nodes - how can one script change gradually over time ?

The answer to this puzzle lies in our ability to re-construe the same external situation in different, but related ways [4.2]. At any time, a word's meaning may be a changing mixture of construals.

For instance, consider Bybee's (1995) analysis of the changing meanings of English auxiliary verbs. She notes that the auxiliaries *should*, *would*, and *could* were all originally the past forms of *shall*, *will* and *can* - but that now their meanings have departed considerably from those past forms, in each case accruing a much more conditional/hypothetical reading. *I will* means I definitely will; whereas *I would* means I would only under some circumscribed conditions. Bybee traces this change through texts in Old English, Middle English and Shakespearean English.

Suppose the meaning script of *will* embodies a definite intention to do. When used in the past tense, *willed* (at first) also embodies this definite intention; but there is an implication (as the speaker says *X willed* and does not say *X did* ) that maybe X did not actually carry out his intention, because it was conditional on something which did not happen; this second construal is represented by a distinct script, and listeners may learn this implication generally as a non-linguistic m-script [1] which transforms between the two construals [4.2].

Thus in understanding a sentence, the original `pure intention' meaning of *willed* can be followed up by the use of a non-linguistic re-construal m-script which generates the second meaning `intention maybe not carried out'. This second meaning may then be learnt as a kind of homonym for *willed*.

As the word becomes a mixture of two closely related meanings, the relative weights of the two senses (their learned rule probabilities [3.2]) may change over time, until the second form dominates (and meanwhile the pronunciation changes to *would*). Throughout all this change, speakers and listeners can use the re-construal m-script to understand one another.

So any word in a language may change meanings between two or more `pure script' forms by being a mixture along the way. The rule probabilities of the word m-scripts, learnt by children of each generation, measure the progress of the change over time.

**(B9) Children's over-extensions of word meanings may have a prototype structure:** An example comes from Bowerman (1978) who reported that Eva (at around 18 months) had a prototype meaning of *kick* with three main characteristics: (a) waving a limb (b) sudden sharp contact between the body and an object, and (c) propulsion forward of the object. In various extensions, she used the word *kick* when only one or two of these applied - a kitten with a ball near its paw, a moth fluttering on a table, pushing an object against her sister's chest. Some extensions had nothing in common with others, but all overlapped with the prototype.

In the m-script theory, prototype-like over-extension arises as a result of production with limited vocabulary. Language production is always a best-fit process, making use of the words you know to say something useful - even if you have to include things which are not part of your intended meaning [2.5]. For a child with limited vocabulary, this is a hard problem; so if you have any word which overlaps strongly with what you want to say, you may use it, even if parts of its meaning do not apply. Thus the word *kick* conveyed some useful meaning in all of its over-extended uses - which, as a group, have a prototype structure.

**(B10) Word meanings may have a prototype structure**: The meanings of some adult words are well described by such a prototype structure, in which all senses share some properties with a prototype (or central sense), but where some peripheral senses have nothing in common with other peripheral senses. A typical case is the word *game*, first discussed as a prototype structure by Wittgenstein.

The production mechanism described above explains how this arises, when we try to express new meanings with our limited vocabulary. A word m-script describes a central sense, but the word may be used when only parts of this central meaning script apply. This extension of word usage may be done by speakers of any age. If one of these extended uses is commonly made, then listeners will start to learn it; the word then becomes polysemous, with several distinct meaning scripts. If they all originate from some narrower `prototype' meaning script, each taking on different parts of the original meaning script, then the set of meanings will have a prototype structure.

If you hear six good learning examples of a word with broader-than-usual meaning, you will learn a new m-script for that meaning [3.4]; so the speed of the learning mechanism enables words to rapidly become polysemous with a prototype structure.

**(B11) The set of meanings which are grammatically encoded in any language is quite limited**: This clear cross-linguistic regularity has been described by Talmy (1983) :

*[Grammatical forms] represent only certain categories , such as space, time (hence, also form, location and motion), perspective-point, distribution of attention point, force, causation, knowledge state, reality status, and the current speech event, to name some main ones. And, importantly, they are not free to express just anything within these conceptual domains, but are limited to quite particular aspects and combinations of aspects, ones that can be thought to constitute the "structure" of these domains.*

If you had to form a list of properties to abstract from general cognition for use in social reasoning (about peers and the interactions with them for food, space, alliances, dominance, and so on) then that list would look much like Talmy's list. In this theory, those meanings which are grammatically encoded (e.g. in closed-class morphemes) consist of small atomic elements of the script meaning representation, which can be easily added to or taken from any script. These elements are the basic coinage of social interaction [4.1]; and they are worth encoding compactly for efficient communication.

Different languages may encode different subsets of this limited set of encodable notions (Bowerman 1985). This does not constitute evidence against the existence of a limited set. For communicative efficiency, the words of a language should not try to encode too much (otherwise the meaning of each word would be too narrow; too many words will be required), so the language must be selective. These selective choices of what to encode are often language-wide choices; this happens because word m-scripts evolve to make large domains of regularity in a language [4.4].

**(B12) Children separate mutually less relevant elements of meaning into distinct words**: Just as languages differ in what they encode grammatically, so they differ in what they group together in open-class word meanings. Bybee (1985) has introduced the notion of the mutual `relevance' of two meaning elements, noting that languages tend to encode mutually relevant elements of meaning in the same word; but where languages do not do this, children's' production errors often tend to group them back together (Slobin 1985). (This finding is closely related to the finding (C14) below).

An example is Talmy's (1985) observation that while verbs of motion may encode motion, direction, form of the moving object, and manner of motion, in any one language the verbs only encode only two of these elements. In English, verbs encode motion and manner. Some American languages encode motion and form of the moving object. Spanish encodes motion and direction; manner must be encoded separately by an adverb. For `roll down', in Spanish one must say `descend rollingly'. Spanish children, however, often group meaning according to the English analysis as in **correr abajo* `run down' (Slobin 1985).

Suppose verbs of motion do not encode all elements of motion, manner, direction and shape, because doing so would make verbs too specific; too many different verbs would be required. The choice of which meaning elements are encoded in the verb is a language-wide choice, because of the weak forces for regularity in languages; once a few main motion verbs encode some set of meaning elements, other motion verbs are forced to `line up' with them in a domain of regularity [4.4]. (This also stabilises and perpetuates the differences between languages.)

We can interpret Bybee's Relevance as proximity in script meaning structures. If two elements of meaning are slots on the same node, they are maximally mutually relevant; if they are on neighbouring nodes, less so, and so on [2.1]. The advantage of encoding mutually relevant meaning elements in the same word is not one of learnability (the m-intersection learning method can project out a large script structure spanning the mutually irrelevant meaning elements [3.3]), but of ease of language generation. Generation carves a meaning script apart into pre-defined building blocks [2.5], and is more easily done `piece by piece' using small local building blocks, than by carving out large, sparse, overlapping component meaning scripts. So languages generally evolve [4.3] to have compact building blocks, which encode mutually relevant meanings, making generation easier.

Of the elements commonly encoded in verbs of motion, manner and motion have greatest mutual relevance - being both intrinsic to the verb meaning script, while other elements (the shape of the moving object and the direction of movement) are more separable in the script meaning and so less relevant to the motion. However, some languages, such as Spanish, somehow started to encode motion and direction in the verb (Talmy 1985), and this choice then became frozen into the language by the `domains of regularity' mechanism [4.4].

This faces children not with a learning problem, but with a generation problem; Spanish verbs of motion do not encode the local, mutually relevant parts of the meaning structure which are easiest to pick apart in individual words. Any bias to encode small parts of the meaning structure in individual words will tend to produce English-like forms such as **correr abado*.

As another example, children easily acquire verb affixes encoding tense/aspect and person/number, but have more difficulty learning to mark verbs for gender or definiteness of the direct object (Slobin 1985). Again, I interpret this as a bias in language generation, rather than in learning.

A closely related notion is Bowerman's (1985) `hierarchy of accessibility' of meaning elements which, she proposes, affects (a) what children learn easily, and (b) what most of the world's languages tend to encode grammatically. Accessibility might depend on some script-related notion (e.g. proximity to the root node) or may be influenced by many other factors (eg perceptual salience). Bybee's relevance is a kind of relative accessibility; in practice the two notions may be very hard to distinguish, and partially interchangeable.

**(B13) True synonyms are very rare**: This fact, which aids language learning, is sometimes thought to be an innate property of the human language faculty - proposed as a uniqueness principle (Pinker 1984) or principle of contrast (Clark 1987) which children impose on the meanings they learn.

This theory is able to have its cake and eat it on the uniqueness issue. On the one hand, the observed lack of synonyms can be accounted for by other means, without a uniqueness principle; and on the other, there are independent grounds in the theory to expect some kind of uniqueness principle, as observed in experiments.

The observed lack of synonyms can be accounted for without a uniqueness principle, as follows:

1. Lack of synonyms emerges naturally from language change: Having two words with the same meaning is unnecessary, and will be wiped out by language change; one or other word will soon disappear. In terms of the `m-script evolution' picture [4.3] it has a close biological analogue - the fact that two species never occupy exactly the same ecological niche (word = species; niche = portion of meaning space).
2. In theories of language learning, the uniqueness principle is often motivated by a supposed lack of negative evidence; without negative evidence, how can children `unlearn' overlapping meanings of words (Pinker 1984) ? In this theory, negative evidence is readily available [3.6, 3.9], so complex patterns of overlapping meaning - where one word `occludes' some part of the initially-learnt meaning of another - can be learnt.

However, although a uniqueness principle is not required by language data, there is independent evidence for it.

**(B14) Children apply a uniqueness bias in learning new words**: Markman (1990, 1992) finds experimental evidence that 3- and 4-year olds use a uniqueness bias when faced with new words, tending to believe that some new word must have a meaning distinct from a word they already know.

While the learning mechanism described in [3.3] -[3.9] seems to have no intrinsic bias against learning two words with identical meanings, the bias may be applied when the child comes to store the m-script for a newly-learnt word. For the purpose of language generation, word m-scripts must be stored in some structure which indexes them by meaning, so that a speaker may find the word with the best meaning at every moment. The particular form of storage proposed in this theory is an inclusion graph, so that by descending the graph we find meanings successively closer to the desired meaning [2.8].

Whatever meaning-indexed storage structure is adopted, it will encounter a difficulty when trying to store two words of exactly the same meaning, as they will try to occupy the same place in this structure. We may suppose the structure is not designed to support this, as it serves no useful communicative purpose; there is no point in having two equally good words to say one thing. Therefore a storage difficulty will force the child to try to find a distinct meaning for a distinct word.

(This raises the question of how bilinguals learn words in two languages with the same meaning. It is likely that they learn very early a socially-conditioned `language context flag' which serves to distinguish the lexical entries for their two languages, and allows them to have distinct storage locations)

**(B15) Nouns predominate in the first 100 words**: (Bates et al 1995) A noun m-script can be learnt in the absence of other linguistic knowledge, by inferring what entity is being referred to through non-linguistic means [3.1]. Full verb m-scripts cannot be learnt without knowing some nouns [3.1], and so we broadly expect noun learning to precede verb learning.

Although the bare meaning of a verb might be learnt in the same way as a noun (requiring no other vocabulary), in practice verb senses are not so easy to pick out. Adult speech tends not to refer to a present action in the same way as it refers to a present thing (verbs typically refer to a near-future, desired or near-past action). A knowledge of nouns is therefore needed to give clues about which event (past, future or present) a verb refers to.

**(B16) Verbs and Adjectives are learnt more rapidly after the first nouns are learnt**: (Bates et al 1995) To learn a full m-script for a verb or adjective, you need to know some nouns [3.8]. Knowledge of entity words also helps the child work out what action (near-future or near-past) is being referred to; so we would expect an acceleration of verb learning after, say, the first 50 nouns are known.

**(B17) Learning closed-class morphemes accelerates at 400-600 words vocabulary**: (Bates et al 1995) To learn closed-class morphemes, you cannot use the raw data of adult sentences and their inferred meanings [3.9]; it requires a secondary learning process, which depends on knowing some number of word m-scripts already [3.11]. So we expect closed-class morphemes to be learnt only when the child has a significant vocabulary.

**(B18) Early meanings tend to be over-specialised rather than over-generalised** : While it was once believed that over-extension of word meanings was common (Clark 1973), more recent evidence (Huttenlocher et al 1987) suggests that over-extension is much rarer than was thought; and that when it does occur, it is more likely to be fairly late in the developmental history of a word. In the early use of any word, underextension is more likely than over-extension (Barrett 1995; Dromi 1987).

The script intersection learning mechanism starts from rich meaning structures of observed situations, and prunes them by comparison with one another - rather than building up a script from meaning elements [3.3]. It approaches the true meaning of a word from above, rather than from below. So we would broadly expect early word meanings to be under-extended (from intersecting a few examples with coincidental similarities, or learning highly context-specific words as in B5) rather than over-extended.

**(B19) Children over-extend some words in production after having used them correctly:** Children are observed to use some words with over-broad meanings.

Usually, this arises by the mechanism discussed under (B9) above; under pressure to communicate with limited vocabulary, children use words when only parts of their learnt meaning scripts apply. However, there are interesting examples of late over-extension, observed by Bowerman (1985), whose explanation may be a little more complex.

Bowerman (1985) noted an interesting form of over-extension between pairs of words with closely related meanings, such as make/let (as in *make me watch TV* versus *let me watch TV* ) and give/put (as in *give the plate onto the table*). In these cases, the two meaning scripts may be quite complex and have a lot of structure in common. What is interesting is that these errors emerge only after each word has been used correctly for some time.

There is a possible interpretation in this theory - that, like many other over-extensions, these arise from the secondary learning process. M-scripts for *make* and *let* are initially learnt and used correctly. Then, at some later stage, the child m-intersects these together (looking for some broader generalisation, and forming a higher node on the inclusion graph structure which stores words for fast retrieval in generation [2.8]).

Before this m-intersection is formed, there is no easy way to make the error; but once it is formed, when producing a sentence the child navigates down the inclusion graph, looking for meanings as close as possible to her intended meaning. If the make/let distinction is not an important part of the meaning for the child, she may take a `wrong turn' at this stage, pragmatically adding a small element of extra meaning in order to find a word. Only later does she learn that this element of extra meaning is important in the socially regulated world of adults, and has to be got right.

**(B20) Names for basic-level categories are learnt first**: Many nouns are names of categories. When children learn these category names, the first names learnt are those for `basic-level' categories (e.g *dog*) rather than superordinate (*mammal*) or subordinate (*dachshund*) (Brown 1958; Lakoff 1987).

The basic level is not just the middle level of some category hierarchy; it has important distinguishing characteristics. It is, for instance, the highest level at which we readily form a mental image of the category, and the highest level at which we have motor routines for interacting with members of the category (e.g *chair* versus *furniture*). It is the level of distinctive actions, both physical and social (Rosch 1978; Lakoff 1987).

In the m-script theory, the script meaning of a word is not just an isolated structure; it has links to other meaning structures in the brain, including other scripts for social processes and routines, mental images, and physical movement routines [4.1]. These links are an essential part of the meaning; for instance, *menu* has not much meaning without a link to the well-known restaurant script.

Our pre-linguistic meaning representations contain such links, defined at the broadest, most widely applicable category for which those links work well; this is the basic level. Therefore children tend to learn first the words for basic-level categories, rather than superordinate categories (for which the links don't work) or subordinate categories (whose names occur more rarely in the learning input).

**(B21) Word meanings change by metaphor and metonymy** : Many word meanings depend on metaphor or metonymy, and this has been identified as an important source of systematic meaning changes (Traugott). Is metaphor compatible with a script-based account of word meaning ?

The m-script theory has plenty of room for metaphor and metonymy (although the details have not yet been fully worked out). The reason is that word meaning scripts are strongly linked to other meaning representations in the brain [4.1], such as procedural scripts and spatial models.

These links must be quite diverse and flexible, invoking domain-specific linking procedures (e.g to convert between script and a spatial model). We may suppose that the preferred form of linking for a script is specified by a special slot on its root node. In Jackendoff's (1990) terminology, this slot defines a semantic field in which to interpret the script; call it the *semantic field slot*. By changing the value of the semantic field slot (e.g from `spatial path' to `time interval' or `possession') while keeping the rest of a script unchanged, we make metaphoric relations between different domains - social, spatial, temporal, etc.

This form of script-mediated metaphor may have been useful before language. For instance, linking social rank to vertical displacement enables some useful reasoning about ranks - implying, for instance, that they are transitive.

Strict word meanings have the semantic field slot fixed. Unfixing this slot is the form of over-extension which leads to metaphor. Metaphoric over-extensions and meaning shifts occur in the same way as other extensions and shifts (see B9, B10); people extend word meanings in production by using the word when only part of its meaning applies [2.5] (i.e by changing the semantic field slot) and listeners learn the new meaning with a different `metaphoric' value of the semantic field slot - one which makes sense in the context.

Again, the speed of the learning mechanism enables metaphoric meaning shifts to happen rapidly in a speaking, learning population; hearing six examples is enough to learn a new metaphoric sense of a word [3.2]. A similar account applies to metonymy.

**(B22) Children confuse names for parts of arms and legs**: This is a particular case of (B15) above, with evolutionary significance. Learners of many languages often substitute `finger' for `toe', `wrist' for `ankle', etc.; and the two are not distinguished in some adult languages (Bowerman 1989). If the meaning representation which underlies language evolved after bipedalism, this is puzzling - as hands and feet seem to have little in common. However, if language is based on a script representation of primate social situations, evolved over the last 20 million years [4.1], then it is not surprising; for most of that period, primate hands and feet have had very similar properties and uses, so might well have the same representation.

**(B23) NP-type nouns describe social routines**: As well as mass and count nouns, there is a third category with distinct syntactic properties - words such as `breakfast' and `church' which seem to behave more like noun phrases than nouns. For instance, they can be used `bare' without quantifiers, as in `Come down for breakfast'. NP-type nouns all seem to describe social routines. Children aged 4-5 productively expect new words describing social routines to behave syntactically in this way (Burns & Soja 1994).

In this theory, the linguistic meaning representation is based on a representation of social situations [4.1]. So we might expect that entities which are social routines have some rather special and `complete' meaning representation in scripts, which in turn leads to special `complete' NP-like syntactic properties. This is consistent with the finding that children apply the distinction productively.

**(B24) In disambiguating homonyms, we favour common word senses** : It is a commonplace observation, and an established psycholinguistic finding, that it is easier to understand a sentence containing a homonym when one of the common word senses of the homonym is used. This raises the question : how is the fact that it is a common or uncommon word sense represented in the brain, and how is it learned ?

The rule learning mechanism of [3.2] automatically learns a rule probability s(R) (which is the probability of the effect, given the cause) from the observed frequencies of cause and effect. In learning homonyms, s(R) is the probability of different word senses, which is learned from their observed frequencies [3.7]. Then, when finding the appropriate sense of an ambiguous word or phrase, the Bayesian maximum likelihood comparison of the different possible senses includes this factor s(R) for each sense [2.4]. For rare word senses, which have s(R) near zero, it is harder to find enough other evidence to counterbalance this small factor - making rare word senses slower to understand.

**(B25) Gender has little to do with sex** : In many languages such as German, the gender system seems to correlate with biological sex in a few core cases, but for most words in the language, gender is arbitrary - or perhaps partially predictable by purely phonological regularities. (Maratsos 1982)

We may expect that scripts, being a social representation, have a `sex' slot for people and some other animate entities [4.1]; this may be involved in representing grammatical gender for people, but what about inanimate objects with arbitrary gender ?

If we assume that there is a degree of flexibility in mapping the script representation onto other meaning representations (and thus onto the world), then the extension of gender to inanimate objects can be understood as a domain of regularity in language, extended by selection of m-scripts [4.4].

Suppose initially that proper nouns, improper nouns and pronouns for people all have a `sex' slot defined in their meaning script. This `sex' slot may be correlated with inflectional markers, which then have a number of uses for resolving ambiguities in complex sentences - for instance, in helping to determine the referents of pronouns or the agents of verbs. In all these cases, even one bit of sex information can be very useful (e.g. halving the number of possible referents for a pronoun).

The language can then become less ambiguous if the sex marker is extended to other words which do not semantically need it. Meaning scripts for inanimate objects can be artificially marked with a sex slot - which does not interfere with the interpretation of the script, but makes sentences involving those object easier to interpret, because of the gender clues.

The presence of disambiguation mechanisms, which depend on the sex slot, provides a selection pressure on the word m-scripts for inanimate objects, forcing them to acquire an artificial sex slot - which then becomes their gender. This is a domain of regularity [4.4] which is so useful that it extends across the whole language. Furthermore since, for inanimate objects, the value of the sex slot does not matter, it can be correlated with phonological properties of the words - so making the language more regular and easier to learn.

So the extension from sex to gender is easily accounted for as a result of m-script evolution to form a domain of regularity. The selection pressure is ease of disambiguation.

# 5.3 Phrase Structure Rules

It is in the learning of phrase structure rules that this theory differs most markedly from many theories of language acquisition, such as Pinker's (1984) theory, or the Principles and Parameters approach, in both of which:

1. A few key rules or parameters play a central role in a language
2. They are learned in a discrete, all-or-nothing manner.

In this theory, by contrast:

1. There are no distinct phrase structure rules or parameters; each word carries its own phrase structure rule in its m-script [2.3], so the phrase structure of a language is learnt incrementally along with the words
2. Each word m-script is learnt incrementally by accumulating evidence, with a Bayesian statistical criterion for sufficient evidence [3.2].
3. The basic style of language learning is bottom-up, through individual lexical items, with rather conservative induction of broader generalisations [3.5].

So broadly, while other theories predict a few key, all-or-nothing milestones in language acquisition, this theory leads to an incremental, undramatic form of learning, in which syntax is acquired gradually, `safely' , and linked to the lexicon. For these reasons, the predictions of this theory for broad syntax acquisition tend to be rather negative and undramatic; several of the comparisons in this section have a `negative evidence' flavour.

This theory does not, however, differ markedly from those theories in the continuity assumption. Pinker's theory, and the Principles and Parameters approach, assume that the child is learning the adult grammar, and prefer (for economy of hypothesis) to assume continuity: that the child does not make any major diversions into special `child grammars'. In this theory, too (with a few local exceptions) the child acquires adult word m-scripts, rather than some disposable intermediate forms, from the start.

**(C1) The syntax of any part of speech can be represented and learnt:** Different parts of speech have characteristically different ways of connecting with their arguments and fitting together. All these ways must be somehow represented in the brain, and learnt; any theory needs to explain how.

In this theory, the syntactic constraints of any word are contained in the structure of the left branch of its m-script. I have illustrated by examples how this can embody the main constraints of word order, agreement, and semantic restriction. Then trump links between left and right branches convey the meanings of arguments into the full meaning structure. In the program which can generate, understand or learn a fragment of English, I have verified this general statement for nouns, verbs, adjectives, adverb, auxiliaries, articles, quantifiers, pronouns and prepositions. I have also verified in general terms that the approach extends easily to other broad types of language (strongly case-marked, weak word order, agglutinating, isolative, ergative, etc.).

If all syntax can be embodied in m-scripts, then it can be learned, because the general learning mechanism can acquire any m-script [3.8 - 3.11]. The primary and secondary learning algorithms are the ends of a spectrum, variants of one fundamental learning method which can acquire all syntax.

**(C2) Early syntax centres on verbs**: The verb-centred nature of early syntax is confirmed by several studies (e.g. Bates et al 1988; Tomasello 1992).

In the m-script theory, nouns are learnt first, and then the way is open to learn any part of speech which can combine simply with nouns to express new meanings [3.1]. Verbs qualify on this count; so do a number of other morphemes which children use in `verb-like' ways (Tomasello 1992). With nouns and verbs alone, the child can say many useful things; so the m-script theory is fully consistent with the way children's early syntax centres on verbs.

**(C3) Children link arguments to verbs correctly from the start of verb learning**: While verb arguments are often omitted in early speech, `argument swapping' errors are extremely rare, and comprehension/ differential attention experiments confirm that children learn verb-argument linkages correctly from a very early age. The basic m-script learning mechanism can learn verbs' argument frames correctly, given only (in some minority of occasions) a knowledge of the meanings of the nouns which fill those frames, and a correct construal of the intended meaning script [3.8].

This is a form of semantic bootstrap, which requires no innate knowledge of `linking rules' between grammatical functions and verbs' semantic roles (Pinker 1984, 1989) - because it works entirely in terms of semantic roles, with no mention of grammatical functions.

Many theories of language learning aim to account for the direct learning of an adult grammar (the continuity hypothesis), and so learn rules which use the grammatical functions of subject and object. For instance, in Pinker's (1984) theory, the child learns phrase structure rules with subjects and objects. She must then also know a set of `linking rules' for each verb, to define which thematic roles are filled by the subject and the object. These must either be innate and universal, or learnt simultaneously.

In young children's language there is little evidence for the grammatical functions as distinct from semantic roles. All the evidence from learning first verbs is consistent with learning direct verb-noun linkages involving just semantic roles, with no intermediate concept of subject and object; that is the way verb argument structure is learnt in this theory [3.8].

Evidence for subject and object only emerges later, when children start to use to complex constructions (and the economy devices in them) which define those grammatical functions [4.6].

This theory is also consistent with a continuity hypothesis - in that the child learns adult word m-scripts from the start - but the semantic roles are primary in this process, and grammatical functions like subject and object emerge later, as part of the economy devices. These can only be learnt later, when the child has a large vocabulary and syntax (so can get the evidence for them, and also needs to use them), by separate means.

**(C4) Syntactic constraints also guide verb learning from an early age**: Gleitman (1990) has argued that children have `syntactic' expectations about the linking of thematic roles to specific argument positions from an early age. They can use SVO order to understand which is the agent and which is the patient of a transitive verb from 17 months (Hirsh-Pasek et al 1985), can use sentence form to distinguish a transitive from an intransitive meaning when learning a new verb at 20 months (Naigles 1990), and can use word order to fix the non-obvious verb argument assignments in learning new `chase/flee' type verbs at age 3-4 (Gleitman 1990).

In this theory, there can be no `syntactic' expectations about argument order until a few verbs have been learnt. From then on, m-intersection of those verb m-scripts (the secondary learning process) can form a few broad m-scripts which embody these expectations (e.g. that agents come before the verb, patients after) [3.11]. Gleitman's results, while showing that children can use word order very early, do not conflict with this interpretation. Since comprehension precedes production, at 17 months (when the first evidence for use of word order occurs) children may well have learnt a few verb m-scripts well enough for comprehension, and well enough to have made word-order generalisations.

**(C5) There are no sharp, language-wide transitions observed in language learning**: In a theory where syntax hinges on a few key phrase structure rules or parameters, one might expect some noticeable language-wide changes (e.g. changes discernible in the manner of use of most or all verbs) on or around the day when the child learns a key rule, or fixes a key parameter.

No such sharp or language-wide changes are observed; rather, as emphasised by Tomasello (1992) in his `verb island' hypothesis, different verbs (which are the locus of early phrase structure) seem to be learnt independently, so that the best predictor of a child's usage of any verb is her recent use of the same verb. Each verb matures at its own pace. This is just as we would expect in this theory, where each verb m-script is (at least in the early stages of language learning, before secondary learning begins) learnt independently from examples of that verb in use [3.8].

In a theory with discrete, language-wide changes, it is of course possible to find mechanisms whereby the effects of the change are not seen as a sharp transition, or are blurred out over time; but those theories have to be saved from their own predictions, rather than agreeing naturally with the data.

**(C6) There is no dissociation between syntax and vocabulary size**: If learning of key phrase structure rules (or parameters) were a distinct process from learning words, one might expect some children to set the parameters early or late, compared to their vocabulary development - leading to a statistical dissociation between vocabulary size and syntax.

While many aspects of language learning show strong statistical dissociations across populations (eg between comprehension and production), in studying a sample of over 1000 children Bates et al (1993) found no detectable dissociation between syntax and vocabulary.

This is just what we expect in the m-script theory, where the syntax of a language is embodied in the m-scripts for its words [2.3], so syntax must be acquired along with vocabulary, by m-script learning [3.8]. The m-script theory cannot predict a dissociation between syntax and vocabulary.

**(C7) Early analysed noun vocabulary predicts later syntactic ability; rote production does not**: In a longitudinal study of 27 children , Bates et al. (1988) found that analysed vocabulary at 13 and 20 months correlates very highly with syntactic ability at 28 months.

This is just what we expect in this theory - since early syntactic ability is largely a matter of verb mastery; and to learn verbs, you need to know some nouns to work out how the arguments are filled [3.7].

In the same study, Bates et al found, following Nelson (1985) and others, a spectrum of styles in early language learning; at 13 and 20 months there were two main `styles' of language production - analytic (short forms made up of analysed words) and unanalysed rote production of groups of words. They found that the second `rote' style had very little correlation with 28-month syntactic ability (as measured by Mean Length of Utterance), while the first `analytic' style was strongly correlated (C3 above)

This again is easily interpreted in the theory. To learn language involves at least two separate abilities [3.7, 3.8]:

1. To record in memory `learning examples' - stretches of speech paired with script descriptions of situations.
2. To m-intersect these together, analysing the stretches of speech down to individual words and their meanings.

If children vary independently in their ability to do (1) and (2), this will account for the observed dissociation. For unanalysed rote production, only (1) is required; whereas to develop syntax, both (1) and (2) are required.

**(C8) There is scant evidence for unmarked parameter values in language learning**: In theories which centre on a few key rules or parameters, the learning problem can be eased by assuming that some of these rules or parameters are the `unmarked case', or default, which can be assumed until contradicted by evidence. For instance, Pinker's (1984) theory starts from deep, narrowly branching phrase structure rules (as in X-bar theory) and only learns broader and flatter phrase structures (as required for languages such as Latin or Warlpiri) when this hypothesis fails. In Principles and Parameter theories, it is commonly assumed that each parameter starts at some unmarked value (typically the value leading to a more restrictive language, if there is one) looking for `triggers' in the input which might re-set it.

These theories broadly predict that some languages (or aspects of them) should be learnt faster than others; in a language with an unmarked parameter value, that value should be known correctly from the start, whereas children whose language has the marked value might be expected to make characteristic early `parameter switched' errors before they hear the trigger.

I believe that the evidence for these favoured parameter values or rules is scant (although I do not claim it is non-existent). In principle, to prove the case, pairs of languages (with opposite parameter values) should be looked at together. Some examples are:

- The head parameter, which determines dominant word order, seems to be learnt correctly from the start both in head-first languages (like English) and head-last languages (like Japanese); word order errors are very rare in both cases. German has a hybrid word order, and does show some early word-order errors.
- The null subject parameter has been the focus of much interest; early omission of subjects in languages such as English might look like evidence for an unmarked setting allowing null subjects. However, this is the more permissive of the two settings, and theorists are reluctant to allow this as the default, as it would require negative evidence to re-set it. It seems more likely that children tend to omit many parts of sentences early, and that subjects are particularly omittable because the subject is often obvious.

- There is some evidence that children learning a free word order language tend to used fixed word orders at first. If fixed/free word order is a parameter, this may imply that fixed order is the unmarked case. However, the actual use of word order freedom on a free order language serves a variety of pragmatic and discourse purposes; it seems equally plausible to assume that young children have not yet learnt these subtle uses of the freedom, so stick to a few word orders out of habit.
- Roeper argues that overtensing with auxiliary `do' is evidence for a default parameter, but does not base this on a pairwise comparison of languages

So in spite of the theoretical attractions of default parameter values, no clear-cut evidence for them has been found. In all of these cases (and others) the m-script theory makes the `neutral' prediction that children learn different `parameter settings' - embodied in different forms of word m-script - with equal ease. This seems broadly consistent with the bulk of the data.

**(C9) There are no major blind alleys in language learning**: Theories which hinge on discrete, all-or-nothing learning of a few key parameters or phrase structure rules tend to be haunted by a `one false move' prediction. Suppose the child sets some parameter wrongly; will that send her along some blind alley of language learning, and if so, how will she ever recover ?

Since these theories also tend to use categorical rather than statistical learning mechanisms, this problem is particularly tricky; there can be no gradations depending on weight of evidence. They may need elaborate mechanisms to reset parameters or unmake generalisations, with complex criteria about when resetting can happen.

While children do over-generalise, leading to mistakes with particular words, there is, as far as I know, no evidence for any major `blind alleys' taken by children in learning a language. That is fully consistent with this theory, in which there are no language-wide `strategic' learning decisions to be made; and any word's m-script can be continually re-learnt (and unlearnt if necessary) on the basis of recent evidence.

**(C10) Languages have regularities captured in X-bar syntax** : X-bar syntax, as developed by Jackendoff (1977) embodies two main insights:

1. Different sentence subunits, such as NP, VP, etc., have similarities of behaviour which warrant a common treatment as XP.
2. For each of these subunits, there can be several `bar levels' such as N, N', N'' etc. with different syntactic privileges depending on level.

In summary, the m-script theory readily accounts for (1), but does not yet have any neat account of (2).

The fact that in any language, all the various clause types tend to be either head-first or head-last, uniformly across the language, is accounted for by the strong force towards regularity needed to handle ambiguities [4.5], leading to the emergence of the Greenberg universals.

Other similarities may be understood as general similarities of form between m-scripts for different parts of speech.

The regularities under (2) do not yet have any pleasing account in this theory. In this theory, m-unification matches meaning structures. If the meaning structures are simple scripts as used for illustratoin in this paper, then there is little in the scripts to distinguish the different bar levels identified by Jackendoff and others. That is not to say that we could not find elements of script structure which are identifiable as markers of bar level; just that it has not been done yet. The existence of bar-level regularities is probably an indication that script meaning structures are, indeed, more complex than the illustrative structures I use here.

**(C11) In assigning agent roles, cue strength depends on overall cue validity** : Bates, MacWhinney and co-workers (MacWhinney & Bates 1989) have performed experiments where speakers of many different languages and ages hear simple transitive sentences with conflicting cues about which noun is the agent. The cues manipulated include word order, case-marking on the noun, animacy, and verb agreement. These experiments robustly reveal fascinating cross-language differences, and differences between ages, in the importance hearers assign to different cue types in choosing the agent.

For instance, Italian children under 7 give priority to animacy, followed by SVO word order; whereas Italian adults give priority to SV agreement, followed by clitic agreement and animacy. English speakers of all ages give priority to SVO word order.

They find that across many languages, young children assign cue strength on the basis of overall cue validity. Bates and MacWhinney (1989) summarise data which qualitatively confirm this in English, Italian, French, Spanish, German, Dutch, Serbo-Croatian, Hungarian, Turkish, Hebrew, Warlpiri, Chinese, and Japanese.

Cue validity is defined independent of the experiments which measure cue strength, as

Cue validity = Cue availability * Cue reliability

Availability is the proportion of occasions the cue is there to be used, and reliability is the proportion of those occasions when it gives the correct role assignment; so both can be approximately measured for a language from corpora or texts.

The strong correlation between the cue strengths seen in young children and overall cue validity is an important finding, which MacWhinney (1989) interprets in a connectionist learning model of cue competition. It is equally possible to interpret it in the Bayesian learning and processing model.

There has been concern that, because the test sentences often give conflicting cues about the agent assignment, subjects switch to some other (perhaps non-grammatical, conscious) processing strategy. However, the robustness of the findings across different experimental conditions argues against this. In the m-script theory, I propose that hearing of a conflicting set of cues acts like a signal that something has been misheard or garbled, and so tends to greater activation of the strategies for handling ambiguities. (This is not a switch in strategies, as ambiguity handling is continually necessary).

The core ambiguity-handling strategy is (a) to use broad general m-scripts (gathered by secondary learning) to get some clue about what is going on [3.11] , then (b) use Bayesian maximum likelihood estimation to choose the best possible interpretation [2.4].

For each possible cue, we suppose the child has learnt, by secondary learning, a broad m-script which relates the cue to the choice of agency; for instance, English has a very strong SVO word order cue, which can be embodied in a simple m-script - the m-intersection of many verb m-scripts. The rule probability s(R) of this m-script is acquired from the many examples, and is equal to the cue reliability. Then, when Bayesian maximum likelihood inference is used to choose between the different possible interpretations, the rule strength s(R) of each conflicting cue enters into the competition.

For instance, suppose there are three cues, R1 , R2 and R3 , and R1 indicates a certain agent assignment in contradiction to R2 and R3 . The maximum likelihood calculation is to compare s(R1)[1- s(R2)][1-s(R3)] with [1-s(R1)] s(R2) s(R3); thus R1 competes with R2 and R3 on the basis of their cue reliabilities.

In this theory, therefore, we predict that cue reliability, rather than cue validity, is the best indicator of cue strength in the child. However, cue availability (the other component of cue validity) does enter into the predictions. While the eventual rule strength depends only on reliability, the speed of learning depends on cue availability. Since secondary learning depends on broad generalisations from a lot of evidence, we may assume it proceeds slower than primary learning, and this slower learning may affect the observed results. Detailed examination will be required to decide whether cue validity, or cue reliability paced by cue availability, gives a better fit to the data.

The main point is that a Bayesian maximum likelihood choice of interpretations, with broad cue m-scripts learned by secondary learning, gives a good overall interpretation of cross-linguistic cue competition results in young children.

**(C12) In many languages, cue strengths change markedly between ages 6 and 16** : The same researchers have measured cue strengths into adulthood in the same set of languages, and found slow but profound changes over the age range from 6 - adult. They interpret these data in terms of the conflict validity of the different cues, where conflict validity is a single number for each cue, defined in terms of how reliable the cue is when in conflict with other cues.

In this theory, the interpretation of these results involves something like conflict validity, but it is not just a single number per cue.

For instance, consider Kail's (1989) results on French. For children under 6, the order of cue strength is SVO word order first, then animacy, then VSO and SOV order. For adults, however, the order of cue strength is completely different : SV agreement, then clitic agreement, then animacy, then SVO order, then word stress. In this theory, we would interpret such slow but profound changes above age 6 as arising from two sources:

(A) Changing Cue Reliability : SVO word order is not a reliable cue to the agent in adult French, largely because of the extensive use of clitics and pronouns (which have different word orders) and the use of diverse word orders for pragmatic/discourse purposes. The child under 6 has probably not yet differentiated or learnt most clitic pronouns, and has not yet learned the pragmatic/discourse markers for other word orders. She works at sentence level, rather than discourse level (Karmiloff-Smith 1979).

If the child restricts her learning set to the sentences she can interpret, this may exclude many sentences which use clitics and exotic word orders; so in this selected sample, the cue reliability of SVO word order is rather high. Later, when the child understands clitics, SVO reliability drops. So cue reliability cannot be simply estimated from text corpora; it depends on how the child filters her learning input.

(B) Learning Exception Rules : If two cues A and B both predict an agent assignment, the child will first learn individual rules (A agent) and (B agent); when these rules conflict, she can use the Bayesian maximum likelihood inference to choose between them. However, given time the child will accumulate enough information to learn an exception rule (A&B agent), which may have a different rule probability s(R) from that given by maximum likelihood combination. This process of learning exception rules is slow [3.6] and is discrepancy-driven - a new rule is learnt only when the old rules make systematically wrong predictions.

This learning of exception rules will slowly acquire something related to the `conflict validity' of cues proposed by Bates, MacWhinney and their collaborators, and can be expected to account for some of the same effects; but conflict validity will not be a single number per cue.

In summary, the m-script learning theory has within it mechanisms which seem able to account qualitatively for the changing cue strengths which have been observed in several languages; but a detailed analysis and comparison will be tricky, because of the complex overlapping effects of (A) and (B) above.

**(C13) Meaning elements encoded locally in the sentence are learnt most easily**: Slobin (1973, 1982, 1985) has summarised evidence that greater separation in the sentence of morphemes encoding an element of meaning leads to slower learning and more errors. For instance, German gender is marked on articles rather than on the nouns themselves, and is learnt slowly; part of the cause may be the distance between the article and the noun whose gender is at issue. Slobin (1982) notes that children understand causative constructs earlier in Turkish and Serbo-Croat than in English or Italian, because in the former two languages the causative construct is indicated by a local cue, and suggests that a local cue effect also explains the early acquisiton of object inflections in Turkish.

While some of these examples may have multi-faceted explanations, there is a clear causal link from separation in sentences to difficulty of learning, which may contribute to all of them.

A construct is learned by collecting some six or so clear learning examples, forming the SMS for each one, and m-intersecting them together [3.9]. The speed of learning is paced by the time taken to gather these learning examples.

When a child hears a sentence or fragment L phonemes long, the probability P that she hears it clearly, understands all the words and infers the intended meaning can be roughly modelled as $P = \exp[-\mu L]$; every extra phoneme diminishes P by a multiplicative factor. The exponent $\mu$ gets smaller as the child's command of language increases, but for young children, $\mu$ is large, leading to a strong cutoff in P with increasing L. Therefore it is harder for the child to gather long learning examples.

A very local construct can be learnt from short learning examples, so can be learnt rapidly; but any construct which extends across many phonemes will require longer learning examples, which (in the early years) take longer to gather. That is why local constructs can be learnt earlier. (Another example of this effect is that nouns tend to be learnt before verbs; nouns require only the noun sound in learning examples, whereas verbs also require their arguments)

**(C14) Children tend to mark individual meaning elements explicitly and separately**: Slobin (1985) has summarised evidence from many languages that children tend to be explicit in their speech, preferring free morphemes over bound or contracted morphemes (e.g in English auxiliaries, Bellugi 1967), and under-using ellipsis (e.g in Japanese, Clancy 1985). Children tend to separate elements of meaning which in adult language are fused; for instance, in French, using *de moi* in stead of *mon*, separating possession from the pronoun (Karmiloff Smith 1979; Clark 1985). Similar examples are found in Hungarian (MacWhinney 1985) and Hebrew (Berman 1985). Slobin (1985) summarises this and other evidence in his two operating principles, `Maximal substance' and `Analytic form'.

In the m-script theory, these effects arise not from a preference for a particular `child grammar' or from a production bias; they arise because evidence for the separate forms accumulates faster, so the separate forms are learned earlier. For instance, in French the possessive *de* is used with high frequency in many places; whereas a pronoun-linked possessive such as *mon* occurs more rarely. Therefore it takes a child longer to accumulate the necessary clean learning examples for *mon* than for *de* and *moi*; there will be a period during which *de* is securely known as a possessive, and *mon* is not; during this period the child is likely to use *de moi* and similar `analytic' constructs in stead of adult-like fused constructs such as *mon*.

**(C15) Negation, interrogatives and conditionals are moved outside clause boundaries**: It is a robust finding across many languages (Slobin 1985) that in early uses of negation, children tend to put the negative particle outside the boundary of an unmodified clause, rather than `fused in' to the clause at a variety of places (eg on the verb) as it typically is in adult language. Slobin cites evidence from English (Bellugi 1967), Polish (Smoczynska 1985), Turkish (Aksu-Koc & Slobin 1985) and French (Clark 1985).

There is a similar `clause external' placing tendency in interrogatives (English:Bellugi-Klima 1968; Hungarian: MacWhinney 1973) and conditionals (Hungarian: MacWhinney 1973).

This effect can be understood, in the m-script theory, as arising from a difference in the speed of learning of different constructs, dependent on the placement of negation (or conditional status or interrogation) in the script meaning structure.

For instance, the simplest representation of negation seems to be some form of `negation slot' on the top node of a whole scene, whatever the content of that scene. If we assume that negation is represented in meaning structures in this way, and if the adult language has any clause-external form of negation at all, then that word is represented by a very simple and broad-range m-script, shown in figure 5.1 below.

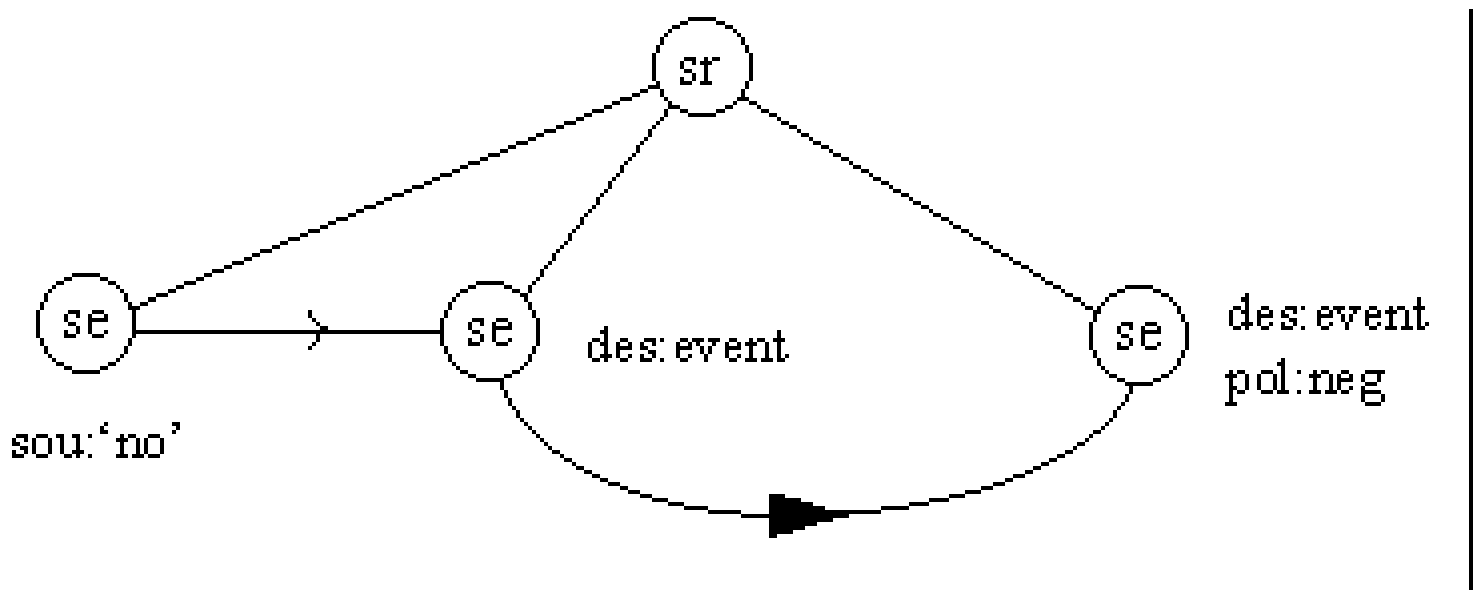

*Figure 5: word m-script for a clause-external negator*

This m-script for `no' means that if any sequence of words which designates an event or process (i.e results in a slot [des:event] ) is preceded by the sound `no', then the polarity of the resulting event scene is negative - it is asserted not to happen.

The clause-external placement of the negator (in the left branch of the m-script) allows any clause to be negated; just as the top-node positioning of the `polarity' slot allows any event meaning structure to be negated. This m-script is so simple and so easily applicable that, we may suppose, given any opportunity the child will learn it and may then use it as a universal negator.

In contrast, many adult forms of negation - e.g fused into the verb, or into an auxiliary, or in an agent such as `nobody' - are more complex as m-scripts, and each one is not so universally applicable. Their use by the child is, in the early years, doubly inhibited - there are more of them to learn, and as the evidence for each distinct form of negation accumulates more slowly (only when its special cases occur) each one takes longer to learn. It is then not surprising that, under pressure to communicate, children often fall back on a universal negator.

The question then arises: if the universal clause-external negator is so easily learnable and so adaptable, why do adult languages use many specialised, hard-to-learn forms? I believe that the evolution of these specialised negation m-scripts has been driven by two selection pressures:

- **Economy**: an isolated, clause-external negator is at least another morpheme to say - whereas many of the clause-internal negators replace, rather than add to, morphemes inside the clause.
- **Reliability**: For communication in noisy environments, there is a possibility that the clause will be heard entire, while its external negator is lost in noise - communicating exactly the opposite to the intended meaning. But if the negator is attached to the verb, then when a noise obliterates the negator it is likely to obliterate the verb too - causing the listener to seek clarification, or at least not be misled. Clause-external negators can cause dangerous misunderstandings.

For adults, these two advantages of specialised within-clause forms overcome the learnability disadvantage.

**(C16) Double-marking of negation and other constructs is a later error**: It has been noted in several cases (e.g Hungarian conditionals: MacWhinney 1973; English negation: Bellugi 1967; English past tense morphology: Kuczaj 1978; French possessives: Karmiloff-Smith 1979) that redundant marking of the same meaning element tends to occur more in older children, not when children first learn to express that meaning element.

Redundant marking is not a generally-forbidden feature of adult language; for instance, it is regularly used in many forms of agreement such as noun-adjective agreement, verb-agent agreement, etc. The m-script language generation method [2.5] supports redundant marking (and the learning mechanism builds it into the m-scripts of words which have agreement constraints),and generally gives speakers the option of doing it - although the mechanism naturally favours economy over redundant marking. Therefore children may easily over-mark meaning elements, especially when they are concerned to convey those elements clearly. They need to learn the conventions of adult language about what can be redundantly marked, and what cannot.

For negation, over-marking will not happen when the child has learnt just one, clause-external negator - but may happen later when she has also learned some clause-internal negators, and has the option to use them as well. Perhaps not yet being confident of the clause-internal forms will increase the tendency to double mark `for safety'.

How do children learn to correct these over-markings ? As for many other errors, the theory has a direct mechanism to gather implicit negative evidence [3.9] - discrete occasions where the child can observe `where I would have used a double negative, an adult used a single clause-internal negative'. This is an intrinsic part of the primary learning mechanism [3.11] . Accumulating enough of these examples, the child learns that specific double negative forms are not used; but this is a slow learning process [3.6], and so children correct these errors slowly.

# 5.4 Morphology

The issue of closed-class morphemes and over-regularisation has been a test case for theories of language acquisition. The key evidence under debate is the fact that children learn regular and irregular morphology (for instance, of the English past tense) in a `U-shaped curve' - first learning inflections of both regular and irregular verbs correctly, then learning the productive rule and over-regularising the irregular forms, then finally (and slowly) removing errors in the irregular forms.

One difficulty has been that the over-regularisation error can only be corrected by the use of negative evidence - to establish that forms like `hitted' or `goed' cannot be used. The consensus is now that explicit negative evidence (e.g. correction by adults) cannot account for the correction (Marcus 1993); and many theories are reluctant to use implicit negative evidence - for fear of ruling out other rare, but legal, forms. The Bayesian theory of learning can use implicit negative evidence [3.6, 3.10], so can in principle account for the correction of over-regularisation - but as we shall see, some extra assumptions are needed to do so.

**(D1) Individual word morphology is learnt before productive morphology rules**: It seems clear across all languages that individual word morphologies are known before any productive rule is known.

The morphology of individual words is learnt by the primary learning process [3.9], which can start as soon as word learning begins. Productive regularities, which allow children to coin regular inflections they have never heard before, are learnt by the secondary process, which takes the individual word morphology m-scripts as its input [3.11]. This secondary process must be able to segment the inflection from the root, which it does by the normal segmentation learning process [3.7, 3.8].

Therefore in this theory, productive morphology cannot be learnt before a set of (typically six or more) individual word morphologies have been learnt.

**(D2) Productive regular inflections can be learnt even if regular forms are not in a majority** : In the English past tense forms which children are initially exposed to, irregular forms are a majority, yet the regular past tense rule is learnt.

The Bayesian learning mechanism is very robust, able to learn a regularity even if the learning examples which obey it are in a minority [3.3, 3.4]. Therefore even when the majority of a child's known verbs have irregular past forms, the regular inflection can still be learnt (by the secondary learning process) from the minority; the rule will be learnt provided it gives a more economical account of the data as a whole - that is, if more than about six verbs conform to it.

Then, when the child is asked to produce a past form of a new verb (where he has heard only the present form), the regular inflection rule, being the only known way to produce a past form of this verb, will be used.

**(D3) We learn which dimensions of meaning are encoded by inflectional morphemes** : An inflection can convey meaning in several different dimensions, such as gender, case, number, tense, or modality - along any one or several of these together. The task of learning which dimensions are signified, and how, is complicated by syncretism; an inflectional system may describe several dimensions at once, but particular inflections in that system may collapse one or more dimensions.

The learning examples for the primary learning process (to learn individual inflected words) have all possible dimensions of meaning represented within them; sex, number, semantic role etc. will all appear in the examples (as values of some slot on some node) because the child may put anything she observes about the scene into the right-hand branch of her learning example [3.7, 3.8]. If an inflection has no connection with some dimension of meaning (some slot), these learning examples will have randomly distributed, contradictory values of the slot, and so script intersection will remove the slot [2.1]. If, however, the inflection denotes some value of a slot, that value will survive the script intersection, to become part of the meaning of the inflected form.

So if a particular inflection denotes one, two or more dimensions of meaning, those pieces of meaning will all be learnt as part of individual word inflections, and so (by the secondary process) will then be learnt in the productive inflection rule. The learning of multi-dimensional inflectional paradigms (as proposed in Pinker's (1984) theory) emerges naturally from the general word learning process.

**(D4) The speed of learning of inflectional morphology varies between languages**: The age at which children first apply inflections productively to new stems varies across languages. Turkish children make productive inflections by the age of 2, and in other languages such as English (Kim et al 1994), children make productive inflections from age three.

There is probably no single explanation of this fact in the theory, because it arises from a variety of factors - phonological prominence, regularity, communicative need, agglutinative versus synthetic and so on. One of these factors is analysed in (D5) below.

Slobin (1982) cites 12 possible reasons why the inflectional morphology of Turkish my be expected to be learnt particularly early - that Turkish inflectional morphemes are (1) postposed, (2) syllabic (3) stressed (4) obligatory (5) tied to the noun (6) consistent with the verb-final typology of Turkish (7) ordered plural-possessive-case (8) nonsynthetic in their mapping of function into form (9) expressing only grammatical roles, not pragmatic information (10) exceptionless (11) applied consistently to all pro-forms (12) distinct (no homonyms).

It is consistent with the earning model of this theory that at least the factors (2), (3), (4), (5), (8), (9), (10), (11) and (12) should all lead to faster, more reliable learning; these effects could be modelled quantitatively, although I have not done so. So we have a basis for understanding the different speeds of learning productive inflectional morphology.

**(D5) Productive inflections are learnt faster in agglutinating languages than synthetic languages** : For instance, Slobin (1982) finds that affixes are learnt faster in Turkish (agglutinating) than in Serbo-Croat (fusional). In English, the more features encoded in a grammatical morpheme, the slower the child comes to use it (Brown 1973; de Villiers and de Villiers 1973, Pinker 1981).

Consider an agglutinating language in which there are typically N inflectional affixes to a stem, each with two possible forms; compare this with a synthetic language, having just one affix with 2N possible forms. Of the order of six individual words (complete with inflections) are needed for the secondary learning process to learn one productive regular inflection rule [3.11].

Then for the agglutinating language, about 12 word m-scripts must be known to drive the learning of one regular inflection; but for the synthetic language, about 6.2N+1 word m-scripts must be known, which may be much larger than 12. Thus in the agglutinating language, learning of regular inflections can start faster, because the required number of learning examples are gathered sooner.

**(D6) In Agglutinating languages, inflections are learnt from the outside inwards**: Inflectional and derivational morphemes nearest the word stem are the last to be used productively. There is a fairly natural explanation of this in the theory - that the child first learns words with several morphemes attached, by the primary learning process [3.9], then m-intersects several of these together to strip off the outer inflectional morpheme by the secondary learning process [3.11], and then by further application of the secondary process (a tertiary process ?) strips off the next inner morpheme, and so on.

However, while this is a plausible account within the theory, it is not unique; alternatives which do not make the right prediction are also possible. It seems that m-intersection is powerful enough to extract some `inner' inflectional morpheme straight away by secondary learning - so why does it not sometimes do so? In the verbs of semitic languages, there are only `inner' inflections between the consonants, and these must be learnt somehow.

I suspect that the true account involves a mixture of factors, including:

- M-intersection proceeds most `naturally' from the outside in, as discussed above; although it can work otherwise.
- Phonemes at the ends of a word are most phonologically salient, so most learnable
- Languages evolve so that the inner morphemes are most tightly bound to the core meaning (e.g. imply the most far-reaching changes to the core meaning) - for instance, derivational morphemes are always nearer the word stem than inflectional morphemes. So outer morphemes represent more easily detachable and learnable pieces of meaning.

Why do languages change in this way ? There are many possible reasons. For instance, if tightly bound elements of meaning are not separated, this makes for more reliable communication in the presence of noise. (seealso (C13)).

**(D7) In agglutinating languages, children make no errors in ordering affixes**: (Slobin 1985) This looks at first like a simple enough prediction of the learning theory; if affixes are learnt from the outside in, as above, there seems to be no mechanism available in the theory to mislearn their order. However, it is not that simple, and it seems that an extra assumption (albeit a fairly innocent one) is needed to account for children's lack of affix-ordering errors in production.

Consider a hypothetical language in which the `inner' affix defines number and the outer affix defines case. The m-script for a particular case affix must pass across the number slot unmodified from its left branch to its right branch, so that the number information from the inner affix is not lost. It does this by a variable slot value [number:?N] on the entity node in left and right branches, which is easily induced from the learning examples. The m-script for the inner `number' affix needs no such variable slot value to pass across the case, as the case slot in the meaning will have been stripped off by the case affix.

In generation, these m-scripts can be used in two different ways: either they can be `pre-unified' with the noun stem to form an m-script for the full inflected noun before use, or each affix m-script and the stem m-script can be m-unified with the SMS on the fly. In either case, to prevent the two affix m-scripts from being used in the wrong order, the case affix m-scripts must have the constraint of requiring some value of the number slot to be defined, although it does not matter which value it is. As it stands, a variable identity such as ?N does not impose this constraint; we need to postulate a special kind of variable value which does. With this extra assumption, the affix-ordering constraint can be learnt and reliably applied.

**(D8) Irregular forms are initially learnt correctly** : It is well established that before the dip of the U-shaped curve, irregular forms are used correctly (Marcus et al 1992; Marcus 1995).

Before any regular morphology rules are learnt, irregular past forms such as *hit* or *went* are not different in any way from regular forms such as *counted* or *waited*; so they are all equally learnt correctly by the primary process, direct from primary learning examples [3.9].

**(D9)There is transient over-regularisation of irregular forms** : The dip in the U-shaped curve is now well established for many types of over-regularisation of irregular forms, such as English past tenses (Marcus et al 1992) and English noun plurals (Marcus 1995).

As soon as the child has learnt around 6 examples of the regular inflection, she may start to learn the productive rule by the secondary mechanism [3.11]; but will subject the rule to a test of statistical significance before acquiring it [3.11]; so possibly rather more than 6 examples are needed.

A regular inflectional rule, once learnt, may be applied to all words of a class (e.g. to all verbs, to form the past tense); at this point, the child knows two ways to produce a past form of an irregular verb: (a) Use the previously learnt irregular form m-script (e.g. *went*), or (b) m-unify together the stem m-script for *go* with the inflection m-script for *-ed* to produce an m-script *goed*, which is then used for generation in the usual way.

Possible reasons why the over-regularised form is only used sporadically are discussed under (D10) below.

To correct the error, the child uses the primary learning process to gather negative evidence [3.10]. That is, when learning any words from a sentence where an adult said *went*, the child may do both partial sentence understanding of words heard, and partial generation from the inferred meaning script (and in so doing, silently generate *goed*); so on these occasions the child may observe that “I would have said *goed* where the an adult said *went*”. These constitute negative evidence for *goed*, and so can be used by the Bayesian learning process to learn a negative rule - that “the regular past inflection of *go* never happens” (a negative rule is one with rule probability near zero)

If it is so easy to gather negative evidence, why does over-regularisation take a long time to eradicate ? There are three reasons:

1. The Bayesian analysis shows that it may require a large number of learning examples to gain confidence in the exception rule [3.6].
2. Over-regularisation is rare in the first place. Therefore a child does not `silently' generate over-regularisations very often; she does not use all her opportunities to gather negative evidence.
3. As above, rate of learning depends on the rate of errors. If x denotes degree of belief in an over-regularisation, its rate of change (through unlearning) is governed by an equation of the form $dx/dt = -lx$ ; so x is expected to undergo slow exponential decay.

Therefore this theory can broadly account for the slow correction of over-regularisations.

However, the theory does not yet fully define the mechanism whereby a negative rule (the rule that a certain form does not occur, in places where the child can apparently generate it by a productive rule) actually prevents the productive rule from being applied. To have a detailed account of how the prevention happens, we need to build a more detailed procedural model of the language production process. This could be done, but would involves extra assumptions about how language production works.

**(D10) English noun plurals and past tense verbs are over-regularised with low frequency**: Both for English past tensesand English noun plurals it is now known (Marcus et al 1992; Marcus 1995) that the over-regularisation is made only at low frequencies (less than 10%) even in the trough of the U-shaped curve.

The m-script theory does not give any unique reason for this low frequency of over-regularisation, but is not inconsistent with it; several accounts of this fact can be devised. The probabilistic nature of the theory (both for learning and for production) does not force it to make a categorical prediction that one or the other form will be used exclusively at any stage. Some possible components of an explanation are:

- In general, we would expect a specific `exception' rule to override a general rule whenever the more specific rule is applicable. Therefore when the child definitely wishes to convey past tense, the specific irregular form may overrides; but in a few cases where tense information is not well defined, the specific form is not clearly preferred and the over-regular form may `sneak by'.
- Use of the specific rule involves fewer m-unifications, as an extra m-unification is required to construct the regular form; so the irregular form is easier to produce.
- The child has used the irregular form correctly before the over-regular form is producable - so continues to use the irregular form most of the time from habit, even when an alternative is available.

None of these accounts are particularly crisp or compelling, but the finding of sporadic over-regularisation certainly does not contradict a prediction of the theory.

**(D11) Specific Language Impairment affects regular morphology**: Although Specific Language Impairment (SLI) is probably not a homogeneous disorder, because of a lack of clear idendificatory criteria (Fletcher & Ingham 1993), nevertheless it does seem to show a strong tendency for impaired morphology, often in the absence of other clear deficits. This is so particularly in the cases of familial SLI studied by Gopnik et. al.

Because of the particular link with morphology, there is a possible interpretation in this theory, that *SLI is an impairment of the secondary learning process* [3.11] which is used to learn productive inflectional morphology. Although the primary and secondary processes are not entirely distinct - being points along a spectrum - it seems quite possible for the secondary end of the spectrum to be preferentially impaired by a genetic disorder.

Evidence on SLI in several languages suggests that phonological salience of morphology also plays a role - for instance, Italian SLI children show rather few deficits of morphology (Leonard et al 1987, 1992); but the `secondary learning' interpretation of SLI cuts across this distinction in interesting ways:

- In languages with prominent but simple morphology (such as Italian) it may be possible to learn the morphology of many words individually by the primary learning process, and so SLI children may show few deficits. Productive morphology with new words is the real test.
- In agglutinating languages, the secondary learning process is required to `factorise' the agglutination; so we would predict high SLI impairment of morphology in these languages. In the Greenlandic language Inuktitut, which has very rich verbal and nominal morphology, SLI leads to frequent omissions of verb inflections marking person, number and modality (Crago, Allen and Ningiuruvik 1993).

An `SLI = impaired secondary learning' picture seems broadly consistent with the rather confusing data. It also makes an interesting prediction - that SLI children should make fewer over-productivity errors (of several types, not just of morphology) than other children, when these follow from secondary learning - for instance, in alternating verb argument structures.

**(D12) High use of closed-class morphemes at 20 months leads to low use at 28 months**: Bates et al. (1988) found in their longitudunal study of 27 English-speaking children that the children who made highest use of closed-class morphemes at 20 months tended to be those who used them least at 28 months. This puzzling fact illustrates the difficulty of interpreting simple measures of child language.

The interpretation given by Bates et al (1988) for their findings can be couched in the m-script theory as follows: most of the closed-class forms produced at 20 months are unanalysed - part of larger rote-learned forms. These are narrow, context-specific m-scripts, learnt by the primary learning process [3.9]. Children who use more of these forms tend to analyse their input less into individual word m-scripts, and so are less ready for the secondary learning process [3.11] which must precede the productive use of closed-class morphemes seen at 28 months; thus there is an inverse correlation between use of closed-class morphemes at 20 and 28 months.

**(D13) Ergative and accusative case markers are initially under-extended** : In languages which use case marking to distinguish the two main arguments of a simple transitive verb, the case marking scheme is either nominative/accusative (in which case the accusative, rather than the nominative, is marked) or ergative/absolutive (where the ergative subject is marked). For intransitive verbs, the subject (being either absolutive or nominative) is not marked in either type of language. The markings of ergative or accusative achieve economy, in each case only marking one out of the three commonest verb arguments.

Children learn these markings soon and fairly reliably, and do not significantly over-extend them (e.g in Mayan Quiche: Pye 1979,1980). However, they do tend to under-extend the use of both the accusative and ergative markers, initially using them correctly only in simple `manipulative action scenes' (Slobin 1985) where the `who does what to what' is very clear; only later extending them to other transitive verbs such as `see', or `call-out'.

This initial under-extension has been observed in Russian accusative markers (Gvozdev 1949), and ergative markers in Kaluli (Schieffelin 1985). In Kaluli, the ergative marker is more reliably used with past-tense verbs, and is more likely to be omitted for future or negated verbs.

These facts have a fairly simple and direct explanation in the m-script theory. The key roles in transitive and intransitive verb scenes are marked by two distinct slots: an `actor' slot which defines who initiated the action ( = the subject for both transitive and intransitive verbs) and a `changed state' slot which defines which entity undergoes a change of state (object for a transitive verb, subject for an intransitive verb). Nominative/accusative markings are linked in m-scripts to the `actor' slot, while ergative/absolutive case markings are linked just to the `change state' slot [4.6].

Since these case markings correlate directly to simple, easily observable aspects of the meaning structure, they can be learnt early and reliably [3.4] (whether for individual nouns by the primary learning process [3.7], or productively by the secondary process [3.11]) ; and once learnt, there is no reason to expect any over-extension of either the accusative marker (= `did not initiate the action') or the ergative marker (= `did not undergo a change of state').

However, we would expect these markings to be used reliably only in cases where it is semantically clear who initiates the action and what undergoes a change of state. For manipulative action verbs, it is clear, especially in the past tense (where the state change has definitely happened); but for a verb like `see' it is less clear, for a future verb it is less clear, and for a negated verb it is particularly unclear. In all these cases, therefore, the child is likely to take the safe course of omitting any marker, as is observed.

So this form of under-extension arises not (as some others do) from the child initially learning a meaning which is too narrow and specific. The meaning is learnt correctly from the start, but the conditions for applying it are not always obvious.

# 5.5 Complementation and Control

Many English verbs such as *want*, *tell*, *try* and *seem* take a sentential complement, in which the subject is omitted. The rules for filling in the missing subject need to be acquired, and children do this rapidly and reliably; they start using complement-taking verbs quite soon after their first verbs, and make very few errors of control of the missing subject (Pinker 1984).

Learning of control relations occurs straightforwardly in the m-script theory, as a part of the normal m-intersection primary word learning process [3.9]. To see this, we need to see how control relations are represented and used in the m-scripts for complement-taking verbs.

The m-script for a typical control verb, *wants* , is shown in figure 5.1. In its left branch, it requires an animate entity scene (the person who wants something), the sound *wants* , and an event scene - the event that he wants to happen, which is the complement. In *Charlie wants to eat cake* this event scene will have been derived by m-unifying the complement verb *to eat cake* without a subject.

The right-hand meaning branch of *wants* represents the wanter (the entity defined by a trump link from the entity in the left branch) having a desire that a certain scene take place (the scene defined by a second trump link, from the complement event scene in the left branch; so this event is inserted into the full meaning). The control relation specifies who is the agent in this desired scene. The identity of this agent is given by a variable identity ?A, which is the same variable as the identity of the wanter. So in *Charlie wants to eat cake*, the m-unification automatically sets the identity ?A to Charlie, in all the places where it occurs - ensuring that it is Charlie who does the eating in the desired scene.

How is the *wants* m-script learned ? In the primary learning process, the child collects a small number of learning examples - where she hears the words and infers the intended meaning by other means - and m-intersects them together [3.8]. In each learning example, the identity of the agent in the desired scene is the same as the identity of the wanter. The first step in m-intersection is script intersection; this automatically detects equal slot values at different places in a script, and represents them by a shared variable value [2.1]. So the m-intersection automatically detects the control relation and embodies it in the m-script.

For *Charlie wants Lucy to go away*, there is a second, distinct *wants* m-script which consumes two entity scenes and the complement event scene in its left branch - and similarly, by a shared variable identity, ensures that it is Lucy who goes away, in the desired scene. These two m-scripts are learnt and used entirely separately.

**(E1) Children acquire some complement-taking verbs early**: Empirically, complement-taking verbs are not the first verbs learned, but they follow on soon after; several different types of complement-taking verb (subject-equi, object-equi, and raising to subject) are acquired within a few weeks of each other, typically at an age range 18 - 24 months (Pinker 1984; Tomasello 1992). They then form an important part of the child's language ability.

A complement-taking verb such as *wants* cannot be acquired until the child has enough knowledge of a few other verbs to apply each of them, in comprehension of the complement, without a subject. As soon as she has, the complement-taking verbs can be acquired straightforwardly by the primary learning process, as described above. Thus complement-taking verbs cannot be amongst the first verbs learned, but they can follow very soon after.

**(E2) Children make few errors of control with complement-taking verbs:** There is widespread evidence that (with the exception of tough-movement and the exceptional *promise*; Chomsky 1969) for most complement-taking verbs, children make very few errors in assigning the subject of the controlled verb. They seem to learn this correctly from the very start of learning complement-taking verbs (Pinker 1984; Sherman & Lust 1993)

The equality of identity slots, between the wanter and the agent in the wanted scene, will hold in every learning example which the child gathers when learning the m-script for *wants*. The script intersection learning process discovers this equality of identities [2.1], and encodes it as a shared variable identity value [id:?A] in both the `posessing' scene and the subordinate `what is desired' scene [3.8]. The shared variable identity slot embodies the control relation: whenever the word m-script for *wants* is used in generating or understanding a sentence, this control relation will be enforced by m-unification [2.3].

Learning a shared identity is a very basic and necessary part of the primate social learning mechanism [4.1], and is robustly built into the m-intersection word learning mechanism; so it is not something we would expect children to get wrong. Once learned, this shared variable will ensure that the child only uses the word where the same identity - e.g. of wanter and of agent in the wanted scene - is intended. So the m-script theory predicts that such errors of control are very rare, as is found empirically.

**(E3) `Tough-movement' complements are acquired more slowly:** Carol Chomsky (1969) reported that children do make errors of control in tough-movement verbs, such as *Donald is hard to see* , at much later ages - up to age 10 they may think that Donald is doing the seeing. However, the control relations for these verbs may involve long-range dependencies (as in *Donald is hard to fool yourself you like* ), so cannot be learned by the direct mechanism outlined above. Since there can be arbitrary levels of embedding, there is no fixed place on the meaning script where the shared identity may occur; so a shared variable identity cannot be learned by script intersection of learning examples.

Learning to fix long-range dependencies requires different learning mechanisms, which generally require more semantic and pragmatic knowledge, and take much longer to master.

**(E4) Children make mistakes in inflection of embedded verbs:** Pinker (1984) has summarised evidence that, while children are using control relations very reliably, at the same time they are making frequent errors of inflection of the embedded verb - either using the bare form , or sometimes over-tensing.

Children can start to learn complement-taking verbs at a very early stage, when they have learnt just a few other verbs by the primary learning process [3.9] - possibly before they have started on the secondary learning process which will give them productive control over verb inflections [3.11]. In this case, it is not surprising that a patchy knowledge of verb inflections prevents them from learning which complement-taking verbs require which inflections - and they learn m-scripts which are very permissive about complement verb inflection.

**(E5) Verbs acquired intially without complementisers take some time to acquire them:** Bloom et al (1984) observed that early-acquired complement-taking verbs - which require the complementiser *to* but were acquired without it - carry on being used without the complementiser, some time after other later-acquired verbs are used correctly with *to*. The complementiser is not retro-fitted to all verbs which require it at the same time.

For the early complement-taking verbs, which are acquired before *to* is recognised either as a preposition or a complementiser, it seems that the sound *to* is either ignored or incorporated in the verb sound, as in *wanna* . Recognition of *to* as a preposition may help its recognition elsewhere as a significant sound. Once *to* is recognised in the sound stream, we would expect it to be correctly acquired in the m-scripts of new complement-taking verbs which require it, as is observed.

Its slower incorporation in the early complement-takers can be accounted for much like the over-regularisation of morphology. For each `old' verb, when he starts to hear *to*, the child can learn a new m-script which has the complementiser - but initially regards the old and new m-scripts as alternatives. It is only later, by accumulating negative evidence [3.10] of the form `where I would have used *want* , an adult said *want to*`, that he learns that the form without complementiser is not used. This may require a lot of learning examples, which (as for inflectional over-regularisation) may take some time to gather [3.6].

**(E6) Verbs with optional complementisers are correctly learned:** A challenge for theories which use logical, all-or-nothing learning `choice points' is to frame the choice criteria so they can both (a) correct early learning errors or over-generalisations, and also (b) learn that for some verbs, such as *help*, the complementiser *to* is optional. It is hard to formulate a categorical learning rule which manages both these at the same time (Pinker 1984). For the Bayesian learning theory, this presents no difficulty, as it accumulates evidence incrementally without discrete choice points:

Optional complementisers: Examples of *help* both with and without the complementiser *to* will be heard; so two distinct m-scripts (with and without the complementiser) are learnt by the usual primary mechanism [3.9]. As both forms continue to be heard, no negative evidence will accumulate to eliminate either of them [3.6]. Both forms will persist.

Obligatory complementisers: If some verb with an obligatory complementiser is erroneously learnt without it (e.g. before the child has segmented the complementiser), then as soon as the child is sensitive to the complementiser, negative evidence against the no-complementiser form will accumulate [3.10], leading to its unlearning.

**(E7) The `wanna' contraction is not made over a gap**: Crain (1991) has shown that while children readily contract *want to* as *wanna* in questions such as (a) `*What do you want to_ eat _ ?'*, they do not contract it in questions of the form (b) `*Who do you want _ to brush your* *hair*?', because the trace of the missing subject in the complement (shown as _ ) separates *want* from *to.*

As described above, there are two distinct m-scripts for *want*. One, for same-subject wishes, matches partially understood sentence fragments of the form <[entity] *want to* [event]>; while the second, for other-subject wishes, matches <[entity] *want* [entity 2] *to* [event]>.

The same-subject *want* is used in (a); since it is learned from examples in which *want* and *to* are contiguous, they can be learned as an undivided, contractible sound *wanna*. The other-subject want is used in (b); but since it is learned from examples in which *want* and *to* are separated by the other subject, they must be learned uncontracted; so they are much more likely to be said uncontracted.

**(E8) The rare `promise' control structure is learnt more slowly**: This phenomenon does not have any neat and crisp explanation in the theory, but nevertheless can be understood. A clear analysis of the various issues has been given by Pinker (1984). It seems that in this case, the true account may be rather multi-faceted and messy, including factors such as:

- The form which children get wrong until age 10 or so (*Sam promised Fred to leave*) is a rather rare use of the word, compared with simple `indicative' forms such as *I promise she will* or the usual *I promise to pay....*
- *Promise* is a `theory of mind' word which requires at least a sophisticated theory of mind, and a rather sophisticated construal of the situation, to learn correctly. It may be usefully compared with the *imply/infer* pair, which many adults do not learn `correctly' all their lives; so much so that the meanings of these words may be said to have changed from the current `dictionary' forms.

In summary, the key facts of acquisition of control relations are accounted for directly and naturally in the m-script theory.

# 5.6 Auxiliaries

The English auxiliary system is complex and irregular, a product of fairly recent language change. It is a challenge for any learning theory to explain how the child learns to use modal auxiliaries (such as *can, could, will, must* , and *should*), perfect *have*, progressive *be,* and passive *be* - including the complex and irregular rules about how they can, and cannot be used together, and in questions. As Baker (1979) and Pinker (1984) have noted, any simple syntactic theory, which makes broad generalisations, tends to over-generate many illegal forms.

In this theory, the account of `what is to be learned' about auxiliaries is very close to Pinker's (1984) account - except that while Pinker retains separate phrase structure rules, in this theory the phrase structure rules are built into the verb m-scripts.

Pinker proposes that the allowed used of each auxiliary verb are learnt in `paradigms' with two key dimensions - the verb morphology (which can have four values : infinitive, finite, perfect participle, and progressive participle) and sentence modality (which also has four possible values - neutral, inverted, negative and emphatic).

Each auxiliary selects rigidly for just one morphology of the verb in its complement (infinitive for *can*, perfect participle for *have*, etc.) ; but the auxiliaries also differ in their own possible verb morphologies, as summarised by Pinker (1984):

| **Infinitive** | **Perf-part** | **Prog-part** | **Finite** | |
|---|---|---|---|---|
| **Can** | - | - | - | can |
| **Have (perf)** | have | - | - | has |
| **Be (progr)** | be | been | - | is |
| **be (passive)** | be | been | being | is |
| **walk** | walk | walked | walking | walks |

The entries in this table (if learnable) would account for many of the regularities about which combinations of auxiliaries are allowed, such as *John must have left* or forbidden , such as *John must can leave*.

Auxiliaries and normal verbs also differ in the sentence modalities they can take part in. For the `finite' verb morphology these are (again from Pinker 1984):

| **Neutral** | **Inverted** | **Negative** | **Emphatic** | |
|---|---|---|---|---|
| **Can** | can | can | can't | can |
| **Walk** | walks | - | - | - |
| **Do** | - | does | doesn't | does |
| **Need** | - | need | need | - |
| **Better** | better | - | better | better |
| **Used to** | used to | - | - | - |

The irregularity of the tables shows the irregularity of the auxiliary system. But if these tables, and their entries, can be learnt without over-generalisation, then auxiliaries can be used correctly.

To see how the table entries are learnt in the m-script theory, consider the word m-script for a typical English auxiliary, *can*, shown in figure 5.2.

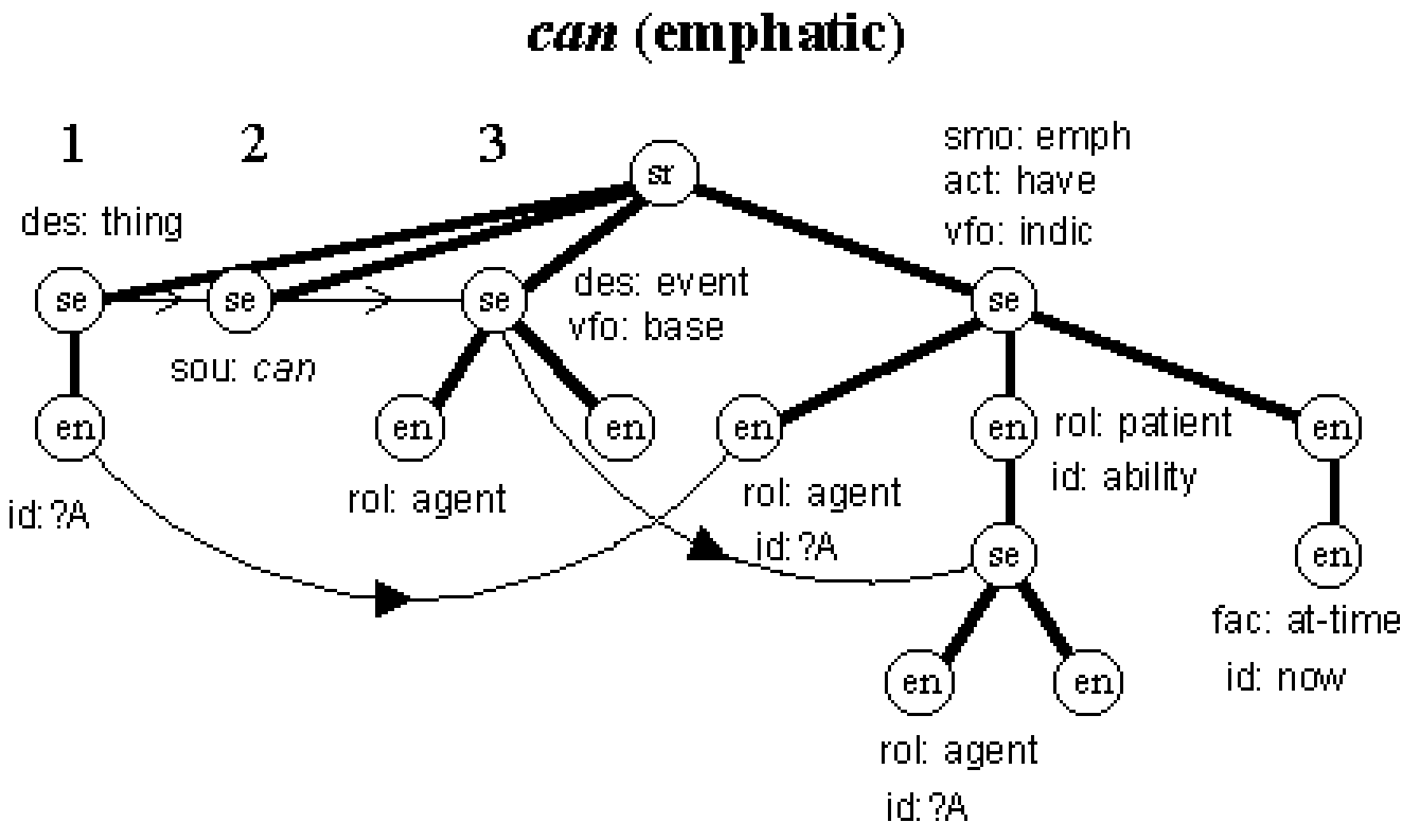

*Figure 5.2: m-script for `can'*

The left branch of this structure defines the syntax of the word *can*, which requires three input scenes. In left-to-right order, scene (1) describes an entity (the subject), scene (2) has just the sound `can', and scene (3) describes an event (the meaning of the verb complement).

The left branch describes the meaning of *can* - that the subject of *can* has an ability to partake in an event scene (the legs of the man) as agent. The shared variable identity `?A' embodies the control relation - that the entity appearing before `can' in the sentence (scene(1)) is the same as the owner of the ability, and the same in the agent on the event.

In understanding the emphatic sentence `My friend *can* see you', the sequence of events is:

- the words *my friend* are processed by m-unification to give a scene (A), describing a person.
- the sequence *see you* is processed, first applying the m-script for *you* and then m-unifying the m-script for *see* (with a patient *you*, but without a defined agent entity) to give a scene (C) describing a seeing event. Since *see* is the base form, this has a slot `verb form = infinitive'.
- The m-script for *can* is applied, matching scenes (A)(1), and (C)(3), and creating the full meaning scene (D) from (4) (its right-hand branch). Since *can* is emphatic in this m-script, the result has a slot `sentence modality = emphatic'.

For generation, the m-unifications would go in reverse order, right to left. Therefore we can find all the information required for Pinker's paradigms in the m-script for *can* (emphatic):

1. The slot (vfo:inf ) on scene 3 means the subordinate verb (*see* above ) must be infinitive.
2. The slot (vfo:indic) on scene 4 in the right branch means that *can* itself delivers a finite indicative result.
3. the slot (smo:emph) means that the result is emphatic.

Therefore the *can* m-script defines one table entry in the multi-dimensional paradigm table.

Auxiliary m-scripts like that for *can* are learnt by the primary learning mechanism described in section 3 - collecting learning examples (SMS) in which emphatic *can* is the only unknown word, and m-intersecting them together. The full meaning and constraints of emphatic *can* - including the paradigm slots above, and the trump links - are projected out by this process.

Note that the m-intersection also learns the control relation for the auxiliary (that if X can do something, it is X doing it) just as it does for verbs like *want*; the shared identity `?A' appears on the three nodes of the m-script, because in every learning example, the same individual appears on those three nodes.

The interrogative form of *can* is learnt as a separate m-script, with different constituent order on the left branch. So three m-scripts are learnt for *can* - neutral, interrogative, and emphatic.

Therefore the English auxiliary system is acquired by learning a large-ish number of independent m-scripts - filling out just the entries in the paradigms which are heard in adult speech.

**(F1) Highly irregular English auxiliaries are learnt reliably**: Children learn auxiliary verbs fairly early, and do so reliably (Pinker 1984). They obey Pinker's selection rules, summarised above, form the start. These rules can be encapsulated in m-scripts for each English auxiliary, with the verb morphology and sentence modality reflected as semantic slots in the meaning scripts.

The m-scripts for auxiliaries can be learnt by the usual primary learning mechanism, from (of the order of) six examples each [3.1]. It is not hard for a child to gather six learning examples for each auxiliary verb; from then on, she can use it reliably and effectively in comprehension or production.

**(F2) Over-generalisation of auxiliaries does not occur**: Children rarely seem to make auxiliary errors of the `John must can go' variety which would follow from over-generalisations - either amongst auxiliaries, or between auxiliaries and other verbs.

One may ask - having correctly learnt the auxiliaries by primary learning, does the theory predict that children might form broader generalisations by secondary learning, which could lead to over-generation ? In the Bayesian learning theory, any further generalisation (learnt by the scondary process) must pass a test of statistical significance [3.2, 3.11]. While I have not worked out the numbers for English auxiliaries, it seems almost certain that any broader generalisation would not pass this test, given the small number of auxiliaries and the absence of any `true' regularity across them (apart from the similarity of all modal auxiliaries, which might well be learned).

**(F3) Errors of Auxiliary control almost never occur**: For instance, children are never observed to say `he can see' meaning `he can be seen'. In this theory, the control relations of auxiliaries are learnt just like the control relations of other complement-taking verbs; if two entity nodes have the same identity in all learning examples, then the equal identity of those entity nodes is assured in the learned result, discovered by m-intersection [2.2]. Learning shared identity values is a core aspect of the learning theory, needed to learn any verb; so it is not surprising that it works reliably for auxiliary control relations.

**(F4) Children often fail to invert subjects and auxiliaries in Wh-questions**: Compared with virtually non-existent errors of control, the significantly-occurring failures of inversion in wh-questions such as `how you did that?' call for explanation. I have not found any neat or decisive account, but nevertheless the error can be understood. Compare the sentences:

(a) He told her he was hungry.

(b) He was hungry.

(c) Tell me who did it.

(d) Who did it ?

(e) Tell me where you hid it.

(f) *Where you hid it ?

From the comparison of (a) and (b), or (c) and (d), it may seem to the child that any sentence or question can be freely nested inside matrix verbs like `He told her', `I know', or `Tell me'; this knowledge can be embodied in a simple m-script, learned by m-intersection. Such an m-script, given (e), would then licence (f). In this account, non-inversion of auxiliaries is just a special case of non-inverted questions (like (f), which involves no auxiliaries). To correct this error, the usual slow process of accumulating negative evidence [3.10] is required.

**(F5) Complement verbs are sometimes overtensed**: Pinker (1984) notes a significant level of over-tensing errors in complement verbs after auxiliaries, as in *Can you broke those ?* - particularly with the auxiliary *do* .

Again, there is no single neat or satisfying account of this effect, but there are several possible contributory factors. One of these, as in (E4) and following Pinker (1984) is that when the child is first learning auxiliaries, he has not yet mastered regular verb morphology - and so cannot reliably distinguish between tensed and infinitive forms of all verbs. This means that constraints such as `*can* always requires an infinitive verb' may not be learned reliably at this stage - and even if it is, the finite and infinitive forms of some verbs may be confused. This agrees with the finding that overtensing errors are more common for irregular verbs.

# 5.7 Alternating Verb Argument Structures

The fact that some verbs have alternating argument structures (as in *give me the book* and *give the book to me*) has been the origin of a fascinating puzzle in language acquisition - the fact that children make these alternations productively, and while they do sometimes over-generalise, they eventually correct this.

The account of alternating verb argument structures in this theory draws heavily on Pinker's (1989) account - in particular, using his analysis of the various verb meaning structures essentially unchanged. It differs in the following ways:

- Pinker's account uses innate linking rules between grammatical functions and semantic roles. This account works directly with semantic roles, having no linking rules.
- Negative evidence is allowed, so the child can eventually correct over-generalisations, and learn idiosyncratic non-alternating forms.

The core of this account is the use of m-scripts to embody the *non-linguistic* knowledge of possible alternations.

When the same external situation can be construed in two different ways Ñ when it can be represented by two different but related scripts Ñ there is an m-script which describes the relation between the two scripts. This m-script is a function from scripts to scripts, which can be applied by m-unification. Applying it to one construal delivers the other as result, in either direction. Such m-scripts are a part of our general, pre-linguistic, social intelligence, helping us to represent and reason about social situations [4.1, 4.2].

Consider, for instance, the locative alternation, as in the pair of sentences *John sprayed paint on the wall* and *John sprayed the wall with paint*, whose meaning scripts are Sand Srespectively. Sdenotes that John acted on the paint, making it move onto the wall, and Sdenotes that he changed the state of the wall, using the paint as an instrument. We recognise that these two different meanings can be exchanged, and the verb *spray* has two interchangeable argument structures.

Levin and Rappaport (1986) have identified 152 locative verbs, some of which have both locative forms, and some do not. Pinker (1989) has shown how these verbs can be grouped into 'narrow conflation classes' based on semantic criteria, so that some classes have the alternation, and others do not.

The m-script A which relates the two construals Sand Sis shown in figure 5.3. This m-script embodies the locative alternation in meaning structures; its two branches contain the respective meaning structures as proposed by Pinker, translated into the script notation.

The left-hand branch will match a script such as Sof the form "?A propels ?B on a path which ends in a spatial relationship ?R with object ?C", while the right-hand branch matches a script such as Sof the form "?A acts on ?C, using ?B as an instrument, with effect that ?C becomes covered in spatial relationship ?R".

A can be used to transform reversibly between Sand S. If A is m-unified with S, so that the left branch of A matches S, then the right branch of the result will be S; writing this as a function, S= A(S). Similarly by m-unification in the reverse direction, S= A.

This m-script A which embodies the locative alternation can be learnt by the same mechanism of m-intersection which is used for language learning Ñ by observing a number of occasions in which scripts Xand Xare alternative construals of the same situation, combining Xand Xbelow a common script node, and m-intersecting these learning examples together.

Although the mechanism for learning A is very like the language learning mechanism, it is in fact completely non-linguistic. A child with no vocabulary can learn A by observing situations, making the two construals, and m-intersecting the results. It is a piece of general social knowledge, learnt by the mechanisms which evolved to learn such things before language existed. Knowing A has nothing to do with linguistic competence.

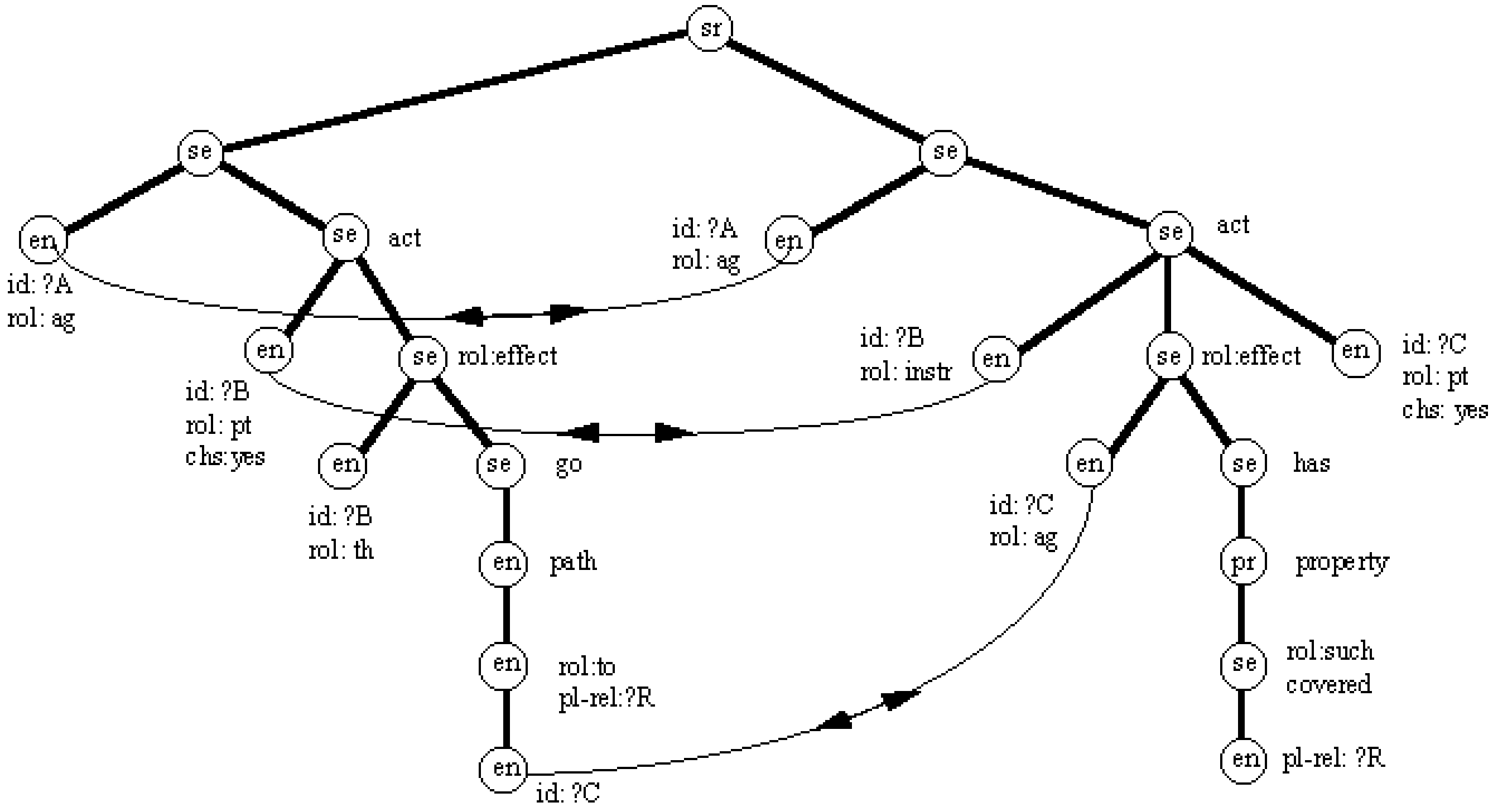

*Figure 5.3: M-script A which embodies the locative alternation for one of its narrow conflation classes.*

It is not yet clear how broad or narrow are the alternation m-scripts which we learn. Do we learn just one broad m-script to do all locative alternations, or do we learn several narrow ones for different types of locative alternation ? I shall assume for simplicity that we learn just one broad alternation m-script, like figure 5.3 above.

What then distinguishes verbs which do alternate argmuenr structures, andverbs which do not ? The answer relates to the question of what elements of meaning are commonly encoded in verbs (and therefore justify coining a new verb). English transitive verbs may encode something about the path of a thing caused to move (as in the left branch of figure 7) and they may encode something about the resultant state of a direct object (as in the right branch of figure 7).

As Pinker (1989) has noted, the verbs which alternate do both of these at the same time. *Spray the wall with paint* tells us both about the manner of motion of the paint and the final state of the wall; so *spray* alternates. *Fill the bucket with water* tells us about the final state of the bucket, not about the manner of motion of the water; so *fill* does not alternate. In order to alternate, a verb must say something about its direct object (beyond what is said by some simpler verb such as *move* or *change*) in both argument configurations.

With this background, we can understand some empirical findings about children's use of alternating verbs:

**(G1) Alternate argument structures for the same verb are learnt early and without confusion**:(Pinker 1989) The basic learning process is proposed to be conservative - simply learning each alternate form from examples, by the usual primary learning process [3.8, 3.9]. For many verbs, both alternate forms will be learnt this way.

**(G2) Alternations of argument structure are in broad classes, yet respect narrow-range rules** : (Levin & Rappaport 1986; Pinker 1989) The broad-range classes include all verbs whose meaning scripts can be alternated (by m-unifying with the broad, construal-changing m-script [4.1]), to make an alternate construal of the same situation. These verbs have two `object-like' arguments (e.g. the wall and the paint, in the examples above). In each construal, one is the direct object and the other is an adjunct.

However, for many verbs in such a class, in one of the construals the verb says nothing interesting about its direct object, beyond what is already said by some simpler verb. Without its adjunct, in this construal it is a rather useless verb. In these cases there is no alternation; for instance, *fill the glass with water* cannot become *fill the water into the glass*, because *fill the water* says nothing more (about the water) than *move the water*. The narrow conflation classes (Pinker 1989) are defined by this semantic criterion.

In language generation, the adjunct meaning is stripped off first, so children never have occasion to say *fill the water into the glass*; for the meaning without the adjunct, the simpler verb *move* would be sufficient [2.3].

There may also be a uniqueness principle which prevents children from storing an m-script for the forbidden alternation, because it has the same meaning as *move*.

**(G3) Children use the alternations productively**: There is widespread evidence that children coin new alternations, for the dative, locative and causative alternations (Pinker 1989). Gropen et al (1991) have shown that children will use either alternate form for novel locative verbs, with preferences depending on the detailed meaning.

Two general (broader than single-word) m-scripts are involved:

- Knowing one meaning construal of an alternating verb (in a broad conflation class), children can make the other construal by applying an alternator m-script to the verb meaning script.
- Having learnt several verbs in a narrow conflation class with one argument structure, children may use the secondary learning process [3.5] to learn a broad `argument structure' m-script expressing the fact that any verb in that conflation class may have that argument structure. They can then apply this rule productively to new verbs in the narrow class.

Thus the children in Gropen et al's (1991) experiment may use (a) an alternation m-script to get the meaning they observe into the required form, where they can then (b) apply the broad `argument structure' m-script to infer that a novel verb *keat* has an argument structure (which they use in production) for that meaning.

However, it is consistent with the general `bottom-up' nature of learning in this theory that children are rather conservative in their learning and application of these argument structure m-scripts, as is observed.

**(G4) Children make over-extensions, but correct them**: While children are fairly conservative, nevertheless they do apply the alternations productively, and sometimes over-generalise (particularly the causative) in examples like *Daddy giggled me*.

In this theory, all such errors are corrected by gathering negative evidence, when children internally generate a part-sentence and compare it with what an adult said [3.10]. In this way, they observe that “Where I would have said *Joe giggled her* Mummy said *Joe made her giggle*”. After gathering enough examples, the child learns that *giggle* has no causative alternation.

**(G5) There are `idiosyncratic' non-alternators, which children learn** : In spite of the fact that most alternator verbs can be understood as belonging to Pinker's narrow semantic classes, giving a semantic guide to alternation, nevertheless we expect that over time the frontier between alternation and non-alternation may change. At any time there are bound to be a few `idiosyncratic' verbs whose alternation (or not) seems to have no good semantic basis. Compare `give' and `donate'; or for the causative alternation, `He burped the baby' versus `he cried the baby'. Some alternators are just a matter of usage.

Children need to learn these eccentric non-alternators They can use the mechanism of negative evidence [3.10] to do this.

**(G6) Children learn passives of `action' verbs before others**: Cross-linguistic data on acquisition of the passive shows children acquiring it fairly early in many languages, such as English passives late in the third year (Bowerman 1973), and earlier in some languages where the passive is prominent - e.g. Sesotho in the second year (Demuth 1989,1990). However, it seems that in both these languages, children acquire the passive form for prototypic action verbs (*hit*, *kissed*) earlier than non-action verbs (*like* or *surpass*) (Maratsos, Kuczaj and Chalkley 1979; Pinker, Lebeaux and Frost 1987; Demuth 1990)

In Pinker's analysis, which this theory follows, the passive is another (rather broad-range) alternation. It is so broad that it seems it must be learnt in some productive form; we could not learn all passives by hearing each verb in passive form several times. Like the other alternations, it rests on a piece of pre-linguistic knowledge - roughly that `If A does X to B, then B goes into a state of having had X done to him' - which can be expressed as an m-script. Therefore the mechanisms for learning it, for making productive generalisations of the passive, and for unlearning the passives of non-passivisable verbs, are all as described above.

It is likely that children can acquire the pre-linguistic m-script knowledge that `If A does X to B, then B goes into a state of having had X done to him' more easily and earlier for action verbs (where X may make a visible difference to B) than they can for others; this would account for the earlier acquisition of action passives.

# 5.8 Pronouns, Gaps, Quantifiers and Movement

In considering the use of pronouns and long-range movement phenomena, we come close to the limits of this theory as currently formulated [2.6]. Its account of these phenomena goes beyond the basic m-unification process of generation and production, requiring extra processes (to resolve pronoun identities, quantifier scopes and gap identities) going on in parallel with m-unification, to construct a sentence meaning. We need to make extra `procedural' assumptions about how people do this, before analysing how they learn to do it.

However, there are still some useful insights, and the challenge should not be ducked. A focus of interest in learnability of anaphora, pronouns and long-range movement has been the status of universal constraints, such as Chomsky's principles A,B and C. If these are universal, does the child need to learn them at all ? What must she learn in order to use them ? It turns out that in this theory, there are reasons to expect universal constraints, with consequences for learning. First we sketch how the phenomena are handled in language use.

First consider **anaphora**. The m-script for a reflexive pronoun such as *himself* has a form like a noun m-script - with no trump links, but with a right-hand branch entity describing a male of unknown identity. This identity is represented by a variable such as `?A'. There is an `instruction' slot on the m-script (which becomes an instruction slot in the meaning script) to be activated as soon as the entity is m-unified into any verb meaning scene, as one of the verb's arguments. At that point, the identity `?A' is (by instruction) to be equated with the identity of the agent of the verb meaning scene. So in *Charlie hurt himself*, immediately after the m-script for *hurt* is applied, the identity `?A' is set to Charlie's identity - giving the required effect. This implements Principle A.

Third person **pronouns** such as *him* are treated in the same way, but now the instruction is that the variable identity must not be unified with the verb agent - or with any other entity immediately beneath the verb meaning script. It must, at some time later, be unified with the identity of some other entity, usually from outside this script. This achieves the effects of Principle B. However, finding the right entity to unify with the variable entity requires some heuristic search of likely `nearby' entities.

Next consider **relative pronouns** like `who'. In the expression *the boy who liked fish*, the sequence of m-unifications in understanding is :

(1) *the boy who liked fish*

(2) [entity 1] *who liked fish*

(3) [entity 1] *who liked* [entity 2]

(4) [entity1] *who* [event]

(5) [entity 3]

In the step from (3) to (4), the m-script for *like* (which normally requires both agent and patient arguments) is m-unified without an agent argument - consuming a `gap'.

Relative pronoun m-scripts like *who* carry an instruction, which says in effect `look for any variable identity inside the [event] script, and unify it with the identity of the initial [entity]. Thus the boy becomes the agent in the *liked* scene, as required.

Had the phrase been *the boy who I thought liked fish*, as before the *liked* scene has a variable identity for its agent; and because the instruction is to look for any variable identity inside the event scene, even when *liked* is nested inside *thought*, its agent is equated to the boy - resolving a long-range dependency.

Finally consider **quantifiers** such as *every*, *each*, and so on. The meanings of quantified sentences are captured by attaching quantifier slots to script nodes, and equating certain identity slots within these script trees (as bound variables); for instance, in *every boy loves his mother*, the quantifier `every ?X' appears above a script meaning tree for "boy ?X loves ?X's mother". To produce this meaning, a quantifier slot meaning `every' must move up the script tree (from where it was attached to the `boy' entity) and three identity slots ?X must be equated.

Handling any of these phenomena requires significant procedural extensions to the basic m-unification method of language understanding. They all require some kind of heuristic tree-searching procedures to be triggered by specific word m-scripts. Corresponding extensions are required for generation. I have implemented these extensions in the program which handles a fragment of English, and they work - that is, they do a fairly good job on most typical sentences. They do not yet handle all the hard examples studied by linguists.

It is a plausible assumption that a small number of these tree-searching and matching procedures are an innate pre-linguistic part of our social intelligence; representing things another agent does to himself, or searching for unknown entities, or quantifying over individuals, may often be required in social reasoning. Some language universals can be understood as innate limitations of the tree-crawling operations - for instance, that they cannot cross certain barriers (Chomsky 1986) in the script trees. We also assume that instructions to do these procedures can be attached to certain script nodes by word m-scripts.

To make these learnable, we require extensions to the learning theory. Possibly whenever the child finds a variable identity in a meaning script (with little else known about the entity) this automatically triggers a trial of some of the `search and match' procedures; and if one of them works, the instruction to do it is attached to an appropriate node. Then m-intersection of these learning examples will leave an `instruction' slot on some node of the word m-script, as appropriate for `himself' or `him' or `who'.

That is a sketch of an extended learning theory which I have not fully developed. To summarise:

- Handling these constructs requires heuristic tree-crawling operations, which may have innate limitations.
- These operations (and their limitations) have not been worked out as fully or as neatly as the m-script algebra operations, or validated against extensive language data
- Instructions to do the operations are attached to the m-scripts for pronouns, quantifiers, etc., and so get put onto scripts.
- We have only a sketch of how these instruction slots might be learned.

However, even with just this sketch, we can start to understand some learning phenomena:

**(H1) Anaphors and pronouns have complementary binding domains:** Across many languages, it seems broadly that anaphors (reflexives) must be bound within some domain, and pronouns must be free within the same domain - while the extent of the domain may vary across languages (Kapur et al 1993). The m-script theory enables us to understand two aspects of this phenomenon: its existence and its learnability.

- **Existence:** It is communicatively efficient to have two sets of constructs with touching, but not overlapping, domains. By not leaving a gap in the middle, languages do not leave speakers unable to express some meanings; and by not overlapping, they do not create an unnecessary source of ambiguity. Therefore we may expect the pronouns and anaphors of any language to evolve language-wide towards complementary domains, by the process of m-script evolution described in [4.3, 4.4].
- **Learnability**: If the domains of anaphors and pronouns are complementary, then each may be learnt using the other as negative evidence. Since non-uses of *him* are often uses of *himself* the child may gather this evidence, by the mechanism of [3.9], effectively observing "where I might have said *him*, an adult said *himself*" - and consolidate several examples into an exception rule [3.6].

However, we need to assume that there is some way of specifying, in the m-script for a pronoun or reflexive, what script domain is to be searched for possible referents, and what possible referents within it are eligible; and that this specifiable information is also learnable by the m-intersection mechanism (or possibly an extension of it)

**(H2) Reflexives are used correctly before pronouns**: This has been established by observations in English, Chinese, Dutch, .....

Even though languages differ in their use of reflexives (e.g. in some languages they may be bound to subjects or objects, in other languages only to subjects), in all languages the reflexive must be bound within some domain (= subtree of the meaning script) while pronouns must be bound outside it. Therefore finding the right referent for a reflexive is always an easier task than finding a pronoun referent, and is likely to provide more successful positive learning examples early on, when the child's linguistic knowledge makes it hard for him to construct large meaning scripts. So we expect reflexives to be learnt first.

It is even possible that fully learning the pronoun rules requires a knowledge of reflexives, so that Principle B is discovered from negative evidence - that adults use reflexives (where the child may silently generate a pronoun), gathering negative evidence as in [3.9].

**(H3) Pronoun reference principles have irregular edges**: All attempts to systematise the rules for pronoun reference seem to have an unsatisfactory feel about them - that we will always be discovering some new exceptions and complexities; and that in any case, usages will change over time. It seems wrong that the rules of pronoun usage should have to have neat edges, just because of restrictions in our learning theory.

In this theory, as it is possible to learn from negative evidence, any idiosyncratic fact of usage (e.g. that pronoun X cannot be identified with entities of a certain type in certain circumstances) can in principle be learned in this way - removing an artificial constraint to `neat edges'. We need to assume that the negative evidence mechanism extends to the rather special `referent finding' information attached to pronoun m-scripts.

**(H4) Some constraints on long-range movement are known from an early age**: For instance, de Villiers has shown how children from age three onwards can recognise that in `*When did she say she ripped her dress*?' the *when* may refer to the saying or the ripping; but that in `*When did she say how she ripped her dress*?' it can refer only to the saying. They do this at an age where, it seems, they cannot possibly have gathered enough evidence about this (rather rare) type of question to have learnt the constraint. However, equally perplexing, in the example above many young children will answer the nested *how* question rather than the intended *when* question; this is so in several languages. De Villiers, Roeper and others have explored a host of similar issues in laboratory experiments, revealing fascinating limits to children's' performance.

If we frame the procedural, tree-crawling aspects of the m-script theory in certain `obvious' ways, then some of these constraints emerge naturally as things not doable by the mechanisms. However, I have not developed a single, compelling and economical account of how the tree-crawling operations are done, and it is quite possible that other formulations would make `forbidden' operations feasible. This is an area for further investigation.

**(H5) Ross' Island Constraints are always obeyed**: Two of Ross's (1967) island constraints are that relative clauses and coordinate constructs act as barriers to *wh*-movement. This forbids sentences such as *What did you see the man that wore _* ? and `*I like fish and chips. What do you like _ and chips* ?'. It seems that children almost never violate these constraints in spontaneous speech; and De Villiers and Roeper (1995) have confirmed experimentally the relative clauses act as barriers to *wh*-movement.

In the m-script theory, when a gap is encountered in a relative clause, the m-script for the verb of the clause is applied with a missing argument entity. This leads to an unidentified entity (with a variable identity slot) in the verb meaning scene. Later, after the m-script for the relative pronoun is applied, another variable identity (from the pronoun meaning script) must be equated with this identity; there is a tree-search of the meaning script below the relative pronoun to find the appropriate entity to equate (so allowing long-range movement).

Although there is not space to give details here, both coordinate constructs and intermediate relative clauses interfere directly with this tree-search, making it infeasible in any simple form; so it is likely that these barrier phenomena (forbidding *wh*-movement from relative and coordinate constructs) arise from innate, pre-linguistic constraints on the tree-searching process. In this case, we would not expect children ever to violate the barrier constraints.

# 5.9 Bilingualism and Language Change

It is estimated that nearly half the world's population is functionally bilingual, and that most of these are `native speakers' of their two languages (Wolck 1987,1988). So bilingual language acquisition is not a peripheral, esoteric issue; it is a concern for any learning theory.

**(I1) Two or three languages can be learnt simultaneously**: Many bilinguals are exposed to both languages from birth, and acquire them at normal rates. There seems to be no limit on how different are the two (or sometimes three) languages acquired. This presents difficulties for some theories - for instance, for Principles and Parameters theories - where multiple sets of parameters must be postulated, losing much of the attractiveness of the theory.

In the m-script theory, acquisition of multiple languages is straightforward; no change to the mechanisms is required. Hearing sentences in one language, children learn the m-scripts for words, and so learn the syntax and semantics of the language; similarly for the other language. Apart from the word m-scripts, there is very little to learn. Two separate sets of m-scripts are acquired, and children use social/pragmatic knowledge to decide which m-scripts to use on which occasion. There is no extra difficulty even if the two sets of m-scripts embody very different grammars.

**(I2) Children learn overlapping vocabularies for two languages:** It was once believed that bilingual children go through an early `one language' stage in which, for instance, they do not learn words with the same meaning in both languages. Following more detailed and careful studies, this is no longer believed; young bilinguals' vocabularies contain pairs of words with the same meaning, from the earliest stages.

As was described above, the m-script theory can account for the lack of synonyms in languages without requiring a uniqueness principle, but there are independent grounds, both theoretical and empirical, for supposing that there is in fact a uniqueness principle at work in language learning.

This raises the question of how bilinguals learn words in two languages with the same meaning. It is likely that they learn very early a socially-conditioned `language context flag' which serves to distinguish the lexical entries for their two languages, and allows them to have distinct storage locations.

There is evidence that bilingual children distinguish their two languages phonetically very early, before they know any words; so throughout the course of word acquisition, this phonetic distinction can define two distinct contexts, and word meanings can then carry an implicit `context flag' which effectively breaks the uniqueness.

**(I3) There is no evidence for a single grammatical system early in learning two languages**: The Single System hypothesis (Volterra & Taescher 1978) proposed that bilingual children initially acquire just a single syntactic system along with a single vocabulary. Over the years, evidence has accumulated against this hypothesis, leading in stead to the Separate Development Hypothesis (SDH), that bilingual children acquire two syntactic systems from the start. There is currently little or no evidence against the SDH, and much evidence for it (de Houwer 1993).

This theory inevitably predicts the Separate Development Hypothesis. Since the syntax of each language is entirely embodied in the m-scripts of its words [2.3], as the child acquires the words for each language, she must inevitably acquire each syntax as well.

**(I4) The course of bilingual language learning is very similar to the course of monolingual learning** : All evidence suggests that the bilingual child learns both languages by going through just the same stages as a monolingual child, in the same order. This is so even for children bilingual in a spoken language and a sign language.

This result again is an inevitable (if vanilla-flavoured) prediction of the m-script theory, where the course of language acquisition is defined by the acquisition of word m-scripts. There is no good reason why a bilingual child should acquire the m-scripts for words of one language in any different sequence from that of a single-language child; so the sequence of language development stages is expected to be the same. (In a theory with a separate syntax-learning component, the presence of two syntaxes might well delay syntax relative to word acquisition.)

**(I5) Code-switching is done most frequently with nouns**: Bilingual children are generally sensitive to the needs of their listeners, using the appropriate language and only mixing the two when a mixed language context is appropriate (e.g. with another bilingual). When doing this, they code-switch for a variety of reasons, often within sentences. It is observed that code-switching occurs most with nouns.

Since (unlike other parts of speech such as verbs) noun m-scripts have zero `valency', requiring no other meaning elements in order to m-unify, they are the most easily substitutable elements in a language. The presence or absence of case inflections is the only impediment to noun-switching, and even that can be tacked on.

**(I6) Neighbouring languages do not completely intermix**: It is a little-remarked puzzle that neighbouring languages, which may share a long `frontier' of bilinguals, do not just diffuse into one another like two gases in a bottle. Why does this not happen ? Why are French and Dutch still distinct ?

In terms of this theory, if word m-scripts can propagate freely through generations by the m-intersection learning mechanism, why do they not propagate freely across language frontiers and beyond, intermixing the two languages ?

Part of the answer is social and behavioural - that people view their language as part of their identity, and often will not use words from other languages in order to be `one of us' rather than `one of them'. This is reflected in the fact that language change often diffuses outwards from areas of high socio-economic status - in a direction where the `social identity' resistance is least. However, this is probably not the whole story.

The m-scripts of a language evolve so as to align themselves into domains of regularity - sets of words with common syntactic structure, which work well together. They do this because any word which does not conform to the common structure will thereby be less usable [4.4] and will tend to die out. This same selection pressure therefore acts as a kind of `immune system' for a language, rejecting alien words which do not fit into the regular structures; so French word m-scripts cannot freely diffuse into a Dutch-speaking area. As with code switching, this resistance to alien forms is weakest for nouns. The domains of regularity extend over geographic space as well as over the space of meanings.

**(I7) Creoles form very rapidly from Pidgins** : When people of many different languages are thrown together, they soon develop a pidgin, which is a second language for all of them. The pidgin has very limited syntactic resources, and is very inefficient for communicating complex meanings. However, within as little as one generation, there starts to form a creole which is a very different language, and is the first language for its speakers. It is only learnt by children, while adults stick with the pidgin. The creole has a regular and productive syntax, and gives its speakers much better means to express complex meanings.

What is remarkable is the rapidity and reliability of the formation of creoles. This has been interpreted as evidence for a human biological endowment for language (eg Pinker 1994), and more specifically for a particular `bioprogram' form of language (Bickerton ). It may be evidence for such a large conjecture; however, it is also (much more directly) evidence for the rapidity and robustness of the language learning mechanism.

Productive creole forms can only spread rapidly through a population if they can start from small beginnings - if every learner can learn a construct, even at the stage when only a few speakers use it; when the signal may be masked by a lot of noise.

The learning mechanism of this theory evolved to learn useful social regularities in a noisy social milieu [4.1]; fierce social selection pressure over 20 million years has honed it to a highly robust Bayesian form, which can pick out a syntactic regularity even when it is diluted 20:1 or more in spurious noise [3.4]. It is just this rapid, robust learning that would be required to form a creole from a pidgin in a very few generations.

**(I8) Creoles use simple analytic forms to express meanings** : The remarkable resemblances between creoles in different parts of the world are evidence (in spite of controversy about the extent of influence of the `superstrate' languages) for some common processes going on in their formation. Bickerton (1984) has argued that this common influence is the `bioprogram' - an innate endowment for language which biases towards a certain form of language, seen clearnly in the creoles.

In this theory, the intepretation is different, and can be best understood in the `m-script evolution' picture [4.3]. Each language is a population of word m-scripts, and each word m-script replicates through a population of speakers by the learning mechanism. Language change is a process of evolution and selection of word m-scripts, analogous to natural selection of species - but operates much faster. The criteria for `fitness' of a particular word m-script are various, and include:

1. Ease of learning
2. Economy of communication
3. Reliability of communication
4. Fit with other common word m-scripts (regularity).

In just the same way, there are many different criteria for fitness of a species, and the criteria vary over time. When a piece of land has been devastated, certain fast-growing plant species can re-invade it rapidly, and thrive for a few generations until a more typical diverse ecology, with slow and fast-growing species, gradually establishes itself.

A pidgin is the language equivalent of devastated land, inhabited by only the most primitive word m-scripts. The development of a creole depends on which m-scripts can most rapidly invade and colonise this linguistic waste land. I suggest that, while for most mature languages the key factors determining word m-script fitness are (2) and (3), for a creole the dominant fitness criterion is just (1) - ease of learning. Ease of learning determines speed of invasion of the waste land. It is this altered form of m-script competition, rather than any innate bias in the human language endowment, which leads to the distinctive form of creoles.

Creoles are notable for the use of analytic forms which convey meanings in simple, separate chunks. For instance, tense, mood and aspect are conveyed by separate particles, rather than being built into verb morphology. Negation is conveyed by a simple pre-verbal particle.

The speed of learning in this theory is determined by the `six clean examples' heuristic [3.4]. If, for instance, past tense is conveyed by a separate morpheme for every verb, it can in principle be learnt from six good examples of past tense sentences - which will be heard much sooner than six examples of a particular verb in the past tense, as would be required if the past tense were built irregularly into verbs; and much faster than the secondary learning process, which would be required to learn a productive verb inflection for the past tense [3.11].

Children show a bias to these analytic forms, just because they are more easily learnable (Slobin 1985). The bias towards analytic forms, for easy learning, is even stronger in creoles.

**(I9) Tense, Mood and Aspect are expressed in order TMA for creoles, MTA for most languages** : Bickerton (1981) described a core Tense/Mood/Aspect system for creoles, which consists of pre-verbal particles marking A=[+nonpunctual] aspect, M=[+irrealis] mood and T=[+anterior] tense. These particles can be used in any combination, but always appear pre-verbally in the order TMA-verb. While individual creoles can have extensions to this scheme, the basic core TMA scheme applies to a high proportion of creoles of diverse origin (Arends et al 1994). However, for most mature languages this is not the preferred order; in a cross-linguistic survey Bybee (1985) found the dominant order to be MTA-verb, with tense closer than mood to the verb stem in 88% of cases. Why do creoles systematically differ from mature languages in this way ?

The difference can be understood as arising from three factors:

1. Easy learnability is the dominant factor which defines the form of creoles.
2. Constructs are easily learnable when meaning elements which are semantically close (in meaning scripts) are kept close together in sentences.
3. The creole realis/irrealis mood is a simple binary distinction on the top node of a verb meaning script; but in mature languages, modality is a more complex concept closely linked to the agent.

The creole irrealis marker seems to denote a binary distinction between an event that has actually happened and one which has not (which may be future, conditional, imagined, etc.). The tense marker denotes tense not relative to the time of speech, but relative to the event under discussion. Therefore it seems likely that, in a sentence meaning structure:

- Aspect (punctual/nonpunctual) denotes the internal time structure of the verb's action, and is therefore denoted internal to the verb meaning script.
- Mood (realis/irrealis) is a binary slot on the top node of the verb meaning script
- Tense (anterior/non-anterior) places the verb meaning script relative to som other time of reference.

In this case, aspect is most closely entwined in the verb meaning, next comes mood, and last is tense. In a sense, aspect is inside the meaning of a verb, mood is on the surface of the meaning, and tense is outside it. Aspect is like the shape of an object, and tense is like its position relative to something else; shape is clearly a more intrinsic part of the object itself.

For learnability, the semantically closest meanings should be expressed nearest the verb stem. This accounts for the TMA order of the creole verb-modifying particles; but then why do mature languages tend to have MTA order ?

In most languages, the irrealis side of the realis/irrealis distinction is expanded into a wide range of modalities - of desire, intent, potential, obligation, and so on. This range of meanings cannot be expressed by a simple binary slot on the top node of the verb scene. As we saw for English modal verbs [5.6], the modalities are expressed by more complex script structures, exemplified by the m-script for *can* in figure 5.X. *Fred can swim* is expressed by a meaning script which can be roughly paraphrased as `Fred possesses an ability for Fred to be in a swimming scene'; *Fred should swim* can be expressed as `Fred has an obligation....' , and so on.

These more complex scripts can express a wide range of modalities, as mature languages require; but in these meaning scripts, the type of modality (obligation, ability,...) is detached from the verb meaning scene, and is now closer to the entity nodefor the agent. Therefore, for ease of learning, complex modalities should be expressed in MTA order - as they are in most mature languages.

This pervasive difference in structure between creoles and mature languages arises from the different range of meaning scripts they can express, combined with a script structure/learnability correlation.

# 5.10 Extra Assumptions

Table 5.1 implies that 15 of the 101 comparisons with data require some extra assumptions, beyond the core theory, to get agreement with the data. The distinction between core theory and extra assumptions is clearly rather arbitrary; if some extra assumption is involved in the account of several pieces of data, and so starts to achieve some economy of hypothesis, then it may be incorporated in the core theory.

However, if for the moment we take the `core theory' to be that described in sections 2-4 of the paper, we can summarise the extra assumptions which are needed to get agreement with various pieces of data:

| **DATA** | **EXTRA ASSUMPTION** | |
|---|---|---|
| B19 | Children over-extend some words in production after having used them correctly | Secondary learning forms new nodes in the subsumption graph storage of word meanings, which make it easier in production to confuse words below that node. |
| B21 | Word meanings change by metaphor and metonymy | Meaning scripts have a `semantic field' slot which defines how they map onto other cognitive models (eg spatial). Changing the value of this slot creates metaphors. |
| B25 | Gender has little to do with sex | We can freely add a `sex' slot to the meaning scripts for inanimate objects. |
| C10 | Languages have regularities captured in X-bar syntax | Meaning scripts are more complex than the examples in this paper, in a way that supports semantic distinctions between different X-bar levels |
| D6 | In agglutinating languages, inflections are learnt from the outside inwards | Scripts with a time-order constraint `A immediately follows B' are easier to learn (or to use in production) than scripts with a time order constraint `A is somewhere within B'. |
| D7 | In agglutinating languages, | There is a value `not yet defined' (as opposed to |
| | children make no errors in ordering affixes | `unknown') for inflection-controlled slots on meaning scripts, which prevents reassembly of agglutinated inflections in the wrong order. |
| D9 | There is transient over-regularisation of irregular forms | There is a procedural mechanism whereby a negative rule `form X never occurs' can block the application of a productive rule which gives form X. |
| D10 | English noun plurals and past tense verbs are over-regularised with low frequency | Several possible accounts: e.g. the irregular exception overrides the general rule, except when tense information is not well-enough defined in the child's meaning script to trigger the overrule (when tense is an afterthought). |
| D11 | Specific Language Impairment affects regular morphology | SLI is a deficit in the secondary learning mechanism |
| E8 | The rare `promise' control structure is learnt more slowly | Several possible accounts: e.g rarity of the form in adult speech, or that the meaning depends on `theory of mind' variables which cannot be directly observed. |
| F4 | Children often fail to invert subjects and auxiliaries in Wh-questions | Children learn and apply a broad m-script which implies that any statement or question can be freely extracted from forms such as `tell me' (e.g. *tell me why you did it*). |
| F5 | Complement verbs are sometimes overtensed | When children are learning auxiliaries and other complement-taking verbs, they have not yet mastered the morphology of the complement verbs. |
| H1 | Anaphors and pronouns have complementary binding domains | Each reflexive or pronoun m-script has attached information which defines what parts of the meaning script may be searched for referents, and what referents are eligible; this attached information is learnable. |
| H3 | Pronoun reference principles have irregular edges | As for (H1) above; and the attached `referent-finding' information is learnable from negative evidence |
| H4 | Some constraints on long-range movement are known from an early age | Tree-crawling procedures for matching *wh*-gaps are innate, and have intrinsic constraints which forbid certain kinds of long-range movement. |

*Table 5.2: extra assumptions used in comparisons with data*

Several of these extra assumptions concern the form of the script meaning representation; further systematic investigation of script meaning structures may clarify some of these assumptions. Other assumptions centre on procedural aspects of the theory, particularly for language production.

# 6. Other Theories of Language Acquisition

This section contains brief remarks on other theories of language acquisition, comparing them with the m-script theory.

## 6.1 Pinker's Theory of Language Learning

Pinker's (1984, 1989) theory of language acquisition is, to my knowledge, the only other broad-range theory which has been defined in some computational precision and systematically compared with a wide range of acquisition data. The 1984 theory is a broad treatment of many aspects of acquisition, and the 1989 theory works within this framework to address the puzzle of alternating verb argument structures. The relevant data are always examined critically and thoroughly, especially where they seem to challenge the theory - much more thoroughly than I have been able to examine the data in this paper.

Pinker's theory has many features in common with this theory, but there are also important differences. To list first the main points of resemblance:

- Pinker's theory is based on a phrase-structured, non-transformational approach to syntax (LFG).
- It addresses mainly the early issues of language acquisition from ages 1- 4
- It assumes that the child learns by listening, in situations where he can infer the intended meaning.
- Meanings are represented as tree-like script structures, and are key to the learning mechanism
- Both theories embody a continuity hypothesis; Pinker's because he assumes that the child is acquiring an adult grammar (without going through any `child grammar') as null hypothesis, and the m-script theory, because the child can acquire adult-like m-scripts from the start.
- The treatments of auxiliary verbs and alternating verb arguments in the m-script theory are directly based on Pinker's treatments.
- The treatments of inflection, complementation and control in the two theories have much in common.

So the m-script theory owes a large debt to Pinker's work. The main differences are:

1. Pinker's theory is (as he says) a collection of sub-theories for different phenomena - phrase structure rules, inflection, auxiliaries, lexical acquisition, and so on. The m-script theory accounts for all these phenomena by one learning technique (m-intersection) applied in primary and secondary learning mechanisms.
2. In Pinker's theory, a language consists of both lexical entries and phrase structure rules. This theory is fully lexicalised, with no separate phrase structure rules; phrase structure is embodied in the m-scripts of each word.
3. Pinker's theory is based on `one-memory-limited' learning, where the child uses only the current utterance and already-known rules to infer new rules. This theory uses almost unlimited storage and retrieval of scripts (sentence-meaning pairs) to infer word m-scripts; although for any one word sense, only about six learning examples are needed.
4. In Pinker's theory (like others) learning events are triggered based on categorical, all-or-nothing criteria. This theory uses Bayesian statistics to weigh the evidence for any new m-script, so can learn it incrementally, with increasing levels of confidence as evidence accumulates.
5. Pinker assumes that the child uses essentially no negative evidence. In this theory, there is a well-defined mechanism for the child to gather negative evidence about all kinds of linguistic constructs, as a part of the normal learning process.

I will claim - inviting readers to judge for themselves - that in each of the differences (1) - (5) this theory is an advance over Pinker's.

(1) is a difference of economy of hypothesis. One broad unified theory is to be preferred over a collection of sub-theories, if it can fit the data. The comparisons of section 5 show that this theory does fit the data, with many confirmations and very few counter-examples. Whether it can continue to do so, after the kind of in-depth comparisons with the data that Pinker has made, remains to be seen.

Again in (2), a fully lexicalised theory seems to score over `separate syntax' theories in economy of hypothesis. We know that children have to learn very many word meanings; if they can learn the syntax of their language by the same mechanism, rather than needing separate mechanisms, this is more economical both for the theory and for them. But again, the real question is: does a fully lexicalised theory fit the data ? In general, a non-lexicalised theory tends to predict some sharp `watershed' events on the day a child acquires some key rule of syntax, and there is no evidence for such watersheds (C5). The fully-lexicalised nature of this theory makes other significant, confirmed predictions in (C6), (C7), (C9), (F1), (I3) and (I4).

In difference (3), one-memory theories of learning have, I believe, usually been adopted as a theoretical convenience, and not for any cognitively or biologically-motivated reasons. The one-memory restriction makes computational theories of learning simpler to think about, but has no other rationale. There is evidence that many animal species can make abstractions and generalisations from experience. This usually requires the comparison of several learning examples; why not allow it for language learning ?

Human memory capacity is very large, and there is no reason why children should not store hundreds or thousands of learning examples (scripts for sentence-meaning pairs) in their minds - and selectively retrieve just the few which are relevant to one word, when they are ready to learn it.

The advantage of this multi-example learning is seen most clearly seen in difference (4). In its categorical, all-or-nothing learning procedures, Pinker's theory is haunted by a `one false move' prediction. If the acquisition of a strategic piece of syntax - such as a major phrase structure rule - can be triggered by a single example, what is to stop a single mis-analysed or misheard example from leading to a false move which sends the child off down a blind alley ? Learning examples need to be `vetted' very carefully; too rigorous vetting would inhibit all learning, while too loose vetting leads to many false moves. The most effective kind of vetting (waiting for several examples to accumulate) is ruled out by the one-memory assumption.

In contrast, the Bayesian criterion built into the m-script theory allows the child to wait until several examples show that a rule is statistically significant, in the presence of noise, before adopting it; and even then, there are no strategic, language-wide learning decisions to take, because syntax is fully lexicalised. This makes language learning very robust.

There is evidence that Bayesian-like learning criteria are used in many aspects of human and animal cognition (Anderson 1990; Gallistel 1990) - as one would expect, because Bayesian learning is provably optimal (see Appendix B). There seems to be no good reason to exclude this robust Bayesian weighing of the evidence from language learning.

The `one false move' problem is exacerbated by the assumption (5) that the child uses no negative evidence. This forces Pinker to postulate a system of marking some parts of rules (but not others) as provisional, able to be retracted later on the basis of other evidence. The retraction rules often seem complex and contorted. A false retraction can be just as harmful as a false move, so the vetting criteria for retractions need to be framed with great care. Can retractions themselves be provisional ? If so, Pinker is coming close to weighing the evidence of multiple examples by probabilistic criteria.

It is now well established that children do not rely on explicit negative evidence such as parental corrections, and many theories, like Pinker's, have struggled with the supposed lack of negative evidence. In this theory, the primary learning process requires the child to `silently generate' sentences describing the inferred meaning of an adult's sentence, comparing them with what she hears. This not only picks out the sounds and meanings of new words from those of known words, but also provides negative evidence on supposedly-known words; the child may often observe `where I would have said X, an adult said Y'. Enough of these negative learning examples can change or retract an m-script.

The negative evidence mechanism allows the m-script theory to account straightforwardly for many disparate pieces of evidence, such as (B10), (B15), (D8), (G4), (G5), (H2), and (H3), where other theories (including Pinker's) struggle. These ,and other examples where we easily learn the jagged edges of our language, constitute a strong case to allow negative evidence (A).

In summary, while Pinker's assessments of the language data are unmatched, and have inspired several features of this theory, the m-script theory has a more soundly-based, robust and unified learning mechanism, which allows it to account for much of the evidence more easily than Pinker's theory.

## 6.2 Principles and Parameters

The Principles and Parameters (P&P) approach to language acquisition is based on a core assumption which this theory does not share - that the language learning problem is so hard (and the child's learning data so poor) that it must be reduced to a problem of setting a few parameters.

The m-script theory shows (in a working computer model) that there are robust and powerful learning mechanisms, which can rapidly learn the complex structures of language in the presence of noise. Around six examples per word are sufficient, and it seems that children can gather these examples easily at the required rate. Therefore the m-script learning mechanism is a working counterexample to the core assumption of P & P theories.

If we do not need to accept the P&P approach because of the supposed poverty of the stimulus, how do P&P theories fare in comparison with the child language data ? I believe that the data have not, over the years, given any striking confirmation of any of the core predictions of P&P models; rather, they have posed a series of problems which have necessitated successive dilutions of the original P&P idea - by maturation hypotheses, lexicalisation of parameters, etc. Without having counted the score, I very much doubt whether a systematic comparison of P&P with a broad range of child language data (as in Table 5.1) would yield anything like the measure of agreement shown by the m-script theory.

However, there is reason to hope that much work in the broader generative grammar framework (as opposed to P&P learning theory) is complementary to the m-script theory. Generative grammar has addressed issues, such as gaps and movement, where the m-script theory has (so far) little to say; and it has gained many useful insights into these phenomena. These insights have previously been expressed in terms of several levels of structure (such as D-structure and S-structure) which are hard to reconcile with the one-level m-script formalism. However, in Chomsky's (1992) minimalist programme, the underlying structures are now simpler and more compatible with the m-script approach.

If you believe that the m-scripts have merit as a theory of basic language learning, then maybe it will be possible to translate the insights of generative grammar on gaps, movement and other complex phenomena into m-script terms, to the benefit of both theories.

# 6.3 Connectionist Models

Ultimately, it seems likely that our theories of language learning will be connectionist, since that is how most computation is done in the brain. However, the question at issue is whether (a) neural nets will just provide components in the implementation of a language engine in the brain, or whether (b) language learning itself uses something like today's connectionist learning schemes. Pinker and Prince (1988) have characterised (a) as `Implementational Connectionism' and (b) as `Eliminative Connectionism'; it is the eliminative version that I will discuss first.

Rumelhart and Maclelland's (1985) neural net learning model for verb past-tense morphology raised hopes that this form of raw connectionism might encroach the symbolic-AI heartland of language. This model stirred up a lively controversy, leading ultimately (I believe) to the result that simple neural net learning does not adequately fit the facts of verb over-regularisation.

My aim here is not to comment on that specific issue, but on the wider prospects for eliminative connectionist models of language. There has, to my knowledge, been no published demonstration that neural nets can make any inroad on the core problem of human language - which is the richness and productivity of its meaning structures - and there are two reasons to expect that none will be made.

**(1) Productivity**: Today's generation of neural nets encode knowledge in a finite (typically small) number of connection weights. While these may easily encode small bounded symbolic structures (such as the relation between verb stems and their past inflections), nobody has yet found a good way to make these connection weights encode an unbounded symbolic structure such as a script tree. Connectionist nets do not have the unbounded, productive representational power of language (Fodor & Pylyshin 1988).

All demonstrations of neural nets in language have confined themselves to low-dimensionality sub-problems such as verb morphology or the simpler facets of syntax. They have never ventured (in print) out onto the high-dimensionality ocean of productive, meaningful language.

**(1) Learning Speed**: Bayesian learning has optimal performance (learning the required rules from a few examples) when the inbuilt prior probabilities match the actual probabilities in the environment. The reason why neural nets can learn many different patterns, (but typically learn them very slowly) is that neural nets do not have strong prior probability biases built into their structure; so they are open to many patterns, and fast learners of none (Denker et al. 1987). It is at present hard to see how the required prior probabilities could be built into a neural net.

When neural net learning results are reported, the learning times are measured in epochs, or thousands of trials. This is far too slow for a child, making it highly unlikely that these learning mechanisms have anything to do with human language learning.

While the prospects for eliminative connectionism are bleak, an exploration of implementational connectionism may be much more fruitful:

- In the m-script theory, there is a need for associative retrieval of m-scripts on the basis of their word sounds and of their meanings. Connectionist models of associative memory can give good performance, and might easily be adapted to this task.
- The core operations of m-unification and m-intersection may each be regarded as a type of energy minimisation problem. There are connectionist architectures (such as Boltzmann machines) which tackle these types of problem very effectively.

So a connectionist implementation of the m-script theory might be constructed out of rather a few network types, and might give good performance - in contrast to eliminative connectionist models.

# 6.4 Slobin's Model of Acquisition

Slobin (1973, 1985) has put forward a model of language acquisition consisting of about 40 Operating Principles (OPs) abstracted from extensive cross-linguistic study of acquisition in over a dozen languages (Slobin et al 1985). A typical OP is:

*OP (POSITION): FIXED WORD ORDER. If you have determined that word order expresses basic semantic relations in your language, keep the order of morphemes in a clause constant*.

Slobin's operating principles are, as he emphasises, derived bottom-up as an attempt to understand actual language acquisition data, rather than top-down from any overriding theoretical expectations. They are an excellent summary of much that we know about language acquisition.

However, beyond that level, they are (on their own) hard to intepret theoretically. Being stated verbally, they leave room for interpretation, especially when two or more OPs make different predictions. If, for instance, we took the OPs as specifications of language learning sub-modules in the brain, and tried to build computational models of those sub-modules, there would be many questions to answer (for instance, about how competition between OPs is resolved) and we would probably end up with a highly complex theory. Much extra detail would need to be added to realise each individual OP computationally (Pinker 1986), as well as `glue' to bind the OPs together as a working theory (Bowerman1985).

It is perhaps more useful not to go from OPs to a theory, but from a theory to OPs. For any computational theory of language learning, we may ask - does each OP emerge from the workings of the theory ? If not, how does the theory account for the data which are summarised by that OP ? Or does the theory actually contradict an OP ? In this way, Slobin's OPs can serve as a very valuable link between computational theories and data, highlighting gaps in any theory, and potential conflicts between theory and data.

I have not yet done this systematically for the m-script learning theory, but it may be very worthwhile to do so. From a preliminary review which I have made, many of the OPs seem to have a clear and direct interpretation in the m-script theory; others require more careful analysis (e.g. to see what the Bayesian probability theory predicts in specific cases); and a few clearly go beyond the scope of the m-script theory as currently formulated.

# 6.5 Siskind's Model of Lexical Acquisition

If precisely-stated algorithms of language acquisition are rare, then working computational models are even rarer. One of these is Siskind's (1996) model of lexical acquisition, which has interesting similarities and differences from the lexical acquisition part of the m-script theory.

As in the m-script theory, Siskind assumes that word meanings are represented by some kind of feature structure (~ script), which can be inferred from learning examples. In each example, the word is heard along with some inferred sentence meaning structure which contains (~ includes) the word meaning. Siskind presents an algorithm which, like that of section 3, can infer a word meaning structure from a small number of learning examples. He presents results from running the algorithm on artificial meaning examples.

The major differences between Siskind's algorithm and the m-intersection algorithm of this theory are :

- Like Pinker's theory, Siskind's algorithm is one-memory limited. At each step it uses only the current example, together with summary information on all previous examples. M-intersection picks from a large set of possible examples.
- It uses no probabilistic inference, but uses purely categorical all-or-nothing reasoning.
- It separates the problem into two steps - first isolating all the meaning elements (~ slot values) in the meaning of a word, then assembling them together in a tree structure.

The logic by which Siskind's algorithm converges on the correct word meaning, using only a few learning examples, is quite similar to the logic of m-intersection. When two meaning scripts are intersected together, the only slots which survive are slots which appear on both input scripts; thus the slots on the result are something like a set intersection of the slots on the inputs. Random coincidences between learning examples are rapidly eliminated by more examples. The first stage of Siskind's algorithm uses essentially this set intersection of slots in the inputs - and therefore rapidly converges on the set of slots in the word meaning, as the number of examples increases.

However, script intersection converges on the true meaning from above, and stops when a Bayesian criterion of sufficient evidence is satisfied (typically after 6 or so examples). Siskind's algorithm converges on the true meaning set both from below and from above, and stops when the two versions coincide.

By separating the two steps, Siskind's algorithm does not use structural information in the first step. If one example has a slot in position A on a script tree, and a second example has the same slot in position B, then script intersection will (correctly) eliminate the slot from the word meaning - while Siskind's algorithm does not do so until later. This difference in speed between the two algorithms is probably not very significant, and would be hard to discern in child learning data.

While Siskind's algorithm has similarities to the m-script algorithm, and probably has comparable performance, I would claim some advantages for the m-intersection algorithm:

- By learning the left branch of a word m-script and the trump links as well as the right (meaning) branch in one operation, it learns the syntax at the same time as the meaning.
- Using probabilistic inference with a Bayesian criterion of sufficient evidence, it is provably optimal, and more robust.
- It finds both content (slot values) and tree structure in one step, which is theoretically more economical than Siskind's two-step procedure.
- It solves the word segmentation problem in acquring each word

# 7. Discussion

The theory described in this paper has, I believe, a transparency, simplicity and power which commend it theoretically:

**Transparency**: Each word in a language is represented by an m-script structure which can be easily drawn or envisaged, and understood. The core operations of m-unification and m-intersection can be done with pencil and paper, and can be understood by an analogy to chemical reactions.

**Simplicity**: The structure of every language is completely embodied in the m-scripts of its words, with no other parameters or rules. One operation of m-unification is used for both language understanding and production. All the syntax and semantics of a language are learnt by the m-intersection mechanism. Language regularities arise from m-script evolution.

**Power**: the m-script formalism can express meanings of unbounded complexity, and the syntax of adult languages. M-intersection learning is fast, robust, and can learn the m-script for any word. Negative evidence can be gathered to correct any error.

However, these would be of little consequence if the theory did not agree with the data. The most important result in this paper is the comparison with empirical data on language learning, summarised in table 5.1. When compared with many diverse facts of language learning, the theory gives a natural and unforced agreement, without requiring extra *ad hoc* assumptions, in 84 out of 101 cases. Extra assumptions are required to fit the other 17 comparisons. I have not yet encountered any major, theory-threatening difficulty.

Even allowing for my possible selectivity in the choice of comparisons, and blindness to the faults of my own theory, this is an encouraging result. There is no other theory of language learning (with the possible exception of Pinker's (1984) theory) which claims this level of success in accounting for such a broad range of facts. Why does this theory do well in comparing with child language data?

The single most important answer is that **the Bayesian learning mechanism is powerful enough to do the job**. Other theories have generally used much less powerful learning mechanisms, and therefore have problems comparing with children's consummate ability to learn. The power of the Bayesian m-script mechanism spans four distinct dimensions:

1. **Capability** : It can learn an unlimited range of complex structures
2. **Speed**: It needs only a few examples per structure
3. **Robustness**: These examples may be diluted amongst a large majority of non-examples, and may each themselves have large amounts of random noise
4. **Reparability**: It can gather negative evidence to learn the exceptions to some generalisation.

It is an obvious fact that children's language learning exhibits all of (1) - (4). Pure symbolic and P&P theories generally fail on (3) and (4), while connectionist models fail on (1) and (2).

We can confirm that the Bayesian learning theory delivers all of (1) - (4) in two ways: either by doing the mathematical analysis to show that it does, or by building it into a computer program and observing its performance. I have done both, and am confident that others can reproduce the result.

The coherence of the learning theory rests heavily on the claim that **to learn a language is just to learn a set of word m-scripts**. In other words, **a language is just a set of word m-scripts**. This claim is supported by the computer program which handles a non-trivial sample of English in this way, and by the correspondence of m-scripts with unification-based grammars such as LFG.

The view of language underlying this theory has much in common with cognitive linguistics (Lakoff 1987; Langacker 1990) - in its grounding of language in semantics rather than syntax, in the biological origin of scripts in social cognition, and in the strong links between scripts and other cognitive structures. However, cognitive linguistics has adopted a connectionist model of language learning (Langacker 1990) which, I argue in this paper, cannot fit the facts. Perhaps this theory can be the learning theory for cognitive linguistics.

A key departure of this theory from the mainstream is the proposal that language regularities do not reflect regularities in the mind, but result from a process of m-script evolution for efficient communication. Thus the deep structure of language does not tell us as much as we thought about the deep structure of the mind; it tells us how language evolves on a neutral substrate of mind. For those who want to use language to learn about the mind, this may may be a disappointment.

However, in telling us less about the mind, it may actually be telling us more. In science, less is more; a theory with fewer assumptions and simpler structures is better than a more complex theory. So if a theory such as this one, without elaborate language-specific structures in the brain (but with general social learning mechanisms in stead) can fit the data, then it is to be preferred; it gives us a simpler, more transparent theory of the mind.

If this sounds dangerously empiricist after thirty years of Chomskyan nativism, so be it. The test of a theory is not how well it caters for our empiricist or nativist leanings; but how well it fits the data. On the test of language acquisition data (which was the primary motivation for Chomskyan nativism) this theory does better than any existing nativist theory.

The all-round agreement with the data leads me to claim that maybe **this is the way we learn our first language**. If others wish to prove this claim wrong - by an in-depth examination of the data on one language, or by working out the theory in greater depth, or by carrying out new experiments - they have my support; because if this theory is at all on the right track (as the data seem to show it is) then proving it wrong will yield new insights and progress.

**Acknowledgements**: I gratefully acknowledge my debt to the many researchers whose empirical results on language learning make child language such a vital and stimulating field. Some are cited here, many who should be are not. This theory would be nothing without their results.

# Appendix A: Script and M-Script Algorithms

This appendix gives brief descriptions of the algorithms for script and m-script operations which are the basis of language processing and learning in the theory. You may wish to run through these algorithms `by hand' with a pencil and paper for some simple examples, to get a feel for what the key operations do - particularly for m-intersection, which is the core operation of the learning theory.

All these algorithms have been implemented in a Prolog program which does language processing and learning for a small subset of English, and which can check the implementation of the operations by checking script algebra relations. The program is described in Appendix D.

## A.1 Script Operations

There are three key operations: script inclusion (the inverse of subsumption), unification and intersection (generalisation). They are closely equivalent to the corresponding operations for feature structures, as studied in computational linguistics (Scheiber 1986). They are all used as building blocks for the m-script operations described in section A.2. There is a fourth operation, which I call script subtraction, which is required as a sub-operation for m-unification.

For each of the three main operations, we need to match two tree-like script structures together node by node. This is done from the root downwards (as scripts are drawn with the root node at the top). Since there are choices as to how the two scripts are matched up, all three operations involve an element of search (over possible matchings) to find a satisfactory matching (for inclusion) or the best matching (for unification and intersection). In the Prolog implementation, this search is done using the Prolog failure-driven backtracking mechanism.

For the descriptions which follow, label the different nodes of a script by indices i,j etc.; then node i of script A is denoted by A(i), and the subtree of script A which is rooted at node i is also a script, which will be denoted by A[i].

Variable slot values are called `aces' and are denoted by capital letters preceded by a question mark, as in `?A', `?B', etc.

**Script Inclusion**: To check that script A includes script B (i.e. to check that A >s B) do the following steps:

1. Match the root node of A with the root node of B.
2. For the matched pair of nodes A(i) and B(x), if B(x) has any slot *s* with a defined value *v*, check that A(i) also has the same slot *s* with the value *v* or a more specific value *w*. If *v* is a plain constant value, *w* must be the same value; if *v* is a class-valued constant, *w* must be the same class or a subclass of it; if *v* is an ace value, *w* may be an ace value or a constant.
3. If there is any slot on B(x) not matched by a slot in A(i) as above, fail and try another node matching, if there are any left.
4. If two nodes A(i) and B(x) are successfully matched as in (2) above, try out all possible pairing sets of their child nodes. A pairing set is a set {(j,y), (k,z) ...}such that every node B(y) amongst the child nodes of B(x) is paired with some A(j) from the child nodes of A(i). If (j,y) are paired, then nodes j and y must be of the same node type (node types are holder, script, scene, entity and property, abbreviated as ho, sr, se, en, pr).
5. Pairing sets must respect time-order constraints: if there is a time-order arrow from node B(y) to B(z), then for a pairing set {(j,y), (k,z),...} there must be a time-order arrow from A(j) to A(k), either directly of via some intermediate A(l), A(m),....
6. If there is no such pairing set, fail and try a different matching at the parent level, if there are any left.
7. For each pairing (j,y) of the pairing set, do step (2) to check that the slot values in A(j) include all slot values in B(k).
8. Having traversed the tree structures, all nodes and slot values in B must be matched by some node and slot values in A. If some ace value ?V occurs on more than one slot in B, then it must everywhere be matched by the same value *w* (either the same constant, or the same ace value ?W) in A. Otherwise, the inclusion test fails.

In practice, the testing of time-order arrows is made easier if, before comparing A and B, you take the transitive closure of all time-order arrows in A, and separately for B. Then in a pairing set {(j,y), (k,z),...}, if there is a time-order arrow from B(y) to B(z), there must be one directly from A(j) to A(k). For the other script operations of unification and intersection, assume the transitive closure of time-order arrows is also done in advance.

**Script unification**: To form the unification C = A Us B of two scripts A and B, do the following steps:

1. Match the root node of A with the root node of B.
2. For a matched pair of nodes A(i) and B(j), form a node C(k) of the result. For each slot s: if neither node has a value, put nothing on C(k); if only one of the nodes A(i) and B(j) has a value *v*, then put the slot-value s(*v*) on node C(k); if both nodes have the same value *v*, put s(*v*) on C(k); if A(i) has a class value *v* and B(j) has a class value *w*, where *w* is a subclass of *v* (or vice versa) put the subclass value s(*w*) on C(k); if one node has a constant value, and the other has an ace value, put the constant value on C(k); if both nodes have an ace value, put an ace value on C(k) ; if the two nodes have inconsistent values *v* and *w*, fail and try another node pairing.
3. If two nodes A(i) and B(x) are successfully matched as in (2) above, try out all possible pairing sets of their child nodes. A pairing set is a set {(j,y), (k,z) ...}such that at least one node B(y) amongst the child nodes of B(x) is paired with some A(j) from the child nodes of A(i). If (j,y) are paired, then nodes j and y must be of the same node type.
4. For each node pair (j,y), do step (2) to check that the nodes A(j) and B(y) are consistent, and to form the result node C(l).
5. If a node from A(i) or B(x) is not paired with some node from the other input script, then that node and all its descendants goes into the result C unchanged. Thus every node in A or B must appear in C. Attempt to maximise the number of paired nodes, to minimise the total information content of the result C.
6. Pairing sets must respect time-order constraints: if there is a time-order arrow from node B(y) to B(z), then for a pairing set {(j,y), (k,z),...} there must not be a `reverse' time-order arrow from A(k) to A(j). (Time-order arrows on both A and B have been transitively closed). The time-order arrows on the result C include all time-order arrows from A and B.
7. Variable slot values must be unified consistently. If a variable value ?V appears more than once in A, and is unified with constant values *v*, *w*, .. from B, then these constant values must be consistent (eg *v* = *w*, or *v* is a subclass of *w*). If ?A is anywhere unified with ?B, then they are then treated as equal; they must nowhere be unified with mutually inconsistent constants. If any inconsistency occurs, fail and try other node pairings, if there are any.

Therefore script unification may fail. If it does, this means that the scope sets s(A) and s(B) have no overlap.

Script unification can always be checked from its definition in terms of script inclusion. The result C must obey C >s A and C >s B; and it should not be possible to take any information away from C (to remove any nodes or slots), keeping both these relations true.

**Script intersection**: To form the intersection D = A ∩s B of two scripts A and B, do the following steps:

1. Match the root node of A with the root node of B.
2. For a matched pair of nodes A(i) and B(j), form a node D(k) of the result. For each slot s: if neither node has a value, put nothing on D(k); if only one of the nodes A(i) and B(j) has a value, put nothing on node D(k); if both nodes have the same value *v*, put s(*v*) on D(k); if A(i) has a class value *v* and B(j) has a class value *w*, where *w* is a subclass of *v* (or vice versa) put the superclass value s(*v*) on D(k); if one node has a constant value, and the other has an ace value, put the ace value on D(k); if both nodes have an ace value, put an ace value on D(k) ; if the two nodes have inconsistent values *v* and *w*, put an ace value on D(k).
3. If two nodes A(i) and B(x) are matched as in (2) above, try out all possible pairing sets of their child nodes. A pairing set is a set {(j,y), (k,z) ...}of nodes A(j) and B(x); it may be empty. If (j,y) are paired, then nodes j and y must be of the same node type.
4. For each node pair (j,y), do step (2) to form the result node D(l).

5. If a node from A(i) or B(x) is not paired with some node from the other input script, then that node and all its descendants do not appear in the result D. Attempt to maximise the number of paired nodes, to maximise the total information content of the result D.
6. For a pairing set {(j,y), (k,z),...}, where A(j) and B(y) match to give node D(l), and A(k) and B(z) match to give node D(m), there is a time-order arrow from D(l) to D(m) only if there is one from A(j) to A(k), and one from B(y) to B(z).
7. If a slot s(*v*) from A matches with a slot s(*w*) from B to give a slot s(?X) on D, then wherever else a slot s(*v*) from A matches with a slot s(*w*) from B (with the same values *v* and *w*, which may be either constants or aces), it gives a slot s(?X) with the same ace value ?X on D.
8. Any slot s(?X) on D, whose ace value ?X is not repeated elsewhere on D, gives no information and can be removed.

Script intersection always gives a result.

The result of a script intersection can be checked from its definition in terms of script inclusion. The result D must obey D <s A and D <s B; and it should not be possible to add any information to D (to add any nodes, slots or time order arrows), keeping both these relations true.

The script subtraction E = B- A is defined as the script E with smallest information content which obeys E Us A = B. Thus E only exists if B >s A. An algorithm to find E is:

1. Check that B >s A; if not, fail. Otherwise set E = B.
2. Start at a leaf node A(i) of A, and find the node B(x) of B which matched it in the inclusion test. For any slot s(*v*) on A(i), if the same slot s(*v*) occurs on B(x), remove this slot and value from E(x).
3. If all slots have been removed from a leaf node E(x), remove the node.
4. When all children of a node A(j) have been treated in this manner, do (2) for the parent node A(j) and its matching B(y).
5. If all slots have been removed from a non-leaf node E(y), and all its child nodes have been removed, remove the node E(y).

The result of a script subtraction can be checked from its definition E Us A = B.

# A.2 M-Script Operations

An m-script M represents a set of scripts, which is called its scope, $\sigma(M)$. M-scripts look like scripts, in which some nodes are trump nodes (denoted by labels !A, !B etc. on the node), and there may be a trump link between any pair of trump nodes.

In the descriptions which follow, the script which is derived from an m-script M by removing all its trump nodes and trump links will be denoted by Ms. This is always within the scope of M, Ms e $\sigma(M)$. (Here, the letter e denotes set membership; it is an attempt at epsilon)

We shall consider only m-scripts where each trump node has at most one trump link attached to it; so every trump node is either a lone trump node, or one of a pair attached by a trump link. These m-scripts are apparently sufficient for language. For m-scripts not obeying this constraint, the algorithms are more complex to state.

To define precisely the scope of any m-script, we need to define *truncated subtrees* of the m-script. For every node of a type which may be a trump node, there can be certain slots on nodes of that type. We divide these slots into `up' slots (which are considered to be above the trump node, and this unaffected by the presence of that trump node and its links) and `down' slots, which are below the trump node. Most slots are `down' slots, but it is useful for language processing to make one or two slots `up' slots.

Then for an m-script with N trump links, there are (N+1) truncated subtrees. The first of these, M[1] is rooted at the root node (conventionally node 1) of M, and is truncated at any trump nodes; it contains only the `up' slots on those nodes, and no child nodes below them. For every trump node i, there is a truncated subtree M[i] whose root is the trump node, and which is truncated at any trump node below i. It contains only the `down' slots of trump node i, and only the `up' slots of lower trump nodes where M[i] is truncated.

Thus the information in M is partitioned between its truncated subtrees M[i]; every slot from M appears on precisely one of the M[i]. Although any trump node appears on two of the M[i] (it is the root node of one of them, and a leaf of another) its slots are partitioned between the two.

An m-script M is well-formed if, wherever it has a trump link from node i to node j, M[j] >s M[i]. Assume the input m-scripts in all of the algorithms below have been checked to be well-formed.

The definition of the scope of an m-script is implicitly given in the algorithm to determine whether some script, S, is in the scope of M:

**Scope test**: To check that script S is in the scope of M, or that S e σ(M), do the following steps:

1. Check that S >s M . If not, fail. If S >s M, label the nodes of S so that M[i] is matched by S[i] in the script inclusion test.
2. Form the truncated subtrees M[i] of M, and the corresponding truncated subtrees S[i] of S.
3. Above the top trump nodes of M, S and M must be equal; so check that S[1] = M[1]. If not, fail.
4. If there is a trump link from node i to node j of M, the subtrees of S must obey S[j] = S[i] Us M[j].

N m-includes M, or N >m M, if any script S in the scope of N is also in the scope of M.

**M-inclusion test**: To check that N >m M, do the following steps:

1. Check that N >s M; if not , fail. If so, label the nodes of N so that M[i] is matched by N[i] in the script inclusion test.
2. Above the top trump nodes of M, N and M must be equal; so check that N[1] = M[1].
3. To ensure that Ns e σ(M), if M has a trump link from node i to node j, then the truncated subtrees of N must obey N[j] = N[i] Us M[j].

4. To ensure that σ(N) < σ(M), wherever N has a trump node i, M must also have a trump node on node i, or its parent node, or its grandparent node, etc.
5. To ensure that σ(M) σ(N), wherever M has a trump link from node i to node j, either N must have a trump link from i to j, or N must have no trump nodes on or above i or j.

If any of these tests fail, then N >m M is not true.

(In what follows, I am using the notation Ps for the script part of P - i.e P without any trump nodes or links)

**M-unification**: P = M Um N, is the P with largest possible scope obeying P >m M and P >m N; therefore σ(P) < σ(M) and σ(P) < σ(N). P may not exist, if σ(M) and σ(N) do not overlap. To find P, do the following steps:

1. Calculate R = Ms Us Ns. If this unification fails, the m-unification fails. If P exists, then Ps >s R. Label the nodes of M, N, and R so that matching nodes have the same node labels.
2. If P exists, then Ps e σ(M) and Ps e σ(M). So if M has a directed trump link from node i to node j, P must obey Ps[j] = Ps[i] Us M[j]. R might not obey this relation, but it may be possible to `round out' R by adding extra information to make it do so. This is done by two calculations: Ps[j] = R[i] Us M[j] (forward rounding), followed by Ps[i] = R[i] Us (Ps[j] - M[j]) (backward rounding) in succession.
3. Do (2) for all trump links on N or M. If any of the unifications fail, then the m-unification fails. (When the trump links of M and N overlap in complex ways, this step may involve quite a complex iteration to satisfy all the trump link equations of M and N together; but this rarely happens). We now have a `minimal' Ps which is in the scope of both M and N, but it has no trumps; we need to expand its scope as far as possible, by adding trumps to it.
4. To enlarge the scope σ(P) as much as possible, if both N and M have a trump node i, put a trump on node i of P. If N has a trump on node i, and M has a trump on the parent or grandparent (etc.) of i, put a trump on node i of P.
5. To ensure σ(P) does not extend beyond either σ(N) or σ(M), if either N or M has a trump link for node i to j, put a trump link from node i to j of P. (again, there are subtleties here from overlapping trump links of M and N, but they are rarely important)

The resulting m-script P can be checked from the definition of m-unification. P should obey both P >m M and P >m N; but it should not be possible to extend the scope of P (remove nodes or slots; add trump nodes; or remove trump links) without making this untrue.

M-unification is the key operation for language generation and understanding; for both of these, there is one m-unification per word m-script. The initial unification in step (1) tests the applicability of the word (syntactic and semantic constraints). In step (2), the m-unification requires one unification for each trump link in the word m-script:

- For *language generation*, the right branch of the word m-script matches an existing meaning structure, and a backward rounding operation (script subtraction and unification) is required to pass meaning scripts back along the trump links to the left branch of the m-script.

- For *language understanding*, the left branch of the word m-script matches existing meaning structures, and a forward rounding operation (unification) is required to pass meaning scripts along each trump link into the right-branch meaning script - combining the meanings of the arguments (left branch) with the intrinsic meaning of the word (right branch).

**M-intersection**: Q = M ∩m N, is the Q with smallest possible scope obeying M >m Q and N >m Q; it always exists. To find Q, do the following steps:

1. Calculate Qs = Ms ∩s Ns. Label the nodes of M, N, and Q so that matching nodes have the same node labels.
2. Put a trump on any node i of Q where (a) Q(i) differs from M(i) or N(i) in any slots or child nodes, or (b) either M or N has a trump on node i, or (c) where some ancestor node of Q(i) has a trump because of (a) or (b). The scope of this Q includes the scopes of both M and N, but it can possibly be narrowed by adding in trump links.
3. For any two trump nodes i and j on Q, add a trump link from i to j if both M[j] = M[i] Us Q[j] and N[j] = N[i] Us Q[j], and if neither M nor N has any trump nodes without the same trump link on node i or j.

(step 3 must be replaced by a rather more complex procedure in some cases with complex trump configurations)

M-intersection is just script intersection, with extra steps (2) and (3) to discover trump nodes and links. Step (3) discovers a trump link wherever both the input m-scripts M and N obey the trump link relation.

M-intersection is the key operation used to discover word m-scripts in language learning. However, for this purpose there is a key difference from the definition given above, in the discovery of trump links.

When forming `learning example' scripts by partially understanding sentences and observing the meaning they refer to, the child may observe not just the true meaning script X, but also some extra information; she infers a script Y which includes X, Y >s X. Therefore the trump link relation M[j] = M[i] Us Q[j] becomes just M[j] >s M[i] because of extra information on the `downstream' end M[j] of the trump link. (As the `upstream' end M[i] comes from previously learnt words, it has no extra information.)

So if we did m-intersections of these example scripts as above, we would not discover the trump links. We need in stead to use `broad m-intersection', which differs from plain m-intersection, replacing (3) by (3') :

3' For any two trump nodes i and j on Q, add a trump link from i to j if both M[j] >s M[i] and N[j] >s N[i], and if neither M nor N has any trump nodes without the same trump link on node i or j.

In practice a word is learnt by m-intersection of several learning examples, which because of the commutative and associative property of m-intersection (an m-script algebra identity) can be done in any order.

# Appendix B : Proof of Bayesian Optimality

This appendix contains a proof that, within a certain mathematical model of learning, a Bayesian learning procedure gives the best possible fitness; so that natural selection will tend (other things being equal) to produce animals with Bayesian learning in preference to any other.

The mathematical model of learning uses some abstractions and approximations (such as discrete time steps) but is nevertheless quite a good approximation to the problem of learning social causal regularities in a primate group. Primates are under intense selection pressure to compete socially, and need to learn these regularities as fast as possible. If Bayesian learning is optimal, this implies that primate selection by social competition leads to near-Bayesian learning of social regularities. In this theory, social learning is the forerunner of language learning, which will therefore inherit its Bayesian character.

We characterise animal learning as follows:

- There are one or more sequences of **time steps** t = 1....T. During each time step, some **event** E(t) takes place, and is observed to take place by the animal. The events E(t) may have complex structure, e.g. they may be described by scripts.
- For each time step t, the probabilities of different E(t) are typically independent of t and independent of the previous events E(t-1), E(t-2), etc. However, in any environment (eg a primate social group) there is a small number of causal regularities, or **rules** (denoted by R). Each rule R states that `if E(t-1) is in a certain set of `cause' events, then with a **rule probability** $\sigma$(R), E(t) will be in some other set of possible `effect' events'
- The animal does not know which rules R actually hold in its environment or group; the task of learning is to find out, by observation of events, which rules are probably true, and then to use that knowledge to predict events and choose actions.
- The set of rules {R1 , R2 .....} which actually hold in one environment - each Rj defined with its cause events, effect events and rule probability - constitute the **state** Si . There is a complex multidimensional space of possible states Si , with a **prior probability** $P_A$(Si) defined for each possible state.
- For states Si in which just three rules R1, R7 and R10 hold, $P_A$(Si) is actually a probability density r in the three rule probabilities s1(R1), s7(R7) and s10(R10); the integral of this density is the (finite but small) probability that just those three rules, and no others, hold in the environment. Prior probability densities such as r(s1 , s7 , s10 ) may be modelled as fairly uniform in s1 ,s7 , and s10 - implying that even if rule R1 is true, we do not know much *a priori* about its rule probability s1(R1), which might be near 0, near 1 or anywhere in between.
- The animal observes the events E(1) ... E(T-1); these will be called its **sense data** Dm . The sense data will have different probability distributions depending on the rules, i.e. depending on the actual state S ; these are described by a conditional probability P(Dm | Si), which can be calculated from probability theory. If the sense data give good evidence for one of the rules (e.g. R7 ) the animal should learn this rule, and then be able to use it to predict outcomes and choose actions.
- The event E(T) are considered as part of the state Si. Thus the event E(T+1) depends only on Si (i.e. on the event in the previous time step, and on the causal rules R which hold between successive time steps), and not directly on Dm .

- Just before time step (T+1), knowing E(1) ... E(T), the animal chooses some action Ak , out of a set of possible actions, to take in time step (T+1). There is then some outcome On , which depends only on the action Ak and the event E(T+1) which was happening at the same time. Since E(T+1) depends only on Si , not on Dm , the probabilities of different outcomes are described by a conditional probability P(On | Ak Si ).
- Each possible outcome O has a value V(O ) to the animal. Values V are measured on some scale which can be translated into lifetime fitness.
- The task of cognition (which includes learning) is to choose actions such that the average value of outcomes (the average increment to lifetime fitness, for each set of time steps) is as large as possible. We shall show that Bayesian learning maximises this average value of outcomes - i.e. leads to best possible fitness.

To make this abstract framework a little more concrete in the context of primate social learning: There may be three distinct causal regularities in some primate social group: R1 : `When Brutus bites Cassius, Cassius becomes frightened', R2 : `If you threaten Cassius when he is frightened, he will give you food, and R3 : `If you threaten Cassius when he is not frightened, he will bite you'. Suppose R2 and R3 are well known, and the task of learning is to discover, by observation, whether R1 holds.

The prior probability that R1 holds is quite small, p(R1) = 0.1; and even if it does hold, there is no a priori knowledge about its cause-effect probability s1(R1); so r(s1) = 0.1 in the range 0<s<1.

Suppose a monkey observes 49 events E(1) .. E(49); in 12 of these, Brutus bites Cassius, and on 10/12 occasions, Cassius is then frightened. Normally, Cassius is only frightened 20% of the time. Now, in time step 50, Brutus again bites Cassius. Should our monkey go and threaten Cassius ? If the rule does hold (and has high rule probability s) then Cassius will be frightened and our monkey will get some food (fitness increment +0.01 on some scale); but if the rule does not hold (those 10/12 occasions were just flukes; Cassius was frightened for some other reason) then our monkey will get a bite (fitness increment -0.1).

The proof that Bayesian learning gives the optimum decision rule is now quite simple:

The way an animal chooses an action Ak, given sense data Dm , is summarised in a **decision function** Fk(Dm ). Fk gives a value for each action index k and sense data Dm **.** The animal acts as if, when having sense data Dm , it calculates all the decision functions Fk for each possible k, and chooses the action Ak with the largest Fk .

Decision functions can be used to describe any relation between sense data and actions ; that is, to describe the input-output relation of any cognitive system. So they give an external, functional specification of any possible brain. Our problem is: what is the best possible set of decision functions Fk ?

Figure 2 shows the relations between these concepts.

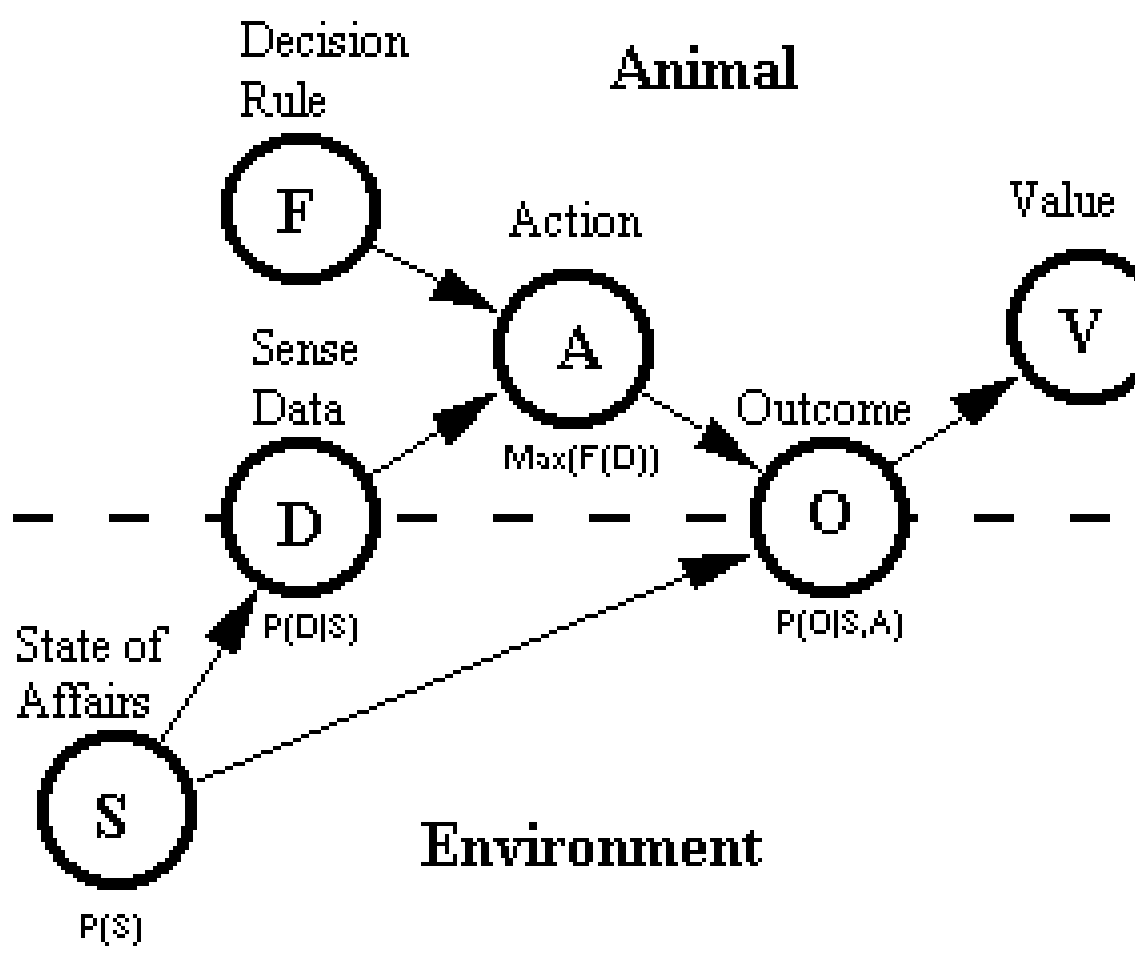

*Figure 2: Flow of causation for the mathematical model of animal cognition. Time ordering is from left to right, and arrows flow from causes to effects. The probabilities P of states of affairs S, of sense data D and of outcomes O are discussed in the text. The horizontal dashed line denotes the interface between the animal and its environment.*

The **requirement equation** defines the best possible decision functions Fk(Dm) - those which give best average outcome V, and which therefore give greatest possible fitness.

The expected value from taking action Ak in the state of affairs Si is the average value of all possible outcomes, weighted by their probabilities :

$$W(S_i, A_k) = \sum_n P(O_n | S_i, A_k)\, V(O_n) \qquad (2.1)$$

Using the definition:

$$\theta_k(x_1, \ldots x_k \ldots x_N) = 1 \text{ if } x_k = \max(x_1 \ldots x_N)$$
$$= 0 \text{ otherwise} \qquad (2.2)$$

we can calculate the expected value from one state of affairs Si , which gives sense data Dm , if the animal uses decision functions Fk to choose an action:

$$U(S_i, D_m) = \sum_k \theta_k(F_1(D_m) \ldots F_N(D_m))\, W(S_i, A_k) \qquad (2.3)$$

The theta function picks out one term from the sum over actions Ak - the term which has the largest Fk. So it embodies the decision rule to pick out one action by calculating the Fk and picking the largest. U is the average value resulting from that choice.

The average value for a single encounter is got by summing over all states of affairs and sense data sets, weighted by their probabilities:

$$
\begin{aligned}
V_{avg} &= \sum_i P_A(S_i) \sum_m P(D_m|S_i) \;\; U(S_i, D_m) \\
&= \sum_i P_A(S_i) \sum_m P(D_m|S_i) \sum_k \theta_k(F(D_m)\ldots) \; W(S_i, A_k) \\
&= \sum_k \sum_m \theta_k(F(D_m)) \sum_i P_A(S_i) \, P(D_m|S_i) \; W(S_i, A_k) \\
&= \sum_k \sum_m \theta_k(F(D_m)) \; G_k(D_m) \qquad (2.4)
\end{aligned}
$$

where :

$$
\begin{aligned}
G_k(D_m) &= \sum_i P_A(S_i) \;\; P(D_m|S_i) \; W(S_i, A_k) \\
&= \sum_i P_A(S_i) \;\; P(D_m|S_i) \sum_n P(O_n|S_i, A_k) \; V(O_n) \quad (2.5)
\end{aligned}
$$

Gk depends only on things outside the animal's control Ñ on probabilities of states, probabilities of sense data, probabilities of outcomes and values of outcomes. Gk(Dm) is effectively the average value of doing action Ak when having sense data Dm .

In the design of a cognitive system, Gk cannot be varied, but the decision functions Fk can.

(2.4) is a sum over all Dm, where (because of the q function) for each Dm only one Gk enters the sum. The maximum $V_{avg}$ is got by setting

$$
F_k(D_m) = G_k(D_m) \quad \text{for all k and m} \qquad (2.6)
$$

since this choice means that for every Dm , the q function picks out the largest of the Gk and adds it to the sum. Any other Fk gives a worse average outcome, by sometimes picking a smaller Gk .

Therefore the optimum decision rule is given by

$$F_k(D_m) = \sum_i P_A(S_i)\, P(D_m|S_i) \sum_n P(O_n|S_i, A_k)\, V(O_n) \qquad (2.7)$$

Thus equation (2.7) is the requirement for animal cognition Ñ the optimum form of cognition in all circumstances. I shall call it the **Requirement Equation**.

We can multiply all the Fk by any positive common factor, and it will not alter the choice of action. This enables us to rewrite (2.7) in two parts :

$$P_p(S_i) = \frac{P_A(S_i)\, P(D_m|S_i)}{\sum_j P_A(S_j) P(D_m|S_j)} \qquad (2.8)$$

$$F_k(D_m) = \sum_i P_p(S_i) \sum_n P(O_n|S_i, A_k)\, V(O_n) \qquad (2.9)$$

(2.8) is recognisable as Bayes' theorem, applied to find the most likely state of affairs Si in the light of the sense data Dm; then (2.9) chooses the best possible action, in the light of the likely states Si. For many problems, finding the state by (2.8) is the hard part, and then choosing an action by (2.9) is comparatively easy.

# Appendix C: A Fundamental Theorem of Language Learning

This fundamental theorem of language learning, stated in section 3.11, is:

**If speakers compose sentences by m-unification of word m-scripts, and people learn words by m-intersection of sentences they hear, then any set of word m-scripts (a language) will propagate stably through generations of speakers**.

This theorem is important because without it, there could not be so many diverse yet stable languages in the world; just as without DNA replication, there could not be diverse and stable species. Its proof was sketched in section 3.11; this appendix gives a more detailed (though still not rigorous) derivation. The derivation gives insight into how the whole theory of language generation, understanding and learning hangs together.

We assume (a) that every m-script has a trump on its top script node, and (b) that each word m-script has, in its left branch, a sequence of `sound scenes' one per phoneme, rather than one sound scene for the whole word sound.

I shall first derive the result under simplifying assumptions, then show how they can be relaxed. The simplifying assumptions are:

- Sentences are unambiguous (eg there are no homonyms)
- No `gap' constructs are used
- Speakers only express meanings which are exactly and unambiguously statable in the language

To understand a simple unambiguous sentence with no `gap' constructs, the processing is as follows:

1. Form a phoneme sequence script P, with a top holder node and one sound scene per phoneme heard, with time order arrows between the sound scenes.
2. Select word m-scripts W1, W2 , W3 etc. whose sequences of phonemes match the sequences of phonemes heard, and m-unify them with P, forming successively (P Um Wi ), ((P Um Wi ) Um Wj), etc.
3. Since two m-scripts will, in general, not m-unify, the order of m-unification is highly constrained. Each word m-script must be m-unified with all scenes of its left branch matched - both the sound scenes, and the meaning scenes (which must match meanings produced by previously m-unified words). All words except nouns require one or more meaning scenes to match their left branch; so the first word to be m-unified must be a noun.
4. Each word m-unified adds the meaning scene of its right branch to the growing structure. Continue until all phonemes have been consumed by some word, and all meaning scenes except one have been consumed by a word. If you cannot, the sentence is ungrammatical.
5. The one remaining meaning scene is the meaning of the sentence.

The script algebra implies that m-unification is exactly commutative, A Um B = B Um A, and partially associative (A Um (B Um C)) approx = ((A Um B) Um C). It is `partially' associative in two senses:

- It may be that A Um (B Um C) exists, but (A Um B) Um C does not (for instance, if A has no common structure with B, but C shares features with both A and B); but if both do exist, they are approximately equal.
- `Approximately equal' means that the great majority of nodes and slot values are equal. Occasional small differences will be ignored.

This means that if a product (((A Um B) Um C) Um D)... can be formed at all, in any order, then it will be the same irrespective of which order it is formed in, and can be written without all the brackets as (A Um B Um C Um D..), with the constituent m-scripts in any order.

(Note this order of m-unification has nothing to do with the order of words in the sentence. That is defined by time-order constraints between scenes of the SMS, regardless of how the SMS was formed).

For a grammatical sentence, with phoneme sequence script P and words W1 , W2 , ... W3 , the m-unification

Z = (((P Um Wi) Um Wj) Um Wk)... Um Wn)

exists for some order of m-unification, and obeys

Z >mWi

for every word Wi in the sentence. Furthermore, since every phoneme in P has been consumed by some word W, P itself adds no information to Z which is not in the product of the W; so Z can be rewritten without P:

Z = ((Wi Um Wj) Um Wk)... Um Wn)

= (Wi Um Wj Um Wk... Um Wn)

The meaning M of the sentence can be found within Z as the right-hand meaning scene of the last-unified word Wn; it can be identified as the only scene which has no trump links going out of it to other scenes (it has trump links coming into it).

Since this full meaning scene is a part of Z, it follows that

Z >m M

so

Z = Z Um M

= (Wi Um Wj Um Wk Um ...Wn Um M)

Suppose some speaker starts with exactly the meaning M and wishes to express it. M was the result of m-unifying the last-applied word Wn with the penultimate structure ((Wi Um Wj) Um Wk)...); this guarantees that we can form (M Um Wn), which is what the speaker does, as the first step in generation. Subsequent steps in generation are to m-unify the m-scripts for all the words in the sentence, forming

Z' = ((((M Um Wn) Um Wk) Um ....Wi)

= (Wi Um Wj Um Wk m ...Wn Um M)

= Z

Here we have used the `partial associative' property of m-unification to show that the speaker and the listener form the same m-script structure Z in their heads.

The same result can be shown in a less abstract manner, by considering the action of individual word m-scripts in generation and understanding. Suppose a word W has a sound scene and N (typically N= 0, 1, or 2) meaning scenes in its left branch, and one meaning scene in its right branch. There is a trump link from each left-branch meaning scene to the right-branch meaning scene.

- In understanding, W is m-unified left-to-right, given up to N meaning scenes and creating one new meaning scene M in its right branch.

- In generation, W is m-unified right-to-left, given just the right-branch meaning scene M and creating the N left-branch meaning scenes.

In either case, the structure Z so created has N `left branch' meaning scenes S1...SN each connected by a trump link to the `right branch' meaning scene M, and each obeying a trump link relation:

M[i] = W[i] Us Si

These trump link relations define 1:1 relations between their ends; if all the Si are fixed, then M is fixed, and conversely, if M is fixed, then all the Si are fixed.

The full m-script structure Z above is a network of meaning scenes connected by trump links. Most scenes have trump links coming in to them, and one trump link going out. The only scenes with no trump links coming in are noun meanings; the only scene with no trump link going out is the full meaning M.

Because each trump link defines a 1:1 relation between its ends, the structure is over-determined. If you start at the full meaning scene M and work right-to-left (backwards over trump links), you must get the same result as by starting at the noun meaning and working left-to-right (forward over trump links) - because the trump link equations, relating the scripts, are the same in both cases. That is why `Z formed left-right' in understanding must be the same as `Z formed right-left' in generation.

Now consider a child hearing a sentence in which she does not know just one word, but knows all the other word m-scripts; and where she can infer the meaning M from non-linguistic clues. The child can construct the same m-script structure Z by working from both ends towards the middle. If the missing word is a noun, its meaning script can be constructed back from M along the trump links of other words. Otherwise, all meaning scenes in Z can be found by working along trump links from either end.

The only part of Z which cannot be built in this way is the trump links of the unknown word W; Z obeys the trump link equations, but does not have the relevant trump nodes or a trump link. So the child actually constructs a near-copy Z' of Z which, because it does not have the trump nodes, obeys Z' >m Z. Since Z >m W, this implies that Z' >m W.

Suppose the child hears a number of sentences containing this unknown word W. For each sentence Si she constructs an m-script structure Z'i obeying Z'i >m W.

If you m-intersect together a number of structures Z'i, all of which obey Z'i >m W, then the result W' = (Z'1 ∩m Z'2 ...Z'n) must also obey W' >m W. As explained in section 3.8, as the number of examples Z' increases, the amount of extraneous information (which survives in W', but is not a part of W) decreases exponentially; and the m-intersection automatically discovers the trump links in W. After about six examples, we can expect all extraneous information to have been eliminated, leaving W' = W; the unknown word will have been correctly learned.

This faithful learning applies to any word W; so if adults use any arbitrary set of word m-scripts W1 , W2 ,... for communication by m-unification as above, and if the child learns by m-intersection, she can learn all the Wi by starting from the nouns and working upwards.

This proves the fundamental theorem in the case of no ambiguities, homonyms of gap constructions.

The extension of the theorem to take account of these other complexities is not difficult. Principally, we rely on the robustness of the learning mechanism in rejecting `bad' learning examples Z ,and in being able to pick out a small signal from a large amount of noise.

**Homonyms**: If the child ever tries to combine examples Z1 and Z2 which refer to different senses of a homonym, the result of this m-intersection will have very small information content in its right-hand meaning branch, so will be rejected by the learning mechanism.

**Other Ambiguities**: If the child somehow selects the wrong sense of an ambiguous sentence, this too will be rejected by the learning mechanism for too small information content. Generally, it is safest to avoid including ambiguous sentences in the learning set, for slower but safe learning.

**Gap Constructs**: Here, the m-script structure Z contains scenes which are not determined by the meaning of some noun, and have no incoming trump links; the only meaning they have (in the first stage of understanding) comes from the argument restrictions of the verb. However, if the child has inferred the correct full meaning M, the identity of the gap filler can be got by tracing information backwards along trump links from M to the gap entity, in the `generation' direction.

# Appendix D: The Language Learning Program

This appendix describes a program which can understand, generate and learn a fragment of English using the mechanisms of this theory.

The program is written in Prolog, and uses some specific features of LPA Prolog as implemented on the Mac. The source code is no longer available.

A graphical editor can be used to build and display scripts and m-scripts, and was used to generate some of the diagrams in this paper.

The program is based on an implementation of the script algebra and m-script algebra operations, using algorithms similar to (but more complete than) those described in Appendix A. The implementation can be checked by testing that the results of the operations obey the relations of the script algebra and the m-script algebra. The m-script operations can handle the trump link structures needed for word m-scripts, but cannot correctly handle trump link structures of arbitrary complexity.

The program is implemented using m-scripts which differ from the ones shown in this paper in one respect. In stead of having one script node at the top with several scene nodes hanging from it (as in this paper), a word m-script has one 'script holder' node at the top with two script nodes below it - one for the left branch and one for the right. The scenes then hang from these script nodes.With that form, the understanding process is slightly more complex than I now think it needs to be; but that is the way I did it, and I haven't yet had time to modify the program. I used a chart parser mechanism to keep track of time order constraints which I now believe can be done automatically as part of m-unification; but apart from that, there are no real changes in how understanding, generation or learning are done.

In word m-scripts, the sound of one word is represented as one `sound scene' containing the whole sound of the word, rather than as a sequence of `phoneme sound scenes'. Semantic representations of word meanings broadly use the principles of lexical and conceptual semantics as explored by Jackendoff, Levin , Pinker and others, but are rather crude in places; it is not often that the precise details of semantic representation affect the broad principles of language processing and learning (although of course they affect its realisation in many ways).

**The Language Understanding Component** takes as input typed sentences, typically containing only words whose m-scripts have been defined to the program. It proceeds by m-unification and other operations until a single meaning structure has been found, or some form of failure (eg from an ungrammatical sentence) has occurred. It digests sentences at a roughly linear time cost of 10 - 100 seconds per word on the Mac SE30.

It uses a chart parsing mechanism to control the selection of sound scenes, and meaning scenes from previous word m-unifications, to m-unify with the left branch of the next word to be processed. Word sounds are initially arranged as arcs between neighbouring nodes i and i+1 in a chart of nodes from 0 to N (for an N-word sentence). Nouns are processed first, each giving an entity-describing scene. This scene is inserted as an arc spanning the same two nodes as the noun sound. For homonyms, two or more entity scene arcs span the same pair of nodes.

To m-unify any word, its left-branch scenes (including the word sound scenes, and other meaning scenes) must be matched with contiguous arcs of the chart, and all the constraints of m-unification must be satisfied by the scenes on selected arcs. Then the right-hand branch of the result (the meaning of the word, with its arguments filled in) is put in the chart as a new arc spanning from the start to the end of the contiguous arcs which the word matched.

Verb m-scripts can, when necessary, be m-unified with one or more of their noun argument scenes missing (so consuming a gap). Higher priority is given to verb m-unifications which do not require gaps.

In this way, meaning scenes are built up which span longer and longer stretches of the chart. The sentence is fully understood only when a unique meaning scene spans the whole chart.

The m-unification of words is interspersed, in quasi-parallel fashion, with various other operations:

- When two or more distinct meaning scripts span the same two nodes, this is an ambiguity. The main strategy is to take the script intersection of these two or more meanings, and to use that for further m-unifications rather than the two individually. This avoids a combinatorial explosion of possible readings, and keeps sentence processing time approximately linear in sentence length.
- The m-scripts for third-person pronouns and reflexive pronouns define the entity's identity as a variable value. The program attempts to equate these variable identities to other known identities, using heuristics which search for the closest good match subject to appropriate constraints (eg Chomsky's principles A and B).

- Variable identities arise from gaps in m-unifying verb m-scripts. Other word m-scripts (eg those for relative pronouns) trigger procedures to match these gap variable identities appropriately - eg to resolve a long-range dependency.
- Conjunctions and disjunctions are handled in a manner similar to ambiguities - by taking a script intersection of the two conjoined entity scenes.
- Ambiguities must eventually be resolved by a simple Bayesian maximum likelihood heuristic.
- There are various crude heuristics for fixing quantifier scopes.

With this rather ragbag set of techniques, the program is able to handle a wide range of the constructs of adult English, generally getting a sensible meaning structure out at the end - even when two or more complex features combine in the same meaning. It certainly cannot handle all the difficult and intricate cases studied by linguists; but there is no reason to suppose that it could not be progressively refined to do so.

The set of words known to the program is currently about 400, which includes about 80 non-verbs and 10 different m-scripts per verb for each of 30 verbs; and multiple senses for many words, particularly prepositions. Typical of the sentences it can successfully understand are:

*You can give the biscuit to me. Fred told Lucy the lamp was broken. Charlie wanted to go home. The boy who Lucy thought ate her cake went home. John will not have hurt himself. The big boy should not have eaten lucy's biscuit. Who did Lucy think ate her biscuit ? Lucy's biscuit was eaten by Fred. Fred cooked and ate the fish. Who went to London ? John. Lucy told Anna to go home when Charlie left. Has every bishop been seen with an actress ?*

For the **Generation Component** of the program, the same word m-script are arranged in a subsumption graph according to their right-hand meaning branches; searching down this graph from the root can give successively better and better approximations to the required meaning at any stage of generation.

The generation component starts with a meaning structure, which may be one got by the understanding component, or may be constructed by hand (and therefore have meanings not fully expressible with the program's limited vocabulary). A search is made down the subsumption graph for the word whose meaning W best captures the meaning M to be expressed, in that:

- M has as little as possible structure not in W and above the incoming trump links of W (extra structure in M below the trump links of W may be captured by words later in the generation process; anything above will be lost by use of W)
- W has as little as possible extra structure not in M (this adds meaning which was not in the intended meaning; the extra meaning could be right or wrong)
- No slot values in W actually contradict slot values in M.

Sometimes, when no appropriate word is to hand, the program has to choose some very bland word such as `do' or `thing'; but it carries on anyway. The m-script for W is m-unified with M, which creates the word sound scene of W (to be said later) and some new meaning scenes M'. Meaning in M but not in W has been carried back along the trump links into these new meaning scenes. The process is then repeated on each of the new meaning scenes M' ,

selecting other words to match them, until all meaning has been consumed or lost. At this stage you have a set of word sound scenes, with time order constraints between them (from time-order arrows in the word m-scripts). The words can then be said (or rather, typed out by the program). Various choices have to made using appropriate heuristics and methods:

- If some entity has been mentioned recently, it may be described by a pronoun.
- Generally, the speaker must decide how fully an entity needs to be described to define it uniquely for the listener - distinguishing it from other `nearby' candidates.
- Entity descriptions can sometimes be left out altogether (eg subjects within the scope of a relative pronoun).

This generation component can satisfactorily generate the same set of sentences as the understanding component can understand. It behaves gracefully when it cannot fully express the meaning - coming up with something grammatical, useful and not misleading.

The generation and understanding components are both used by the **learning component**, which we can now describe.

In a typical run of the learning program, one or more words are excised from its full vocabulary, and it is presented with a few sentences, each of which contains an unknown word. It is also given a meaning structure, which is the actual meaning of the sentence, but with some amount of random extra meaning slots and nodes tacked on. These simulate things the child may notice or infer about the scene, which are not part of the speaker's meaning. For each such sentence-meaning pair, the learning procedure does these steps:

1. From the provided meaning scene, generate as far as possible, using only m-scripts for known words which were heard in the sentence. This consumes words at either end of the sentence, leaving a stretch of words in the middle which are not used up (the unknown word is within the stretch of unused words); and one un-consumed meaning scene which corresponds to those words. This meaning scene forms the right branch of a learning example script.
2. From the stretch of unused words, understand as far as possible using the m-scripts of the known words. This yields a number of meaning scenes, and a word sound scene for the unknown word, which form the left branch of the same learning example script.

The program constructs one of these learning example scripts for each of the given sentence-meaning pairs, and m-intersects them together. It uses the `broad' m-intersection criterion, inferring a trump link between nodes i and j whenever S[j] s S[i] for all learning examples.

(Note that this differs from the procedure used in the proof of the fundamental theorem of language learning in Appendix C, in that a heavily pruned structure is constructed from each sentence-meaning pair, rather than the full sentence-meaning m-script Z containing all other word m-scripts, as in the proof of the theorem; but this pre-pruning just makes the m-intersection operation easier, without changing the result. Without pruning, the m-intersection operation will still remove all the extra example-specific structure.)

The result of running the program is that this learning algorithm reliably discovers the correct m-script for words of any part of speech. M-intersection finds just the structures in the left and right branches (given about 4 or more learning examples) with occasional extra slots, and then finds all the trump links - never proposing spurious ones.

The program has also been run in `bootstrap' mode starting from zero vocabulary, like an infant. When fed carefully-chosen example sentence-meaning pairs, starting with nouns, it has bootstrapped itself to a vocabulary of 50 words, and could easily go further.

This learning simulation has not simulated the Bayesian significance criterion in any detail, but has relied only on the `six examples' rule of thumb which emerges from the Bayesian calculation. Nor has it yet simulated the selection of appropriate examples from the many inappropriate examples a child hears during the day, and the information-theoretic criterion for rejecting inappropriate examples (eg homonyms). However, I believe that if these components of the program were to be built, they would work. The Bayesian statistics which show how they work are not complex and do not make any implausible assumptions.

The program shows that the tight interconnection between language generation, understanding and learning via the m-script algebra works not just in theory, but also in a running computer implementation. Something like this implementation might also be found in the human brain.

## *Footnotes to chapters (now orphaned)*

[1] This m-script is similar to the verb alternation m-scripts of section (5.7), but, unlike the verb alternators, has no syntactic changes of verb arguments on the left branch.

---

## *Footnotes*

[1] For instance, if a child learns a language by setting a few parameters, how does a bilingual child get by ? If parameters evolved to simplify the learning task, why did we complicate it again by providing multiple parameter sets just for multi-lingual children ?

---

## *Footnotes*

[1] In deriving the form of Bayes' rule above, we have used the fact that $P_A$(not R) = 1- $P_A$(R) 1.

[2] The value of d depends on the relative penalties of (a) believing the rule if it is not true and (b) not believing the rule if it is true.

[3] Using this best estimate is a heuristic to get to the answer; a more rigorous appoach is to integrate over all possible s(R) between 0 and 1, using probability densities. It gives the same answer.

[4] We assume that the outcome F(t+1) always has an average information content $I_{avg}$ bits ; if the rule has operated (with probability s), so that F(t+1) s G, then Ie(R) of these bits are accounted for by the rule, and the remaining bits occur with probability 2**(Ie - $I_{avg}$). Otherwise, if the rule does not fire (with probability (1-s) the $I_{avg}$ bits occur with probability 2**(- $I_{avg}$).

[5] For larger l, the number would be larger, for simple or complex rules. Therefore (perhaps surprisingly) a large value for l does not bias a learner towards simpler rules; it just makes all rule learning a bit slower.

[6] The design of the experiment ensured that the monkeys knew the general rule before learning the exception; it was top-down learning.

[7] The same variable `?A' appears in both branches of the m-script. This means that the person mentioned before the sound `shout' (in the left branch) is the same as the person doing the shouting (in the right branch meaning structure). The variable `?A' can convey a bare identity between the branches, but the trump link can convey much more - a whole script subtree.

## *Footnotes*

[1] To be precise, the process of language generation can sometimes alter the script meaning structure; see below.

[2] I have used the term inclusion (rather than subsumption) by analogy with set inclusion, to emphasise the close link between the script operations and set theory.

[3] I have used the term `intersection' rather than `generalisation' to maintain the clear link to set theory. When two scripts A and B are intersected together, the information in the result is like a set intersection of the information in the inputs. Any slot and value must appear in both A and B to appear in the result.